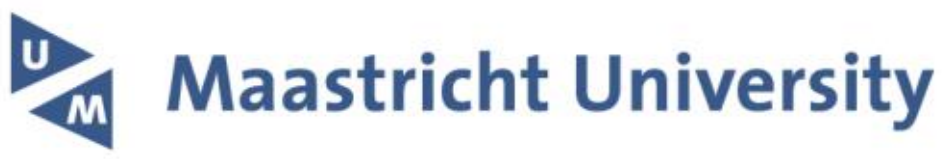

# Connecting online criminal behavior with machine learning

# Connecting Online Criminal Behavior with Machine Learning: Using Authorship Attribution to Analyze and Link Potential Online Traffickers

# Connecting Online Criminal Behavior with Machine Learning: Using Authorship Attribution to Analyze and Link Potential Online Traffickers

## Dissertation

to obtain the degree of Doctor at the Maastricht University,
on the authority of the Rector Magnificus, Prof. Dr. Pamela Habibović
in accordance with the decision of the Board of Deans,
to be defended in public on Wednesday 7th of January, 2026 at 16:00 hours

by

**Vageesh Kumar Saxena**

***Supervisors***:

| | |
|---|---|
| Dr. Gerasimos Spanakis | Maastricht University, Netherlands |
| Prof. Dr. Gijs van Dijck | Maastricht University, Netherlands |

***Assessment Committee***:

| | |
|---|---|
| Prof. Dr. Hans Nelen (***Chair***) | Maastricht University, Netherlands |
| Prof. Dr. Ana Isabel Martins Botto de Barros | Tilburg University, Netherlands |
| Prof. Dr. Floris J. Bex | Utrecht University, Netherlands |
| Prof. Dr. Jill Coster van Voorhout | Free University Amsterdam / Maastricht University, Netherlands |
| Prof. Dr. Anna M. Wilbik | Maastricht University, Netherlands |

***Organizations that contributed to the doctoral research*:**

| | |
|---|---|
| CFLW Cyber Strategies | Netherlands |
| German Research Center for Artificial Intelligence | Germany |
| Bashpole Softwares, Inc. | United States of America |

This research is partially supported by the Sector Plan Digital Legal Studies of the Dutch Ministry of Education, Culture, and Science. In addition, this research was made possible using the computational resources from Data Science Research Infrastructure (DSRI) hosted at Maastricht University.

| | |
|---|---|
| *Keywords:* | Machine Learning, Natural Language Processing, Computer Vision, Multimodal Learning, Responsible NLP, Authorship Attribution, Darknet and Escort Markets |
| *Printed by:* | Print Service Ede - `www.proefschriftenprinten.nl` |

ISBN/EAN:: 978-90-836415-2-2

# SUMMARY

Cyber-enabled trafficking operations exploit the anonymity of online platforms to conduct illicit activities across decentralized markets, presenting critical challenges for law enforcement. The sheer volume of advertisements and vendor activity within and across these platforms makes it difficult to gauge the size, scope, and structure of trafficking operations, hindering investigative efforts. To address this gap, this research proposes authorship attribution approaches—leveraging stylistic (writing) and photometric (visual) patterns in advertisements—as a tool to connect online vendor (author) activities across and within platforms. By identifying behavioral signatures (e.g., linguistic behavior, image composition, and multimodal patterns), these methods can potentially help law enforcement trace illegal vendor accounts and reconstruct criminal networks, providing a critical starting point for investigations. This research advances machine learning methodologies, datasets, and ethical frameworks to empower law enforcement and practitioners in disrupting these cyber-enabled trafficking trafficking networks.

The transition of criminal activities to digital platforms, including Darknet Markets and Online Escort Platforms, demands innovative and adaptive methods to analyze, link, and connect distributed and anonymized advertisements. This research introduces an authorship attribution approach designed to address darknet markets by examining textual advertisements to identify and verify vendors through stylistic patterns. Authorship attribution consists of two key tasks: authorship identification, which focuses on identifying an artifact (text, image, or audio) to a known candidate set of authors (here, vendors), and authorship verification, which determines whether two pieces of artifact originate from the same author based on stylistic similarity. In *VendorLink* (Chapter 2), experiments on three curated datasets from combination of seven darknet markets reveal that deep learning-based transformer architectures, employing contextualized embeddings to encode text ads, outperform traditional stylometric methods and conventional neural networks in vendor identification (a closed-set task focused on classifying text advertisements to predefined vendors) while also excelling in vendor verification (an open-set similar-ity task focused on linking text advertisements in emerging, previously unseen vendors through stylometric and linguistic similarity). To adapt to emerging low-resource scenarios — such as smaller upcoming darknet markets — a transfer learning strategy (similar to the transfer of knowledge from a teacher to a student) enables knowledge transfer from a large-scale transformer model to a more scalable and low-compute BiGRU architecture by initializing its embedding layer with pre-trained representations. The research also extends these principles to *IDTraffickers* (Chapter 3), a dataset composed of text advertisements from an online escort market where phone numbers serve as proxy labels for vendor identity. Instead of linking ads one at a time in the vendor verification setup, the study formalizes a compute-efficient retrieval approach that clusters ads by the vendor through similarity searches. Both studies incorporate error analysis and explainability techniques to validate the role of linguistic and stylometric patterns in linking illegal vendor accounts across Darknet and escort platforms. Collectively, these studies tackle the challenge of associating decentralized advertisements with criminal entities, laying the groundwork for knowledge graphs that map trafficking operations.

Furthermore, online escort advertisements often contain both text and images, necessitating multimodal analysis. The *MATCHED* dataset, introduced in Chapter 4, consists of escort advertisements from seven U.S. cities, providing a collection of textual descriptions and associated images that depict both the advertised services and escort appearances. By capturing writing styles from text advertisements and photometric styles from image advertisements, this dataset enables the development of a multimodal authorship attribution system that leverages text and vision transformer architectures to attribute vendor-specific behaviors. To effectively integrate textual and visual features, multiple latent fusion techniques are combined with advanced training strategies designed to jointly optimize vendor identification and verification tasks. This unified approach enhances the model's ability to recognize known vendors (supporting the vendor identification task) while detecting emerging ones in previously unseen advertisements (supporting the vendor identification task), thereby strengthening investigative capabilities. Improved retrieval performance in vendor verification further facilitates the construction of structured knowledge graphs, equipping investigators with more effective tools to link, analyze, and disrupt human trafficking networks.

While authorship attribution models offer promising potential for sup-porting criminal investigations, their real-world deployment requires careful consideration of the legal and ethical dimensions alongside their technical capabilities. These dimensions become particularly critical when models process sensitive personal data, influence investigative outcomes, or impact individuals' rights—contexts in which errors or biases can lead to wrongful accusations, privacy violations, or unfair treatment. To address these challenges, this research develops a systematic framework (Chapter 5) grounded in four core principles: (1) Privacy and Data Protection; ensuring the lawful processing of personal information, (2) Fairness and Non-discrimination; aiming to prevent biased outcomes that could disadvantage specific demographic groups, (3) Transparency and Explainability; promoting human oversight and transparency in model decisions; and (4) Broader Societal Impact; evaluating the potential harm to participants, stakeholders, and the broader environment. In addition to these four pillars, the framework incorporates accountability by providing role-specific guidelines for each stakeholder involved in the authorship attribution Software Development Life Cycle. This ensures that all parties—from designers and developers to investigators—understand their responsibilities and are equipped to address ethical concerns effectively. The study presents a structured questionnaire that guides practitioners in navigating trade-offs, such as balancing investigative accuracy with privacy protection during model design and deployment. By embedding these Ethical, Legal, and Social Issues into the authorship attribution software cycle, the framework proactively mitigates the risks of misuse while aligning the technology with societal values and legal standards. It is important to note that this framework is specifically focused on authorship attribution applications within the domain of Natural Language Processing.

By advancing authorship attribution approaches, this research provides scalable tools for tracing, connecting, studying, and disrupting cyber-enabled trafficking networks. It also facilitates the creation of IDTraffickers and MATCHED datasets while responsibly developing authorship attribution methods that significantly improve accuracy in identifying connections within darknet and online escort marketplaces. However, the research has several limitations, including the reliance on labeled data, the computational demands of transformer models, and the challenge of adapting to evolving criminal behavior. Future research will focus on exploring the model's generalization capabilities across broader contexts, including data from multiple online marketplaces and countries collected from more recent sources, working on enhancing model explainability, and establishing international collaboration to deploy and test the practical applications of authorship attribution approaches in combating the global nature of online trafficking. By bridging machine learning, ethics, and multimodal analysis, this research provides a robust solution for responsibly disrupting decentralized cyber-enabled trafficking operations, all while prioritizing societal safeguards and legal protections.

# Samenvatting

Cybergestuurde smokkeloperaties maken misbruik van de anonimiteit van online platformen om illegale activiteiten uit te voeren op gedecentraliseerde markten, wat kritische uitdagingen vormt voor de wetshandhaving. Het enorme volume aan advertenties en de activiteit van verkopers binnen en tussen deze platformen maakt het moeilijk om de omvang, reikwijdte en structuur van mensenhandeloperaties in te schatten, hetgeen de onderzoeksinspanningen belemmert. Om deze kloof te dichten, stelt dit onderzoek methoden voor voor auteursattribuering —door gebruik te maken van stilistische (schrijf-) en fotometrische (visuele) patronen in advertenties— als een hulpmiddel om online activiteiten van verkopers (auteurs) binnen en tussen platformen te verbinden. Door gedragskenmerken te identificeren (bijvoorbeeld linguïstisch gedrag, beeldcompositie en multimodale patronen) kunnen deze methoden mogelijk de wetshandhaving helpen illegale verkopersaccounts op te sporen en criminele netwerken te reconstrueren, wat een cruciaal startpunt voor onderzoeken biedt. Dit onderzoek verbetert machine learning-methodologieën, datasets en ethische kaders om de wetshandhaving en betrokken partijen te versterken in het verstoren van deze cyber-enabled mensenhandelnetwerken.

De verschuiving van criminele activiteiten naar digitale platformen, waaronder Darknet-markten en online escortplatformen, vraagt om innovatieve en adaptieve methoden om verspreide en geanonimiseerde advertenties te analyseren, te koppelen en te verbinden. Dit onderzoek introduceert een methode voor auteursattribuering die is ontworpen om darknetmarkten aan te pakken door tekstuele advertenties te onderzoeken om verkopers te identificeren en te verifiëren via stilistische patronen. Auteursattribuering bestaat uit twee belangrijke taken: auteursidentificatie, wat zich richt op het toewijzen van een artefact (tekst, beeld of audio) aan een bekende kandidaatset van auteurs (in dit geval verkopers), en auteursverificatie, wat bepaalt of twee artefacten afkomstig zijn van dezelfde auteur op basis van stilistische gelijkenis. In *VendorLink* (Chapter 2) tonen experimenten op drie samengestelde datasets, samengesteld uit zeven darknetmarkten, aan dat deep learning-gebaseerde transformer-architecturen, die gebruikmaken van contextuele embeddings om tekstadvertenties te coderen, beter presteren dan traditionele stylometrische methoden en conventionele neurale netwerken bij de identificatie van verkopers (een gesloten taak gericht op het classificeren van tekstadvertenties aan vooraf gedefinieerde verkopers) en tevens uitblinken in auteursverificatie (een open taak gericht op het koppelen van tekstadvertenties in opkomende, voorheen niet waargenomen verkopers door middel van stilometrische en linguïstische gelijkenissen). Om in te spelen op opkomende situaties met beperkte middelen — zoals kleinere, opkomende darknet-markten — maakt een transfer learning-strategie (vergelijkbaar met de overdracht van kennis van een leraar naar een leerling) kennisoverdracht mogelijk van een grootschalig transformer-model naar een meer schaalbare en rekenefficiënte BiGRU-architectuur door de inbeddingslaag te initialiseren met vooraf getrainde representaties. Het onderzoek past deze principes ook toe op *IDTraffickers* (Chapter 3), een dataset bestaande uit tekstadvertenties van een online escortmarkt waar telefoonnummers dienen als proxy-labels voor de identiteit van de verkoper. In plaats van advertenties één voor één te koppelen in de auteursverificatieopzet, formaliseert de studie een rekenefficiënte retrieval-aanpak die advertenties per verkoper clustert door middel van gelijkeniszoekopdrachten. Beide studies maken gebruik van foutenanalyse en technieken voor verklaarbaarheid om de rol van linguïstische en stilometrische patronen te valideren bij het koppelen van illegale verkopersaccounts over Darknet- en escortplatformen heen. Gezamenlijk pakken deze studies de uitdaging aan om gedecentraliseerde advertenties te associëren met criminele entiteiten, waarmee de basis wordt gelegd voor kennisgrafieken die mensenhandeloperaties in kaart brengen.

Bovendien bevatten online escortadvertenties vaak zowel tekst als beelden, wat een multimodale analyse noodzakelijk maakt. De *MATCHED*-dataset, geïntroduceerd in Chapter 4, bestaat uit escortadvertenties uit zeven Amerikaanse steden en biedt een verzameling van tekstuele beschrijvingen en bijbehorende beelden die zowel de geadverteerde diensten als de verschijning van escorts weergeven. Door de schrijfstijlen in tekstadvertenties en de fotometrische stijlen in beeldadvertenties vast te leggen, maakt deze dataset de ontwikkeling mogelijk van een multimodaal systeem voor auteursattribuering dat gebruikmaakt van tekst- en visie-transformerarchitecturen om verkopersspecifiek gedrag toe te schrijven. Om tekstuele en visuele kenmerken effectief te integreren, worden meerdere latente fusietechnieken gecombineerd met geavanceerde trainingsstrategieën die gericht zijn op het gezamenlijk optimaliseren van de taken van verkopersidentificatie en -verificatie. Deze geïntegreerde aanpak verbetert het vermogen van het model om bekende verkopers te herkennen (ondersteunend de taak van verkopersidentificatie) terwijl het opkomende verkopers in voorheen ongeziene advertenties detecteert (ondersteunend de taak van verkopersidentificatie), waardoor de onderzoeksinspanningen worden versterkt. Verbeterde retrieval-prestaties in verkopersverificatie vergemakkelijken bovendien de constructie van gestructureerde kennisgrafieken, waarmee onderzoekers effectievere hulpmiddelen krijgen om mensenhandelnetwerken te koppelen, analyseren en te verstoren.

Hoewel modellen voor auteursattribuering veelbelovend potentieel bieden ter ondersteuning van strafrechtelijk onderzoek, vereist hun inzet in de praktijk een zorgvuldige afweging van juridische en ethische dimensies naast hun technische capaciteiten. Deze dimensies worden bijzonder kritisch wanneer modellen gevoelige persoonlijke gegevens verwerken, invloed uitoefenen op onderzoeksresultaten of de rechten van individuen raken — situaties waarin fouten of vooringenomenheid kunnen leiden tot onterechte beschuldigingen, schendingen van de privacy of oneerlijke behandeling. Om deze uitdagingen aan te pakken, ontwikkelt dit onderzoek een systematisch raamwerk (Chapter 5) gebaseerd op vier kernprincipes: (1) Privacy en gegevensbescherming: het waarborgen van de rechtmatige verwerking van persoonlijke informatie, (2) Eerlijkheid en non-discriminatie, met als doel bevooroordeelde uitkomsten te voorkomen die specifieke demografische groepen kunnen benadelen, (3) Transparantie en verklaarbaarheid: het bevorderen van menselijk toezicht en transparantie in modelbeslissingen, en (4) Brede maatschappelijke impact: het evalueren van de potentiële schade voor deelnemers, belanghebbenden en de bredere omgeving. Naast deze vier pijlers omvat het raamwerk ook verantwoordelijkheid door het verstrekken van rol-specifieke richtlijnen voor elke belanghebbende die betrokken is bij de Software Development Life Cycle van auteursattribuering. Dit zorgt ervoor dat alle partijen — van ontwerpers en ontwikkelaars tot onderzoekers — hun verantwoordelijkheden begrijpen en in staat zijn ethische kwesties effectief aan te pakken. De studie presenteert een gestructureerde vragenlijst die praktijkbeoefenaars helpt bij het navigeren door de afwegingen, zoals het balanceren van onderzoeksnauwkeurigheid met privacybescherming tijdens het ontwerp en de inzet van het model. Door deze ethische, juridische en sociale kwesties te integreren in de softwarecyclus voor auteursattribuering, beperkt het raamwerk proactief de risico's van misbruik, terwijl de technologie wordt afgestemd op maatschappelijke waarden en juridische normen. Het is belangrijk op te merken dat dit raamwerk specifiek gericht is op toepassingen van auteursattribuering binnen het domein van natuurlijke taalverwerking.

Door auteursattribuering-benaderingen te verbeteren, levert dit onderzoek schaalbare hulpmiddelen voor het traceren, verbinden, bestuderen en verstoren van cybergestuurde trafficking netwerken. Het faciliteert tevens de creatie van de IDTraffickers- en MATCHED-datasets, terwijl het op verantwoorde wijze methoden voor auteursattribuering ontwikkelt die de nauwkeurigheid in het identificeren van verbindingen binnen darknet- en online escortmarktplaatsen aanzienlijk verbeteren. Het onderzoek kent echter verschillende beperkingen, waaronder de afhankelijkheid van gelabelde data, de computationele eisen van transformermodellen en de uitdaging om zich aan te passen aan veranderend crimineel gedrag. Toekomstig onderzoek zal zich richten op het verkennen van de generalisatiecapaciteiten van het model in bredere contexten, met inbegrip van data van meerdere online marktplaatsen en landen, verzameld uit recentere bronnen, op het verbeteren van de verklaarbaarheid van modellen en op het opzetten van internationale samenwerking om de praktische toepassingen van auteursattribuering benaderingen in de bestrijding van de mondiale aard van online trafficking te implementeren en te testen. Door machine learning, ethiek en multimodale analyse te verbinden, levert dit onderzoek een robuuste oplossing voor het op verantwoorde wijze verstoren van gedecentraliseerde cyber-enabled trafficking operaties, terwijl tegelijkertijd de maatschappelijke waarborgen en juridische bescherming voorop staan.

# Acknowledgments

First and foremost, I express my deepest gratitude to my supervisors, Asst. Prof. Jerry Spanakis and Prof. Gijs van Dijck, for their invaluable guidance and unwavering support throughout this journey. Their steadfast dedication to research that aligns with personal conviction—regardless of trends or external pressures—has been a constant source of inspiration. The autonomy and flexibility they extended to me from the beginning empowered me to delve into my deepest research interests, fueling my motivation and significantly contributing to the successful completion of this PhD.

I am profoundly thankful to my Law & Tech Lab colleagues for the memorable moments and enriching discussions we shared over the past four years. My heartfelt appreciation goes to Andreea Grigoriu, Antoine Louis, Bogdan Covrig, Bram Rijsbosch, Caroline Cauffman, Catalina Goanta, Constanta Rosca, Gustavo Arosemena Solorzano, Hannes Westermann, Hellen van der Kroef, Iris Xu, Johanna Gunawan, Kevin Jacobs, Lucas Giovanni, Lyra Hoeben-Kuil, Manuela Paolini e Silva, Marta Kołacz, Mindy Duffourc, Monika Leszczynska, Pedro Hernández Serrano, Rohan Nanda, Roland Moerland, Ruta Liepina, Samyak Sheth, Shashank Chakravarthy, Tom Vos, Vlada Druță, Wen-Ting Yang, and Yongle Chao. I am equally indebted to Nils Rethmeier, Prof. Aurelia Tamò-Larrieux, and Mark van Staalduinen for their guidance, collaboration, and unwavering support, as well as to Benjamin Ashpole for the enriching experience during my internship at Bashpole Software, Inc.

Special thanks to Chris Kuipers, Jordy Frijns, Marcel Brouwers, and Vincent Emonet from the Data Science Research Infrastructure team at Maastricht University. Their diligent maintenance of computational resources was pivotal to the realization of this research.

My deepest gratitude goes to my family: my parents, Mr. Anand Das Saxena and Mrs. Nidhi Saxena; my brother, Vishesh Saxena; my wife, Pooja Nagendra Babu; my in-laws, Mrs. C.P. Kanaka Durga, Mr. K.N. Nagendra Babu, Mrs. Meghana Nagendra Babu, and Mr. Anant Sharma; and my late grandparents Mr. Jagdish Saran Saxena, Mrs. Prakasha Saxena, Mr. Shrikrishna Das Saxena, and Mrs. Vidyawati Saxena for their unconditional love, encouragement, and unwavering support. I am also grateful to my extended family members: Adarsh Saxena, Aditya Saxena, Akhila V. Dev, Ankit Asthana, Anshika Saxena, Chandan Saxena, Geetha Saxena, Jyoti Saxena, Nikhil Saxena, Nitin Saxena, Ritika Saxena, Roopam Saxena, Sumali Saxena, Surabhi Pradhan Sinha, Varsha Saxena, and Yatin Saxena.

Lastly, I extend my heartfelt thanks to my friends Ami Panchal, Andres Urquizo, Ayushi Sachan, Burak Özmen, Carolina Canencia, Devang Jasani, Fabian Barteldes, Goncalo Mordido, Hatim Bentahar, Karan Kr. Seth, Karthik Ajith Kumar, Mahendra Kumar, Mardhini Vaadwal Sreedar, Nina Harlacher, Rahil Engineer, Rahul Gandhok, Ratandeep Swami, Rishabh Gupta, Roshan Prakash Rane, Sahana Govindappa Premakumari, Sandeep Hanumantha Rayudu, Saurabh Mathur, Shubham Kushwaha, and Taryn Gehman for their support and friendship throughout this journey.

*Vageesh Saxena*
*Maastricht, January 2025*

# PUBLICATIONS

Most of the material presented in this manuscript is based on papers previously published in venues throughout the PhD program:

- **Vageesh Saxena**, Nils Rethmeier, Gijs van Dijck, and Gerasimos Spanakis. 2023b. VendorLink: An NLP approach for Identifying & Linking Vendor Migrants & Potential Aliases on Darknet Markets. In Proceedings of the 61st Annual Meeting of the Association for Computational Linguistics (Volume 1: Long Papers), pages 8619–8639, Toronto, Canada. Association for Computational Linguistics.

- **Vageesh Saxena**, Benjamin Ashpole, Gijs van Dijck, and Gerasimos Spanakis. 2023a. IDTraffickers: An Authorship Attribution Dataset to link and connect Potential Human-Trafficking Operations on Text Escort Advertisements. In Proceedings of the 2023 Conference on Empirical Methods in Natural Language Processing, pages 8444–8464, Singapore. Association for Computational Linguistics.

- **Vageesh Saxena**, Aurelia Tamò-Larrieux, Gijs van Dijck, and Gerasimos Spanakis. 2025a. Responsible Guidelines for Authorship Attribution Tasks in NLP. Ethics and Information Technology 27, 16 (2025), Springer Nature.

- **Vageesh Saxena**, Benjamin Ashpole, Gijs Van Dijck, and Gerasimos Spanakis. 2025b. MATCHED: Multimodal Authorship-Attribution To Combat Human Trafficking in Escort-Advertisement Data. In Findings of the Association for Computational Linguistics: ACL 2025, pages 4334–4373, Vienna, Austria. Association for Computational Linguistics.

For each of these works, the PhD candidate originated the research direction, curated the necessary data, designed the proposed methodologies, implemented and ran all experiments, conducted the analyses, and drafted the papers in full, which were subsequently refined with input from his supervisors.

For completeness, research completed during the PhD program that falls outside the scope of this thesis is reported below:

- Antoine Louis, **Vageesh Kumar Saxena**, Gijs van Dijck, and Gerasimos Spanakis. 2025. ColBERT-XM: A Modular Multi-Vector Representation Model for Zero-Shot Multilingual Information Retrieval. In Proceedings of the 31st International Conference on Computational Linguistics, pages 4370–4383, Abu Dhabi, UAE. Association for Computational Linguistics.

**The Association for Computational Linguistics (ACL) and Empirical Methods in Natural Language Processing (EMNLP):** ACL is a leading international society for researchers in computational linguistics and Natural Language Processing (NLP). Established in 1962 as the Association for Machine Translation and Computational Linguistics (AMTCL) and renamed in 1968, the ACL has played a pivotal role in advancing the field. The organization hosts an influential annual meeting each summer and sponsors the journal Computational Linguistics, published by MIT Press. Both the conference and the journal have showcased significant research breakthroughs and serve as benchmarks within the discipline. Similarly, EMNLP is recognized as a leading conference in NLP and Artificial Intelligence (AI). Along with the ACL and the North American Chapter of the Association for Computational Linguistics (NAACL), EMNLP is considered one of the three primary high-impact NLP research conferences. The following past acceptance rates (in the years of publication) from the ACL (Rogers et al., 2023; Che et al., 2025) and EMNLP (Bouamor et al., 2023) illustrate the association's rigorous peer-review process:

| Conference | Year | Submissions | Accepted (Main) | Accepted (Findings) | Acceptance Rate (Main) |
|---|---|---|---|---|---|
| EMNLP | 2023 | 4904 | 1047 | 1060 | 21.30% |
| ACL | 2023 | 4864 | 1074 | 901 | 22.08% |
| ACL | 2025 | 8300 | 1700 | 1392 | 20.48% |

Table 1: Acceptance rate of Long-Papers in the **ACL-2023**, **EMNLP-2023**, and **ACL-2025** main conference tracks. All the research work presented in this thesis were accepted for long-paper main conference tracks.

# Disclaimer

This research addresses highly sensitive and explicit topics, including human trafficking, illegal trades involving drugs, weapons, and organs, as well as financial frauds conducted on Darknet and escort markets. The research incorporates datasets containing detailed text descriptions of such activities alongside images that may include nude, semi-nude, sexual, and explicit content related to escort advertisements.

Readers are advised to approach the content with caution. If they find any material distressing or triggering, please consider seeking support from appropriate mental health or support services. This research is conducted with the utmost respect for ethical standards, aiming to contribute to understanding, preventing, and disrupting these illegal activities.

# Contents

# LIST OF FIGURES

# List of Tables

# List of Algorithms

# List of Abbreviations

**AA** Authorship Attribution

**ACL** The Association For Computational Linguistics

**ads** Advertisements

**AI** Artificial Intelligence

**BERT** Bidirectional Encoder Representations From Transformers

**BiGRU** Bidirectional Gated Recurrent Unit

**BLIP** Bootstrapping Language-Image Pre-training

**CLIP** Contrastive Language-Image Pre-Training

**CNNs** Convolutional Neural Networks

**Darknet** Dark Web

**DL** Deep Learning

**DPIAs** Data Protection Impact Assessments

**EMNLP** Empirical Methods In Natural Language Processing

**EU** Eurpean Union

**FAISS** Facebook AI Similarity Search

**GDPR** General Data Protection Regulation

**GPT** Generative Pre-trained Transformer

**HR** High-Resource

**ITC** Image-Text Contrastive

**ITM** Image-Text Matching

**LEAs** Law Enforcement Agencies

**LR** Low-Resource

**MAA** Multimodal Authorship Attribution

**ML** Machine Learning

**NLP** Natural Language Processing

**OOD** Out-Of-Data

**Q-Former** Querying Transformer

**ResNet** Residual Neural Network

**RoBERTA** Robustly Optimized BERT Pre-training Approach

**SDLC** Software Development Life Cycle

**SOTA** State-of-the-art

**U.S.** United States Of America

**VGG** Visual Geometry Group

**ViTs** Vision Transformers

# 1

# INTRODUCTION

*"Artificial intelligence is not a substitute for human intelligence; it is a tool to amplify human creativity and ingenuity."*
*– Fei-Fei Li*

## 1.1 CYBERCRIME: DEFINITION AND SCOPE

In an era of digital interconnectedness, cybercrime has emerged as a pervasive threat that exploits the vulnerabilities of global networks and systems. Unlike traditional crimes, cybercrime transcends geographical boundaries, leveraging the anonymity and scale offered by the Internet to inflict harm on individuals, organizations, and states (Borwell et al., 2021). This dynamic and adaptive phenomenon disrupts systems and poses significant challenges for Law Enforcement Agencies (LEAs), researchers, and policymakers. As technology evolves, so does cybercrime, necessitating a comprehensive understanding of its history, mechanisms, and implications.

Cybercrime refers to illegal activities conducted through or targeting digital systems that leverage the interconnectedness and anonymity of the Internet. It encompasses various activities, from sophisticated attacks on critical infrastructure to exploiting online platforms for traditional crimes, and can be categorized into cyber-trespass (e.g., unauthorized system access), cyber-deception/theft (e.g., identity theft, online fraud, digital piracy), cyber-porn/obscenity (e.g., child sexual exploitation materials), cyber-violence (e.g., cyberstalking; cyber-terrorism), cyber-trading/trafficking (e.g., trading and trafficking of illegal products, materials, or individuals), and many more (Bossler and Berenblum, 2019). The extent of these crimes in most nations worldwide is challenging to estimate due to a lack of standardized legal definitions for these offenses and few valid, reliable official statistics (Holt and Bossler, 2015). Either way, these crimes can always be classified into two groups (Furnell et al., 2015):

- **Cyber-Dependent Crime:** These crimes cannot occur without digital technologies. Examples include hacking, ransomware attacks, and Distributed Denial-of-Service (DDoS) attacks. These crimes often exploit software vulnerabilities, targeting systems directly to achieve malicious objectives. For instance, ransomware encrypts critical data and demands payment for its release, crippling essential services such as healthcare and transportation systems.
- **Cyber-Enabled Crime :** Traditional crimes are enhanced or facilitated through digital means. These include fraud, human trafficking, drug trafficking, and child exploitation. Perpetrators leverage online platforms like social media or encrypted communication channels to scale their operations. For example, human trafficking networks often use online advertisements (ads) and messaging platforms to coordinate illicit activities, making it harder for LEAs to trace these operations.

## 1.2 THE GLOBAL LANDSCAPE OF CYBER-ENABLED CRIMES

### 1.2.1 SHIFT FROM OFFLINE TO ONLINE CRIME

The evolution of crime from offline to online has fundamentally altered the nature of criminal activities. Historically, crimes such as fraud, trafficking, and illicit trade were limited by geographical boundaries and physical constraints. Criminals relied on in-person interactions, physical exchanges, and localized networks to conduct their activities. However, the advent of the internet removed these barriers, enabling crimes to scale globally with minimal resources. This shift to online platforms has exponentially increased the reach and efficiency of criminal operations, particularly in cyber-enabled crimes.

For instance, drug trades, which were historically confined to physical locations or in-person distribution, have migrated to online Darknet markets such as Silk Road (Martin, 2014), AlphaBay (Li et al., 2021b), and others (Buxton and Bingham, 2015), transforming how drugs are sold and distributed. These platforms allow vendors to reach customers globally, providing anonymity through cryptocurrency transactions and encrypted communications (Chawki, 2022). These marketplaces have enabled the expansion of the drug trade while reducing the likelihood of physical encounters between buyers and sellers.

Similarly, sex trafficking, the most prevalent form of human trafficking —a crime once confined to brothels or street-level exploitation—has increasingly moved online (Polaris, 2023b,a). Online platforms have become instrumental tools for traffickers, enabling them to advertise services, recruit victims, and coordinate their operations with unprecedented efficiency (Lugo-Graulich and Meyer, 2021; Lugo-Graulich, 2024). Unlike the visible, location-bound nature of traditional trafficking methods, digital platforms offer traffickers the dual advantage of broader reach and reduced risk of detection. Websites, social media channels, and encrypted communication tools now serve as virtual marketplaces where exploitation thrives in concealed corners of cyberspace (Europol, 2021). Before regulatory crackdowns, platforms such as Backpage and Craigslist operated as major hubs for trafficking activities, providing traffickers with a degree of anonymity and access to a vast customer base that was previously unattainable through physical operations (Lugo-Graulich and Meyer, 2021). By leveraging these platforms, traffickers minimized their operational risks while maximizing their reach, exploiting the internet's vast infrastructure to perpetuate harm on a large scale.

The COVID-19 pandemic profoundly reshaped the global cybercrime landscape, accelerating a shift from physical to digital crime due to widespread lockdowns and the sudden reliance on online systems. Cybercriminals seized this unprecedented opportunity to exploit societal fear and digital dependency. Between January and April 2020, over 907,000 COVID-19-themed spam messages, 737 malware incidents, and 48,000 malicious domains were identified globally, all leveraging pandemic-related narratives to deceive individuals and institutions. These cyberattacks targeted personal data, financial assets, and critical infrastructure, particularly healthcare systems, which were already overwhelmed by the pandemic. Ransomware attacks on hospitals, for instance, disrupted critical medical services, highlighting the severity of these threats (Interpol, 2020). This surge in malicious activity reflected a broader trend of opportunism, with cyber-enabled criminals rapidly adapting their methods to exploit the anxieties and necessities of the pandemic.

During this period, darknet markets also experienced a surge in activity, with traffickers exploiting increased demand for medical supplies and personal protective equipment (PPE). A study by the United Nations Office on Drugs and Crime (UNODC) noted a proliferation of listings for counterfeit or non-existent COVID-19-related products, underscoring the adaptability of organized criminal networks to capitalize on crises (UNODC, 2021). Similarly, online platforms became pivotal for human trafficking activities, with perpetrators leveraging encrypted communication tools and social media to recruit victims and advertise services (Coxen et al., 2023). These trends reveal the dynamic nature of cybercrime, where technological innovation is rapidly repurposed for malicious activities. Understanding these shifts and developing adaptive countermeasures remains critical to safeguarding individuals and institutions as society digitizes.

### 1.2.2 Difference between Traditional and Cybercrime

The evolution of crime in the digital age has blurred the lines between cybercrime and traditional crime. While both forms of crime often stem from shared motives, such as financial gain or coercion, their execution, societal implications, and detection challenges differ fundamentally.

**Technological Integration in Criminal Activities:** Criminals increasingly incorporate digital tools into traditional crimes, leading to hybrid offenses. As highlighted by Montoya et al. (2013), 41% of fraud cases in East Netherlands involved digital elements such as phishing, while 16% threats utilized digital communication tools. Conversely, physical crimes like burglary rarely involve Information and Communication Technology (ICT); only 2.9% of residential burglaries included stealing bank cards for digital theft. Darknet markets, accessible via specialized software like Tor, further enable the anonymous trade of illicit goods such as drugs, weapons, and counterfeit documents, complicating LEA's effort (Krylova, 2019; Andrei and Veltri, 2024). Similarly, human trafficking networks exploit online platforms, including escort service websites, to advertise and exploit victims, using digital anonymity to evade detection (L'Hoiry et al., 2024; National Institute of Justice, 2024). These technological integrations highlight the evolving landscape of criminal activities and the challenges LEAs poses in addressing hybrid and purely digital crimes.

**Victim-Offender Dynamics :** In traditional crimes, offenders often have to be close to their victims, but cybercrime enables perpetrators to target individuals across vast distances. For example, phishing scams and ransomware attacks allow offenders to victimize individuals or organizations in different countries without meeting them physically (Montoya et al., 2013). Similarly, in fraud cases, cybercriminals often exploit anonymity to deceive strangers, reducing the risk of exposure and retaliation (Borwell et al., 2024). Darknet markets further exacerbate these dynamics by operating on anonymous platforms, where the true identities of buyers and sellers remain concealed. This anonymity facilitates fraud, as many transactions fail to be completed, with buyers sending payments but never receiving shipments (Bergeron et al., 2022). In human trafficking, victim-offender relationships are shaped by trust, coercion, or pre-existing ties, such as familial or romantic connections over social media websites, which traffickers exploit to establish control (SHI, 2021). Online platforms and anonymity shield traffickers, enabling cross-border operations and complicating victim identification (OSCE, 2023). Furthermore, traffickers often maintain hierarchical networks using technology to monitor victims, ensuring their compliance and sustaining exploitation (Cockbain, 2018).

**Psychological and Emotional Impacts:** Victim experiences differ significantly between cybercrime and traditional crime. Borwell et al. (2024) highlights that cybercrime victims often endure prolonged psychological stress due to the anonymity of perpetrators and the global reach of attacks. By contrast, traditional crimes induce more immediate but localized emotional responses, such as the short-term fear and distress experienced by burglary victims. Cybercrime also amplifies offenders' ability to depersonalize victims, lowering psychological barriers to offending as actions may feel less harmful in the absence of face-to-face interactions. Darknet markets aggravate these dynamics, with victims of fraud or exploitation frequently experiencing anxiety, mistrust, and helplessness due to the anonymity of perpetrators and the difficulty of force seeking justice (Ferrara et al., 2021). Perpetrators often rationalize their actions, detaching themselves from the harm caused by the lack of direct interaction (Gray, 2024). Similarly, victims of online escort markets face severe psychological trauma, including depression, anxiety, and post-traumatic stress disorder (PTSD), particularly in cases of trafficking or coercion (Altun et al., 2017). Feelings of shame and helplessness are common, exacerbated by societal stigma and continuous monitoring by traffickers, which amplifies their distress (Hossain et al., 2010). Anonymity on online platforms allows traffickers to evade detection and fosters a sense of impunity, further lowering psychological barriers to exploitation (OSCE, 2023).

**Law Enforcement Challenges and Resource Allocation:** Policing cybercrime is filled with challenges, including underreporting and resource allocation issues. The lack of clear distinctions between cybercrime and traditional crime often leads to misclassification, with hybrid offenses like online fraud and trafficking crimes being recorded as traditional crimes, thus masking the true extent of cybercrime (Montoya et al., 2013). Addressing these issues requires specialized resources, including digital forensics, international collaboration, and advanced tools to trace anonymized networks, which the traditional policing methods often lag (Zhang et al., 2012). Furthermore, the integration of digital tools into investigative processes remains limited compared to physical tools (Weulen Kranenbarg, 2021). The sheer volume and global scope of cybercrime further exacerbate these challenges, making robust adaptation of LEA's methods imperative for effective classification, investigation, and better resource allocation.

## 1.3 Cyber-Enabled Trafficking Activities

### 1.3.1 Online Trafficking Activities on Surface Web, Social Media, and Darknet Marketplaces

The Internet has profoundly reshaped the mechanisms and scale of illicit trafficking, blurring the lines between digital and physical domains. Online platforms now facilitate activities once confined to offline channels, enabling traffickers to exploit new avenues for anonymity, logistical coordination, and financial transactions—thereby pressuring LEAs and policymakers to develop adaptive detection, disruption, and prevention strategies. Both the publicly accessible surface web and hidden segments of the dark web serve as conduits for these various forms of trafficking. The surface web, or the publicly accessible Internet, allows traffickers to use open channels to reach broad audiences quickly while trusting that LEA's limited resources will hamper rapid identification (Deeb-Swihart et al., 2019). Conversely, the dark web offers specialized anonymity software such as Tor, enabling hidden marketplaces and forums that connect international networks of perpetrators and buyers (Jardine, 2015). These overlapping layers of the Internet facilitate a continuum of criminal activity, from open ads that rely on the crowding-out effect of legitimate content to cloaked forums where illicit goods and services can be exchanged with little fear of exposure.

A significant dimension of this phenomenon is the trafficking of human beings, which manifests across the surface web (DeLateaur, 2016) and darker corners of the Internet, aka Dark Web (Nazah et al., 2020; Graham, 2023). Sex trafficking, for instance, leverages public-facing platforms—ranging from mainstream social media sites (Latonero, 2011; Moyo et al., 2024) to more specialized classified advertisement websites (Ibanez and Suthers, 2014)—to recruit and advertise victims under pseudonymous identities. Labor trafficking follows similar patterns, with job postings and recruitment schemes luring individuals into exploitative working conditions (Volodko et al., 2020). Organ trafficking, also a specialized form of human trafficking, though less visible, also utilizes bio-trafficking and darknet forums to connect individuals willing to sell or procure organs in violation of international laws, capitalizing on desperation and circumventing legal channels (Columb, 2015; Rawat et al., 2022). In all these cases, perpetrators benefit from encrypted communication tools and anonymizing platforms to evade detection. More insidious forms of human trafficking, such as child exploitation, also thrive in these digital environments (Lee et al., 2020; Liggett et al., 2020), as perpetrators use hidden forums and secure file-sharing platforms to exchange illegal materials. This online shift complicates victim identification and amplifies trauma, as content can be rapidly disseminated to a global audience.

Beyond human trafficking, illegal trade in goods and services has similarly migrated to online platforms, reflecting the broader transformation of criminal enterprises in the digital era. Weapons trafficking, once reliant on covert physical exchanges, now leverages hidden discussion boards and encrypted messaging channels to connect buyers and sellers across borders (Rhumorbarbe et al., 2018; Holt and Lee, 2023). The trade-in of illegal drugs—including the sale of synthetic opioids, cannabis, and novel psychoactive substances—occurs on websites ranging from the surface web (Liu and Bharadwaj, 2020), social media (Mackey and Liang, 2013; Hu et al., 2023), to highly protected darknet marketplaces (Buxton and Bingham, 2015; Broséus et al., 2016) accessible only via specialized software (Owenson and Savage, 2015). Similarly, wildlife trafficking—the illegal trade of exotic animals and their parts, such as ivory or tiger bones—also utilizes online criminal markets (Lavorgna, 2014) and social media websites (Xu et al., 2020; Wyatt et al., 2022), reaching a global audience.

Although not exactly a trafficking activity, financial crimes intersect with these illicit crimes in multiple ways. Traffickers dealing with humans, weapons, drugs, or counterfeit documents often incorporate escrow systems and cryptocurrency payments, preventing conventional financial tracing and allowing traffickers to maintain anonymity (Kethineni and Cao, 2020). These cryptocurrencies have emerged as a dominant medium for anonymizing transactions. That said, traditional banking platforms (Van Dijk et al., 2018; Coster van Voorhout, 2020) and shadow businesses (Muller-Tabanera and Huang, 2021; Talbott et al., 2021; Rosenow and Munk, 2023) also remain part of these crimes, making the flow of illicit funds exceedingly complex to trace. Identity theft and fraud further fuel this ecosystem (Hedayati, 2012): stolen personal information is bought and sold on dark web markets, equipping criminals with the tools to commit tax fraud, credit card fraud, and other forms of financial exploitation. The resulting cycle—illicit trade generating proceeds, followed by high-tech laundering—reinforces the resilience of these criminal networks.

In summary, cyber-enabled trafficking spans multiple domains—from human exploitation, weapon sales, and illicit drug markets to wildlife crimes—all thriving on the logistical advantages and anonymity on the surface web and darknet marketplaces. Online illegal trades utilizing encrypted channels, payments through cryptocurrencies, and the accessibility of illegal online platforms have accelerated these criminal operations. Efforts to combat them face limitations from jurisdictional complexities, concealed identities, and the sheer volume of illicit activity across borders and platforms. Together, these aspects underscore the need for innovative tools and strategies to detect, track, and disrupt these networks that profit from the online trafficking of goods, services, and people.

### 1.3.2 MOTIVATION & RESEARCH FOCUS

Given the sheer volume of online ads facilitating illicit trafficking activities, it is increasingly infeasible for LEAs to allocate their resources effectively. These ads, spread across surface web platforms, social media, and Darknet markets, create a vast ecosystem of illegal activity surpassing traditional investigative methods' capacity. Traffickers exploit this overload by employing sophisticated evasion tactics that complicate tracing their operations. For instance, on Darknet markets, vendors frequently migrate between marketplaces, often prompted by law enforcement seizures or as a deliberate strategy to avoid detection. By fragmenting their busi-nesses into smaller operations across multiple marketplaces (Hayes et al., 2018; Bradley, 2019; Broadhurst et al., 2021a; Booij et al., 2021; Chan et al., 2024), traffickers maintain their anonymity, making it challenging to identify consistent patterns of criminal behavior. Additionally, these vendors frequently change usernames or create multiple user handles to post ads under different identities (Booij et al., 2021; Bradley, 2019; Rao et al., 2024), further obscuring their tracks. Consequently, law enforcement often prioritizes larger-scale operations, leaving smaller yet equally harmful activities unaddressed.

| Indicator | Description | Examples | Sources |
|---|---|---|---|
| Trustworthy Language; Language Cues | Language to indicate trustworthiness, client screening language, | 100% professional, photos 100% me, Available 24/7, No law Enforcement, Text only, No blocked calls, Upscale gentlemen preferred; | (Whitney et al., 2018), (Lugo-Graulich and Meyer, 2021), (Lugo-Graulich et al., 2024), |
| Obscured Phone Number | Noisy phone numbers with obfuscation techniques | (4 ! 2) 456 9412, tree1ate nein 48-one7 twenty, 4!7 70! fifty6svn | (Whitney et al., 2018), (Lugo-Graulich and Meyer, 2021), (Lugo-Graulich et al., 2024), |
| Provider Ethnicity | Information of Ethnicity for provider | AA, African American, Brown Sugar, Black (Beauty) Pocahontas, Asian, Pacific Islander Caucasian, White, European Latina, Hispanic | (Ibanez and Suthers, 2014), (Ibanez and Gazan, 2016c), (Whitney et al., 2018), (Lugo-Graulich and Meyer, 2021), (Lugo-Graulich et al., 2024), |
| Discrepencies | Different ages and aliases, inconsistencies in story, | – | (Ibanez and Suthers, 2014) |
| Language Suggesting Youth | Ads for minor victims, emojis | Ages between 18 and 23, emojis: growing heart, cherry, cherry blossoms. | (Whitney et al., 2018), (Lugo-Graulich and Meyer, 2021) |
| Emojis | Emojis | rose, rosette, money bag, money with wings, heavy dollar sign, dollar banknote, hundred points | (Whitney et al., 2018), (Lugo-Graulich and Meyer, 2021), |
| Controlled Movement Language | Restrictions on movement, Specification of venue | new in town, back in town for weekend, out of state area code, ad posting in different locations, mention of hotel room, massage parlours, brothels, and spas | (Ibanez and Suthers, 2014), (Ibanez and Gazan, 2016c), (Whitney et al., 2018), (Lugo-Graulich and Meyer, 2021), |
| Payment Language | Pricing indicators | price indication with words hh/hr after phrases like donations, roses, specials | (Whitney et al., 2018), (Lugo-Graulich and Meyer, 2021), |
| Multiple Providers | Advertising multiple providers | multiple names, words like bringing a friend, shared phone | (Ibanez and Suthers, 2014), (Lugo-Graulich and Meyer, 2021), |
| Ad posted by third party | Third person language used in ad | - | (Ibanez and Suthers, 2014), (Ibanez and Gazan, 2016c) |

Table 1.1: Indicators of Human Trafficking in Online Ads from Escort Marketplaces.

Similarly, in human trafficking—particularly sex trafficking facilitated through escort service websites—researchers and law enforcement have identified a series of indicators (Table 1.1), such as language cues, pricing anomalies, and discrepancies in ads, that help distinguish ordinary escort ads from those linked to human trafficking. However, these identifiers are often buried within thousands of online postings, making them difficult to analyze at scale. Effectively leveraging these markers requires analyzing them in a network of related ads (Ibanez and Gazan, 2016a). However, the anonymity and global reach of online escort markets make it particularly challenging to link such ads to an individual or organization behind the operations.

To address these challenges of high ad volumes, decentralized operations, pseudonymous identities, and dispersed platforms, this research focuses on Authorship Attribution (AA) in cyber-enabled crimes, particularly within anonymized online marketplaces used for trafficking operations involving drugs, weapons, humans, and other illicit goods. [1] Digital content, such as textual ads, messages, or even images, often contains stylistic, linguistic, or photometric patterns that can be analyzed to trace these activities back to their originators (Ranaldi and Zanzotto, 2020; Zhang et al., 2019). AA methodologies can help exploit these unique stylometric, linguistic, and photometric signatures to link, identify, and attribute illegal ads to individual vendors, making AA a critical investigative tool for connecting ads within and across marketplaces. For example, textual descriptions in ads may reveal subtle writing fingerprints, such as specific word choices, syntax preferences, or linguistic abnormalities, that persist even when vendors operate under different usernames. Similarly, images associated with these ads can uncover photographic styles, locations, or pose patterns. Together, these stylometric patterns can link ads across platforms and aliases to a single entity or organization. By constructing a knowledge graph—where nodes represent individual ads and edges represent stylometric or linguistic similarities—LEAs can leverage information between these connections to target specific perpetrators, reallocate their resources, and strategically disrupt trafficking networks.

## 1.4 Online Marketplaces and Trafficking Activities

Cyber-enabled trafficking operations leverage a variety of online platforms to expand their criminal activities, including social media websites, classified advertisement portals like Craigslist and eBay, and specialized platforms such as escort markets and darknet marketplaces. While the methodologies proposed in this research are applicable across any of these platforms, this section focuses on the scope, characteristics, and definitions of darknet and escort markets. These platforms play a significant role in facilitating trafficking activities and are central to understanding the landscape of cyber-enabled trafficking discussed in this research.

[1] While this research specifically applies its methodologies and frameworks to Darknet and escort markets, the nature of these anonymized online marketplaces allows the findings to be extended to other forms of trafficking operations.

### 1.4.1 Darknet Markets and Illicit Trades

#### Dark Web, Tor Browser, and Darknet Markets

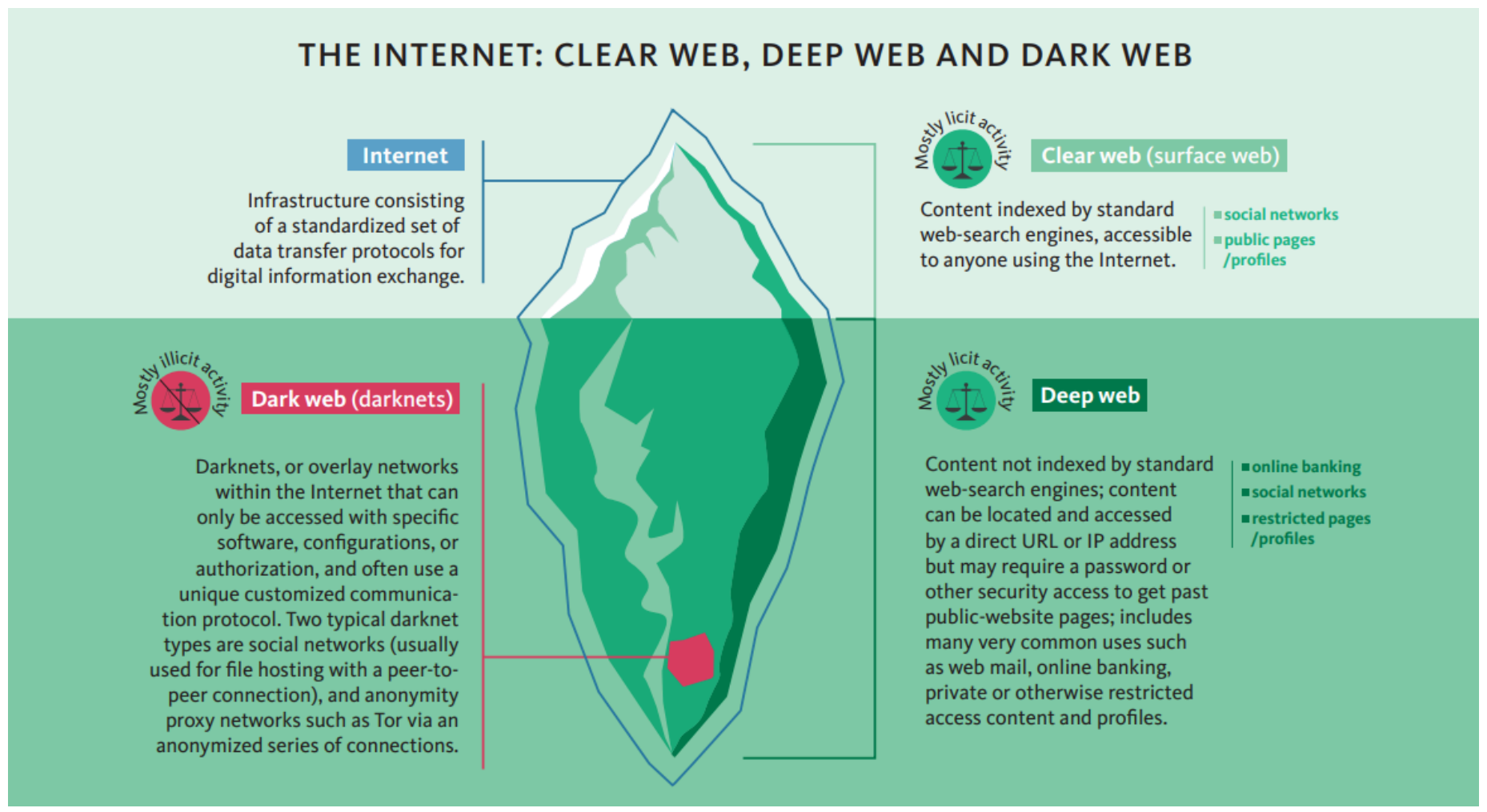

Figure 1.1: Structure of the Internet (UNODC, 2023)

| Activties | Description |
|---|---|
| Illicit Marketplaces | Trading of illicit drugs, malware and exploits, credit cards, identity and stolen information, child abuse media, weapons, etc. |
| Communication platform | Forums of discussions and chats for real-time communication |
| Cybercrime | Malware-as-a-Service for criminals, Command-and-Control hidden servers, and platforms for terrorist operations |
| Source for Threat Intelligence | Scanning of Forums and Marketplaces |
| Anonymous Financial Transactions | Bitcoin transactions for anonymity and money laundering of cryptocurrencies |
| Proxy to the Surface Web | Avoiding censorship and anonymous browsing to evade persecution by local authorities |

Table 1.2: Illicit activities supported on Dark Web pages (Gupta et al., 2021).

As per the iceberg model illustrated in Figure 1.1, the internet is struc-

1

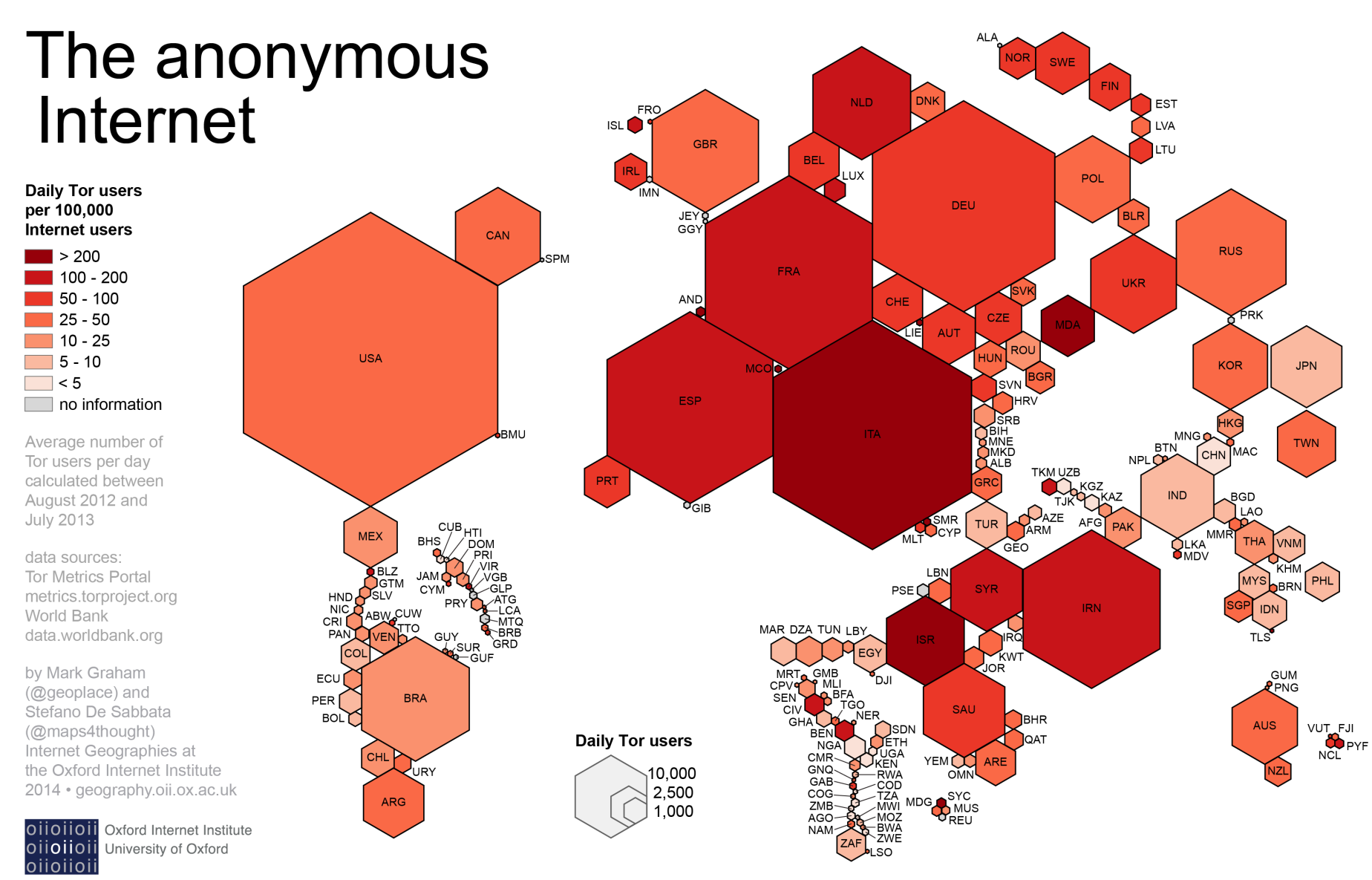

Figure 1.2: Average number of Tor users per country from August 2012 to August 2013 (Loesing et al., 2010).

tured into three distinct layers: the surface web, the deep web, and the dark web, each serving different purposes and accessibility levels (Kaur and Randhawa, 2020). The surface web, which constitutes only about 4% of the internet, includes publicly accessible websites indexed by search engines such as Google and Bing. Beneath it lies the deep web, accounting for approximately 96% of the internet, which contains unindexed content like private databases, academic records, and password-protected information requiring specific credentials or direct links to access. Within the deep web exists the dark web, an intentionally hidden segment accessible only through specialized tools like Tor browsers. This tor network, widely used to access the dark web markets (Owen and Savage, 2016), has millions of global users, exploiting its anonymizing capabilities for illicit activities. As shown in Figure 1.2, Europe stands out as one of the most active regions, with significant contributions from countries like Germany, the Netherlands, and France. Between 2020 and 2025, Russia accounted for the largest share of mean daily users (32.42%), followed by Iran (19.32%) and the United States (12.04%). However, within Europe, the Netherlands ranked among the top users of the Tor network (after Germany and France), with an average of 1,612 daily users, representing 1.66% of the global user base (Loesing et al., 2010). This is further corroborated by Figure 1.3-1.4, which highlight Europe—and particularly the Netherlands—as a hub for darknet vendors involved in the trade of drugs and weapons, emphasizing its role in enabling and combating darknet-facilitated operations.

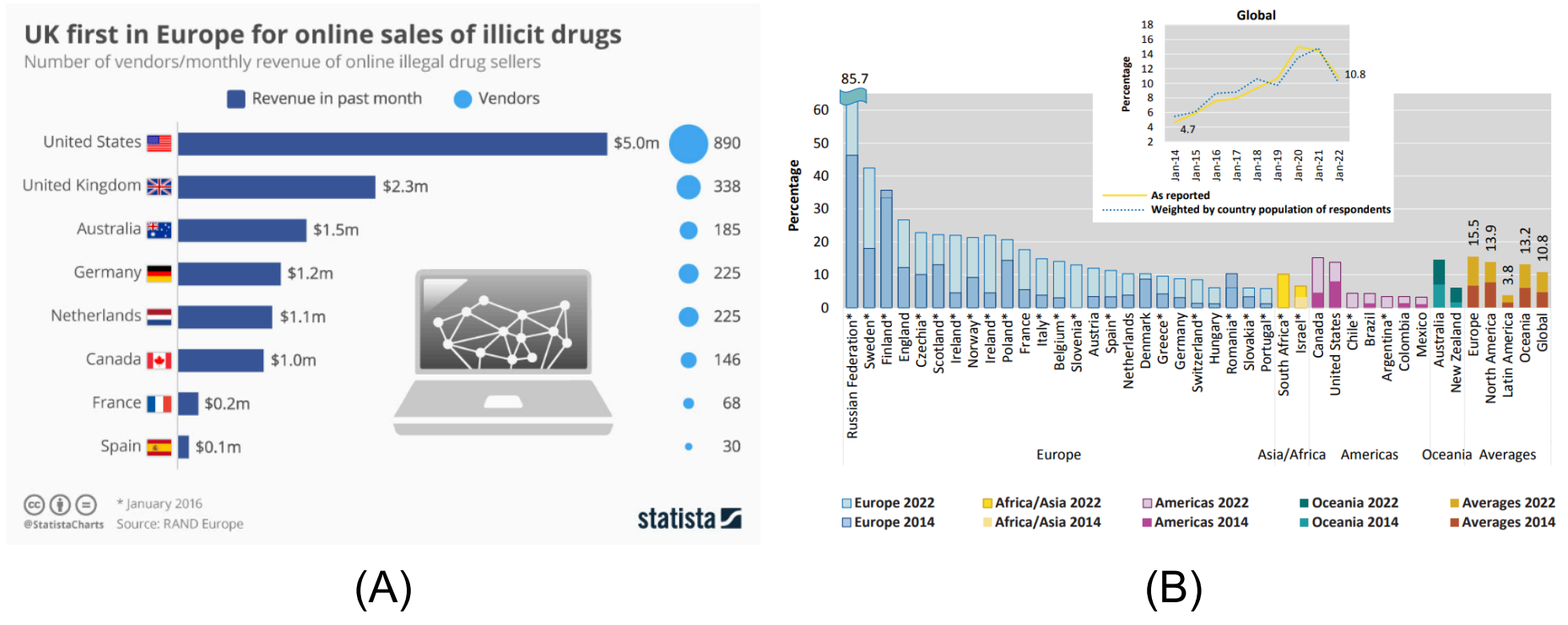

Figure 1.3: (A) Countries with the most number of Darknet Vendors and Monthly Revenue of online illegal drug sellers as per 2016 (McCarthy, 2016), (B) Global average and the proportion of people purchasing drugs over the dark web among surveyed Internet users who used drugs between January 2014 to January 2022 (UNODC, 2022).

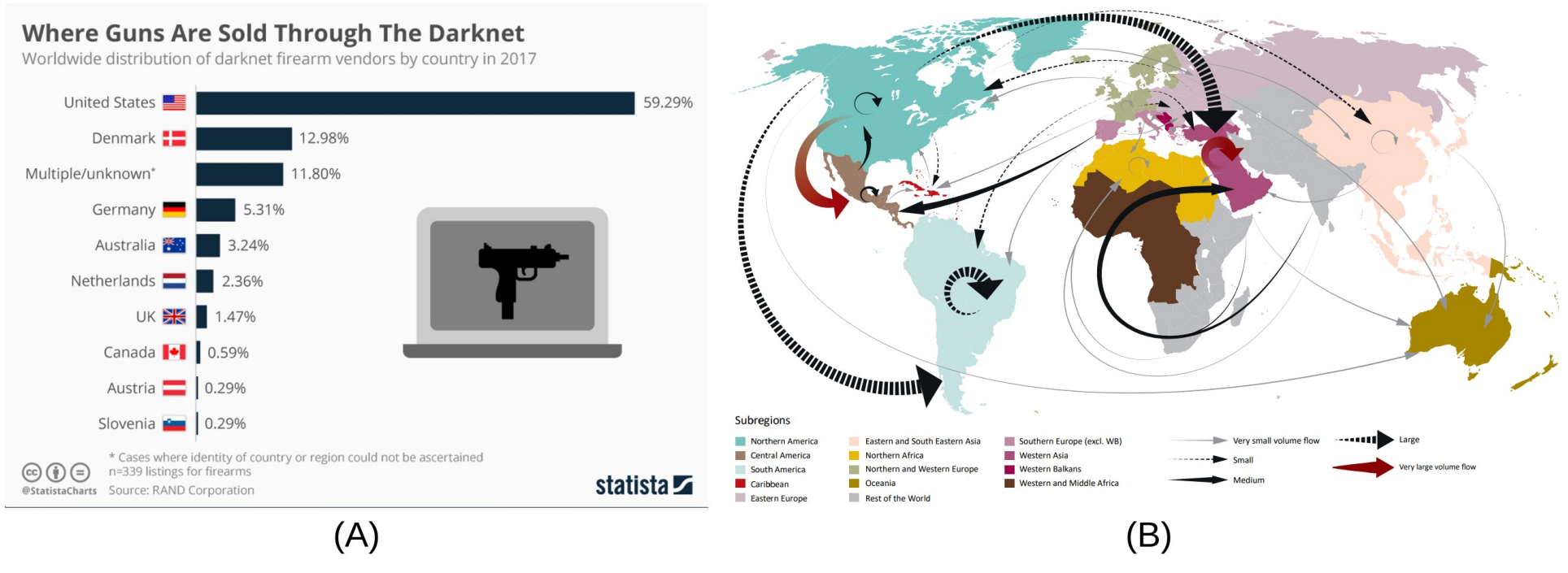

Figure 1.4: (A) Countries with the most distribution of darknet firearms vendors by country in 2017 (McCarthy and Richter, 2018), (B) Transnational firearms trafficking flows (as defined by routes of seized firearms) in 2016-17 (UNODC, 2020).

While the Dark Web facilitates several criminal activities (Table 1.2) including legit and non-illicit operations, this research only focuses on the application of AA approaches on Dark Web (Darknet) marketplaces. Darknet markets are anonymous online platforms operating on the dark web, where anonymous vendors trade goods and services while concealing their identities for legitimate activities, such as protecting privacy in oppressive regimes and illicit operations, including illegal marketplaces, cybercrime forums, information leakage, pornography, frauds, scams, hitman services, and trafficking networks (Kaur and Randhawa, 2020). According to Armstrong (2017) (Figure 1.5), drugs dominate darknet market transactions, accounting for approximately 60% of trade volume. This is followed by fraud-related services, including stolen credit card information and fake IDs, which constitute 16% of activity. Other categories, such as digital services, counterfeit goods, and firearms, comprise smaller trade portions. Finally, Figure 1.6 showcases the most active darknet markets between 2011-22, including AlphaBay, Dream Market, Silk Road, and Agora (Kermitsis et al., 2021). Leveraging these insights and publicly available datasets hosted by Impact Cyber Trust for AlphaBay, Dream Market, Silk Road, Agora, Trade Route, Valhalla, and Berlusconi, this research establishes its foundation on AA methodologies tailored to these seven darknet marketplaces.

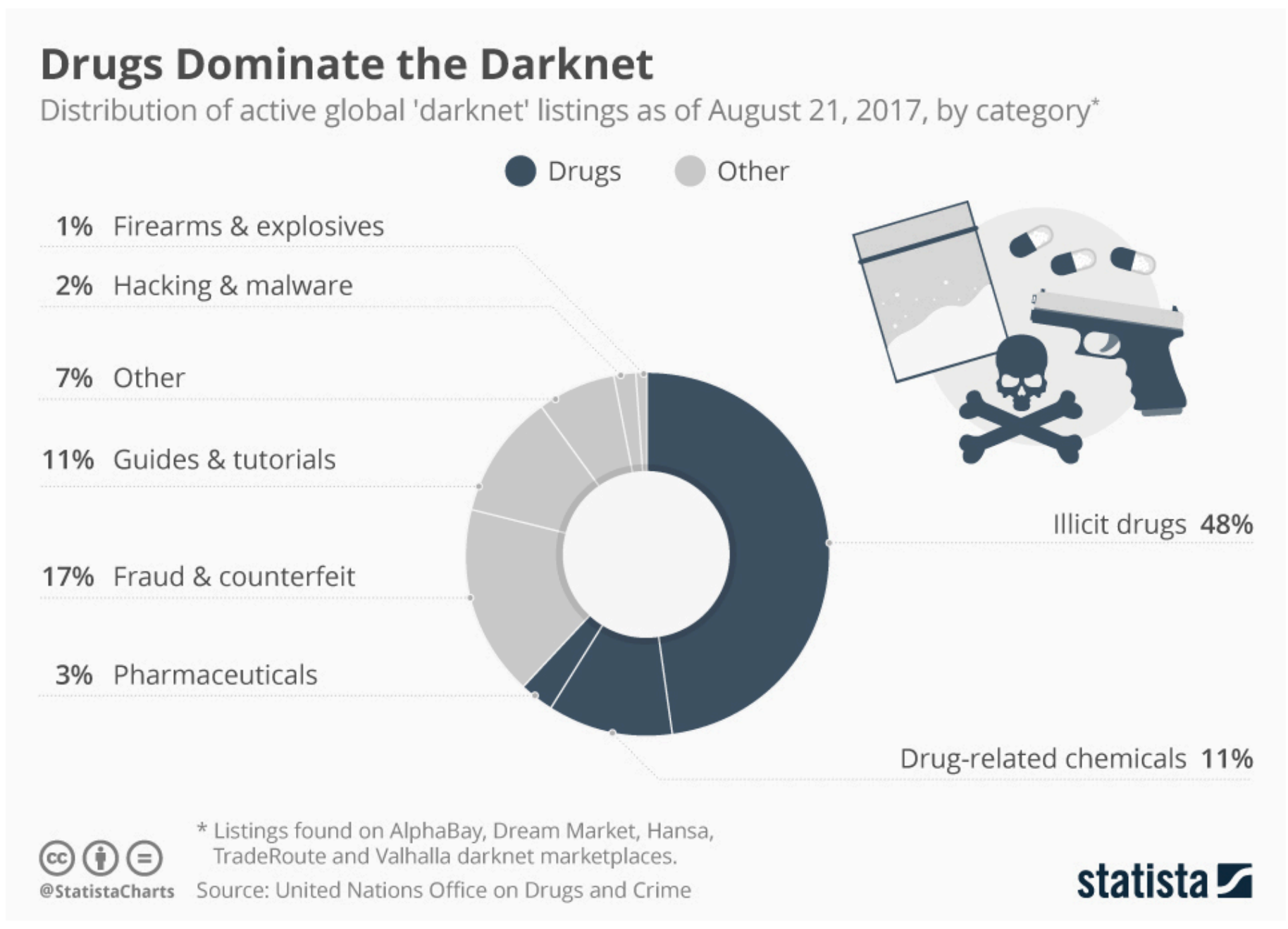

Figure 1.5: Distribution of active global Darknet listing in 2017 (Armstrong, 2017).

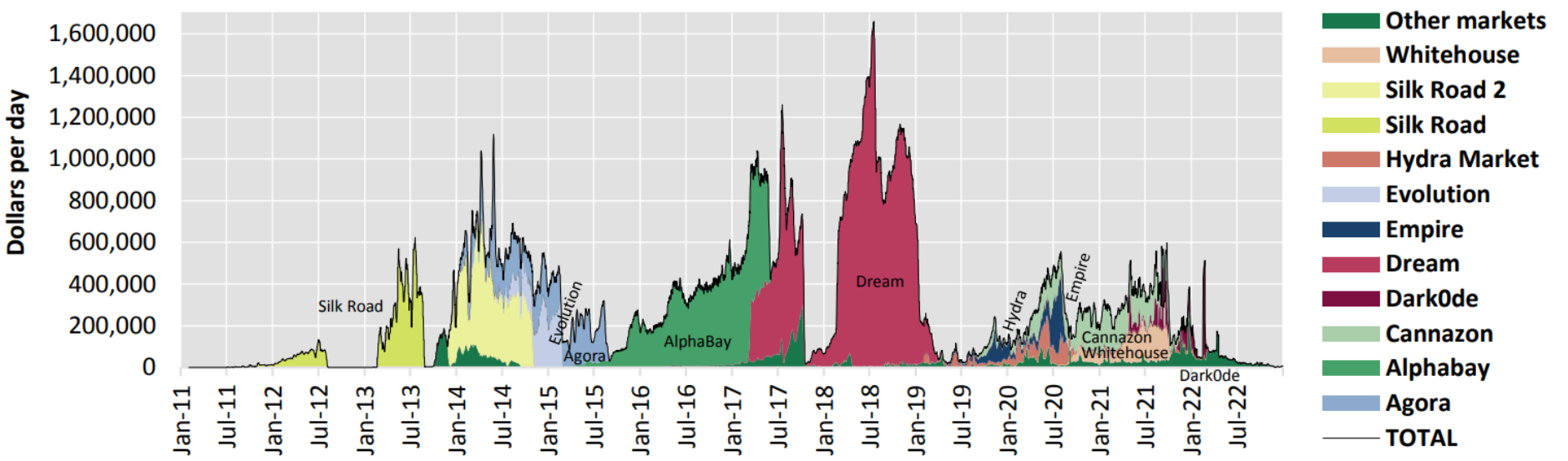

Figure 1.6: Major global darknet markets between 2011–2022 based on their daily minimum sales (more than 90% of these recorded sales are drug-related) (UNODC, 2023).

### 1.4.2 ESCORT MARKETS AND HUMAN TRAFFICKING

Human trafficking is a severe violation of human rights involving the exploitation of individuals through coercion, deception, or abuse of vulnerability for purposes such as forced labor, sexual exploitation, or other illicit activities. Within Eurpean Union (EU) (Eurostat, 2022), 14,311 victims of human trafficking (with an average of 8 victims per million inhabitants) were registered between 2019 and 2020. However, the true number is likely higher due to undetected cases. Sexual exploitation was the most prevalent form of trafficking in the EU, accounting for 51% of registered patients, followed by labor exploitation at 28%. Other forms included forced criminal activities (3%), forced begging (3%), and unspecified or unknown forms (15%). Women and girls comprised the majority of victims, especially in cases of sexual exploitation, with 87% of such victims being female (73% women and 27% girls — Figure 1.7(C)). Children accounted for 22% of victims of sexual exploitation. The countries with the highest number of female victims of sexual exploitation are Romania (901), the Netherlands (878), Germany (786), Austria (435), Italy (429) and Spain (429). The same for children accounted for Romania (477), Germany (140), and the Netherlands (104) Figure 1.7(A).

Geographically, the highest numbers of registered victims were in France (2,709), the Netherlands (2,318), and Italy (2,114). At the same time, the Netherlands, Austria, and Cyprus reported the highest rates of victims relative to their population sizes. Many victims were EU citizens, with Romania, Poland, and Bulgaria being the most common countries of origin. Non-EU victims predominantly originated from Nigeria, China, and Morocco. Furthermore, human trafficking in Europe is closely tied to established migration and refugee corridors, where traffickers exploit vulnerable populations seeking better opportunities. Victims are often deceived with false job promises, coerced through debts, or physically controlled after entering host countries legally. The liberalization of border controls has intensified trafficking flows, creating opportunities for traffickers to operate with reduced logistical risks, and avoid detection and prosecution (Hernandez and Rudolph, 2015). Traffickers are male (78%) and often EU citizens themselves, with Romania, Bulgaria, and Hungary frequently identified as countries of origin for perpetrators targetting vulnerable individuals in labor sectors, leveraging language barriers, limited legal status, and fear of deportation to maintain control over their victims. This exploitation often results in low-wage (cheap) labor in fields such as domestic work, agriculture, and prostitution (Hernandez and Rudolph, 2015). Finally, Figure 1.7(B) reveals that while the number of suspected traffickers and registered victims increased over the years, the number of convicted traffickers remained relatively constant. This discrepancy highlights the urgent need for technology-oriented solutions to assist LEAs in expediting investigations and prosecutions.

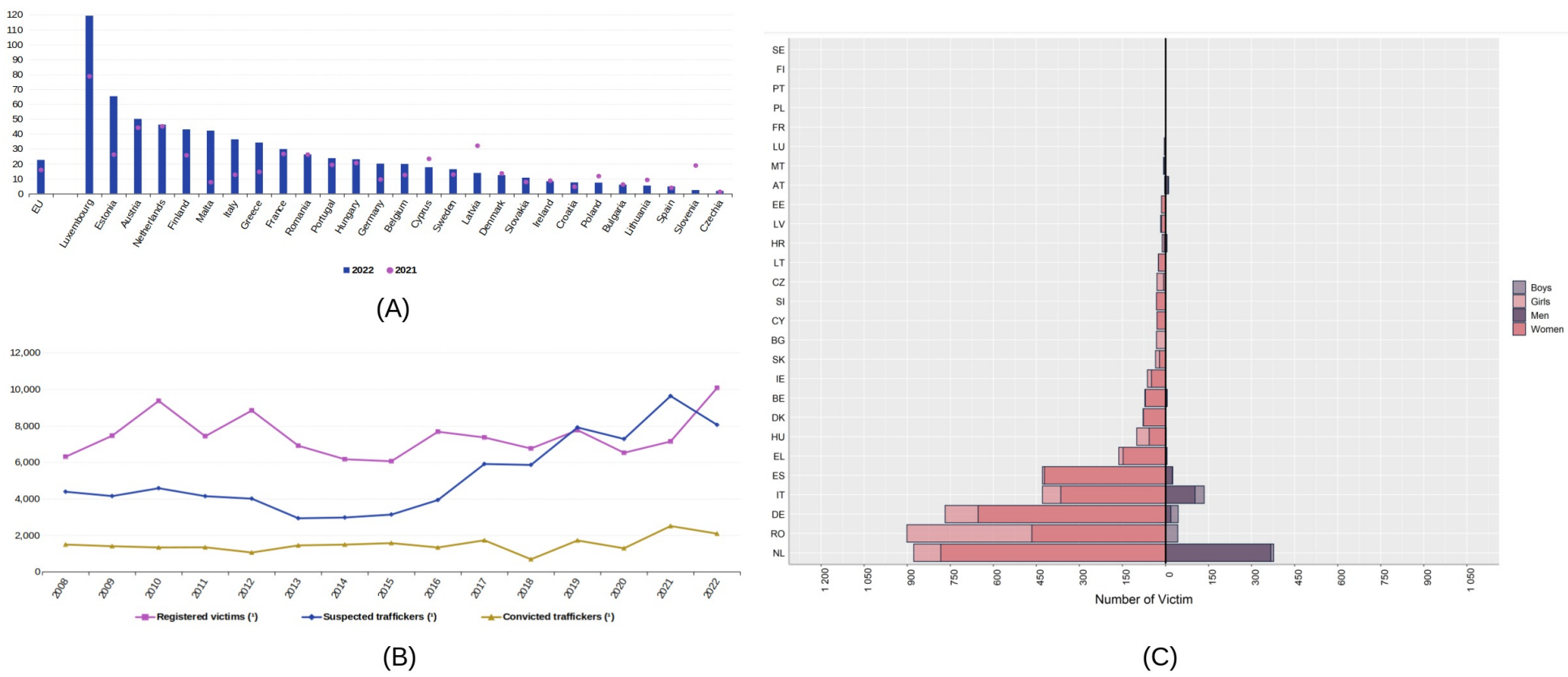

Figure 1.7: (A) Total number of human trafficking victims per EU countries in 2019-20, (B) Registered individuals (victims and traffickers) involved in human trafficking activities between 2008-2022, and (C) Gender and Age of registered victims and traffickers involved in sex trafficking-related activities per EU countries in 2019-20 (Eurostat, 2022).

In the internet age, traffickers increasingly adopt online platforms to expand their reach and evade detection (Hingston and Hingston, 2023). Recruitment often begins on social media or dating apps, where traffickers manipulate and deceive victims into exploitative circumstances (L'Hoiry et al., 2024). Victims are then advertised on classified escort web-sites, with traffickers disguising forced sexual labor as consensual work to broaden their reach while minimizing detection risks (L'Hoiry et al., 2024; Grubb, 2020). Control is exerted through methods like compromising materials and constant digital surveillance, enabling traffickers to operate remotely without physical proximity to their victims (L'Hoiry et al., 2024). Furthermore, encrypted communications, pseudonyms, and VPNs allow traffickers to conceal their identities and adapt swiftly to avoid law enforcement (Grubb, 2020). Given these tactics, identifying and disrupting the digital trails left by traffickers presents a unique challenge. This research addresses these challenges by applying AA methodologies to online marketplaces, identifying and linking traffickers through their stylometric, linguistic, and behavioral patterns.

While existing research extensively highlights the connection between human trafficking and darknet markets (Kaur and Randhawa, 2020; Nazah et al., 2020; Rawat et al., 2022), an analysis of the chosen darknet datasets revealed no such instances. A manual inspection indicated that the majority of sexual content in the chosen darknet datasets pertained to items like toys, books, magazines, and other media forms. To better investigate the application of AA methodologies to potential human trafficking activities—particularly sex trafficking—this research focuses on collecting escort ads from the United States of America (U.S.) Backpage marketplace, a classified platform well-known for facilitating various human trafficking activities (Portnoff et al., 2017; Lugo-Graulich and Meyer, 2021). [2] Although the global prevalence of sex trafficking through online escort marketplaces remains unclear (Giommoni, 2024), numerous studies suggest its existence and propose various solutions to help LEAs address the issue (Ibanez and Suthers, 2016; Ibanez and Gazan, 2016b; Hundman et al., 2018; Giommoni and Ikwu, 2021; Lugo-Graulich and Meyer, 2021; Li et al., 2022b; Vajiac et al., 2023a). This research does not directly assist LEAs in identifying human trafficking cases in online escort ads. Instead, it focuses on linking these ads through stylometric cues and vendor patterns to provide a foundational tool for LEAs to begin examining potential indicators of human trafficking in online escort marketplaces.

[2] Although this research is conducted in the Netherlands, it does not involve data collection within the country. Instead, it relies solely on publicly available datasets or resources shared during the study, without any attempts to gather new data from online platforms. Since the characteristics of escort ads and vendor behavior in online listings are consistent across regions, the methods developed in this study are expected to apply to online escort ads in the Netherlands. Future research will aim to extend the study to Dutch escort markets.

## 1.5 RELEVANT MACHINE LEARNING CONCEPTS

### 1.5.1 AUTHORSHIP ATTRIBUTION

AA is a task in Machine Learning (ML) that seeks to identify the creator of an artifact by associating its distinctive features with the creator's unique style or behavior (Juola et al., 2008). Traditionally, AA has been applied in Natural Language Processing (NLP), where it focuses on textual data by analyzing linguistic and stylistic patterns such as word choice, syntactic structure, and punctuation (for example, recognizing an author by their use of formal versus informal language, or how they structure their sentences) (Stamatatos, 2009a). However, AA extends beyond text to other modalities, including images (e.g., identifying the creator of a photograph based on consistent visual editing styles) (Thomas and Kovashka, 2016), audio (e.g., attributing music to a composer based on vocal patterns) (Kabir et al., 2021), video (e.g., detecting a particular director's editing or narrative style in a movie) (Svanera et al., 2019), and code (e.g., linking a software script to a programmer by examining their coding habits like indentation or variable naming) (Kalgutkar et al., 2019). These diverse applications make AA a versatile tool in areas such as forensics (e.g., linking a threatening letter or manipulated image to a suspect, or authenticating an unsigned painting) (Fobbe, 2021), cybersecurity and cybercrime (e.g., connecting anonymous online blackmailers or identifying vendors behind illegal product listings in darknet markets) (Nirkhi and Dr.R.V.Dharaskar, 2013; Manolache et al., 2022), and anomaly detection (e.g., flagging content that does not match known author profiles as potentially suspicious or from a new actor) (Neme et al., 2011a).

AA can be performed using two primary approaches: authorship identification and authorship verification. Authorship identification (or ***Vendor Identification*** in our research) is a multi-class classification task that assigns an artifact to one of several known creators by leveraging discriminative features (e.g., identifying which known vendor wrote a particular advertisement) (Abbasi et al., 2022). In contrast, authorship verification (or ***Vendor Verification*** in our research) evaluates whether two artifacts belong to the same author. This is typically achieved through similarity-based retrieval (described in Chapter 2) or retrieval tasks (described in Chapter 3) using model embeddings (e.g., checking whether two different ads are posted by the same vendor based on shared writing patterns) (Lei et al., 2022a). Together, these approaches highlight the flexibility of AA across modalities and domains.

### 1.5.2 MODEL REPRESENTATIONS

Representational learning in AA applications enables extracting meaningful and distinctive features from raw data to characterize an artifact's creator (Ai et al., 2022). *Model representations*, also known as model embeddings, are vector-based numerical values that transform raw input data, such as text or images, into a format that Deep Learning (DL) models can process and understand. For instance, a sentence like "A book is proof that humans are capable of working magic – Carl Sagan." would be transformed into a vector of numbers, capturing its linguistic tone, structure, grammar, semantics, and style. Similarly, an image gets converted into a vector, highlighting color schemes, resolution, pixel values, etc. These embeddings distill the most important patterns and features from the input, serving as a compact yet comprehensive version of the original data. By capturing these distinctive characteristics—such as linguistic styles and syntactic structures in text (e.g., how a vendor consistently uses emojis or misspellings for obfuscation) or visual cues like color patterns, shapes, and logo placements in images—model representations allow DL models to focus on task-relevant aspects. In AA, these representations are foundational for identifying and distinguishing authors through their unique patterns.

While traditional AA approaches relied heavily on handcrafted features, such as n-grams (e.g., repeated words), lexical statistics (e.g., word frequency counts), and stylometric markers (e.g., average sentence length or punctuation usage) (Sapkota et al., 2015a; Sari et al., 2018; Sundararajan and Woodard, 2018), these methods often struggled to generalize across diverse datasets and modalities. For example, a handcrafted feature that works well on English blog posts may fail on informal darknet product descriptions or online escort ads. However, the advent of DL has shifted representational learning toward data-driven approaches, where neural networks automatically generate hierarchical embeddings that encapsulate complex patterns across multiple levels. For textual data, models like Bidirectional Encoder Representations from Transformers (BERT) (Devlin et al., 2019a) and Generative Pre-trained Transformer (GPT) (Floridi and Chiriatti, 2020) are widely used to produce contextual embeddings. These models can understand nuances such as sarcasm, hidden codes, or double meanings, which are often used in illicit or anonymized contexts. In image-based AA tasks, Convolutional Neural Networks (CNNs) (O'Shea, 2015) and Vision Transformers (ViTs) (Dosovitskiy et al., 2021) generate embeddings that encode photometric and structural features of visual

data—for example, identifying a vendor based on recurring watermark styles, use of certain image filters, or layout of promotional flyers.

Finally, these model representations enable AA models to compare artifacts for identification and verification tasks. Advanced techniques like transfer learning, contrastive learning, and multitask learning further enhance the quality and robustness of these representations, ensuring adaptability across tasks and domains. Therefore, representational learning is pivotal in equipping AA systems with the ability to generalize effectively and perform reliably in diverse and challenging real-world scenarios.

### 1.5.3 KNOWLEDGE TRANSFER

Knowledge transfer in AA enables DL models to generalize across tasks, datasets, and domains by leveraging prior knowledge. *Transfer learning* involves pre-training a model on a large, general-purpose dataset and fine-tuning it to transfer learned knowledge to domain-specific tasks (Ruder et al., 2019). This approach is particularly effective in low-resource settings where labeled data is limited. Pretrained language models like BERT, Robustly Optimized BERT Pre-training Approach (RoBERTA) (Liu et al., 2019a), and GPT capture general linguistic structures and can be adapted to detect stylistic nuances in textual AA tasks. Similarly, pretrained vision models such as Residual Neural Network (ResNet) (He et al., 2015), Visual Geometry Group (VGG) (Simonyan and Zisserman, 2015), and ViTs can be fine-tuned to analyze images from Darknet or Escort markets.

*Multitask learning*, on the other hand, trains a model on multiple related tasks simultaneously, facilitating the transfer of knowledge across tasks while learning shared representations (Ruder, 2017). In the context of AA, this means a single model might be trained to verify whether two ads are written by the same vendor (verification), assign an ad to a specific known vendor (identification), and detect broader stylistic markers (stylistic analysis) all at once. Together, transfer learning and multitask learning address challenges like data scarcity (e.g., limited labeled data in new escort platforms) and task diversity (e.g., combining textual and visual authorship cues), enabling AA systems to perform robustly in real-world applications such as cyber-enabled trafficking investigations.

### 1.5.4 CONTRASTIVE LEARNING

*Contrastive learning* is a powerful DL technique that enables models to learn meaningful representations by distinguishing between similar and dissimilar data pairs (Chen et al., 2020b). This learning approach maximizes the similarity between model representations (embeddings) of semantically similar instances (e.g., texts or images from the same author) while minimizing the similarity between embeddings of different instances (Petropoulos, 2023; Chen et al., 2023). In the context of AA, contrastive learning is particularly effective in capturing subtle stylistic and semantic features unique to an author across various datasets and modalities (Ai et al., 2022). For example, in escort ads, two ads from the same vendor might use different surface-level content (different descriptions or locations of the escort service) but still share consistent stylistic elements—such as emoji use, style, or visual composition in posted images. Contrastive learning helps the model learn to associate these nuanced similarities while distinguishing them from content produced by other vendors with superficially similar topics but different stylistic traits.

By training on pairs of samples—such as text descriptions, images, or multimodal combinations of text and images—contrastive learning encourages the model to focus on subtle patterns that define an author's style, such as word choice, sentence structure, thematic consistency, photometric patterns, locations, and poses (Valero-Redondo et al., 2024). For instance, a model may learn to associate a particular vendor with consistent backdrops in images (e.g., specific rooms, decor, or watermark styles) or recurring phrase patterns in text. This method enhances the model's ability to generalize across different datasets and adapt to varying textual, visual, or multimodal data forms—making it particularly useful in dynamic environments where new vendors, platforms, or content formats frequently emerge.

### 1.5.5 MULTIMODAL MACHINE LEARNING

Multimodal ML integrates and analyzes information from multiple data modalities, such as text, images, audio, and video (Baltrušaitis et al., 2018). By combining complementary features from different modalities, multimodal systems can uncover richer patterns and relationships often inaccessible when using a single (unimodal) data source. In AA, multimodal ML is particularly relevant when artifacts such as text and images can jointly provide a better understanding of an author's style. For instance, vendors often post ads that include descriptive text and photographs in online escort marketplaces or social media platforms. While the text may reveal linguistic traits—like tone, phrasing, grammar, or specific signature patterns—the accompanying images might contain consistent visual patterns such as watermark placement, poses, image composition, or even the types of environments (background) used in the photos. In some cases, the textual descriptions reference different individuals, introducing inconsistencies that make linking ads based on text alone challenging. However, visual similarities across the accompanying images—such as background settings, photographic style, or repeated use of props—can provide valuable cues for connecting these ads. Conversely, when images depict different individuals with no apparent visual overlap, consistent writing patterns in the text—such as phrasing, tone, or formatting—can help the model identify a common authorship. Taken alone, each modality might seem inconclusive in these cases, but together, they form a more complete and distinctive authorial signature (Zhang et al., 2019).

Integrating text with images in AA requires effective cross-modal alignment, where features from both modalities are encoded into a shared representation space, allowing the model to leverage interactions between visual and textual elements (Gao et al., 2020). For example, a model might learn that a vendor who uses specific slang or abbreviations in the text also tends to use particular lighting styles or props in their images—creating a joint pattern that can be used for attribution. ViTs for images and pre-trained language models (e.g., BERT) for text are often combined using contrastive or joint learning frameworks to align and analyze these modalities (Radford et al., 2021a; Li et al., 2023a). Such architectures can learn multimodal embeddings that reflect semantic and stylistic coherence across modalities.

Multimodal Authorship Attribution (MAA) approaches extend the scope of traditional techniques by capturing nuanced relationships between text and images. This boosts model performance and makes such systems more applicable to real-world investigative contexts, such as uncovering coordinated vendor operations in escort ads or linking profiles across platforms in cybercrime investigations.

### 1.5.6 RESPONSIBLE AUTHORSHIP ATTRIBUTION

Inspired by the ethical Artificial Intelligence (AI) pillars outlined in (Jobin et al., 2019; Fjeld et al., 2020), this research consolidates these foundational ideas into four core principles: *privacy and data protection*, *fairness and non-discrimination*, *transparency and explainability*, and *societal impact*. These principles serve as the ethical framework guiding this work to address the unique challenges AA systems pose in sensitive and high-stakes domains. Privacy and data protection focus on responsible data handling practices to safeguard sensitive information, ensuring compliance with regulatory standards such as General Data Protection Regulation (GDPR) (Voigt and Bussche, 2017) and AI act (Edwards, 2021). Fairness and non-discrimination emphasize the mitigation of biases that could disproportionately affect underrepresented groups, ensuring equitable outcomes across diverse linguistic and cultural contexts. Transparency and explainability are critical for fostering trust by making AA model decisions interpretable and comprehensible. Societal impact considers the broader implications of AA systems, emphasizing their potential to contribute positively to societal welfare while minimizing risks of misuse and harm. [3]

## 1.6 Research Questions and Contributions

The study is guided by the following research questions:

**RQ1: How can authorship attribution methodologies assist in analyzing writing styles to identify and connect online vendors involved in cyber-enabled trafficking operations?**

**Scope:** This research question explores the application of AA methods to illegal online marketplaces, with a focus on darknet and escort platforms. The goal is to develop methods for identifying and linking vendors based on the analysis of writing styles in the text descriptions of online ads.

**Challenges:** Online illegal marketplaces facilitate a wide range of criminal activities, including trafficking of drugs, weapons, humans, counterfeit goods, and other illicit items, by providing anonymity to their vendors. On darknet marketplaces, vendors often adopt multiple aliases or migrate between platforms to evade detection, making it challenging for researchers, LEAs, and practitioners to track their operations or assess the full scope of their activities. In contrast, escort markets present distinct challenges due to the lack of explicit identifiers in ads, such as usernames or contact details. This absence complicates linking ads to specific vendors, impeding efforts to identify patterns and track behaviors across platforms. These barriers hinder LEAs' ability to effectively allocate investigative resources, resulting in fragmented investigations and undetected trafficking operations. Traditional methods for linking these vendor accounts rely heavily on manual investigation, which is labor-intensive, error-prone, and ill-suited for the scale and complexity of modern online illegal marketplaces. To address these challenges, this research question presents two complementary sub-questions outlined below:

**– RQ1 (a): How can authorship identification and verification approaches analyze and compare writing styles to connect existing (known) and emerging (upcoming) vendors across darknet marketplaces?**

[3] These principles are specifically tailored for NLP-based AA applications and are yet to be extended to vision-based AA tasks, a direction that will be addressed in future work.

**Contribution:** ***Chapter 2 – VendorLink for Authorship Attribution on Darknet Markets***

- **Approach:** Utilizes AA methods to analyze linguistic and stylistic patterns across text ads (titles and advertisement descriptions) from seven existing darknet market datasets.
- **Implementation:** Applies existing language models and fine-tunes them to detect distinctive writing styles and classify ads to known vendors. This approach enables the identification of vendors operating across different marketplaces. After fine-tuning, cosine similarity (Steck et al., 2024) is computed between the learned model representations of all text ads, supporting the vendor verification task by linking ads through similarities that likely originate from multiple aliases to their associated parent vendors.
- **Outcome:** Provides a scalable approach that enhances the efficiency of AA applications in tracking vendor accounts across multiple marketplaces, uncovering aliases, and connecting disparate accounts within darknet markets.

**– RQ1 (b): How can authorship attribution approaches be applied in the absence of ground truth on online escort marketplaces? How can the authorship verification approach be systematically designed and evaluated to ensure computational efficiency?**

**Contribution:** ***Chapter 3 – IDTraffickers for Authorship Attribution on Online Escort Markets***

- **Approach:** Introduces a novel AA dataset, IDTraffickers, comprising 87,595 unique textual descriptions of ads linked to 5,244 unique vendor labels collected from Backpage escort markets from 41 U.S. cities. Vendor labels are generated by employing Chambers et al. (2019) to extract obscured phone numbers and network analysis to cluster phone numbers into vendor communities. AA methodologies are then applied to analyze linguistic and stylistic patterns in these textual descriptions, enabling the identification and verification of vendors based on their unique writing styles.

- **Implementation:** Building on VendorLink, supervised learning is used for vendor identification, classifying ads by analyzing linguistic patterns and stylistic features. Unlike VendorLink, where vendor verification relied on exhaustive pairwise comparisons of ads — a computationally expensive and time-intensive process — this study employs the trained ad representations to utilize a retrieval-based approach with Facebook AI Similarity Search (FAISS) (Douze et al., 2024). FAISS indexing enables efficient similarity computation by retrieving nearest neighbors, significantly reducing the computational burden and making the approach scalable for large datasets.

- **Outcome:** A tool to study human trafficking indicators, monitor vendor behaviors, and uncover connections between ads in escort markets. The efficient and scalable methodology supports real-time investigations, offering enhanced capabilities to track and disrupt illicit operations in high-stakes domains.

By emphasizing the significance of distinctive writing styles for connecting illicit ads, these studies establish a comprehensive AA method to link vendors and uncover their operations across multiple darknet and online escort marketplaces, reinforcing the utility of AA methods in addressing cyber-enabled trafficking operations.

**RQ2: How does integrating text with images and advanced training objectives affect authorship attribution generalization and effectiveness within online escort advertisements?**

**Scope:** This research question investigates the integration of multimodal data (text and images) with advanced training objectives, such as contrastive and multitask learning, to improve the generalization of the AA model on both in-sample and Out-Of-Data (OOD) data [4]. The primary focus is on addressing cyber-enabled trafficking by connecting online escort ads through text and images, providing researchers, LEAs, and practitioners (NGO, private investigators, forensic analysts, and legal stakeholders (prosecutor) etc.) with better tools to uncover trafficking networks.

**Challenges:** Existing AA systems rely predominantly on text, overlooking the multimodal nature of online escort ads, where images often provide complementary stylistic cues. They fail to address limited data from vendors with fewer ads and struggle to generalize effectively across unseen or OOD data. By integrating images with text, AA systems can more effectively capture nuanced stylistic patterns, such as recurring stylistic elements, specific locations, or particular poses. This integration can create stronger connections between ads and their vendors, even when the textual content is altered to avoid detection. Additionally, this approach is beneficial in cases where vendors post images of different individuals but use consistent advertisement styles and language, allowing the AA models to link ads in both scenarios.

**Contributions:** ***Chapter 4 - MATCHED for Multimodal AA (MAA) on Online Escort Markets***

- **Approach:** Building upon *IDTraffickers*, a new multimodal AA dataset, *MATCHED*, is curated that contains 27,619 unique text descriptions and 55,115 unique images connected to 3,546 vendors from Backpage escort markets across seven U.S. cities. MATCHED provides a foundation for linking ads to vendors, facilitating the construction of vendor networks that can help researchers, LEAs, and practitioners study human trafficking indicators more effectively.

- **Implementation:** Incorporates advanced training objectives, such as multitask learning and contrastive approaches, along with latent fusion (Pawłowski et al., 2023) strategies. These methods combine features from text and images, aligning their representations to capture complementary stylistic cues, thus improving generalization across unseen or OOD scenarios while simultaneously addressing

[4] In-sample data refers to the dataset used for training and validation, representing scenarios that the model has encountered during training. OOD distribution includes samples not directly seen during training and is utilized to assess the model's generalization capability (Hsu et al., 2020).

the challenge of limited vendor data and low semantic overlap between text and images.

- **Outcome:** The approach demonstrates the potential of multimodal AA systems to outperform unimodal baselines by effectively combining text and image data. It enables robust vendor identification and verification, even in OOD contexts, providing researchers, LEAs, and practitioners with actionable insights for uncovering and disrupting trafficking networks. By addressing real-world challenges, the framework lays a strong foundation for advancing AA research in sensitive domains while enhancing investigative capabilities.

**RQ3: What ethical considerations could guide the development and deployment of authorship attribution systems applied to sensitive (high-risk) domains?**

**Scope:** This research focuses on developing a responsible framework for text-based AA applications within NLP. It addresses the Software Development Life Cycle (SDLC) (Ruparelia, 2010) of AA systems in sensitive domains, ensuring alignment with societal values and ethical standards.

**Challenges:** While NLP-based AA approaches have significant potential for addressing societal and legal challenges, their application in sensitive contexts raises ethical, legal, and societal concerns. Risks include privacy violations, decision-making biases, lack of transparency, and potential misuse of AA systems. Existing AA methods prioritize technical advancements without considering these broader ethical implications, whereas existing ethical AI guidelines do not explicitly account for the unique complexities of AA, creating a gap in frameworks required to ensure responsible design and deployment of NLP-based AA in sensitive contexts.

**Contribution:**
***Chapter 5 - Responsible guidelines for NLP-based AA methods***

- **Approach:** Grounded in established ethical AI principles (Jobin et al., 2019), the proposed framework consolidates and tailors these principles to AA systems' specific challenges and lifecycle, ensuring ethical considerations are systematically addressed across all development phases.

- **Implementation:** A tailored set of guidelines to address the ethical dimensions of AA within NLP. Drawing from established ethical

AI principles—such as privacy and data protection, fairness and non-discrimination, transparency and explainability, and societal impact—the framework adapts these principles to the specific life-cycle of AA systems. It emphasizes role-based responsibilities for SDLC stakeholders, including system designers, project managers, developers, and end-users. The framework's effectiveness is demonstrated through a case study (IDTraffickers), showcasing privacy measures, fairness checks, and transparency techniques applied in a high-stakes context.

- **Outcome:** A structured approach to implement ethical considerations into NLP-based AA applications, addressing risks like privacy violations, biases, and misuse. It promotes accountability through clear role-based responsibilities, enhances trustworthiness via transparency and explainability, and demonstrates practical utility in sensitive domains. By establishing actionable guidelines, the framework supports the responsible development and deployment of AA systems while aligning with societal values and ethical standards in high-stakes applications.

## 1.7 THESIS OVERVIEW

This dissertation is structured into seven chapters in addition to this introductory chapter. The introduction provides the context and motivation for the thesis, the problem statement and the research questions the thesis seeks to address. The rest of the dissertation is organized as follows:

- **Chapter 2** addresses *RQ1(a)* by introducing *VendorLink*, a foundational methodology for text-driven AA to identify and connect vendor migrants and aliases across seven Darknet markets. This research establishes benchmarks for vendor identification, utilizing various techniques, from statistical models to traditional neural networks, as well as contextualized transformer-based models, to ensure robust performance evaluation. Furthermore, it extends the trained model to a vendor verification task that employs cosine similarity on model representations to link aliases. Finally, it features a knowledge transfer task demonstrating how computationally intensive models can be adapted for use in low-resource environments through effective transfer learning. This adaptation is particularly beneficial for emerging markets and vendors, enabling resource-efficient analysis while providing actionable insights to combat illicit activities better.

- **Chapter 3** expands on *VendorLink's* foundation to address *RQ1(b)* by introducing IDTraffickers, a large-scale authorship attribution dataset comprising 87,595 text ads collected from the Backpage escort markets across 41 U.S. cities. This dataset is designed to link and connect 5,244 unique vendors, enabling the identification of potential human trafficking indicators. By capturing the distinctive linguistic and stylistic traits of escort ads, IDTraffickers bridges a critical gap between traditional authorship datasets and the unique challenges of this domain, laying a foundation for further research. Extensive benchmarking is performed using state-of-the-art contextualized models for the closed-set vendor identification task. Additionally, the vendor verification task is formalized, introducing an optimized and lightweight similarity retrieval approach tailored for open-set scenarios, ensuring scalable and efficient linking of vendor connections.

- **Chapter 4** builds on *IDTraffickers* to address *RQ2* by introducing *MATCHED*, a multimodal dataset comprising 27,619 unique text descriptions, 55,115 unique images, and 3,546 vendors collected from Backpage Escort-ads across seven U.S. cities. MATCHED bridges the gap in leveraging multimodal data for AA by integrating textual and visual features to capture richer stylistic patterns. Extensive benchmarking evaluates text-only, vision-only, and multimodal baselines for vendor identification and verification tasks, demonstrating the advantages of multitask training objectives and contrastive learning strategies in improving performance across in-sample and OOD scenarios.

- **Chapter 5** addresses *RQ3* by building upon the ethical challenges of AA in NLP applications by introducing a comprehensive framework of responsible guidelines. Rooted in established ethical AI principles—privacy and data protection, fairness and non-discrimination, transparency and explainability, and societal impact—the framework tailors these principles to the lifecycle of AA systems. The chapter provides role-based guidelines for stakeholders across the SDLC, ensuring ethical considerations are integrated at every phase. Additionally, the framework is validated by applying the guidelines to a case study on *IDTraffickers*, illustrating its practical relevance in sensitive domains.

- **Chapter 6** critically evaluates dataset documentation, methodolog-

ical limitations, ethical implications, and societal impacts of the AA methods proposed for cyber-enabled trafficking investigations, while proposing actionable directions for future research.

- **Chapter 7** provides the conclusion to the thesis by reviewing how the research addresses the three research questions.

# 2

# VendorLink: Authorship Attribution For Darknet Marketplaces

**This chapter is based on the following research:**

- **Vageesh Saxena**, Nils Rethmeier, Gijs van Dijck, and Gerasimos Spanakis. 2023b. VendorLink: An NLP approach for Identifying & Linking Vendor Migrants & Potential Aliases on Darknet Markets. In Proceedings of the 61st Annual Meeting of the Association for Computational Linguistics (Volume 1: Long Papers), pages 8619–8639, Toronto, Canada. Association for Computational Linguistics.

## 2.1 Introduction

Conventional search engines index only surface-web websites, constituting approximately 4% of the internet. The remaining 96% falls under either the Deep Web, which is not indexed, or the Darknet, which operates using advanced anonymity-enhancing protocols (Kaur and Randhawa, 2020). While the Deep Web often serves legitimate needs requiring anonymity, the Darknet is frequently exploited for illegal purposes, including financial fraud and scams (Kaur and Randhawa, 2020), child exploitation (Liggett et al., 2020), terrorism-related activities (Weimann, 2016), and the trade of illicit weapons (Persi Paoli et al., 2017), as well as banned drugs and chemicals (Kruithof et al., 2016).

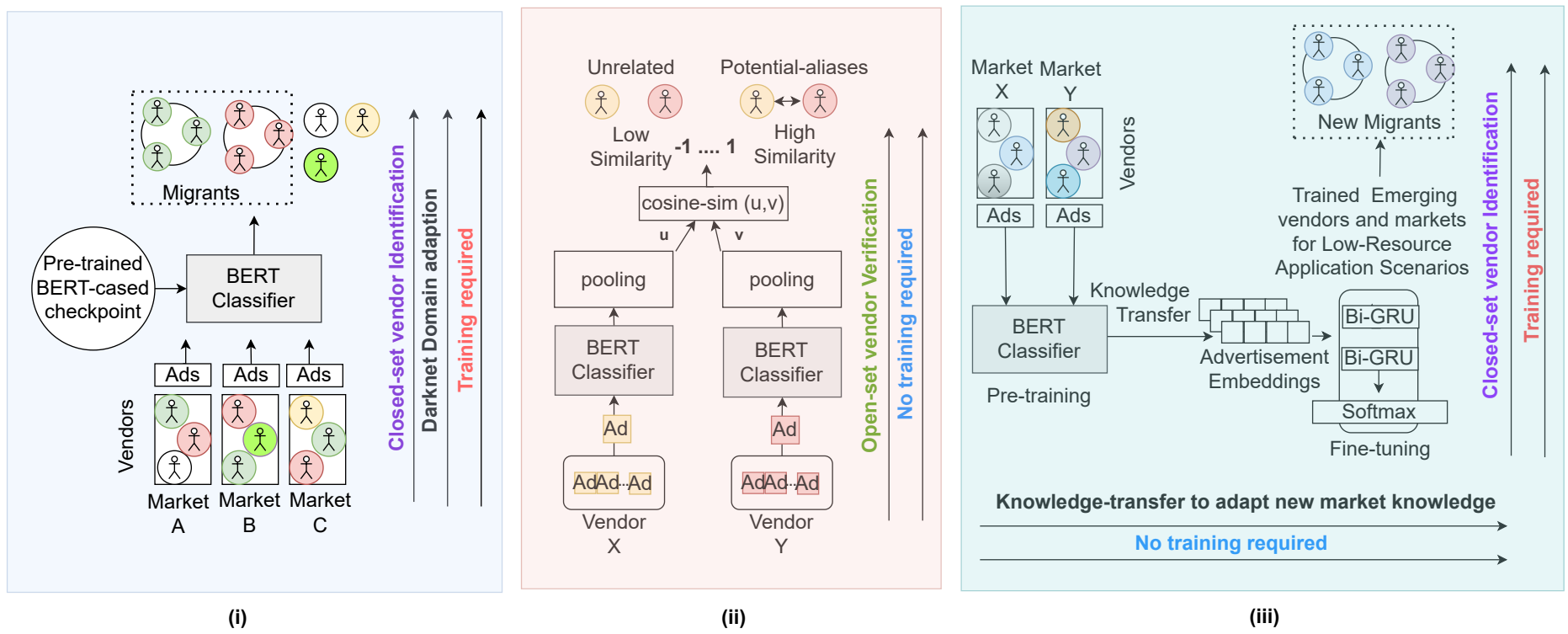

Figure 2.1: **(i) Closed-Set Vendor Identification Task:** A supervised pre-training task that performs classification using a BERT-cased classifier in a closed-set environment to verify unique vendor migrants across existing markets, **(ii) Open-Set Vendor Verification Task:** A text-similarity task in an open-set environment that utilizes style representations from the established BERT-cased classifier to verify known vendors and identify potential-aliases, **(iii) Low-Resource Market Adaptation Task:** A knowledge-transfer task in a closed-set environment to adapt new market knowledge and verify migrants across Low-Resource (LR) emerging markets.

The extensive reach, anonymity, and complexity of the Darknet make it challenging to identify and trace connections between illegal marketplaces (Warner, 2023). While manual efforts to uncover these links are both time-intensive and resource-heavy, recent advancements in online scraping tools (Fu et al., 2010; Hayes et al., 2018) and monitoring systems (Schäfer et al., 2019; Godawatte et al., 2019) have provided researchers and LEAs with the means to analyze (Easttom, 2018; Goodison et al., 2019; Davies, 2020), categorize, and classify Darknet content automatically (Al Nabki et al., 2017; Ghosh et al., 2017; He et al., 2019). Rather than pursuing a similar approach, this study introduces a novel NLP-based AA approach to address vendor identification and verification, aiming to provide insights into the scale and scope of vendor operations and enabling informed decision-making for optimized allocation of law enforcement resources. Unlike prior studies in Darknet research (Ekambaranathan, 2018; Tai et al., 2019; Kumar et al., 2020; Manolache et al., 2022), VendorLink, as outlined in Figure 2.1, focuses on addressing the challenges of identifying and verifying vendors in Darknet markets, contributing the following key contributions: [1]

- **(i) Closed-Set Vendor Identification Task:** Due to limited resources, LEAs prioritize investigating the darknet vendors based on the scale and nature of their operations. To evade detection, these vendors often distribute their activities across multiple marketplaces. Similarly, when a marketplace is shut down, some vendors relocate their operations to other platforms and continue their trade (Booij et al., 2021). For simplicity, this study refers to such vendors as *migrants*. This movement complicates LEAs' efforts to accurately assess the complete scope of a vendor's business. To overcome this challenge, this study employs closed-set multiclass classification (Zhou et al., 2021). This task, termed the *Vendor Identification* task, focuses on identifying a candidate author from a predefined set of known authors by analyzing distinct writing styles in textual advertisements to associate migrant vendors with their corresponding vendor accounts across different darknet markets. Given the significant linguistic differences between darknet and surface web content, as noted in previous research (Choshen et al., 2019; Jin et al., 2022), the supervised training equips our model with domain-specific knowledge tailored to darknet marketplaces.

- **(ii) Open-set Vendor Verification Task:** Darknet vendors often operate under *aliases* or collaborate in groups to distribute their products across multiple marketplaces. This strategy enables them to expand their operations while evading detection by LEAs. Given the vast scope and anonymity of the darknet, manually linking these vendor profiles is infeasible. Additionally, with hundreds of new markets and vendors appearing daily on the Dark Web, the scale of the problem continues to grow. Although the existing research shows promises in vendor identification tasks, these trained classifiers typically fail in real-world scenarios when encountering previously unknown vendors from emerging markets. To address this limitation, this study leverages the style representations learned from the pre-trained classifier (step (i)) representations to compute the cosine similarity between textual ads, enabling the identification of known vendors and the verification of potential aliases or unknown vendors within an open-set environment Zhou et al. (2021).

[1] Our code implementation is publicly available at github.com/maastrichtlawtech/VendorLink.

- **(iii) Low-Resource Market Adaptation Task:** While previous research has shown promising results for vendor identification tasks on the darknet markets (Kumar et al., 2020; Manolache et al., 2022), significant computational and storage demands remain a critical challenge for LEAs. The rapid growth of the darknet markets, accompanied by an influx of new vendors and content each year, highlights the urgent need for systems capable of verifying known vendors from an existing database while simultaneously adapting to novel market knowledge from emerging vendors and marketplaces. Since many LEAs lack the resources to retrain computationally expensive models from scratch each time a new Darknet market emerges, more efficient solutions are necessary. This study, therefore, explores the ability of a pre-trained classifier to leverage transfer learning in low-resource settings characterized by smaller datasets and limited computational resources (Hedderich et al., 2021). The goal is to enable a smaller classifier to adapt to new market knowledge (language and trade categories) while performing closed-set vendor identification for emerging vendors and markets.

## 2.2 Related Research

**Vendor Identification - A Supervised AA Task:** The current advancements in NLP have fueled significant progress in AA, demonstrating the efficacy of techniques such as Term Frequency-Inverse Document Frequency (TF-IDF) - based clustering and classification (Agarwal et al., 2019), CNNs (Shrestha et al., 2017), recurrent neural networks (RNNs) and their variants (Zhao et al., 2018; Jafariakinabad et al., 2019; Gupta et al., 2019), as well as transformer-based architectures (Fabien et al., 2020a; Ordoñez et al., 2020; Uchendu et al., 2020). However, researchers have noted substantial linguistic differences between Darknet and surface web content (Choshen et al., 2019; Jin et al., 2022), underscoring the importance of exploring authorship tasks within the unique context of Darknet language.

Darknet ads typically include a product title, description, vendor name, trade category, price, metadata, and images. Existing studies have employed a range of NLP (Ekambaranathan, 2018; Tai et al., 2019; Manolache et al., 2022) and computer vision (Wang et al., 2018) techniques to leverage these details to identify and connect vendors across Darknet marketplaces. For instance, Zhang et al. (2019) introduced a multimodal approach, uStyle-uID, which utilized writing and photography styles to identify vendors in drug trafficking markets. Similarly, Kumar et al. (2020) proposed leveraging multi-view learning and domain-specific insights to enhance cross-domain performance using stylometric and locational features. However, most of these multi-view details, including the metadata, location, price of the trade, and even images, are embedded within the text description of the ads. Manually extracting these features requires extensive labeling efforts. Therefore, this study adopts an end-to-end approach that relies solely on the ads's title and description to analyze text-based writing patterns for vendor identification and verification tasks.

**Vendor Verification – A Text Similarity Task:** Text-similarity techniques have long been a part of AA research (Castro Castro et al., 2015; Rexha et al., 2018). However, with the rise of transformer-based architectures (Reimers and Gurevych, 2019) and their application in AA (Boenninghoff et al., 2019a; Fabien et al., 2020a; Uchendu et al., 2020), advancements in style representation techniques (Hay et al., 2020; Zhu and Jurgens, 2021) have opened new possibilities for tackling authorship verification tasks. For instance, Wegmann et al. (2022) demonstrated that the effectiveness of these style representations lies in their ability to encode stylistic features by exploiting spurious content correlations. Their findings also emphasize the need for content control in contrastive setups to achieve style representations that are more independent of content. Building on these insights, this study leverages style representations extracted from the ads of the darknet vendors. These representations are obtained by passing the text through a transformer-based classifier trained on the closed-set vendor identification task. Subsequently, the style representations are used to compute text similarity (via cosine similarity) across ads from different vendors.

**Knowledge Adaption - A Transfer Learning Task:** Following Ruder et al. (2019), researchers have explored the effective use of transfer learning in addressing cross-domain and topic-specific authorship attribution (AA) challenges (Barlas and Stamatatos, 2021a; Silva et al., 2023). Inspired by experiments conducted in (Devlin et al., 2019b; Horne et al., 2020), this study employs knowledge transfer to incorporate new market insights from emerging Darknet vendors and markets within low-resource environments. This approach leverages pre-trained style representations to train a computationally efficient Bidirectional Gated Recurrent Unit (BiGRU) classifier (Yu et al., 2021) for the vendor identification task.

## 2.3 Dataset

Several researchers have performed similar analyses using scraped data from active Darknet markets. However, since many of these markets have been seized and shut down by law enforcement, replicating their findings or accessing the original datasets is no longer possible. To address this limitation and promote reproducibility for future research, this study leverages publicly available datasets from well-documented Darknet marketplaces. These include Alphabay (2015–2017) (CMU, 2017), Dreams, Traderoute, Valhalla, and Berlusconi (2016–2018) (AZSecure-data et al., 2017), Agora (timeline unknown) (James, 2017), and Silk Road-1 non-anonymous marketplaces (2012–2013) (CMU, 2012).[2]

| Task | Dataset | Ads. | Vendors |
|---|---|---|---|
| **Baseline / Supervised Pre-Training** | Alphabay | 100,429 | 1,457 |
| | Dreams | 93,586 | 1,422 |
| | Silk Road-1 | 78,681 | 1,392 |
| | **Alphabay-Dreams-Silk** | **272,696** | **4,271** |
| **Low-Resource Supervised Market Adaption** | Valhalla | 2,175 | 110 |
| | Berlusconi | 1,437 | 84 |
| | **Valhalla-Berlusconi** | **3,612** | **194** |
| **High-Resource Supervised Market Adaption** | Traderoute | 19,952 | 612 |
| | Agora | 109,644 | 3,187 |
| | **Traderoute-Agora** | **129,586** | **3,799** |

(a) Number of unique ads and vendor accounts per market.

| Language | Alphabay-Dreams-Silk | Traderoute-Agora | Valhalla-Berlusconi |
|---|---|---|---|
| Catalan | 1369 | 484 | 15 |
| English | 252372 | 120482 | 2997 |
| Estonian | NA | NA | 23 |
| Finnish | NA | NA | 285 |
| French | 2073 | 401 | 22 |
| German | 8813 | 2812 | 112 |
| Italian | NA | NA | 11 |
| Norwegian | 745 | NA | 67 |
| Portuguese | 647 | 271 | NA |
| Romanian | 728 | 759 | 8 |
| Somali | NA | 283 | NA |
| Swedish | 1383 | 639 | NA |
| Tagalog | 1597 | 678 | 20 |
| Vietnamese | NA | 1097 | NA |
| Welsh | 1148 | NA | NA |

(b) Top 10 languages (by langdetect) of ads with their frequency.

Table 2.1: Comparison of the darknet market datasets with their associated tasks and language distribution.

Table 2.1a provides an overview of the unique ads (input sequences) and vendor accounts for each market. To investigate vendor migration and aliases across and within Darknet markets, data from Alphabay, Dreams, and Silk Road markets are merged to form the supervised dataset. This combination is based on the timeline and size of these markets. The table also outlines how the datasets are divided for supervised, low-resource, and high-resource fine-tuning steps. Given that the vendor verification task leverages embeddings (ad representations) derived from the classifier trained on the vendor identification task, the largest possible dataset

[2]These datasets are hosted by the IMPACT Cyber Trust Portal and Kaggle.

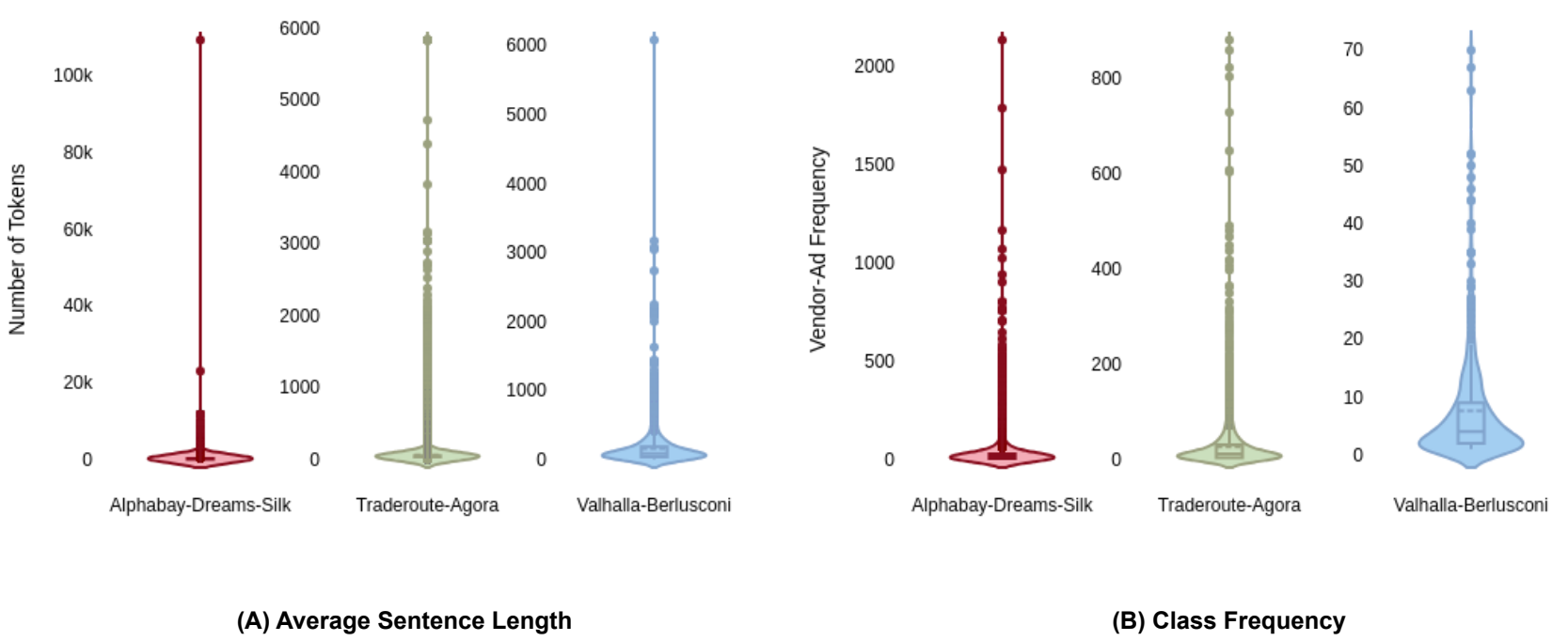

Figure 2.2: (A) Total number of words per ads – Sentence length, (B) Total number of ads per vendor – Class frequency for the darknet datasets.

is chosen for training purposes. [3]

Input sequences are created by concatenating the title and description of ads with the "[SEP]" token. After removing duplicates, Table 2.1b reveals that most ads are written in English, though some contain other languages or multiple languages within the same ad. Interestingly, the noise within the dataset appears to reflect the distinctive writing styles of individual vendors. For instance, the vendor "CaliforniaDreams420" abbreviates "medicines" as "medi...", "SAPIOWAX" frequently uses multiple hyphens ("-") for new lines, and "QualityKing" writes exclusively in uppercase letters. As a result, extensive cleaning or preprocessing could inadvertently eliminate these unique stylistic features, making it counterproductive. However, to streamline vendor labels for classification, all vendor names are converted to lowercase to reduce redundancy, assuming that vendors like "agentq" and "AgentQ" represent the same entity. Vendors with fewer than 20 ads are grouped into a new class labeled "others," enabling the classifier to function effectively in a zero-shot scenario. Although the vendor identification task does not classify vendors with fewer than 20 ads, these ads are analyzed through similarity-based methods in the open-set vendor verification task.

Figure 2.2(A) illustrates the token distribution across all input ads in the

[3] In this context, market data refers to the ads and vendor accounts within a single Darknet market, while a dataset refers to a combination of data from two or more markets.

selected Darknet datasets. The violin plot reveals that the probability density is highest around the median. Since the median token count for the datasets falls between 40 and 100, we truncate ads to the first 512 tokens to enable fair comparisons with other baseline classifiers and transformer-based models. This limit corresponds to the maximum sequence length supported by the transformer architecture used in this study without employing a sliding-window technique (Zhang et al., 2021). Meanwhile, Figure 2.2(B) highlights the class imbalance in the datasets, showing significant variation in the number of ads per vendor account. Some vendor classes and markets exhibit more imbalances than others. Consequently, unlike prior research that predominantly evaluates model performance using accuracy and micro-F1 metrics, this study also emphasizes the model's performance using macro-F1, which assigns equal importance to all classes. Finally, Figure 2.3 illustrates the trade categories in the selected dataset. Notably, drug trade dominates the Traderoute-Agora and Valhalla-Berlusconi datasets, whereas hacking services are the most prevalent in the Alphabay-Dreams-Silk Road dataset.

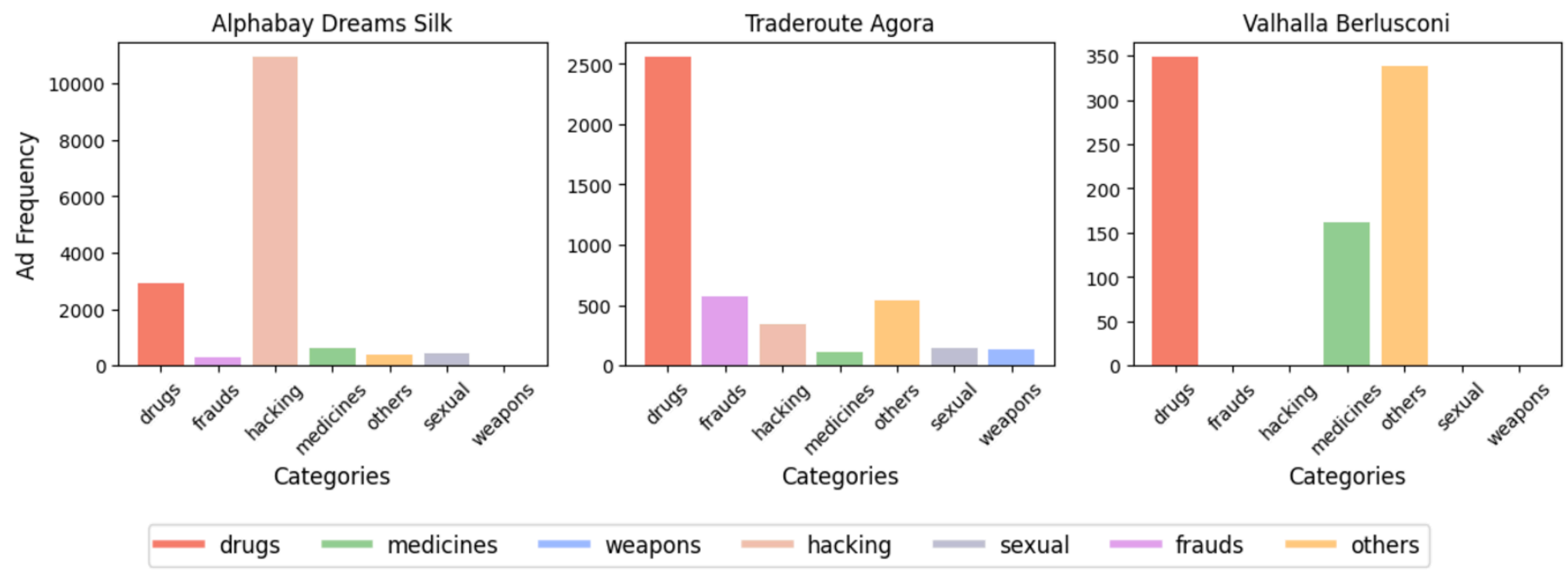

Figure 2.3: Number of Ads with their associated trade categories in Alphabay-Dreams-Silk Road (left), Traderoute-Agora (center), and Valhalla-Berlusconi (right) datasets.

## 2.4 Experiments

### 2.4.1 Sanity Check

Before conducting our experiments, a sanity check is performed to assess the necessity of employing ML algorithms by analyzing the similarity between the darknet ads using traditional stylometric approaches provided by the textdistance library (Voronov, 2019).

**Algorithm 2.1** TextDistance algorithms for stylometric similarity

```
1: Input: Alphabay (A), Dreams (D), Silk Road-1 (S); len(A), len(D),
   len(S) > 1; operation(Op) ∀Op ∈ {within, across}
2: Output: Average similarity
                       // For computing similarity within w and across a markets
3: list_w ← [], list_a ← []

4: function SIMILARITY(text_A, text_B)
5:     return normalized-mean(Levenshtein(text_A, text_B), Jaccard(text_A,
   text_B), Obershelp(text_A, text_B))
6: end function

7: if Op = within then          // Computing average similarity for a vendor
   within a market (e.g., market A)
8:     allVendors ← uniqueVendors(A)
9:     for vendor ∈ allVendors do
10:        for all unique pairs (ad_1, ad_2) in A[vendor] do
11:            if ad_1 ≠ ad_2 then
12:                list_w.append(Similarity(ad_1, ad_2))
13:            end if
14:        end for
15:    end for
16:    averageSimilarity ← MEAN(list_w)
17: else // Computing average similarity for a vendor across markets (e.g.,
   markets A and D)
18:    allVendors ← commonVendors(A, D)
19:    for vendor ∈ allVendors do
20:        for ad_A ∈ A[vendor] do
21:            for ad_D ∈ D[vendor] do
22:                if ad_A ≠ ad_D then
23:                    list_a.append(Similarity(ad_A, ad_D))
24:                end if
25:            end for
26:        end for
27:    end for
28:    averageSimilarity ← MEAN(list_a)
29: end if
```

Since language can be analyzed at the character, token, and sentence lev-els, string, token, and sequence-based similarities are computed between ads using Damerau-Levenshtein distance (character-level similarity), Jaccard Index (token-level similarity), and Ratcliff-Obershelp Pattern Recognition Technique (sequence-level similarity). The overall similarity between two vendor ads is calculated as the average of these three metrics. For a vendor with multiple ads (e.g., Vendor A), the average similarity is computed as the mean similarity between all their ads. For a vendor operating across multiple markets (e.g., Vendor B), ads from Market X are compared with those from Market Y (one market at a time). The average similarity is then computed as the mean of the similarities across all market pairs for Vendor B. Algorithm 2.1 explains the pseudo-code for computing similarity between the ads within and across the darknet markets.

### 2.4.2 Closed-Set Vendor Identification Task

**Architectural Baselines:** To address the task of identifying vendor migrants across multiple markets, this study evaluates various classifiers to analyze the unique writing styles found in Darknet advertisements and establish benchmarks for comparison. Training models on the combined datasets from Alphabay, Dreams, and Silk Road is computationally demanding and time-intensive. Therefore, the baseline architecture is established using a diverse set of approaches:

- **TF-IDF-Based Statistical Models:** These include Multinomial Naive Bayes, Logistic Regression, Random Forest, Support Vector Machines (SVMs), and a Multi-Layer Perceptron (MLP) network.

- **Traditional Neural Networks:** A Bidirectional GRU with FastText embeddings (Gupta et al., 2019) and CNNs operating on character n-grams (Shrestha et al., 2017).

- **Transformer-Based Models:** Pre-trained sequence classifiers such as BERT-base-cased/uncased (Devlin et al., 2019a), RoBERTa-base (Liu et al., 2019b), and DistilBERT-base-cased (Sanh et al., 2019a).

The TF-IDF models, such as Multinomial Naive Bayes and Logistic Regression, rely on the Bag of Words (BoW) technique to represent text data and apply term-weighting to highlight the importance of specific words. These models are computationally efficient and effective for capturing basic stylometric patterns. Traditional Recurrent Neural Networks (RNNs) like BiGRU, with FastText-like word embeddings, capture sequential dependencies and semantic nuances, making them suitable for processing longer text sequences with minimal computational overhead. Similarly,

CNNs operating on character n-grams excel at detecting subtle stylistic variations at the character level, enabling a fine-grained analysis of vendor ads' writing styles. Finally, pre-trained transformer-based backbones like BERT-base-cased/uncased, RoBERTa-base, and DistilBERT-base-cased provide advanced contextual representations by leveraging deep semantic embeddings. Cased architectures are case-sensitive, treating "english" and "English" as distinct tokens, whereas uncased architectures are case-insensitive, interpreting both as identical. Unlike the computational demands of resource-intensive architectures like BERT and RoBERTa-base, DistilBERT, a lighter and distilled version of BERT, employs knowledge distillation to offer comparable performance with reduced computational demands, making it more practical for low-resource environments. All these models are trained on the dreams market dataset comprising 93,586 advertisements to identify 1,422 unique vendor accounts.

**Methodological Baselines:** Building on the architectural baselines, this study establishes methodological baselines to assess the effectiveness of different training strategies on a combined dataset from Alphabay, Dreams, and Silk Road 1, consisting of 272,696 ads and 3,896 unique vendors.

- **BERT-base cased and uncased models:** The first strategy trains BERT-base architectures in cased and uncased variants to determine how uppercase and lowercase patterns in vendor ads influence classification performance.

- **Knowledge transfer with DarkBERT:** The second strategy involves fine-tuning a pre-trained BERT model on Darknet ads for a language modeling task (referred to as *DarkBERT-LM*). This pre-trained model is then used as the backbone for a vendor identification classifier, called the *DarkBERT-classifier*.

- **Adapter layers for parameter-efficient fine-tuning:** Inspired by Houlsby et al. (2019), the third approach involves training a BERT-cased model with adapter layers (*Adapter-BERT*). In this setup, only the adapter layers are updated, while the pre-trained model weights remain frozen, allowing for the evaluation of more efficient fine-tuning techniques.

### 2.4.3 Open-Set Vendor Verification Task

In their research, (Kornblith et al., 2019; Phang et al., 2021) introduced *Centered Kernel Alignment (CKA)* as a similarity metric to reliably measure correspondences between representations in networks trained with different initializations. Building on this approach, this study employs CKA to evaluate the similarity between the representational layers of the trained classifier and an available pre-trained checkpoint that was not trained on Darknet data. Subsequently, the analysis identifies the least similar layers with the lowest CKA similarity, indicating significant changes during training. These layers are further examined to extract semantically meaningful style representations from Darknet market ads. The pseudo-code for computing CKA similarity across layers of the trained classifier and the pre-trained checkpoint is provided in Algorithm 2.2.

**Algorithm 2.2** CKA Similarity Between BERT Layers (Before vs. After Fine-Tuning)

1: **Input:** Darknet ads from Alphabay ($A$), Dreams ($D$), Silk Road-1 ($S$)
2: **Output:** CKA similarity matrix (13x13)
3: Combine all ads: $X \leftarrow A + D + S$
4: Extract embeddings before fine-tuning: $\text{Emb}_{\text{before}} \leftarrow \text{BERT}_{\text{before}}(X)$
5: Extract embeddings after fine-tuning: $\text{Emb}_{\text{after}} \leftarrow \text{BERT}_{\text{after}}(X)$
6: Initialize matrix CKA_similarity[13][13] ← 0
7: **for** $i = 1$ to 13 **do** // layers before fine-tuning
8:     **for** $j = 1$ to 13 **do** // layers after fine-tuning
9:         Get CLS token: $CLS_{\text{before}} \leftarrow \text{Emb}_{\text{before}}[:, i, :]$
10:         Get CLS token: $CLS_{\text{after}} \leftarrow \text{Emb}_{\text{after}}[:, j, :]$
11:         Compute similarity with RBF kernels:
12:         $\text{CKA_similarity}[i][j] \leftarrow \text{CKA}(CLS_{\text{before}}, CLS_{\text{after}})$
13:     **end for**
14: **end for**
15: **return** CKA_similarity

Following the approach of Reimers and Gurevych (2019), the similarity between two vendors is calculated by computing the cosine similarity between the "CLS" token of ad-representations extracted from the trained BERT-base-cased classifier in Section 2.4.2. One vendor is designated as the parent vendor, and this process is repeated for all other vendors in the dataset. However, cosine similarity operates in a linear space, treating all dimensions with equal weight. As noted by Xiao (2018), the focus in such cases should be on the relative ranking of similarities rather than their absolute values. Darknet vendors often advertise products across diverse categories, which lowers the cosine similarity between the ads of two vendors selling under multiple categories. Labeling ads by trade category to enable category-specific comparisons would require significant effort and counteract the goal of maintaining an end-to-end approach. To address this, normalized similarity ($sim_{norm}$) is proposed as a refined measure, which considers the cosine similarity ($sim$) between ads of two vendors while accounting for the self-similarity ($sim_{self}$) within each vendor's ads. This approach is expressed through the following equation:

$$sim_{norm} = 2 * \frac{sim(A,B)}{sim_{self}(A,A) + sim_{self}(B,B)}$$

### 2.4.4 LOW-RESOURCE MARKET ADAPTION TASK

To identify vendor migrants from emerging markets, this study uses the Valhalla-Berlusconi LR dataset, which contains 3,612 ads and 194 vendors. Style representations are extracted from the "[CLS]" token of the pre-trained BERT-based-cased classifier described in Section 2.4.2 for all ads in the dataset. Following the approach outlined by Devlin et al. (2019a), a two-layer bidirectional GRU (Bi-GRU) classifier is initialized with extracted representations to enable knowledge transfer. The Bi-GRU is then fine-tuned to adapt to new market data and identify vendor migrants in the LR dataset. This model is referred to as *transfer-BiGRU*. The use of knowledge transfer allows the BiGRU classifier to evolve and remain effective with emerging vendor and Darknet market data. The performance of *transfer-BiGRU* is compared against both a BERT-based classifier and a Bi-GRU classifier (with FastText embeddings) trained from scratch on the LR dataset. Furthermore, the study also evaluates the zero-shot performance of the architectural and methodological classifiers in comparison to the *transfer-BiGRU* for the closed-set vendor identification task, ensuring a comprehensive assessment of model capabilities.

### 2.4.5 INFRASTRUCTURE & SCHEDULE

**Data:** All the experiments are conducted using a standard splitting ratio of 0.75:0.05:0.20 for the train, validation, and test dataset.

**Training:** Statistical classifiers are trained and evaluated on a server with one Intel Xeon Processor E5-2698 v4 and 512 GB of RAM. All the training and evaluation of neural network models are performed on a single Tesla

V100 GPU with 32 GB of memory. Finally, training on the transfer-BiGRU model for the low-resource setting is conducted on a GeForce-MX110 graphic card with 2 GB of memory. The Adam optimizer is used with $\beta1 = 0.9$, $\beta2 = 0.999$, L2 weight decay of 0.01, and a learning rate of 0.001 is set with warm-up over the first 500 steps, and a linear decay.

**Implementation, Architectures, and Hyperparameters:** Statistical models are trained using unigram and bigram features with balanced class weights. Experiments are conducted on SVMs with both linear and radial basis function (RBF) kernels, Random Forest with *n_estimators* set to 100 and 1000, and *max_depth* values of 5, 10, and 20, and an MLP configured with 100 layers, each containing 100 neurons. These models are evaluated using a 5-fold nested cross-validation approach to ensure robust performance assessment.

The CNN architecture processes sequences of n-gram characters extracted from the darknet ads. This input is passed through six convolutional layers with max-pooling, followed by three fully connected layers. Inspired by (Zhang et al., 2016), the input length is set to 1014 characters (within a max range of 512 tokens), dropout to 0.5 for fully connected layers with 768 neurons, kernel size to 7 for the first two convolutional layers, and 3 for the remaining layers. The filter size is set to 32, and the models are trained with a batch size of 32 until convergence.

For the RNN architecture, a two-layer bidirectional-GRU model with two fully connected layers is initialized with FastText embeddings. Input sequences with variable lengths (but a max of 512 tokens) are first packed and padded using PyTorch utilities before being passed to the embedding layer. The output representations from the Bi-GRU layers are passed through a softmax layer for classification. The hidden units are set to 768, dropout to 0.65, batch size to 32, and training continues until convergence.

Finally, transformer-based architectures, including BERT-base-cased, BERT-base-uncased, RoBERTa-base, and DistilBERT-base-cased, are trained with a sequence classification head. These models are initialized from pre-trained checkpoints and trained with a batch size of 32 (the maximum feasible batch size given resource constraints) for 40 epochs for architectural baselines and until convergence for methodological baselines. Additionally, a BERT-base-uncased model is trained for 20 epochs on the masked language modeling task [4].

All the models are implemented in Python (Van Rossum and Drake Jr (1995)) using Sklearn (Pedregosa et al. (2011b)), PyTorch (Paszke et al. (2019a)), and Hugging-face (Wolf et al. (2020)) frameworks.

**Computational Details:** Table 2.2 and Table 2.3 provide an overview of the number of trainable parameters and the execution time for all the models trained in the architectural and methodological baselines.

| Models (trained on Dreams data) | Trainable parameters | Training time in hrs. |
|---|---|---|
| Multinomial Naive Bayes | - | 53:56 |
| Random Forest | - | 68:27 |
| Logistic Regression | - | 79:42 |
| SVM | - | 81:08 |
| MLP | - | 94:18 |
| Character-CNN | 16M | 0:54 |
| GRU-Fasttext | 39M | 1:12 |
| BERT | 110M | 25:14 |
| RoBERTa | 125M | 23:40 |
| DistilBERT | 68M | 17:57 |

Table 2.2: Number of trainable parameters and training time for VendorLink's architectural baselines.

| Models (trained on Alphabay-Dreams -Silk Road dataset) | Trainable parameters | Training time in hrs. |
|---|---|---|
| BERT-uncased | 111M | 67:02 |
| BERT-cased | 112M | 66:58 |
| DarkBERT-LM | 108M | 156:14 |
| DarkBERT Classifier | 112M | 49:39 |
| Adapter BERT | 4M | 51:00 |

Table 2.3: Number of trainable parameters and training time for VendorLink's methodological baselines.

**Evaluation Metrics:** The evaluation of trained classifiers is performed using metrics such as accuracy, micro-average F1, and macro-average F1 (commonly referred to as micro-F1 and macro-F1), derived from the classification report provided by scikit-learn. Macro-F1 computes the score for each class independently and averages them, ensuring equal treatment of majority and minority classes. Due to the class imbalance in the dataset, a strong emphasis is placed on the models' performance in terms of macro-F1 scores. Additionally, the BERT-base-uncased language model is evaluated based on loss and perplexity. Centered Kernel Alignment (CKA) is also utilized to assess and compute the alignment between the representations of the methodological baselines before and after fine-tuning.

[4] The experiments reveal that the BERT-base-uncased architecture outperforms the cased architecture on the masked language modeling task. Due to the unpredictable noise in data from Darknet markets, the cased architecture fails to predict the case-sensitive next token accurately

## 2.5 Results

### 2.5.1 Sanity Check: Stylometric Baselines

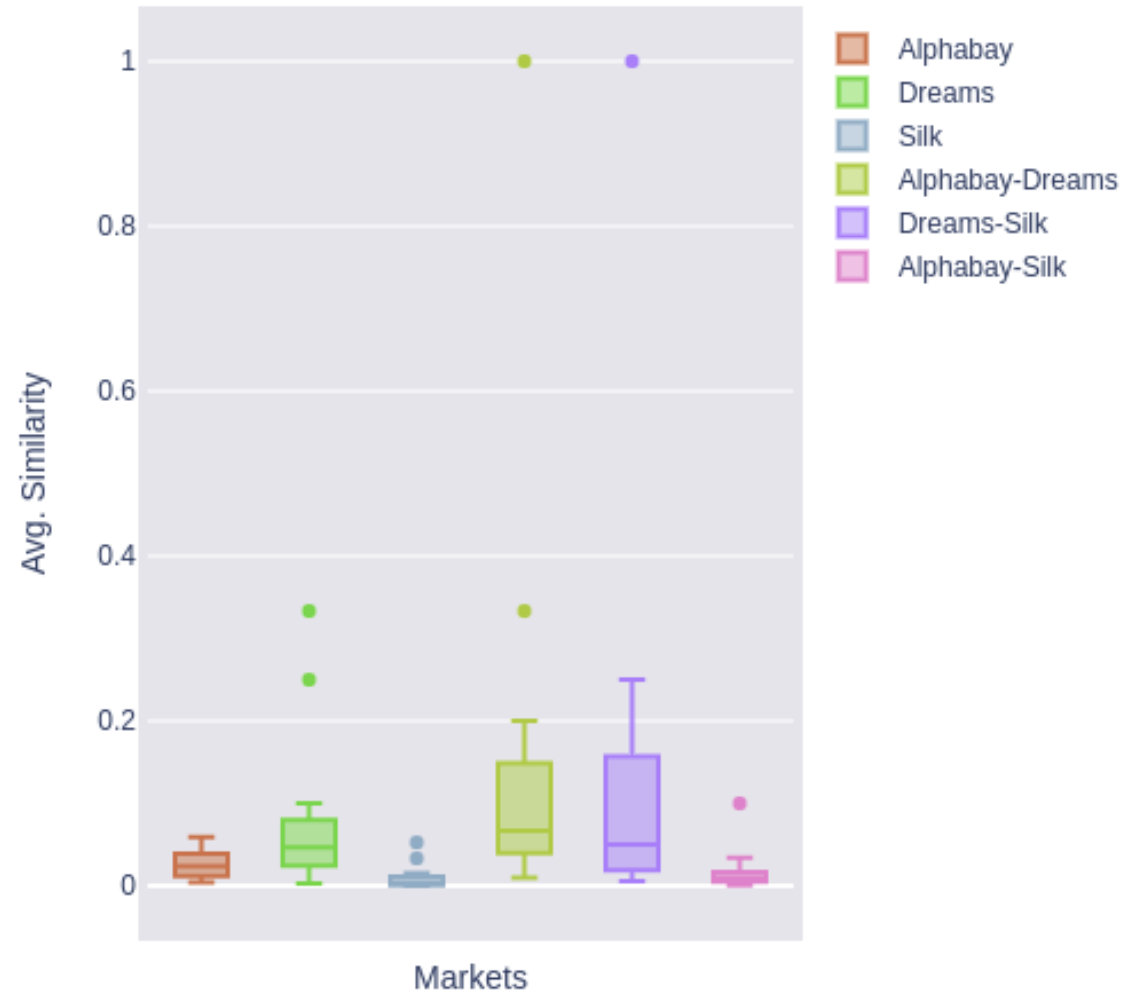

Figure 2.4: Average stylometric similarity (as computed by textdistance) between the vendor ads within and across darknet markets.

In response to the experiment in Section 2.4.1, Figure 2.4 showcases the effectiveness of textdistance-based stylometric methods in identifying similarities between vendor ads, presented through a box plot. The plot displays the average similarity distribution and its skewness for ads within the same markets (Alphabay-Alphabay, Dreams-Dreams, Silk Road-Silk Road) and across different markets (Alphabay-Dreams, Dreams-Silk Road, Alphabay-Silk Road). Due to the significant computational demands of these traditional techniques, results for the Alphabay-Dreams-Silk dataset are unavailable, rendering execution infeasible. Most ads exhibit an average similarity below 0.20, with a few outliers showing higher similarity scores. One notable exception is the vendor "cyanspore," which achieves a similarity score of 1.0 in the Alphabay-Dreams and Dreams-Silk datasets. Since the ads from this vendor are identical, they are excluded from further analysis.

The consistently low similarity scores across datasets reveal the limitations of these stylometric methods in capturing nuanced textual patterns. These findings emphasize the need for advanced models capable of ab-stracting features at a deeper level. Furthermore, the low scores validate that darknet vendors employ distinct vocabularies and writing styles in their ads, both within and across markets, reinforcing the necessity of adopting more sophisticated feature abstraction techniques.

### 2.5.2 CLOSED-SET VENDOR IDENTIFICATION TASK

| Models | Accuracy | Micro-F1 | Macro-F1 |
|---|---|---|---|
| **Statistical Models** | | | |
| Multinomial Naive Bayes | 0.0183 | 0.0144 | 0.0059 |
| Random Forest | 0.0102 | 0.1093 | 0.0449 |
| Logistic Regression | 0.0045 | 0.0090 | 0.0037 |
| SVM | 0.2480 | 0.3974 | 0.3703 |
| **Neural Networks** | | | |
| MLP | 0.6614 | 0.6603 | 0.6594 |
| Character-CNN | 0.7266 | 0.7256 | 0.7248 |
| BiGRU-Fasttext | 0.7374 | 0.7415 | 0.7360 |
| **Transformers Networks** | | | |
| **BERT-cased** | **0.8978** | **0.8978** | **0.9002** |
| DistilBERT-cased | 0.8886 | 0.8885 | 0.8889 |
| RoBERTa-base | 0.8776 | 0.8797 | 0.8736 |

Table 2.4: Performance of architectural baselines on the Dreams market data.

| Models | Accuracy | Micro-F1 | Macro-F1 |
|---|---|---|---|
| BERT-uncased | 0.8947 | 0.8939 | 0.8768 |
| **BERT-cased** | **0.9046** | **0.9066** | **0.9013** |
| DarkBERT-Classifier | 0.9000 | 0.9090 | 0.9073 |
| Adapter BERT | 0.8398 | 0.8330 | 0.8188 |

Table 2.5: Performance of methodological baselines on the combined Alphabay-Dreams-Silk Road 1 dataset.

In evaluating architectural baselines on the Dreams market, Table 2.4 reveals a nuanced performance landscape across different model architectures. The Multilayer Perceptron (MLP) with bigram Bag-of-Words (BoW) features emerges as the top-performing statistical model. At the same time, neural networks like character-based CNN and Bidirectional GRU with fasttext embeddings demonstrate superior performance. The most notable improvement comes from transformer-based architectures, with the BERT-base-cased model significantly outperforming the RoBERTa-base model. The difference in performance is attributed to the RoBERTa model's byte-level BPE tokenizer, which struggles to learn complex features in the noisy, intentionally obfuscated text of Darknet markets. Consequently, the trained BERT-cased classifier on the Dreams market is established as the benchmark for architectural baselines.

Table 2.5 illustrates the comprehensive assessment of the methodological baselines on the combined Alphabay-Dreams-Silk Road-1 test dataset, focusing on writing style influences and model refinement strategies. The experiment reveals that the BERT-cased classifier outperforms its uncased counterpart by approximately 3% across 3,896 class labels, highlighting the significance of preserving uppercase and lowercase patterns. Con-tinued pre-training of DarkBERT-LM on advertisements yields a test perplexity of 2.07 but results in only marginal performance improvements over the BERT-cased classifier. The subtle gains reflect the challenging nature of darknet vendor language, characterized by its unpredictable and noisy linguistic patterns. The Adapter BERT similarly underperforms relative to the vanilla BERT-cased classifier, ultimately reinforcing the BERT-cased architecture trained on the closed-set vendor identification task as the benchmark classifier for the Alphabay-Dreams-Silk Road dataset [5].

#### EFFECTS OF TRADE CATEGORIES AND RANDOM INITIALIZATION ON CLASSIFIER'S PERFORMANCE

| Vendor | Ad Frequencies | Categories | F1-Score |
|---|---|---|---|
| googleyed | 349 | 63 | 0.9340 |
| etizolam | 186 | 59 | 0.9462 |
| gotmilk | 842 | 48 | 0.9893 |
| rinran | 437 | 47 | 0.9816 |
| uhrwerk | 135 | 41 | 0.9925 |
| citizen5 | 35 | 1 | 1.0000 |
| corktech | 35 | 1 | 0.9714 |
| sabinas | 26 | 1 | 0.9615 |
| mrsupermario | 24 | 1 | 0.9615 |
| emperium | 22 | 1 | 1.000 |

Table 2.6: Macro-F1 scores w.r.t vendor advertisement frequency and trade categories.

| Seed | Macro-F1 |
|---|---|
| 40 | 0.8939 |
| 100 | 0.8824 |
| 500 | 0.8812 |
| 1000 | 0.8861 |
| 1111 | 0.9013 |
| **Variance** | $6.46 \times 10^{-5}$ |
| **Std. Dev.** | 0.0080 |
| **Average** | 0.8896 |

Table 2.7: Effects of random initialization on the BERT-cased classifier trained on the Alphabay-Dreams-Silk Road dataset.

The Alphabay-Dreams-Silk Road dataset comprises 272,696 unique ads from 3,896 vendors across 322 distinct categories. Performance analysis of the established BERT-cased classifier in Table 2.6 reveals consistent effectiveness across vendors selling trades in multiple and single categories. This stable performance suggests that darknet vendors maintain similar writing patterns across different trade categories, which the model can successfully leverage to distinguish unique vendor writing styles. Finally, Table 2.7 illustrates the trained classifier's standard deviation, variance, and average performance when initialized randomly. The slight variation in performance (about 1%) demonstrates the robustness of the trained classifier for the vendor identification task.

[5]Since the masked language modeling and fine-tuning of DarkBERT-Classifier did not yield any substantial gain in performance, in the interest of computational expense, further analysis is only conducted using the BERT-cased architecture initialized from a pre-trained checkpoint without any domain-specific masked language fine-tuning.

## MODEL EXPLANATIONS

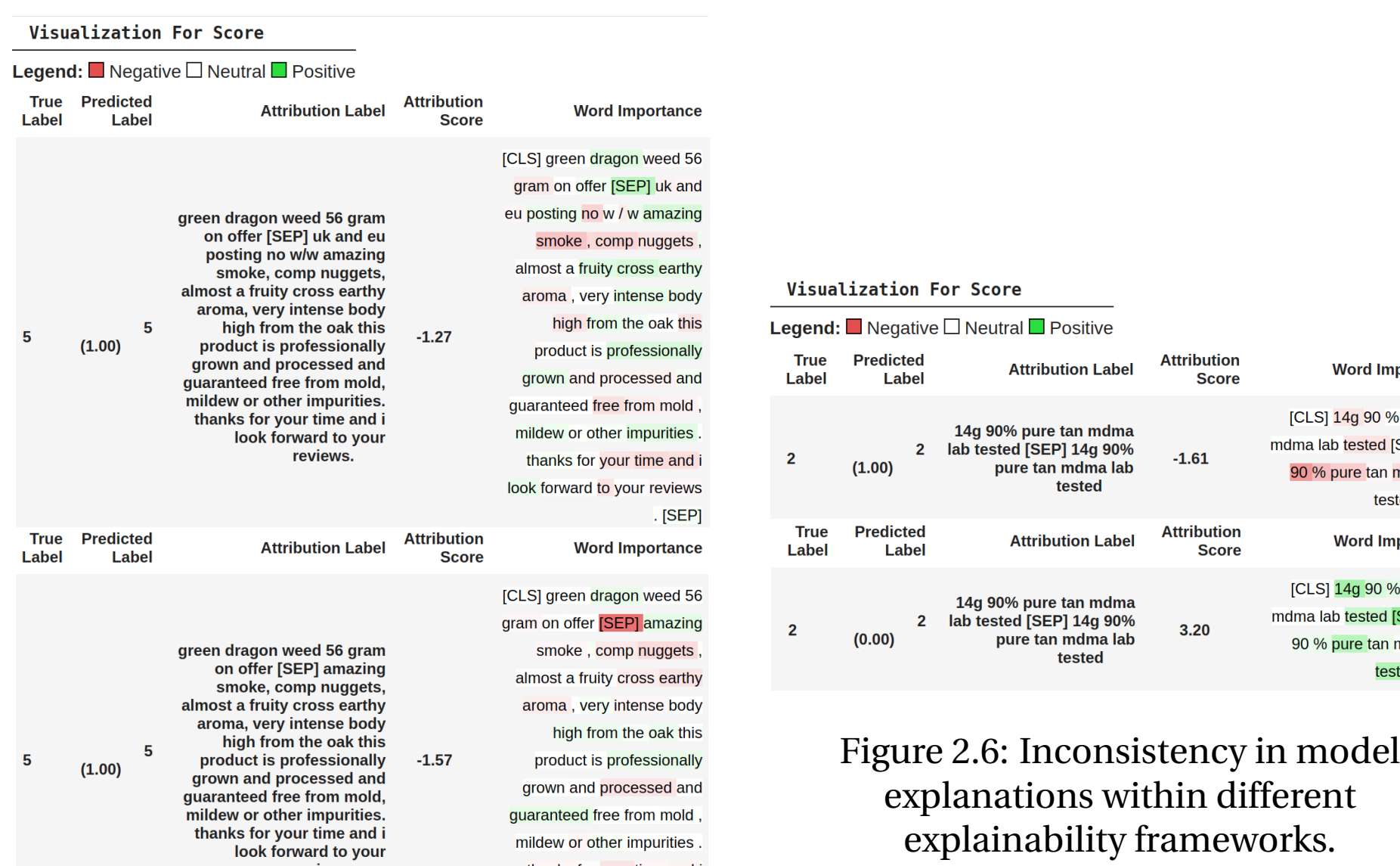

Figure 2.5: Inconsistency in model explanations for similar ads from the same vendor.

Figure 2.6: Inconsistency in model explanations within different explainability frameworks.

Various word attribution-based explainability experiments are conducted on the BERT-cased methodological classifier to gain insights into the model's decision-making process. Figure 2.6 depicts word attributions for the same advertisement from the vendor "pckabml," generated using the captum (Kokhlikyan et al., 2020) and transformers-interpret (Pierse, 2021) frameworks. The results show that, despite identical ads, the two frameworks produce different word attributions, leading to inconsistencies in the explanations provided.

Similarly, Figure 2.5 presents capture-based word attributions for similar advertisements from the vendor "idol." Even with the same explainability framework and high ad similarity, the generated word attributions differ, highlighting inconsistencies in the explanations. These discrepancies may stem from computing word attributions using the [CLS] token rather than analyzing the entire advertisement. While word attribution techniques offer valuable insights, their limitations as an explainability method—such as sensitivity to framework design and token-level computation—underscore the need for further research to develop more robust and consistent approaches.

### 2.5.3 Open-Set Vendor verification task

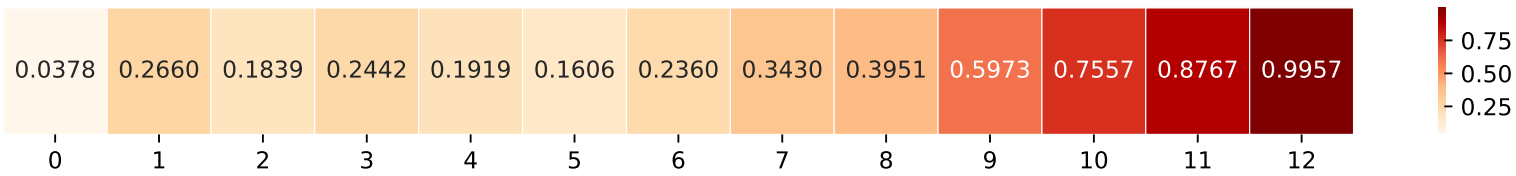

Figure 2.7: CKA distance between layers of the BERT-cased methodological classifier, compared before and after being trained on the Alphabay-Dreams-Silk dataset.

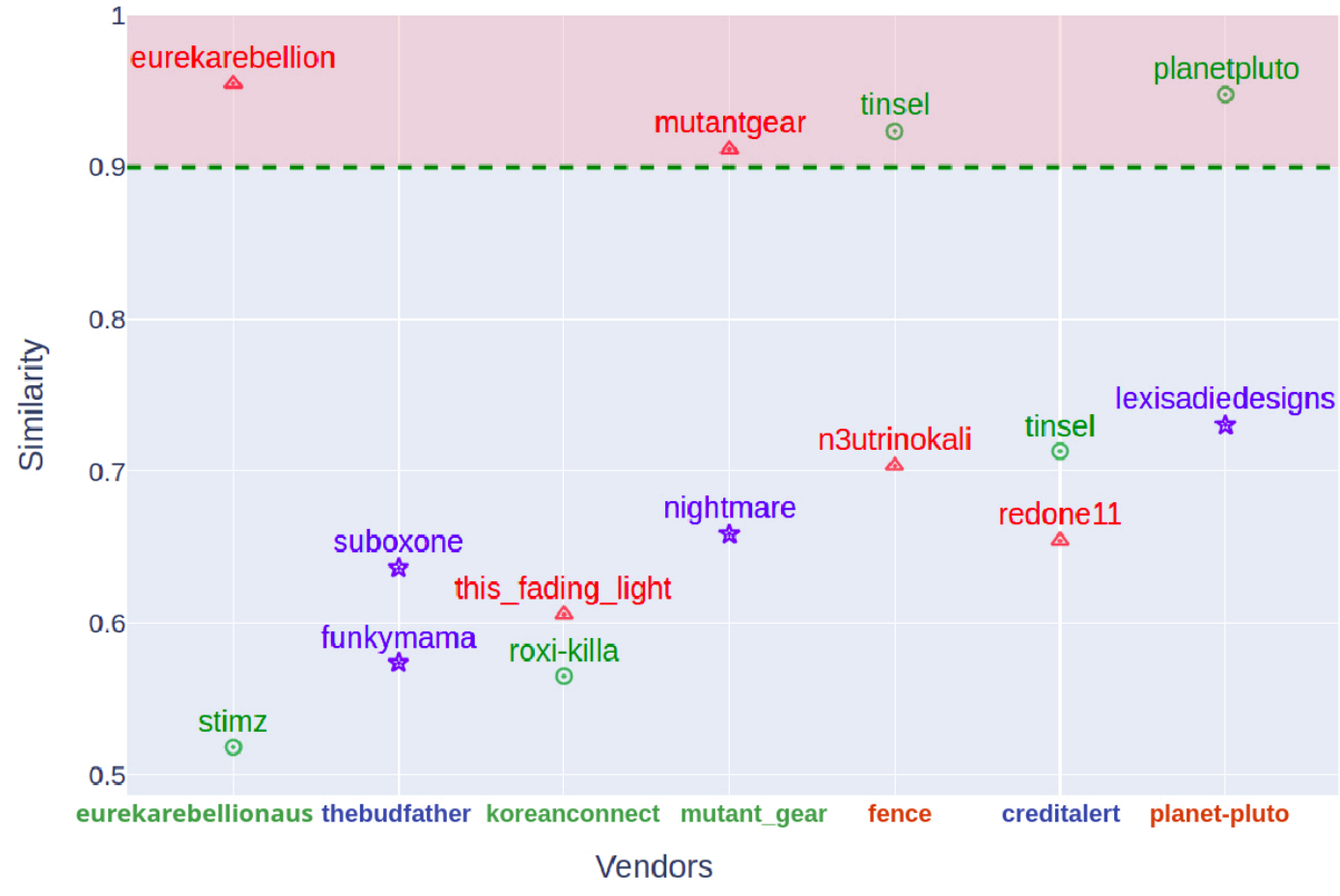

Figure 2.8: Scatter plot between parent vendors (on the x-axis) and their potential aliases (scatter points on y-axis) from Alphabay, Dreams, and Silk Road-1 markets.

Figure 2.7 illustrates an analysis of the BERT-cased classifier's representational layers, revealing a high CKA distance across the final four layers, indicating that the weighted sum of these layers provides the most meaningful style representations for the ads in Alphabay-Dreams-Silk Road 1 dataset. This experiment utilizes these style representations to compute cosine similarities between vendor ads, generating a scatter plot highlighting potential vendor aliases. The analysis identified several vendor pairs with remarkably high ad similarities. Figure 2.8 displays some randomly selected parent vendors on the x-axis and their two potential aliases (scatter points) with a similarity score in their ads on the y-axis. For instance, "eurekarebellionaus" and "eurekarebellion," "mutant_gear" and "mutantgear", "fence" and "tinsel," and "planet-pluto" and "planetpluto" have very high similarity in their ads and can be associated to the same vendor. The higher the similarity, the more likely it is for two vendors to be the same entity. For better visibility, these vendors are highlighted inside the red box of our scatter plot. The scatter plot is generated using Plotly, which allows to zoom infinitely for any vendor. However, only two potential aliases are displayed in the figure for better clarity and visibility. Theoretically, this plot can display the similarity between all the vendors and a parent vendor.

As mentioned, darknet vendors often create aliases to hide from LEAs. However, since numerous vendors appear on darknet markets yearly, it becomes difficult for law enforcement to link these aliases to a parent vendor manually. The unavailability of ground truth poses a challenge in evaluating the existence of these aliases in the datasets. Therefore, this experiment cannot confidently comment upon the accuracy of the similarity-based analyses without the qualitative case study. Practitioners are encouraged not solely to rely on these similarities but to use them as a starting point for their manual investigations. Furthermore, it is strongly discouraged to abide by these analyses as evidence for investigation or prosecution. The sole purpose of this experiment is to help researchers, LEAs, and practitioners bring meaning to the online darknet market data.

| | **Parent Vendor** | **Potential Alias / Copycat** | **Similarity** |
|---|---|---|---|
| High (potential aliases) | houseofdank | houseofdank2.0 | 0.9844 |
| | incorporated | incorporatedv2 | 0.9769 |
| | castro6969 | castro69696 | 0.9541 |
| | thewizard | thewizzardnl | 0.9480 |
| | europills | europills2 | 0.9467 |
| Low (potential Copycats) | topgear | topgear69 | 0.0367 |
| | dutchpirates | dutchpiratesshop | -0.1015 |
| | whitey | whiteyford | -0.1410 |
| | g3cko | gecko | -0.2292 |
| | aussieimportpills | aussieimportpillsv2 | -0.2560 |

Table 2.8: Normalized similarity between parent vendors and their potential aliases/copycats aligned in decreasing order.

Vendor aliases in darknet markets often employ similar-looking handles to establish brand recognition and market dominance. While string-based matching techniques like string_grouper (Berg, 2021) typically detect similar accounts, experimental findings revealed a complex landscape of vendor identities. Specifically, the analysis of the Alphabay-Dreams-Silk dataset uncovered that merely 24% of similar-looking vendor-alias pairs demonstrated a similarity score of 0.7 or above in their advertisements. The experiment exposed the existence of sophisticated copycats characterized by distinctly different writing styles and low advertisement similarities. Detailed examination of 10 parent vendors and their potential aliases, illustrated in Table 2.8, highlights the nuanced challenges in identifying vendor relationships. Vendor relationships cannot be determined through simplistic approaches: some vendor handles appear similar yet exhibit dramatically different writing styles, while others with seemingly unrelated names demonstrate remarkable similarities in their ads. Traditional string-matching tools like string_grouper predominantly focus on superficial handle similarities, potentially overlooking sophisticated vendor alias strategies employed in darknet markets. Furthermore, the experimental approach developed in this study introduces a methodology that can potentially assist in uncovering vendor-alias pairs that would typically escape conventional detection methods—for instance, “fence” and “tinsel,” as shown in Figure 2.8.

### 2.5.4 Low Resource Market Adaption Task

To set the Zero-Shot baselines, the BERT-cased architectural and methodological classifiers are evaluated for zero-shot vendor identification on the LR dataset, Valhalla-Berlusconi. Since the emerging LR dataset introduces new vendors, all emerging vendor accounts are assigned to the class labeled "others." Due to the macro-F1 score being an unweighted mean of F1 scores for all classes, the absence of overlap between vendors in the LR dataset and those from prior markets leads to unreliable macro-F1 results. Therefore, the zero-shot evaluation emphasizes the micro-F1 metrics. Despite not being trained on LR data, the Zero-Shot baselines achieve promising micro-F1 scores of 0.7702 and 0.7388. The decrease in macro-F1 performance from the architectural to the methodological baseline reflects the increase in vendor accounts from 1,442 in the Dreams market to 3,896 in the Alphabay-Dreams-Silk Road dataset.

In addition to Zero-Shot baselines, a BERT-cased and a BiGRU classifier with fasttext embeddings are trained from scratch to adapt to the new market knowledge and vendors in the LR dataset. As illustrated in Table Table 2.9, these End-to-End baselines significantly improve both micro-F1 and macro-F1 scores compared to the Zero-Shot baselines. Knowledge transfer experiments further enhance the performance by extracting style representations from multiple layers of the BERT-cased methodological classifier and using them to initialize the BiGRU before the classification layer. Table 2.9 indicates that initializing with the weighted sum of the last four layers provides the most benefit, enabling the transfer-BiGRU classifier to perform comparably to the End-to-End BERT-cased classifier on the LR dataset. The transfer-BiGRU architecture trained on the closed-set vendor identification task is thus established as the benchmark classifier for the LR, Valhalla-Berlusconi dataset.

| Models | Layer | Micro-F1 | Macro-F1 |
|---|---|---|---|
| *Zero-Shot Baselines* | | | |
| Architectural | - | 0.7702 | 0.2927 |
| Methodological | - | 0.7388 | 0.2401 |
| End-to-End Baselines | | | |
| **BERT-cased** | - | **0.8987** | **0.8148** |
| BiGRU-Fasttext | - | 0.7797 | 0.6957 |
| *Transfer Baselines* | | | |
| **Transfer-BiGRU** | Embedding | 0.7653 | 0.6408 |
| | Last | 0.8590 | 0.7809 |
| | Second-to-Last | 0.8951 | 0.7884 |
| | Weighted Sum All 12 | 0.8928 | 0.7837 |
| | **Weighted Sum Last 4** | **0.8946** | **0.8132** |

Table 2.9: Performance of Zero-Shot, End-to-End, and Transfer baselines on the Valhalla-Berlusconi dataset.

| GPU | Models | Trainable parameters | Training Time (Hrs:Mins) |
|---|---|---|---|
| Tesla-V100 (32 GB) | BERT-cased | 110M | 0:54 |
| | BiGRU-Fasttext | 13M | 0:12 |
| | Transfer-BiGRU | 24M | 0:32 |
| GE-MX110 (2 GB) | Transfer-BiGRU | 24M | 2:40 |

Table 2.10: GPU training details for different models and training configurations.

| Seed | BERT-cased Macro-F1 | Transfer-BiGRU Macro-F1 |
|---|---|---|
| 40 | 0.8039 | 0.7798 |
| 100 | 0.8278 | 0.8005 |
| 500 | 0.7837 | 0.8221 |
| 1000 | 0.8089 | 0.8019 |
| 1111 | 0.8290 | 0.8132 |
| **Variance** | 0.0002 | 0.0002 |
| **Std. Dev.** | 0.0167 | 0.0143 |
| **Average** | 0.8106 | 0.8035 |

Table 2.11: Influence of different initialization on Macro-F1 performance of BERT-cased and Transfer-BiGRU classifiers trained on Valhalla-Berlusconi dataset.

Table 2.10 compares the computational aspects of the trained models on the LR dataset, highlighting significant differences in efficiency. The transfer-BiGRU classifier requires 78% less fewer trainable parameters than the BERT-cased model and reduces training time by approximately half. Additionally, the transfer-BiGRU model demonstrates feasibility on a low-end GPU, such as the GeForce MX110 with 2 GB of memory. This low-compute transfer-BiGRU classifier provides an effective, scalable solution for adapting vendor identification approaches to emerging markets with minimal resource requirements and no significant performance loss. Finally, Table 2.11 demonstrates the effects and robustness of the trained classifiers on the LR dataset when trained from random initialization.

### Adapting to High-Resource (HR) emerging markets

| Models | Layer | Micro-F1 | Macro-F1 |
|---|---|---|---|
| *Zero-Shot Baselines* | | | |
| Architectural | - | 0.7305 | 0.2173 |
| Methodological | - | 0.6498 | 0.1563 |
| End-to-End Baselines | | | |
| **BERT-cased** | - | **0.8750** | **0.8700** |
| BiGRU-Fasttext | - | 0.6577 | 0.6539 |
| *Transfer Baselines* | | | |
| **Transfer-BiGRU** | Embedding | 0.6707 | 0.6698 |
| | Last | 0.7061 | 0.7153 |
| | Second-to-Last | 0.6992 | 0.6911 |
| | Weighted Sum All 12 | 0.6698 | 0.6703 |
| | **Weighted Sum Last 4** | **0.8065** | **0.8177** |

Table 2.12: Performance of Zero-Shot, End-to-End, and Transfer baselines on the Traderoute-Agora dataset.

| GPU | Models | Trainable parameters | Training Time (Hrs:Mins) |
|---|---|---|---|
| Tesla-V100 (32 GB) | BERT-cased | 112M | 32:30 |
| | BiGRU-Fasttext | 31M | 2:25 |
| | Transfer-BiGRU | 42M | 17:23 |

Table 2.13: Computational details of trained classifiers on the Traderoute-Agora dataset.

Following the methodology established in the LR setting experiment, the study extended the experiments to an HR emerging dataset from Traderoute-Agora markets. Table 2.12 and 2.13 present the performance and computational details of the transfer-BiGRU classifier in this context. Despite the transfer-BiGRU's advantages in terms of fewer trainable parameters and reduced training time, the classifier underperforms compared to the end-to-end BERT-cased baseline. This performance discrepancy suggests that the knowledge transfer approach does not universally scale across the emerging HT scenarios.

The transfer-BiGRU model utilizes BERT embeddings and fine-tunes them to enable some domain adaption in LR scenarios, where limited data enables the transfer-BiGRU model to generalize without overfitting. Here, the pre-trained embeddings offer rich semantic priors, and the BiGRU's relatively shallow and efficient architecture enables effective learning with fewer trainable parameters. However, this approach underperforms in the HR scenario due to fundamental architectural limitations. BERT embeddings are deeply contextualized through multiple layers of self-attention, capturing intricate, hierarchical relationships between tokens across an entire sequence. When only these final embeddings are passed to a BiGRU, the transfer-BiGRU model lacks access to the full transformer stack that generated this complexity and cannot replicate the reasoning inherent in BERT's architecture. Despite fine-tuning, the BiGRU processes sequences sequentially and cannot model long-range dependencies or complex relations as effectively as the end-to-end BERT model. As a result, while the transfer-BiGRU provides a lightweight and effective solution under LR scenarios, it fails to scale performance in HR contexts where deeper, end-to-end transformer models can better utilize the abundance of available data.

### 2.5.5 ERROR ANALYSIS

To evaluate the strengths and weaknesses of the trained models, qualitative analysis is conducted on the predictions of the BERT-cased classifier trained on the Alphabay-Dreams-Silk Road dataset. Table 2.14 provides selected examples, displaying only the titles of ads for clarity and space constraints. The first two examples highlight the model's ability to recognize patterns such as "**," "[DRUG1]," "[drug1]," and "greenhouse gown." These examples also demonstrate the similarity between the advertisements of vendors "houseofdank" and "houseofdank2.0," supported by the high similarity scores in Table 2.8. The table also includes false positive examples, where the network confuses terms like "COUNTERFEIT," "quality," "supernotes," source and destination locations, specific drug names ("[drug3]"), and product prices. These instances indicate challenges in distinguishing between subtle variations in vocabulary and context, leading to misclassification.

Further analysis is performed on cases where the BERT-cased classifier fails in the Zero-Shot setting, but the transfer-BiGRU classifier succeeds after knowledge transfer. Table 2.15 showcases vendor ads with significant changes in writing style between the Alphabay-Dreams-Silk Road and Valhalla-Berlusconi datasets. These shifts in style cause the BERT-cased classifier to underperform in the zero-shot setting, failing to verify vendors in the new dataset. After applying knowledge transfer and fine-tuning, the transfer-BiGRU model effectively adapts to the new writing styles in these advertisements, demonstrating its robustness in handling stylistic variations across datasets.

2

| Vendor | Pred | Text A | Text B |
|---|---|---|---|
| house ofdank | TP | ** 1 Lb of Sour [DRUG1] (Greenhouse) ** | ** 1 oz of Greenhouse [drug1] greenhouse grown ** |
| house ofdank2.0 | TP | ** 1 OZ of [drug2] Greehouse grown ** | ** 1 Lb of [drug2] Greehouse grown ** |
| appleinc | FP | 10 x €50 euro COUNTERFEIT notes (Very Good Quality) | 5 x $100 DOLLAR COUNTERFEIT STRIP high quality bills |
| canadian pharmacy | FP | [usa to usa] [drug3] 80mg just 19.99 bucks per pill only | [usa to usa] 30 pills [drug3] 100mg 19.99 usd ultram |

Table 2.14: Qualitative analysis of BERT-cased classifier (trained on Alphabay-Dreams-Silk Road Dataset) for True Positives (TP) and False Positives (FP) predictions.

| Vendor | Alphabay-Dreams-Silk Road | Valhalla-Berlusconi |
|---|---|---|
| cannacorner | [drug1] 3.5g —\|MERCEDES\| | 7g [drug1] —\|lambo\| |
| medicalznl | 5 GRAMS COLOMBIAN [DRUGX] 93% + FREE SHIPPING | 2.5 grams - colombian [drugx] 90+% pure uncut |
| color | Credit Cards Can Be Without Security Code | lasted update credit cards in this file. |

Table 2.15: Qualitative analysis of transfer-BiGRU classifier (trained on Valhalla-Berlusconi Dataset) for True Positives (TP) and False Positives (FP) predictions.

## 2.6 Broader Discussion and Challenges

**Assumption:** This study applies a lower-case transformation to vendor names during preprocessing, treating accounts such as "agentq" and "AgentQ" as belonging to the same entity. While this assumption simplifies analysis, it introduces the risk of misclassifying separate vendors with similar names as a single entity. The classifier is trained in a multi-class setting, where each ad is assumed to correspond to a single vendor account. However, the experimental findings of this study reveal the presence of copycats on Darknet markets—distinct vendors who post similar advertisements under different handles. This discovery challenges the assumption that each ad is uniquely associated with one vendor. Moreover, the possibility of multiple vendors operating with similar names highlights the limitations of supervised approaches in this context. These observations underscore the need for more advanced methodologies capable of effectively addressing ambiguities in vendor identification tasks.

**Unsupervised and HR Settings:** This study relies on the availability of gold labels as a foundational assumption. It employs a supervised pre-training step to facilitate text similarity and knowledge transfer tasks, making the approach dependent on labeled data. Therefore, this approach suffers a significant limitation without these ground truths. Additionally, as detailed in Table 2.12, the approach demonstrates challenges in scaling effectively to identify vendor migrants within HR emerging datasets. This limitation highlights the need for further investigation to develop robust methods to address these scenarios.

**Resource-Intensive Text Similarity Task:** The text similarity task in this study imposes substantial computational demands. Extracting embeddings for each ad associated with a vendor and iteratively comparing them with all ads of other vendors requires significant computational resources. This process, executed across all vendors using cosine similarity, results in an extensive computational load. While effective for analyzing vendor similarities, such a setup proves inefficient and resource-intensive when scaling to large datasets or real-time applications. In subsequent chapters, this study formalizes the vendor verification task in a more streamlined and computationally efficient fashion to enable faster and more scalable linking of aliases and emerging vendors, addressing the limitations of the current methodology.

## 2.7 SUMMARY

In response to RQ1(a), this study develops VendorLink, an NLP-driven approach designed to support researchers, LEAs, and practitioners in verifying, identifying, and linking vendor migrants and potential aliases across text ads from existing and emerging Darknet markets. The approach begins with supervised fine-tuning to incorporate domain-specific knowledge from Darknet markets, enabling the development of a BERT-cased classifier for closed-set vendor identification tasks. Subsequently, style representations are extracted from the trained classifier to compute text similarities between vendor ads in an open-set environment, facilitating the verification of potential aliases. To address scalability and adaptability, knowledge transfer from the BERT-cased classifier is employed to train a low-resource BiGRU classifier, enabling closed-set vendor verification on emerging LR markets. Experiments reveal the discovery of 15 migrants and 71 potential aliases in the Alphabay-Dreams-Silk dataset, 17 migrants and three potential aliases in the Valhalla-Berlusconi dataset, and 75 migrants and 10 potential aliases in the Traderoute-Agora dataset. The findings of potential aliases are based on a cosine similarity threshold of 0.8 or higher between vendor advertisements, underscoring the effectiveness of the proposed approach in identifying stylometric patterns of vendor activity across multiple marketplaces.

# 3

# IDTRAFFICKERS: AUTHORSHIP ATTRIBUTION FOR ONLINE ESCORT MARKETS

**This chapter is based on the following research:**

- **Vageesh Saxena**, Benjamin Ashpole, Gijs van Dijck, and Gerasimos Spanakis. 2023a. IDTraffickers: An Authorship Attribution Dataset to link and connect Potential Human-Trafficking Operations on Text Escort Advertisements. In Proceedings of the 2023 Conference on Empirical Methods in Natural Language Processing, pages 8444–8464, Singapore. Association for Computational Linguistics.

Chapter 2 lays the groundwork for using authorship analysis to connect and trace illegal trafficking activities by linking vendor ads through their writing styles on Darknet markets. It introduces a BERT-cased classifier that identifies vendors by analyzing unique writing signatures in text descriptions, addressing the challenge of tracking vendor migrants across different Darknet platforms. The trained representations (text embeddings) from this classifier are further utilized in a text-similarity task, where cosine similarity is computed to verify vendors by identifying similarities between ads of a parent vendor and their potential aliases, both within and across markets. To address emerging markets and new vendors, a knowledge-transfer task uses the trained representations from the BERT-cased classifier to initialize a BiGRU network and perform vendor identification on an LR dataset.

Despite its success, this methodology faces several challenges. For instance, authorship attribution approaches can be applied to darknet markets, as the darknet ads are posted under anonymous vendor handles (user names). However, extending these approaches to settings without ground truth is more difficult. This is particularly the case in online escort platforms, where ads lack user handles, making it challenging to verify an ad's author directly. Additionally, since the selected darknet ads do not exhibit signs of human trafficking activities, it remains unclear whether the same approach can be effectively applied to online platforms facilitating human trafficking operations. Further investigation is needed to evaluate its applicability in this context. Another challenge, as highlighted in Section 2.6, is the resource-intensive nature of the text-similarity task in VendorLink. Moreover, the vendor verification task, which relies on similarity analysis, requires formalization and proper evaluation to assess the capabilities of the trained models. To address these gaps, this chapter extends the VendorLink methodology to perform authorship attribution on online ads from Backpage Escort markets.

## 3.1 INTRODUCTION

Human trafficking is a global crime that exploits vulnerable individuals for profit, affecting people of all ages and genders (Europol, 2021; GLOTIP, 2024). One of its most prevalent forms is sex trafficking, which involves coercing victims into commercial sex through violence, threats, deception, or debt bondage. Such activities frequently occur in venues like massage parlors, brothels, strip clubs, and hotels (Europol, 2021). Women and girls represent a significant proportion of victims, particularly within the commercial sex industry. While many online escort ads involve trafficking victims, these individuals typically have no control over the content of the ads. The overwhelming volume of such advertisements makes manual identification of trafficking cases impractical, resulting in numerous instances remaining undetected.

Researchers and LEAs rely on sex trafficking indicators (Ibanez and Suthers, 2014; Ibanez and Gazan, 2016c; Lugo-Graulich and Meyer, 2021) to identify advertisements linked to human trafficking. These investigations often require connecting ads to individuals or trafficking networks, typically using contact details such as phone numbers or email addresses.

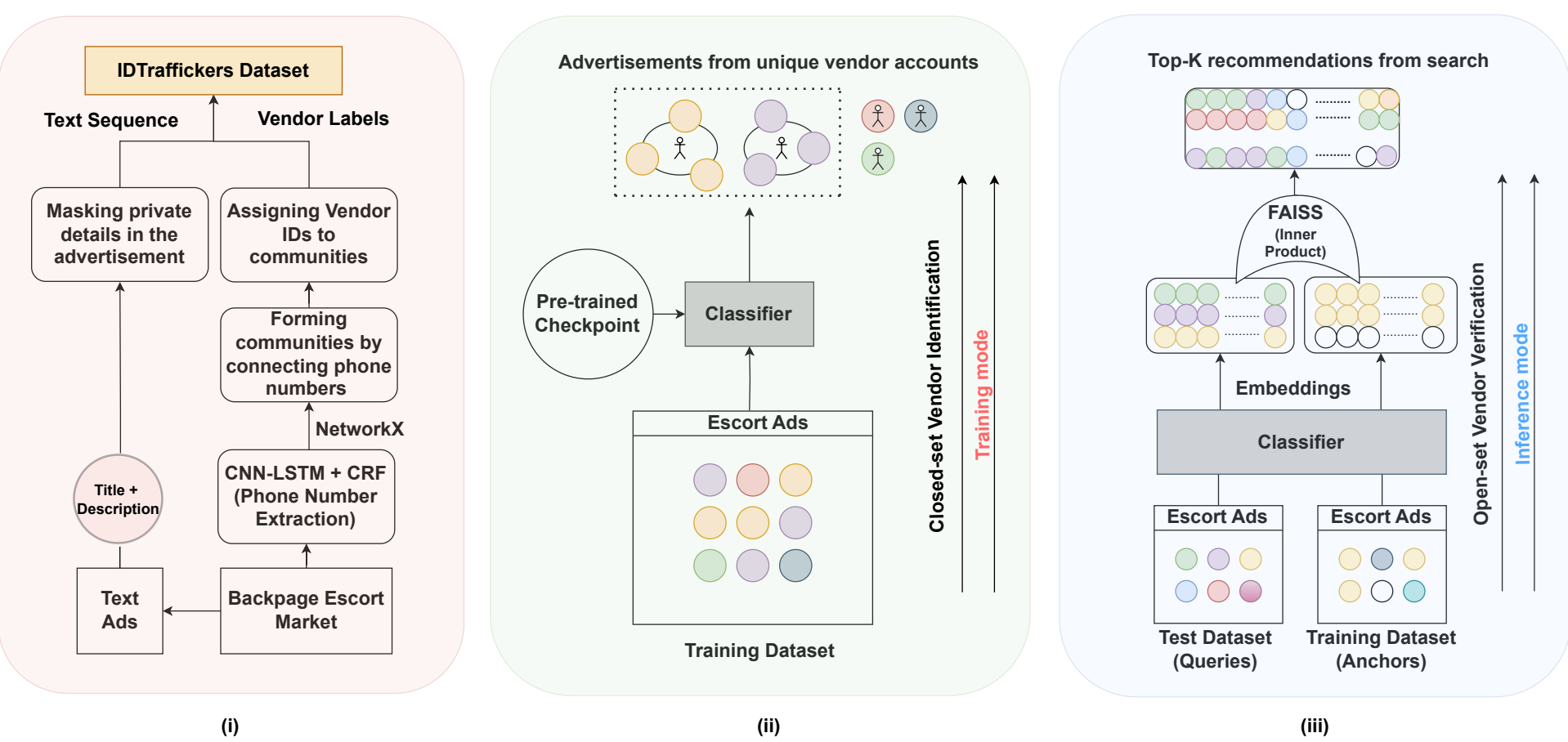

Figure 3.1: **(i) IDTraffickers:** Preparing authorship dataset from Backpage Escort Market, **(ii) Vendor Identification Task:** Identifying human trafficking vendors in closed-set environment, **(iii) Vendor Verification Task:** Verifying human trafficking vendors using similarity-search in open-set environment.

Analysis of the data in this study reveals that only 37% (202,439 out of 513,705) of ads contain such contact information, underscoring a significant limitation in current methods. Furthermore, manual detection of human trafficking cases is both resource-intensive and impractical due to the sheer volume of online escort advertisements. To address these challenges, researchers and LEAs are increasingly adopting automated systems powered by advanced techniques, including data analysis (Keskin et al., 2021), knowledge graphs (Szekely et al., 2015; Kejriwal and Szekely, 2022), network theory (Ibanez and Suthers, 2014, 2016; Kejriwal and Kapoor, 2019; Kosmas et al., 2023), and machine learning (Dubrawski et al., 2015; Portnoff et al., 2017; Tong et al., 2017; Stylianou et al., 2017; Alvari et al., 2017a; Shahrokh Esfahani et al., 2019; Wiriyakun and Kurutach, 2021; Wang et al., 2020a). A recent literature review by Dimas et al. (2022) provides a comprehensive summary of current research trends aimed at combating human trafficking across multiple fronts.

Although most of the aforementioned studies focus on online advertisements from the Backpage escort market, none have explored authorship features as a means to link and connect trafficking operations. Without identifiable information such as phone numbers, email addresses, or private identifiers, authorship analysis techniques can play a pivotal role in connecting vendor communities by examining the language, style, and content of escort ads. As illustrated in Figure 3.1, this study aims to bridge the gap between authorship techniques and human trafficking by introducing the following contributions [1].:

**(i) Authorship dataset:** This study introduces IDTraffickers, an AA dataset comprising 87.5K text ads collected from the U.S. Backpage escort market between December 2015 and April 2016 (Figure 3.1(i)). Analyzing the language and content of these escort ads provides critical insights into the authorship traits and patterns associated with trafficking operations. By developing AA approaches on this dataset, recurring patterns can be identified and utilized to link ads from potential human trafficking communities. This bridges the gap in identifying and connecting individuals or groups involved in trafficking operations.
**Note:** Establishing ground truth for escort ads associated with human trafficking is inherently challenging, as undetected crimes are difficult to identify. Even data derived from arrest records may not account for all undetected ads. To ensure the inclusion of human trafficking-related ads, this study analyzes ads from the Backpage escort website, which previous investigations have linked to human trafficking activities (Portman, 2017). The primary objective is not to directly identify human trafficking ads but to provide a tool that assists law enforcement in prioritizing resources by effectively detecting potential indicators of human trafficking-related content.

**(ii) Authorship Benchmarks:** On online escort markets, multiple vendors and communities often share a single account to post numerous ads. Additionally, some vendors create multiple accounts to evade LEAs' detection and expand their operations. To address these challenges, this study first establishes a vendor identification benchmark through a closed-setting text classification task (Figure 3.1(ii)). In this task, the objective of the classifier is to predict the vendor associated with a given advertisement among a set of known authors. Furthermore, building on the text-similarity task introduced in Chapter 2, the style representations from the trained classifier are utilized to formalize the vendor verification task through an open-setting similarity-search retrieval task (Figure 3.1(iii)). Given an advertisement from a parent vendor, the task involves retrieving all possible ads associated with that vendor.

[1] The code implementation is publicly available at the Github Repository – IDTraffickers.

## 3.2 RELATED RESEARCH

**Linking Escort Advertisements:** Identifying human trafficking operations in escort ads requires a multidisciplinary approach, combining expertise in linguistics, data analysis, machine learning, and collaboration with LEAs. Recent advancements in NLP have enabled researchers to identify trafficking indicators using entity linking approaches (Alvari et al., 2016; Nagpal et al., 2017; Alvari et al., 2017a; Whitney et al., 2018; Li et al., 2022b) for the automatic detection of human trafficking operations. However, these indicators can only be effectively studied within clusters of advertisements linked to individual vendor accounts. Previous work by Chambers et al. (2019) proposed using neural networks to extract phone numbers to connect escort advertisements. However, this study reveals that only 37% of the ads in the dataset contain phone numbers, limiting the applicability of such methods. While clustering approaches (Lee et al., 2021; Vajiac et al., 2023a; Nair et al., 2022b; Vajiac et al., 2023b) can help connect near-duplicate advertisements, they are ineffective in establishing connections between paraphrased or distinct ads. This study uses authorship techniques to analyze unique writing styles within escort advertisements and establish connections to individual vendors to address these limitations.

**Authorship Attribution for Combating Human Trafficking:** Previous research has established numerous ML approaches to analyze text styles and link distinctive writing characteristics to specific authors. As discussed in Section 2.2, these approaches include TF-IDF-based clustering and classification techniques, CNNs, RNNs, and various contextualized transformers with and without contrastive learning objectives. These advancements have led to the development of style representational approaches (Hay et al., 2020; Zhu and Jurgens, 2021; Wegmann et al., 2022), which currently represent the state-of-the-art (SOTA) for authorship tasks. Consequently, several datasets (Conneau and Kiela, 2018; Andrews and Bishop, 2019; Bevendorff et al., 2020, 2023) have been developed to support further research in this domain.

While numerous AA studies have been successfully applied to forensic investigations, cybercrime investigations, spam detection, and linking vendor accounts on darknet markets, none of the existing studies, except Ardakani (2020), focus on connecting human trafficking vendors through escort advertisements. In their work, Ardakani (2020) demonstrate how statistical models can apply authorship attribution to connect escort ads from Louisiana and nearby cities in the U.S. through a classification task. This chapter addresses this gap by introducing a novel dataset, IDTraffickers, highlighting the distinctions between language in existing authorship datasets and escort advertisements. Furthermore, it demonstrates the capabilities of AA approaches in establishing connections between escort ads and human trafficking vendors through vendor verification and identification tasks.

## 3.3 DATASET

The data in this study is collected from online escort advertisements posted on the Backpage Market between December 2015 and April 2016. Backpage, a classified ads website similar to Craigslist on the surface web, hosted listings ranging from apartments to escort services. However, a report by Hackler (2016) indicates that 90% of Backpage's revenue was generated from adult advertisements. Additionally, Portman (2017) highlights that Backpage hosted escort listings associated with sex trafficking operations involving women and children across 943 locations, 97 countries, and 17 languages, making it a suitable dataset for this study. This study focuses on a dataset of 513,705 advertisements spanning 14 states and 41 cities in the United States. While the pre-processing and restructuring of the data for the authorship task were conducted as part of this study, acknowledgment is given to Bashpole Software, Inc. for providing the raw data. To access this data, please visit our GitHub or DataverseNL repositories and follow the instructions provided in the links.

To generate ground truth, i.e., vendor labels, vendor communities are created by extracting phone numbers from the text ads and applying NetworkX (Hagberg et al., 2008) to them. For example, an advertisement may contain multiple phone numbers. Each phone number is treated as a node in the network graph, and matches for these numbers are searched across all other advertisements. If a matched advertisement contains a single phone number, it is assigned to the same vendor community. Suppose the advertisement contains multiple phone numbers, including new numbers that are not in the existing graph. In that case, the knowledge graph is extended with these new nodes, and the matching process is iteratively repeated until all possible matches are identified. Algorithm 3.1 illustrates the equivalent pseudo code for the logic described above.

**Algorithm 3.1** Generating Vendor Communities via Phone Numbers

```
1: Input: Set of advertisements A = {a_1, a_2, ..., a_n}
2: Output: Vendor communities represented as a knowledge graph G = (V, E)
3: Initialize an empty graph G with nodes V and edges E
4: for each advertisement a_i ∈ A do
5:     Extract all phone numbers P_i = {p_1, p_2, ..., p_k} from a_i
6:     for each phone number p_j ∈ P_i do
7:         if p_j ∉ V then
8:             Add p_j to V                    // Add new node to the graph
9:         end if
10:    end for
11:    for each pair of phone numbers (p_j, p_l) ∈ P_i where j ≠ l do
12:        if (p_j, p_l) ∉ E then
13:            Add edge (p_j, p_l) to E // Connect phone numbers within the same ad
14:        end if
15:    end for
16: end for
17: for each phone number p_j ∈ V do
18:     Find all advertisements A_j containing p_j
19:     for each advertisement a_m ∈ A_j do
20:         Extract all phone numbers P_m from a_m
21:         for each phone number p_k ∈ P_m do
22:             if p_k ∉ V then
23:                 Add p_k to V          // Extend the graph with new nodes
24:             end if
25:             if (p_j, p_k) ∉ E then
26:                 Add edge (p_j, p_k) to E // Connect phone numbers across ads
27:             end if
28:         end for
29:     end for
30: end for
31: Return: Knowledge graph G representing vendor communities
```

However, not all phone numbers in the dataset can be directly extracted, as vendors on these escort markets often obscure phone numbers (e.g., (4 ☺ 2) 456 9412, threeoh2FOUR070six22, 6I5 093 93B6, etc.) and inject noise to avoid detection (Chambers et al., 2019). To address this, the TJBatchExtractor (Nagpal et al., 2017) and a CNN-BiLSTM classifier with a CRF head (detailed in Section 3.4.1), as proposed by Chambers et al. (2019), are employed to extract phone numbers from the ads. Following Chambers et al. (2019), the CNN-BiLSTM classifier with a CRF head is trained on a custom escort advertisement dataset (Chambers et al., 2019). This dataset contains 390 real-world noisy phone numbers and 100,000 adversarially formatted phone numbers with obfuscation techniques such as character substitution (e.g., 'I' for '1', 'O' for '0'), word substitution (e.g., "Too" for '2'), confounding characters (e.g., adding separators like '.', '-', '/' randomly), emojis (e.g., ☺ for 0), and Unicode characters (e.g., uncommon symbols).

| Geography | Advertisements | Vendors |
|---|---|---|
| **East** | 24,000 | 5,029 |
| **West** | 22,556 | 2,576 |
| **North** | 3,124 | 254 |
| **South** | 27,871 | 2,291 |
| **Central** | 21,124 | 2,928 |
| **Overall** | 87,595 | 5,244 |

Table 3.1: Total number of unique advertisements and vendors across US geography.

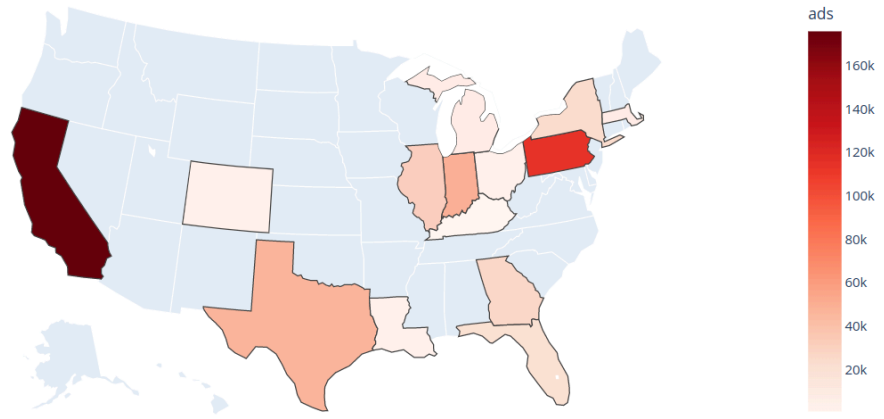

Figure 3.2: Density of unique advertisements collected across American states.

Each community is assigned a label ID as the vendor label. For evaluation purposes, ads without phone numbers are discarded, resulting in a dataset of 202,439 ads. Following the findings of Lee et al. (2021), which indicate that the average vendor of escort advertisements has 4-6 victims, entries from vendors with fewer than five ads (the average of 4-6) are removed. The final dataset comprises 87,595 unique advertisements and 5,244 vendor labels. Then, following the methodology in VendorLink, the title and description of the text ads are merged using the "[SEP]" token, as illustrated in Figure 3.1[i]. Figure 3.3 shows that most advertisements in the dataset (approximately 99%) have a sentence length below 512 tokens and 2,000 characters. Most vendors also have an ad frequency of under 500. Table 3.1 and Figure 3.2 illustrate the total number of unique text ads and vendors collected across the four geographical regions, 14 states, and 41 cities of the United States.

After generating the vendor labels, measures were taken to safeguard privacy by masking sensitive information within the advertisement descriptions. This included masking phone numbers, email addresses, age details, post IDs, dates, and links to ensure that none of these details could be reverse-engineered, thereby minimizing the potential misuse of the dataset. Despite efforts to extract escort names and location information using BERT-NER, RoBERTa-NER-based entity recognition techniques, and the approach described by (Li et al., 2022b), a significant number of false positives were encountered. Unfortunately, attempts to mask this information introduced additional noise into the data. As a result, this approach was ultimately abandoned.

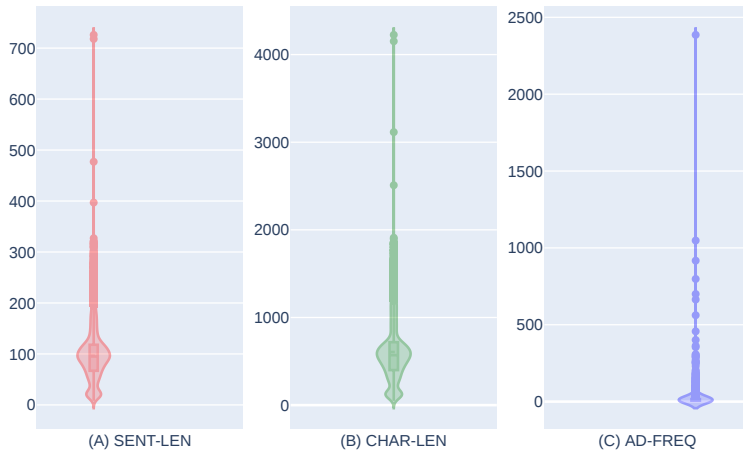

Figure 3.3: (A) Total number of tokens per ad (sentence length), (B) Total number of characters per ad, and (C) Number of ads per vendor (class frequency) distributions.

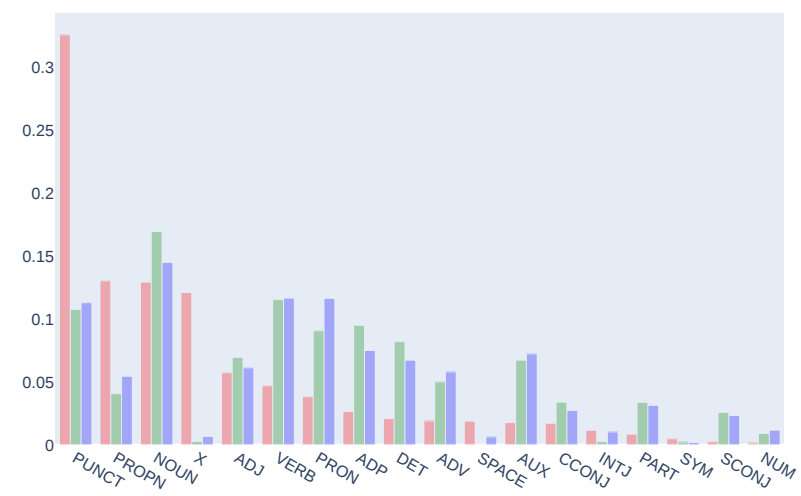

Figure 3.4: **POS-distribution:** Normalized POS-distribution for IDtraffickers, PAN2023, and Reddit-Conversations datasets.

Figure 3.4 presents a comparative analysis of the part-of-speech (POS) distributions across three datasets. The results indicate that the IDTraffickers dataset exhibits a higher frequency of punctuations, emojis, white spaces, proper nouns, and numbers than the other datasets. Punctuations, emojis, white spaces, and random characters collectively account for approximately 47% of all IDTraffickers dataset POS tags. In contrast, these tags represent only 10.6% and 12.4% of all tags in the PAN2023 and Reddit conversation datasets, respectively. This discrepancy underscores the significant noise in the IDTraffickers dataset, emphasizing the need for fine-tuning to achieve effective domain adaptation.

In addition to examining POS distributions, the wikifiability—defined as the presence of entities with corresponding Wikipedia mentions—is investigated on a per-advertisement basis. Figure 3.5 provides insights into the wikifiability across three datasets: IDTraffickers, PAN2023, and Reddit Conversational. Notably, the IDTraffickers dataset exhibits higher wikifiability than the PAN2023 and Reddit Conversational datasets. However, a closer examination in Figure 3.6 reveals that most recognized entities in the IDTraffickers dataset are primarily related to locations, escort names, or organizations. This observation aligns with the nature of the advertisements, as they often include details such as the posting's location, the escort's name, and nearby landmarks.

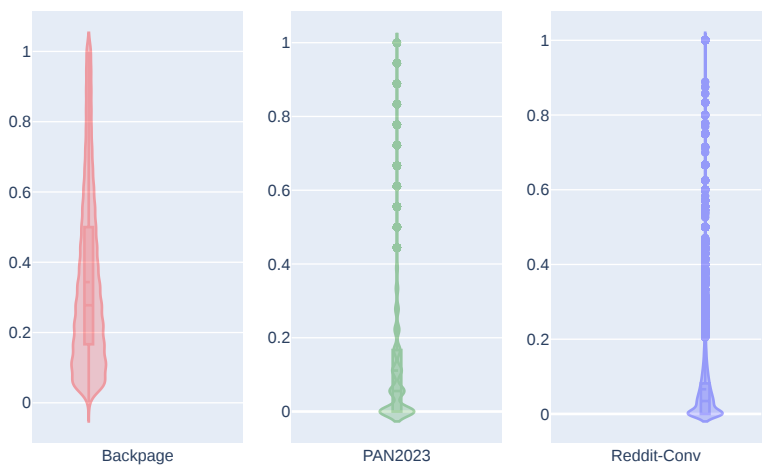

Figure 3.5: **Wikifiability:** No. of entities per advertisement with Wikipedia mentions in the IDtraffickers, PAN2023, and Reddit-Conversations datasets.

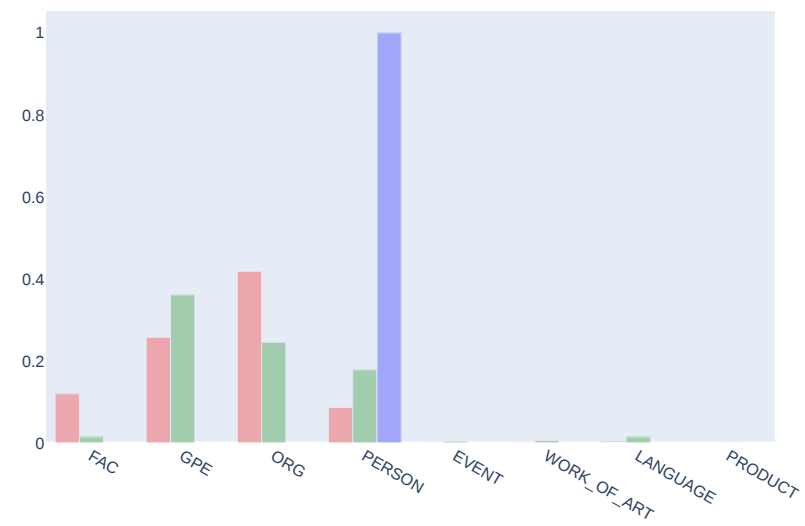

Figure 3.6: **Wiki-entities-distribution:** Extracted entities from the wikification of IDtraffickers, PAN2023, and Reddit-Conversations datasets.

**Language Distribution:** Our analysis reveals that approximately 99.32% of the dataset's vocabulary is English. The remaining portion consists of other languages in descending order: Spanish, Chinese, Japanese, Sotho, French, Bokmal, Dutch, Tswana, Nynorsk, Swedish, Latin, Basque, Swahili, Malay, Xhosa, and Welsh. Given that only a small fraction of the dataset's vocabulary is non-English, the use of multilingual models was not explored in this study. It is anticipated that employing multilingual models would not yield significant performance improvements due to the overwhelming dominance of English in the dataset. These statistics were obtained using the lingua language detection Python package.

## 3.4 EXPERIMENTS

### 3.4.1 GENERATING GROUND TRUTH: EXTRACTING PHONE NUMBERS

To extract phone numbers, we first employ TJBatchExtractor (Nagpal et al., 2017), a tool that utilizes regular expressions and rule-based in-

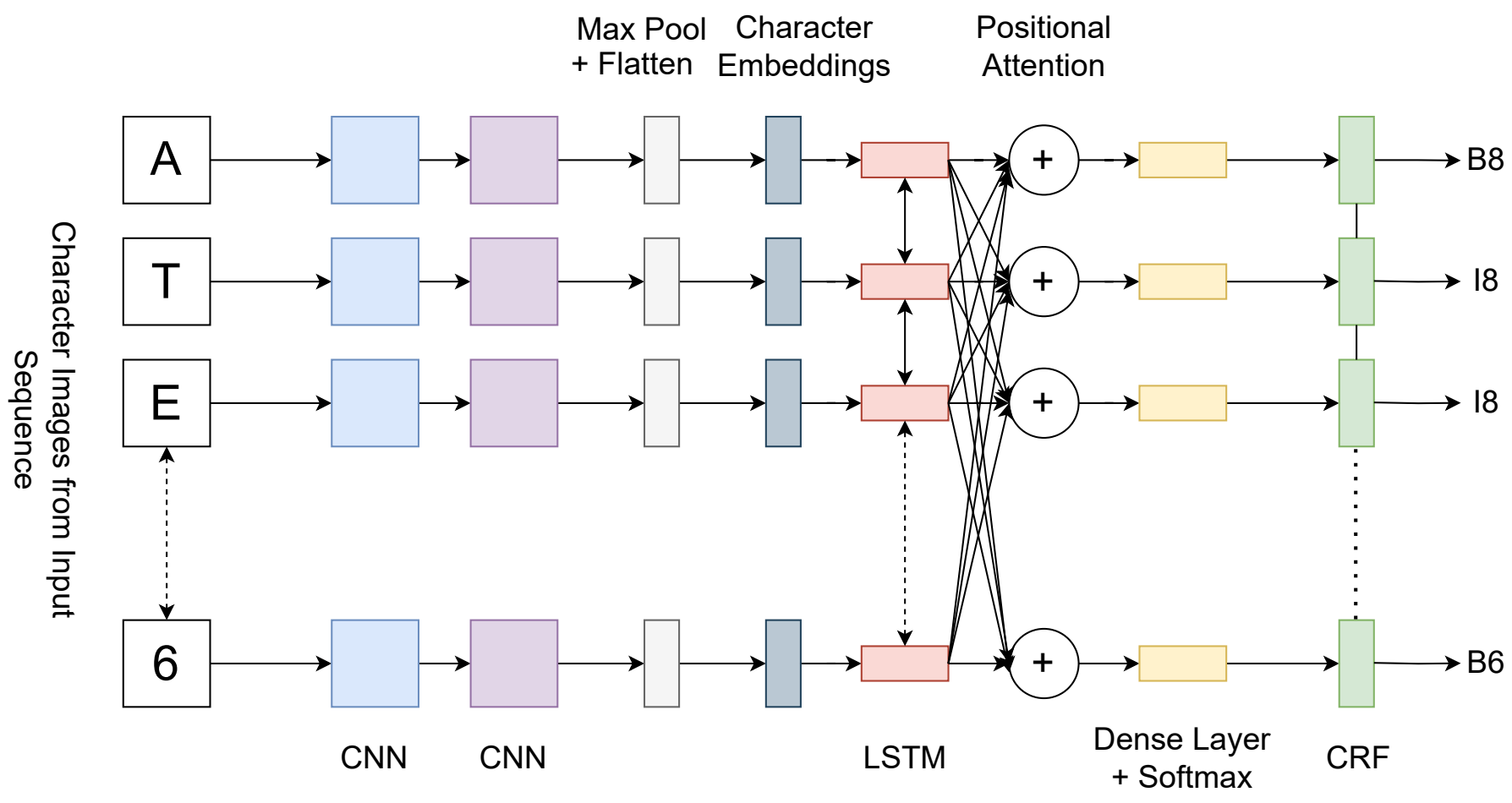

Figure 3.7: Architecture of CNN-BiLSTM classifier with CRF heads for extracting phone numbers from noisy escort ads (Chambers et al., 2019).

3

formation extraction to perform text processing, pattern matching, and feature extraction. For advertisements that do not yield any results from TJBatchExtractor, this study adopts the approach proposed by (Chambers et al., 2019), applying BiLSTM, CNN-BiLSTM classifiers with a CRF head. As detailed in Section 3.5.1, the CNN-BiLSTM classifier with a CRF head performs the best among the three models. This model combines a Convolutional Neural Network (CNN), a Bidirectional Long Short-Term Memory (BiLSTM) network, and a Conditional Random Field (CRF) layer. It is trained on a dataset comprising real-world noisy phone numbers and adversarially generated phone numbers, as described in (Chambers et al., 2019). The architecture of the classifier is illustrated in Figure 3.7.

In the CNN-BiLSTM-CRF model, a text sequence of 70 characters is first encoded at the character level from the escort description to capture obfuscation patterns. For sequences exceeding 70 characters, a sliding window approach is employed, combining 25 characters from the previous window with 45 new ones. The model processes each sequence separately, and predictions are filtered to retain only numbers between 7 and 11 characters in length, corresponding to a 1-digit country code, a 3-digit area code, and a 7-digit telephone number. The CNN processes each sequence, detecting local character transformations such as "O" → "0" or "five" → "5", making it particularly effective for handling adversarial text manipulations. Following this, a Bidirectional LSTM (BiLSTM) captures contextual dependencies, enabling the model to understand long-range relationships between characters and recognize phone number structures even when digits are scattered. Unlike a standard LSTM, which processes text in a single direction, the BiLSTM analyzes forward and backward sequences, enhancing sequence comprehension. Finally, the Conditional Random Field (CRF) layer enforces structured predictions, ensuring coherent number extraction rather than independent character predictions. Using BIO tagging, the CRF assigns labels such as B-PHONE (beginning of a phone number), I-PHONE (inside a phone number), and O (outside any number), preventing errors like partial extraction. Combining these CNN, BiLSTM, and CRF units enhances robustness against adversarial obfuscation while preserving contextual and sequential information.

To evaluate the accuracy of the three classifier models in predicting phone numbers, this study follows (Chambers et al., 2019) and explores levenshtein, perfect, and digit accuracy metrics. Levenshtein accuracy measures the similarity between the predicted phone number and the ground truth by computing the levenshtein distance, which represents the minimum insertions, deletions, or substitutions required to transform one phone number into another. A lower levenshtein distance indicates a more accurate prediction. Perfect accuracy assesses whether the predicted phone number exactly matches the ground truth; even a single incorrect digit results in a score of zero for that instance. This metric strictly evaluates the model's ability to extract phone numbers without errors. Finally, digit accuracy measures the proportion of correctly predicted digits across all numbers, irrespective of their position in the sequence. This metric allows for partial credit when the model correctly predicts some digits but fails to reconstruct the entire phone number. Together, these three metrics provide a comprehensive assessment, balancing strict correctness (perfect accuracy), overall similarity (levenshtein accuracy), and partial correctness (digit accuracy).

### 3.4.2 VENDOR IDENTIFICATION: A CLASSIFICATION TASK

Consistent with the insights presented in Chapter 2, existing research has demonstrated that transformers-based contextualized models surpass traditional stylometric methods, statistical TF-IDF approaches, and conventional RNNs and CNNs in AA tasks (Kumar et al., 2020; Fabien et al., 2020b; Ai et al., 2022). Building on these findings, this study explores closed-set classification using distilled versions of several state-of-the-art language models, including BERT-cased, RoBERTa-base, GPT-2 (Sanh et al., 2019b), and MiniLM (Wang et al., 2020b), smaller variants of AlBERTa (Lan et al., 2019) and DeBERTa-v3 (He et al., 2021), as well as transformers trained on contrastive learning objectives, such as Style Embedding (Wegmann et al., 2022) and DeCLUTR (Giorgi et al., 2021). To address domain-specific variations, a RoBERTa-base backbone is fine-tuned on the IDTraffickers ads for the masked language modeling task. Sentence representations are subsequently extracted from the trained Language Model (LM), and mean-pooling is applied for the closed-set classification task [2]. This model is referred to as the LM-Classifier in the study.

The models are evaluated using several metrics, including balanced accuracy, micro-F1, weighted-F1, and macro-F1 scores. Given the class imbalance present in the dataset (Figure 3.3(C)), performance emphasis is placed on the macro-F1 score to ensure robust performance across all classes. The experimental framework and evaluation metrics are designed to comprehensively assess the effectiveness of the proposed methods in addressing the challenges of AA in imbalanced datasets.

### 3.4.3 VENDOR VERIFICATION: A RETRIEVAL TASK

The closed-set classifier effectively identifies known vendors presented to it during training. However, it cannot handle inference for unknown vendors, a critical limitation in dynamic environments such as online escort advertisements. Given the daily frequency of new ads and the emergence of new vendors, it becomes impractical for LEAs to retrain the network repeatedly. To address this limitation, the trained classifier is leveraged to extract mean-pooled style representations from the ads, capturing the unique stylistic features of each vendor. These representations are then utilized for similarity search using FAISS (Douze et al., 2024), a highly efficient library designed for fast nearest-neighbor search in high-dimensional spaces.

The FAISS similarity-search process efficiently applies k-means clustering to the style representations, which helps partition the data into clusters and identify the nearest neighbors. In this setup, the test set ads serve as query documents, while the training set ads act as index documents.

[2] Multiple pooling strategies, including mean, max, and mean-max pooling, are experimented with for both author identification and verification tasks. However, the mean-pooling strategy yielded the best results.

For each query document in the test set, cosine similarity is employed between the query and the index documents. This process identifies the K closest index documents from the training set for each query, effectively ranking them based on stylistic similarity. Treating the vendor verification task as a retrieval problem allows for identifying potential matches for all the vendors in or outside the training set.

To evaluate the effectiveness of this similarity search operation, several metrics are employed, each providing a different perspective on the retrieval performance (Beitzel et al., 2009; Sujatha and Dhavachelvan, 2011; Jin et al., 2021):

- **Precision@K** measures the proportion of relevant ads among the top K retrieved results. In the context of this study, it reflects how many of the top K retrieved ads belong to the same vendor as the query ad. A high Precision@K indicates that the model effectively retrieves relevant matches within the top K results.
- **Recall@K** evaluates the ability of a model to retrieve all relevant ads within the top K results. It measures the fraction of relevant ads (i.e., ads from the same vendor) successfully retrieved from all relevant ads in the dataset. A high Recall@K suggests that the model effectively captures a large portion of the relevant matches.
- **Mean Average Precision@K (MAP@K)** provides a quality measure across recall levels by averaging the precision scores at each rank where a relevant ad is retrieved. It is particularly useful for evaluating ranking tasks, as it balances both precision and recall. A higher MAP@K indicates better overall retrieval performance, with relevant ads ranked higher in the results.
- Average R-Precision (**R-Precision@X**) calculates the precision at X, where X is the total number of relevant ads for a given query ad. It is particularly useful when the number of relevant ads varies across vendors (class frequency). The average R-Precision score aggregates this measure across all queries, providing a robust evaluation of the model's ability to retrieve relevant ads proportionally to the total number of relevant items.

### 3.4.4 INFRASTRUCTURE & SCHEDULE

**Split Ratio:** The dataset is divided into training, validation, and test sets using a standard ratio of 0.75:0.05:0.20 for all experiments. To ensure reproducibility, the seed parameter is set to 1111 during the splitting process.

**Training:** Network training and evaluation are performed on a Tesla V100 GPU with 32 GB of memory. The Adam optimizer is used for network optimization with specific parameter configurations. The values for $\beta1$ and $\beta2$ are set to 0.9 and 0.999, respectively. An L2 weight decay of 0.01 is included in the optimization process. The optimal learning rate is determined by experimenting with 0.01, 0.001, and 0.0001 values. Upon analysis, the best performance is observed with a learning rate of 0.001. Finally, a warm-up strategy is applied for the first 100 steps, followed by linear decay.

**Architectures & Hyperparameters:** Given the available computational resources, the transformer models are initialized using pre-trained checkpoints from various architectures, including DistilBERT-base-cased, DistilRoBERTa-base, DistilGPT2, all-MiniLM-L6-v2, AlBERT-small, DeBERTa-v3-small, T5-small, DeCLUTR-small, and Style-Embedding. A sequence classification head is added on top of each of these models, and all models are trained until convergence using a batch size of 32. Though experiments with lower batch sizes of 8, 16, and 20 were also conducted, batch size 32 provided the best results. Increasing batch size over 32 causes memory issues during training.

All experiments in this study are implemented in Python 3.9 (Van Rossum and Drake Jr, 1995) using the Scikit-learn (Pedregosa et al., 2011b), PyTorch (Paszke et al., 2019a), Hugging Face (Wolf et al., 2020), and Lightning 2.0 (Falcon and The PyTorch Lightning team, 2019) frameworks.

## 3.5 RESULTS

### 3.5.1 PHONE NUMBER EXTRACTION

Consistent with the findings in (Chambers et al., 2019), Table 3.2 shows that the CNN-BiLSTM classifier with CRF heads slightly outperforms both the simple BiLSTM and BiLSTM classifiers with CRF heads. Since the primary focus is not on processing phone numbers directly, this study prioritizes the consistency of extracted phone numbers rather than their accuracy or quality. The trained models are executed five times at inference, and the extracted phone numbers remain consistent across all iterations. Figure 3.8 presents examples of extracted phone numbers (artificially gen-erated to maintain privacy) alongside the noisy phone numbers found in the text descriptions of escort ads.

| Model | Levenshtein Acc. | Perfect Acc. | Digit Acc. |
|---|---|---|---|
| **BiLSTM** | 0.9951 | 0.9670 | 0.9940 |
| **BiLSTM with CRF heads** | 0.9969 | 0.9750 | 0.9885 |
| **CNN-BiLSTM with CRF heads** | 0.9986 | 0.9892 | 0.9950 |

Table 3.2: Accuracy of classifiers in extracting phone numbers on the combined real-world and adversarially formatted escort ads.

| Advertisements | Predictions |
|---|---|
| e down 9S147 twentynine8 two lk 4 ts , Masc for Masc - m4m (Garland) | 9514729824 |
| QR Code Link to This Post Hey there! Hello hello, 4076 o 59779 looki | 4076059779 |
| o This Post Quicky.. Ready to please ..562Want to 6184 589two 36 cu | 6184589236 |
| e 9seventrEe ochoO45_7to0juan ink to This Post I'm Latin 22, 5'8 I'm | 9738045721 |
| de Link to This Post We are looking fo nue\ J e 398#88 8653 c top here | 9398888653 |
| for a fun experience !!! - mw4w 4 3$28cinco 8256 me...White tres sucke | 4328582563 |
| ooking t 92 --59OCHO 252 \&83 ink to This Post Im looking por a loca | 9259825283 |
| his Post This ad is 100% serious. I'm look 3b088 5&week:0Fore1S Cou | 3608850415 |
| ght g REAL 7eight 6tree37 18SICKStree ant Deepthroat BJ/Fuck Tight Ass | 7863371863 |
| QR Code Link to This Post We aee feeling nau 46forty-four9o13tres8 R | 4644901338 |
| 4m (Be a con Hill/Boston Co 3235-5 gorgeous 🌹 MARIAH 1eightthree $76 fet | 3235518376 |
| oking 4 younger, Sexy p (646)8573fiv&o4 our big natural cu r ves - m4w ( | 6468573504 |
| an {\}Ine39six35 5013 Link to Thls Post Hi! I am visiting Miami o | 9396355013 |
| en! - mw4m (DuPont Clrcl 3 2 5 8 tree seis ൭ \.7 💦💦👅 eighteen IGHT | 3258369718 |
| weekend siete seventy-five2 00nuevetoo0 2 QR Code Link to This Post | 7752009202 |
| your body anywhere 5b38s\/{\}nine5sicks8 svn R Code Link to This Pos | 5638795687 |

Figure 3.8: Extracted phone numbers by the CNN-BiLSTM classifier with CRF head, evaluated on the artificial dataset (Chambers et al., 2019).

### 3.5.2 CLOSED-SET CLASSIFICATION TASK

Table 3.3 presents the performance of the trained classifier baselines evaluated on the IDTraffickers test dataset. The DeCLUTR-small classifier achieves the highest performance for the vendor identification task. Since the success of the Style-Embedding model stems from its ability to capture content-based correlations across various topics, due to similar content in escort ads, the model struggles to effectively handle the noise in the data and differentiate between distinct writing styles. In contrast, DeCLUTR, trained to generate universal sentence representations, excels at identifying stylometric patterns and associating them with specific vendors. Additionally, the language model (LM) is trained until convergence, reaching a perplexity of 6.22. Compared to the end-to-end DeCLUTR baseline (fine-tuned over an existing checkpoint), training a classifier over the LM's learned representations results in only a marginal ~1% performance improvement. The high perplexity of the LM suggests its inability to adapt to the language in escort ads due to the high noise in the dataset. Given this minimal performance gain, further LM training is considered unneces-sary. As a result, the DeCLUTR architecture fine-tuned from a pre-trained existing checkpoint is established as the benchmark for vendor identification on the IDTraffickers dataset.

| Models | Acc. | Micro-F1 | Weighted-F1 | Macro-F1 |
|---|---|---|---|---|
| **Distilled Models** | | | | |
| BERT | 0.9110 | 0.9147 | 0.9143 | 0.8467 |
| miniLM | 0.8888 | 0.8934 | 0.8935 | 0.8101 |
| RoBERTa | 0.9199 | 0.9230 | 0.9229 | 0.8603 |
| GPT2 | 0.9132 | 0.9172 | 0.9166 | 0.8500 |
| **Smaller Models** | | | | |
| ALBERT | 0.7832 | 0.7891 | 0.7925 | 0.6596 |
| DeBERTa-v3 | 0.8703 | 0.8757 | 0.8756 | 0.7825 |
| T5 | 0.9157 | 0.9192 | 0.9190 | 0.8535 |
| **Contrastive Learning Models** | | | | |
| **DeCLUTR** | **0.9230** | **0.9261** | **0.9259** | **0.8656** |
| Style-Emb | 0.8887 | 0.8936 | 0.8932 | 0.8112 |
| **HT Language Model** | | | | |
| LM-Classifier | 0.9294 | 0.9317 | 0.9316 | 0.8726 |

Table 3.3: Balanced Accuracy, Micro-F1, Weighted-F1, and Macro-F1 performances of the transformers-based classifiers on the author identification task.

| Models | Param | Time | Epoch |
|---|---|---|---|
| **DistilBERT** | 66M | 25:44:00 | 63 |
| **DistilRoBERTa** | 82M | 11:13:05 | 27 |
| **DistilGPT2** | 82M | 25:45:00 | 82 |
| **all-miniLM-L6** | 22M | 21:50:04 | 62 |
| **ALBERT-small-v2** | 11.6 | 16:08:52 | 44 |
| **DeBERTa-v3-small** | 44M | 22:13:50 | 16 |
| **T5-small** | 35M | 19:03:35 | 50 |
| **DeCLUTR-small** | 86M | 05:53:33 | 26 |
| **Style-Embedding** | 128M | 18:43:06 | 36 |
| **LM-Classifier** | 86M | 04:47:10 | 19 |

Table 3.4: Total number of trainable parameters, training time, and convergence epoch for the trained classifiers.

| Seed | Acc. | Micro-F1 | Weighted-F1 | Macro-F1 |
|---|---|---|---|---|
| **100** | 0.9256 | 0.9285 | 0.9286 | 0.8694 |
| **500** | 0.9243 | 0.9283 | 0.9286 | 0.8677 |
| **1000** | 0.9172 | 0.9205 | 0.9204 | 0.8559 |
| **1111** | 0.9230 | 0.9261 | 0.9259 | 0.8656 |
| **5000** | 0.9197 | 0.9229 | 0.9229 | 0.8598 |
| **Mean** | 0.9219 | 0.9252 | 0.9252 | 0.8636 |
| **Std.** | 0.0030 | 0.0031 | 0.0032 | 0.0050 |

Table 3.5: Influence of different initialization on DeCLUTR-small classifier's performance.

Table 3.4 details the number of trainable parameters, training time, and epochs of convergence for all classifiers trained in our experiments. While training the LM-Classifier on the author identification task requires the least time, training the language model on our dataset takes 65 epochs and 2 days, 7 hours, and 20 minutes. Table 3.5 displays the mean and standard deviation of the model's performance across balanced accuracy, Micro-F1, Weighted-F1, and Macro-F1 scores. Due to resource constraints, the effects of different initializations on the model's performance are examined only for the established DeCLUTR-small benchmark. The results indicate minimal to no impact on these scores across different initializations. Finally, Figure 3.9 includes additional training details, such as training and validation loss, balanced accuracy, Micro-F1, Weighted-F1, and Macro-F1 scores on the validation set, to ensure reproducibility. These results are generated by exporting reports from Wandb (Biewald, 2020).

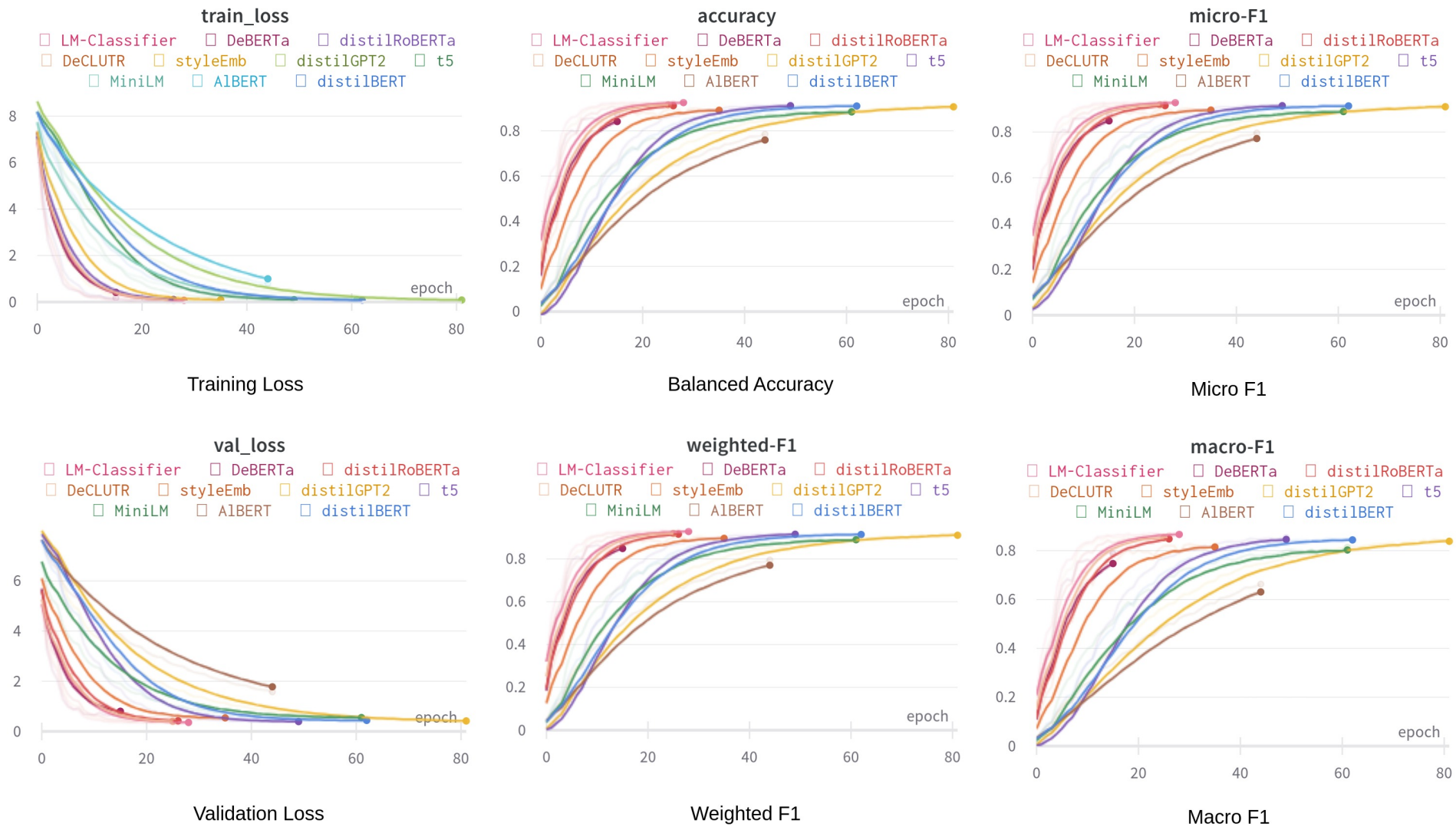

Figure 3.9: Training loss, validation loss, and performance of trained classifiers on the validation dataset.

### 3.5.3 OPEN-SET RETRIEVAL TASK

Table 3.6 presents the open-set author verification results for the DeCLUTR and Style-Embedding models, which achieved the highest performance among the vendor identification baselines. The evaluation compares model performance before and after training on the IDTraffickers dataset. In this experiment, a similarity search is performed on the training dataset for each query in the test set, and the results reflect the average performance across all queries, along with the standard deviation across vendor classes. A significant performance gap is observed between the models before and after training, highlighting the importance of domain-specific adaptation. The zero-shot performance of the pre-trained checkpoints, shown in red, proves inconclusive for the author verification task, reinforcing the necessity of fine-tuning on the IDTraf-fickers dataset to improve retrieval effectiveness.

| K | @1 | @3 | @5 | @10 | @20 | @25 | @50 | @100 | @X |
|---|---|---|---|---|---|---|---|---|---|
| Precision@$K$ | | | | | | | | | |
| Style | 0.0442 ± 0.20 | 0.0410 ± 0.16 | 0.0391 ± 0.15 | 0.0366 ± 0.13 | 0.0329 ± 0.11 | 0.0319 ± 0.10 | 0.0270 ± 0.08 | 0.0227 ± 0.07 | - |
| DeCLUTR | 0.3198 ± 0.46 | 0.2883 ± 0.39 | 0.2671 ± 0.36 | 0.2278 ± 0.32 | 0.1837 ± 0.27 | 0.1693 ± 0.26 | 0.1277 ± 0.21 | 0.0893 ± 0.15 | - |
| Style | 0.9616 ± 0.19 | 0.9437 ± 0.19 | 0.9124 ± 0.21 | 0.8175 ± 0.27 | 0.6818 ± 0.33 | 0.6328 ± 0.35 | 0.4815 ± 0.36 | 0.3551 ± 0.36 | - |
| DeCLUTR | 0.9672 ± 0.17 | 0.9532 ± 0.17 | 0.9221 ± 0.19 | 0.8253 ± 0.26 | 0.6868 ± 0.33 | 0.6367 ± 0.34 | 0.4835 ± 0.36 | 0.3561 ± 0.36 | - |
| Recall@$K$ | | | | | | | | | |
| Style | 0.0023 ± 0.01 | 0.0063 ± 0.04 | 0.0091 ± 0.05 | 0.0146 ± 0.07 | 0.0233 ± 0.09 | 0.0269 ± 0.10 | 0.0394 ± 0.12 | 0.0580 ± 0.15 | - |
| DeCLUTR | 0.0242 ± 0.06 | 0.0567 ± 0.12 | 0.0792 ± 0.16 | 0.1136 ± 0.20 | 0.1539 ± 0.24 | 0.1676 ± 0.25 | 0.2122 ± 0.29 | 0.2590 ± 0.31 | - |
| Style | 0.0828 ± 0.09 | 0.2348 ± 0.24 | 0.3485 ± 0.32 | 0.5092 ± 0.37 | 0.6552 ± 0.37 | 0.6945 ± 0.36 | 0.7909 ± 0.32 | 0.8600 ± 0.27 | - |
| DeCLUTR | 0.0836 ± 0.09 | 0.2397 ± 0.25 | 0.3563 ± 0.32 | 0.5192 ± 0.37 | 0.6653 ± 0.37 | 0.7041 ± 0.36 | 0.7988 ± 0.32 | 0.8664 ± 0.27 | - |
| MAP@K | | | | | | | | | |
| Style | 0.0442 ± 0.20 | 0.0562 ± 0.21 | 0.0598 ± 0.21 | 0.0640 ± 0.21 | 0.0673 ± 0.21 | 0.0681 ± 0.21 | 0.0700 ± 0.21 | 0.0712 ± 0.21 | - |
| DeCLUTR | 0.3198 ± 0.46 | 0.3587 ± 0.45 | 0.3681 ± 0.45 | 0.3750 ± 0.44 | 0.3794 ± 0.44 | 0.3803 ± 0.44 | 0.3823 ± 0.44 | 0.3833 ± 0.44 | - |
| Style | 0.9616 ± 0.19 | 0.9687 ± 0.16 | 0.9698 ± 0.15 | 0.9706 ± 0.15 | 0.9709 ± 0.14 | 0.9710 ± 0.14 | 0.9710 ± 0.14 | 0.9710 ± 0.14 | - |
| DeCLUTR | 0.9672 ± 0.17 | 0.9735 ± 0.14 | 0.9746 ± 0.14 | 0.9752 ± 0.13 | 0.9755 ± 0.13 | 0.9755 ± 0.13 | 0.9756 ± 0.13 | 0.9756 ± 0.13 | - |
| R-Precision@$X$ | | | | | | | | | |
| Style | - | - | - | - | - | - | - | - | 0.0199 ± 0.07 |
| DeCLUTR | - | - | - | - | - | - | - | - | 0.1641 ± 0.23 |
| Style | - | - | - | - | - | - | - | - | 0.8601 ± 0.22 |
| DeCLUTR | - | - | - | - | - | - | - | - | 0.8850 ± 0.20 |

Table 3.6: Precision@$K$, Recall@$K$, MAP@$K$, and R-Precision@$X$ scores for the DeCLUTR and Style-Embedding models before and after being trained on the IDTraffickers dataset

As explained earlier, the Precision@$K$ metric evaluates the model's effectiveness in retrieving relevant vendor ads among the top-$K$ predictions, whereas Recall@$K$ assesses its ability to retrieve relevant ads from the entire set of relevant ads in the training dataset. A high precision at lower $K$ values indicates that the model successfully retrieves relevant ads within the top-ranked results. Since all vendors in the dataset have at least five ads, precision performance is primarily examined for $K \leq 5$. Conversely, an increasing Recall@$K$ at higher $K$ values suggests that the model retrieves more relevant ads from the total available set. Since every vendor in the dataset has at least five ads, recall performance is emphasized for $K \geq 5$. In addition to these metrics, the evaluation also includes MAP@$K$, which accounts for ranking retrieved items by computing the average precision across all relevant items within the top-$K$ results. This metric ensures that relevant ads appear in higher-ranked positions. The results indicate that the retrieval system prioritizes ads from the same vendor when presented with a query advertisement from vendor A. The consistently high MAP scores across all $K$ values demonstrate the model's ability to rank retrieved ads effectively, ensuring that the most relevant ones are placed at the top.

The trained models are further evaluated using the R-Precision metric, which measures precision when the number of retrieved ads matches the number of relevant ads ($X$) for a vendor. This metric strictly focuses on retrieving relevant items while disregarding irrelevant ones. The results indicate that the DeCLUTR-small models achieve approximately 88% pre-cision, indicating that the model successfully retrieves 88% of all relevant ads among the top-ranked results. This demonstrates the effectiveness of the vendor verification setup in identifying and presenting the most relevant ads for a given query, achieving high precision at a rank equal to the number of relevant ads. Based on these findings, the DeCLUTR-small model is established as the benchmark for vendor verification task on the IDTraffickers dataset. However, the results exhibit substantial variation in standard deviation, suggesting that precision fluctuates depending on the vendor and the number of available ads. Further analysis confirms this observation, revealing lower Precision@$K$, Recall@$K$, MAP@$K$, and R-Precision@$X$ values for vendors with fewer ads.

### 3.5.4 MODEL EXPLANATIONS AND ERROR ANALYSIS

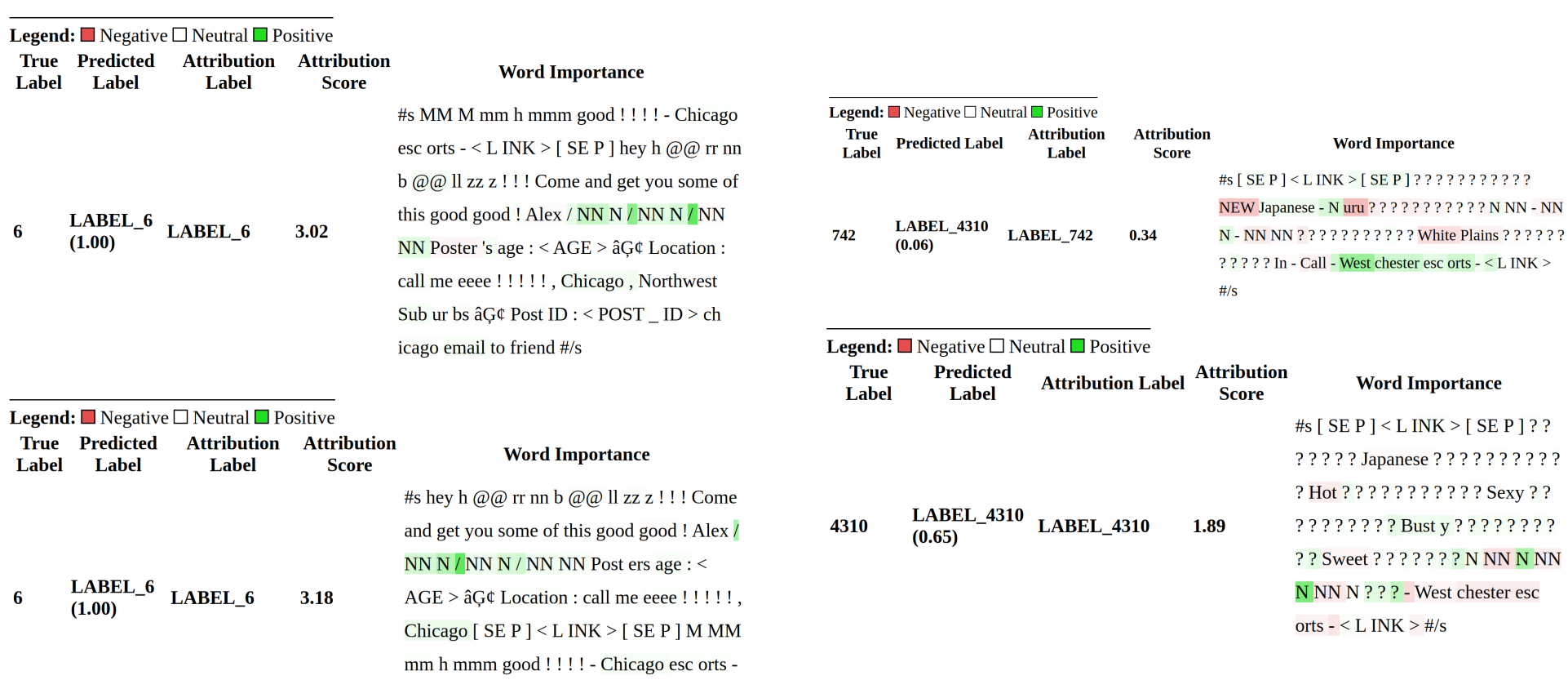

Figure 3.10: True-Positives model attributions for Vendor 6

Figure 3.11: False-Positives model attributions for Vendor 742

Figure 3.12: Model Explanations from the trained DeCLUTR-small classifier

To understand model learning, the transformers-interpret framework (Pierse, 2021), built on Captum (Kokhlikyan et al., 2020), is utilized to compute local word attributions in the text ads using the DeCLUTR-small classifier benchmark. This approach highlights the contribution of individual words to vendor classes (AA predictions). Figure 3.10 – 3.11 illustrates True Positive and False Positive predictions generated by the trained classifier. To compare model explanations, FAISS is employed to retrieve the most similar advertisement from the training dataset for each query in the test dataset. The figures present positive (in green), neutral (zero), and negative (in red) word attribution scores for queries and their corresponding anchor ads associated with the predicted vendor label.

The True Positive explanations reveal consistent writing patterns and word attributions, confirming that the retrieved ads originate from the same vendor. This observation is supported by the consistent use of "@" and "/" preceding the masked phone number digits "/NN N / NN N/ NN NN." In contrast, the False Positive explanations highlight instances where the model incorrectly assigned vendor labels, likely due to substantial content and writing style similarities between vendors 4310 and 742. Both vendors frequently included continuous sequences of "?" and references to Japanese services in their advertisements. Further analysis identifies cases where incorrect classifications stem from strong textual resemblances between ads. This finding underscores the necessity of critically assessing the reliability of classification labels. Vendor labels are assigned based on phone numbers extracted from ads, allowing connections between ads and forming vendor communities. In some cases, multiple phone numbers within an ad facilitated these connections, while in others, the absence of such information led to creating a new vendor label. The hallucinations observed in model explanations stem from this labeling process, suggesting the possibility that certain vendors may represent the same entity.

Finally, inspired by Rethmeier et al. (2020), Figure 3.13 leverages global feature attributions to analyze differences in writing styles between two specific vendors, 11178 and 11189. This analysis extracts word attributions and part-of-speech (POS) tags from all advertisements associated with both vendors. The resulting bar plot illustrates the normalized POS density within vendor ads, reflecting their grammatical structures and linguistic preferences. Simultaneously, scatter points highlight the two most attributed tokens for each POS tag linked to the vendor. The plots are generated using Plotly, which enables infinite zooming for any number of scatter points. However, the study displays only the two most attributed tokens for clarity and readability. The visualization reveals distinct grammatical structures and word attributions between the vendors, indicating they represent separate entities. This analysis can help provide researchers, LEAs, and practitioners with deeper insights into vendor connections without solely relying on ground truth labels.

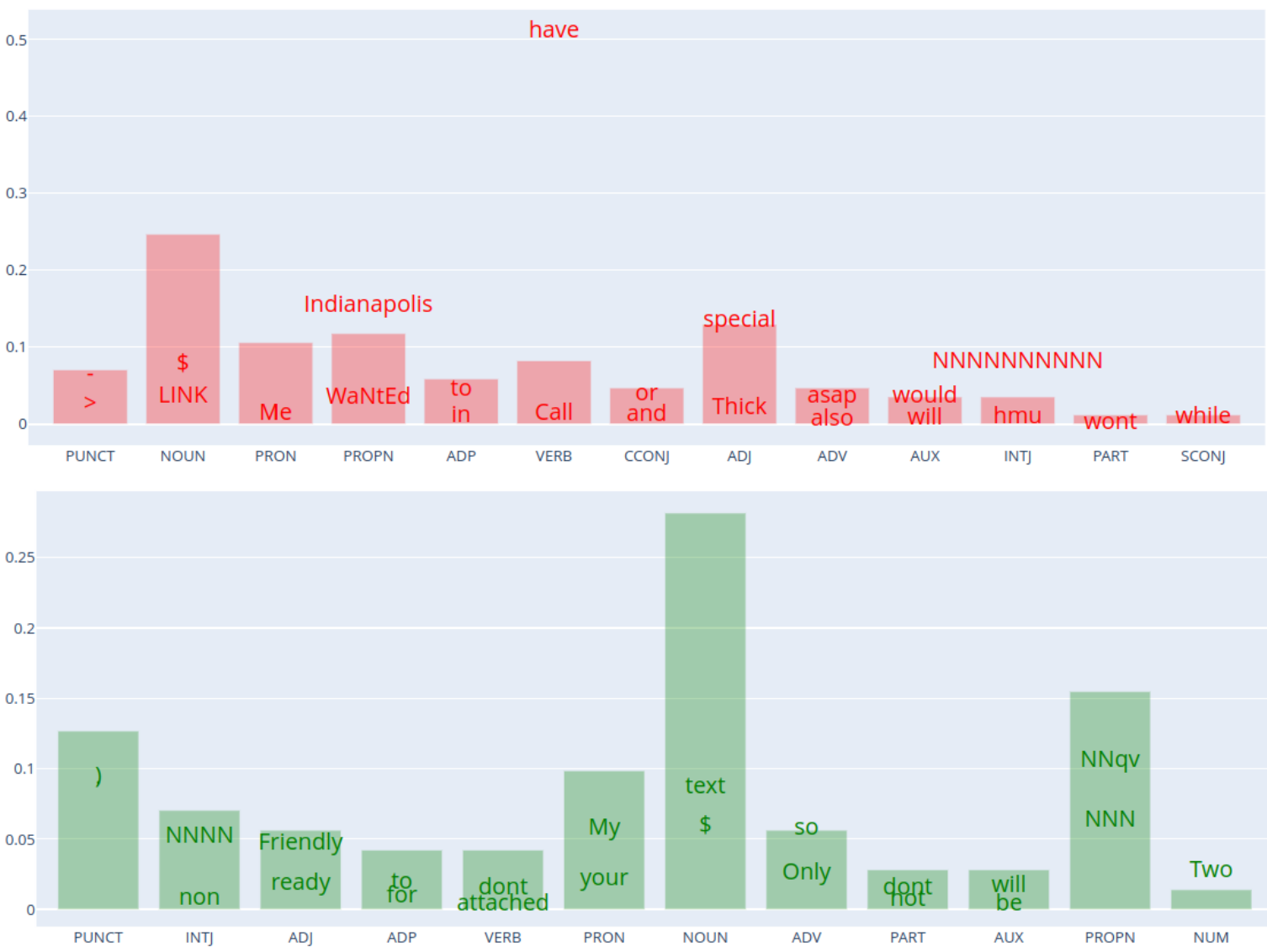

Figure 3.13: Word attribution over POS-distribution for ads of vendor 11178 and 11189.

## 3.6 Broader Discussion and Challenges

**Assumption:** This study assumes that each class label in the classification task corresponds to a distinct vendor, leveraging domain knowledge for adaptation. However, the error analysis reveals misclassifications due to strong writing style and content similarities, suggesting that some vendors may be the same entity. While establishing an absolute ground truth to confirm this hypothesis presents a significant challenge, improving vendor label quality would enhance the performance of the benchmarks.

**Dataset Age:** This study focuses on Backpage ads from December 2015 to April 2016, despite the existence of several active online escort platforms such as Craigslist, YesBackpage, and Classified Ads. The selection of Backpage stems from documented human trafficking activities on the platform (Portman, 2017). Other active websites are not included due to the lack of verified human trafficking information in their data. It is important to note that this research does not aim to track or assist in ongoing human trafficking investigations. Instead, it demonstrates the potential of AA techniques in aiding researchers, LEAs, and practitioners with resource prioritization. Given this objective, the dataset's age is considered a minor limitation.

**Scalability:** AA approaches rely on identifying distinct writing styles, yet human trafficking perpetrators span diverse racial, ethnic, and gender demographics (GLOTIP, 2024). As a result, generalizing the approach to vendors operating years apart presents challenges. While the methodology can theoretically extend to other escort websites, its effectiveness depends on several factors, including variations in writing styles, AI-generated text (e.g., chatGPT-like automated systems), language diversity, and data noise. These variables influence the model's ability to generalize beyond the Backpage dataset.

**Model Generalization:** The ranking task ensures that query ads are not drawn from the training dataset. However, the vendor identification classifier is trained on the same dataset, which limits insights into the model's zero-shot capabilities. To evaluate model generalization, further work assesses whether the model generates stylometric representations that remain effective across OOD datasets.

**Limited Modalities:** This study applies AA methodologies exclusively to text descriptions in escort ads. However, these ads contain additional information, such as images, which can provide crucial context for vendor identification. Integrating multimodal data—combining text and images—could enhance the model's effectiveness and provide a more comprehensive understanding of vendor behaviors. Further research explores multimodal approaches to improve AA applications in online escort ads.

## 3.7 SUMMARY

Extending the research in Chapter 2, this study collectively responds to RQ1(b) by identifying key challenges in applying AA to online escort advertisements. First, this study introduces IDTraffickers, an AA dataset comprising 87,595 Backpage escort advertisements with 5,244 vendor labels, generated by linking phone numbers mentioned in the ads. To establish a benchmark for the vendor identification task, a DeCLUTR-small classifier is trained, achieving a macro-F1 score of 0.8656 in a closed-set classification setting. Furthermore, to formalize the vendor verification task, this study utilizes a similarity search on the classifier's style representations to perform an open-set retrieval task, obtaining an R-Precision score of 0.8852. To generate insights from the trained model, local and global feature attribution techniques are employed, uncovering misclassifications that suggest multiple labels may correspond to the same vendor. Despite these challenges, the classifier effectively captures stylistic patterns, enabling vendor identification and verification tasks. The findings, dataset, and benchmarks presented in this study aim to support future research, LEAs, and practitioners in linking escort ads to potential human trafficking networks.

# 4

# MATCHED: Multimodal Authorship Attribution For Online Escort Markets

**This chapter is based on the following research:**

- **Vageesh Saxena**, Benjamin Ashpole, Gijs van Dijck, and Gerasimos Spanakis. 2025b. MATCHED: Multimodal Authorship-Attribution To Combat Human Trafficking in Escort-Advertisement Data. 2025b. Accepted at Association for Computational Linguistics, 2025.

Chapter 3 establishes the foundation for authorship attribution in online escort ads by formalizing and evaluating vendor identification and vendor verification tasks. It introduces a novel dataset, IDTraffickers, which leverages phone numbers as proxy identifiers to construct vendor communities—serving as ground truth labels. The dataset contains 87,595 escort ads collected from Backpage, spanning 41 cities across 14 U.S. states and 5,244 unique vendor labels. This enables the linking of escort ads to underlying vendors based solely on the textual descriptions of the Backpage marketplace. Through extensive experimentation, the study sets DeCLUTR-small as the benchmark model for closed-set vendor identification and proposes a retrieval-based approach as the benchmark for open-set vendor verification.

Despite the strong performance demonstrated in Chapter 3, several limitations remain unaddressed. Most notably, the reliance on textual data alone overlooks the rich visual information embedded in escort advertisements—images that often carry distinctive photometric styles and cues that can further inform AA models. Additionally, while Chapter 2–3 benchmark a range of SoTA text-based methods, they do not evaluate models trained specifically for the vendor verification task, prioritizing representation learning over classification. Notably, both vendor identification and vendor verification offer complementary benefits: the former encourages learning fine-grained discriminative features, while the latter promotes clustering of stylistically similar authors in a shared representation space. To address these limitations, this chapter builds upon Chapter 3 by introducing MATCHED, an MAA dataset, MAA benchmarks incorporating text and visual information, and leveraging advanced model training strategies such as contrastive learning and multitask learning. These enhancements improve generalization across both in-sample (seen) and out-of-distribution (unseen) vendor data.

## 4.1 Introduction

Human trafficking, particularly sex trafficking, continues to exploit vulnerable individuals through coercion, deception, and the digital anonymity afforded by modern platforms (Europol, 2021; GLOTIP, 2024). Although end-to-end classification methods have shown promise in detecting trafficking activities in online escort ads (Alvari et al., 2016; Tong et al., 2017; Alvari et al., 2017b), their dependence on expert-generated labels can lead to overfitting and reduced generalizability. In response, LEAs and researchers have developed trafficking indicators to detect suspicious advertisements (Ibanez and Suthers, 2014; Ibanez and Gazan, 2016c; Lugo-Graulich and Meyer, 2021). However, these indicators typically require the grouping of ads based on links between individuals or networks, relying on explicit identifiers such as phone numbers and email addresses (Chambers et al., 2019). As reported in Chapter 3, only 37% of the Backpage ads contain these identifiers. Furthermore, supervised (Nagpal et al., 2017; Li et al., 2022b; Liu et al., 2023) and unsupervised entity-linking techniques (Rabbany et al., 2018; Nair et al., 2022a; Vajiac et al., 2023a) that depend on explicit textual similarities (e.g., names, phrases, near-duplicates) lose effectiveness when vendors modify details to avoid detection.

To overcome these challenges, Chapter 3 proposes AA methods that ex-ploit latent language and stylistic patterns shared among ads from the same vendor or group on online Backpage escort markets. The findings indicate the success of these NLP-based AA approaches in linking ads by identifying subtle textual consistencies even when explicit markers are obscured. However, most existing AA research has focused exclusively on text, neglecting the multimodal nature of escort ads, which typically combine text (e.g., titles and descriptions) with images. Incorporating visual cues can capture additional stylistic consistencies, such as recurring backgrounds, poses, or attire, which uniquely characterize a vendor's profile. Moreover, while the AA methods presented in Chapter 3 require a minimum of five ads per vendor, this text-only focus inherently excludes vendors with fewer postings or those who frequently alter textual content.

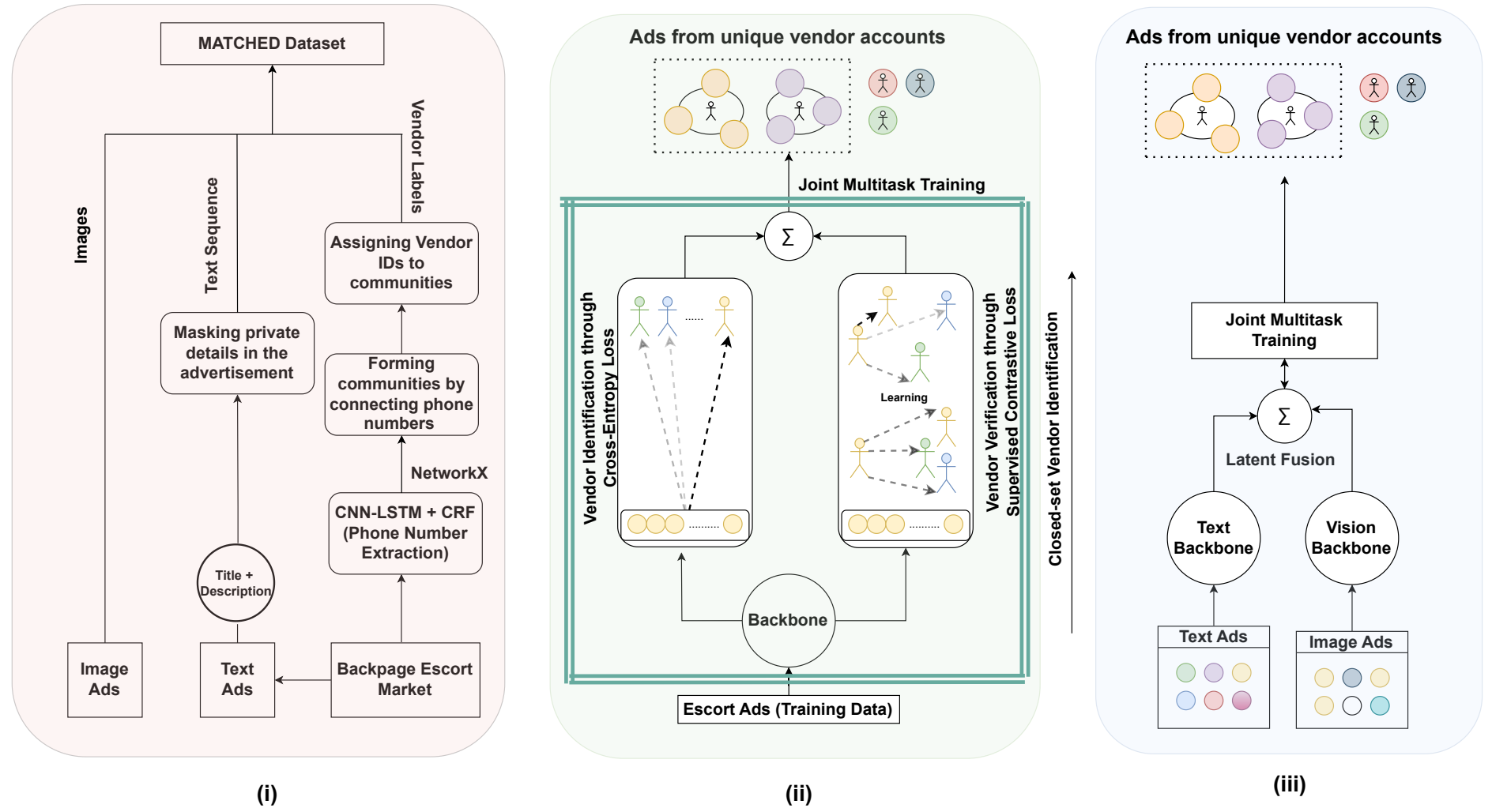

Figure 4.1: (i) Collection process of MATCHED dataset, (ii) Joint multitask training objective for closed-set vendor identification task, (iii) Multimodal fusion with multitask training objective for the closed-set vendor identification task.

Addressing these gaps, this chapter introduces a MAA approach that combines textual and visual features to enhance vendor identification and verification (Zhang et al., 2019). Finally, this integrated approach supports the construction of AA-driven knowledge graphs for researchers, LEAs, and practitioners and facilitates targeted investigations across vast collections of raw escort advertisements. As illustrated in Figure 4.1, the key contributions of this work are as follows:

- **(i). MATCHED Dataset and Comprehensive Benchmarking:** This study introduces MATCHED, a novel MAA dataset comprising 27,619 unique text descriptions and 55,115 images collected from Backpage escort advertisements across seven U.S. cities between December 2015 and April 2016. Benchmarks are established for text, vision, and multimodal domains, with performance evaluated on both in-sample and OOD distribution datasets. MATCHED serves as a robust foundation for future MAA research. Due to the sensitive nature of the data, anonymized metadata is shared via DataverseNL, while the full dataset remains restricted and accessible only through formal requests. The associated code is available at MATCHED.

- **(ii). Enhanced Performance through Multitask Training:** A joint multitask framework is proposed, simultaneously optimizing vendor identification and verification tasks. This approach outperforms traditional single-task models by 1.61% (text-only baseline) and 1.52% (vision-only baseline) on macro-F1 scores for classification and by 1.68% (text-only baseline) and 6.75% (vision-only baseline) on R-Precision for retrieval tasks. While these improvements may appear subtle, the dual-focus approach significantly enhances the ability of researchers, LEAs, and practitioners to identify known vendors and uncover emerging ones in OOD advertisements, thereby improving investigative capabilities.

- **(iii).Advancements in Model Performance through Multimodal Training:** Traditional AA methods predominantly rely on textual data, often overlooking valuable stylistic cues from images and excluding vendors with fewer advertisements. This study integrates text and image data through a multimodal approach, improving performance even for vendors with limited postings. Pairing a single text description with multiple images (e.g., one text with five images produces five samples) expands the training set and enriches feature representation. Although text remains the dominant modality, incorporating images enhances text-only results by 5.43% on retrieval R-Precision, marginally improves vision-only results by 0.75% on retrieval R-Precision, and increases classification macro-F1 by 32.62%—ultimately providing a more comprehensive and robust AA framework.

## 4.2 RELEVANT RESEARCH

### 4.2.1 MULTITASK AUTHORSHIP ATTRIBUTION

Existing AA research predominantly focuses on either authorship identification or verification tasks. Authorship identification, framed as a classification task, aims to assign artifacts to predefined authors based on stylistic cues such as lexical patterns, syntactic structures, or thematic consistency (Mohsen et al., 2016; Kale and Prasad, 2017; Yülüce and Dalkılıç, 2022). This approach assumes a closed-set scenario where authors are known a priori and represented in the training data. In contrast, authorship verification operates in an open-set paradigm, evaluating whether two artifacts originate from the same author by measuring stylistic similarity through metrics like cosine similarity or probabilistic scoring (Koppel and Schler, 2004; Bevendorff et al., 2022; Lei et al., 2022b). While identification excels at linking ads to known entities, verification is critical for detecting emerging or previously unseen vendors, a common challenge in dynamic online environments.

To bridge this gap, Ai et al. (2022) introduces a joint training framework that unifies author identification and verification by combining cross-entropy loss for identification with contrastive learning for verification, enabling models to learn robust, discriminative features that account for variations in writing style across text from the same author while enhancing the separation of text from different authors. By simultaneously aligning text from the same author and diverging dissimilar pairs in the embedding space, the joint approach improves generalization to OOD distribution samples.

This joint multitask training fosters robust feature learning by integrating discriminative (classification) signals with contrastive similarity constraints, capturing author-specific and cross-author stylistic patterns. It further mitigates overfitting to known authors by refining embeddings through verification objectives, enhancing adaptability to OOD distribution. Finally, it combines closed-set identification accuracy with open-set verification capabilities, aligning with the dual operational needs of researchers, LEAs, and practitioners to monitor known, unknown, and emerging entities. For human trafficking investigations on online escort platforms, where escort vendors frequently change writing styles or imagery to evade detection, this joint approach offers a scalable solution for linking ads across platforms. By unifying identification and verification, this study advances AA's applicability to real-world investigations, providing with tools to effectively disrupt trafficking networks.

### 4.2.2 Multimodal Machine Learning

Recent advancements in multimodal ML have led to various fusion strategies designed to integrate heterogeneous data sources effectively. These fusion methods are commonly classified as early, late, and latent (intermediatory) (Huang et al., 2019; Zhang et al., 2020a). Early fusion merges raw features from different modalities at the input level but often suffers from challenges such as modality-specific noise and scale discrepancies (Gallo et al., 2018, 2020). Similarly, late fusion, which combines predictions from independently trained unimodal models, can miss subtle cross-modal interactions critical for tasks demanding fine-grained integration. In contrast, latent fusion assigns modality-specific features into a shared latent space during the intermediate processing stages (Yuan et al., 2014; Zhang et al., 2019; Bakkali et al., 2020). Several techniques have been developed in the latent fusion paradigm to combine textual and visual features. A straightforward approach concatenates individual representations, where modality-specific vectors are directly appended to form a joint representation (Gallo et al., 2018; Li et al., 2024b). While this method preserves the distinct characteristics of each modality, it does not fully capture the complex interdependencies between them. Mean pooling offers an alternative by averaging the features to reduce noise; however, this approach can dilute distinctive signals crucial for effective discrimination (Sleeman et al., 2022; Bakkali et al., 2020). Alternatively, self-attention mechanisms have been introduced, which dynamically weight each modality's contribution based on contextual relevance, thereby enhancing the model's capacity to focus on salient cross-modal cues (Kiela et al., 2019; Zhang et al., 2020a; Gan et al., 2024). Finally, adaptive auto-fusion approaches allow neural networks to learn optimal fusion weights tailored to the specific characteristics of the input data, enabling a more nuanced integration of complementary signals (Sahu and Vechtomova, 2021).

Advanced multimodal ML architectures like Contrastive Language-Image Pre-Training (CLIP) (Radford et al., 2021b), Bootstrapping Language-Image Pre-training (BLIP) (Li et al., 2022a), and BLIP2 (Li et al., 2023b) further exemplify the potential of latent fusion. CLIP uses an Image-Text Contrastive (ITC) objective to align visual and textual features, ensuring semantically related pairs are closely embedded. It employs two parallel encoders—typically a vision transformer for images and a text transformer for text—to project both modalities into a shared latent space.

During training, the model learns to associate matching image-text pairs by pulling their embeddings closer together while pushing apart non-corresponding pairs. This contrastive learning paradigm enables CLIP to perform zero-shot transfer across various vision-language tasks, effectively capturing high-level semantic alignments. Building on this, (Villegas et al., 2024) propose extending the ITC objective with an Image-Text Matching (ITM) objective, which enforces finer-grained correspondence between modalities. While ITC focuses on aligning overall semantic content, ITM involves a classification task to determine whether a given image-text pair truly corresponds. This dual-objective approach enhances the model's ability to capture nuanced cross-modal relationships.

The BLIP architecture extends these ideas by integrating contrastive and generative pre-training strategies. Unlike CLIP, which relies solely on contrastive learning, BLIP introduces a generative component that leverages captioning or language modeling tasks, typically using Flan-T5 (Chung et al., 2022), to enrich joint representations. The contrastive component, similar to CLIP's ITC, aligns image-text pairs in the latent space, while the generative component generates textual descriptions from images, enabling the model to learn richer, context-aware embeddings. This dual approach allows BLIP to excel in tasks like image captioning and visual question answering, where understanding fine-grained relationships between modalities is critical. Unlike the BLIP architecture, which trains the image encoder and language model from scratch, BLIP2 leverages frozen pre-trained components: a vision transformer for images and Flan-T5 for text. These components are connected by a lightweight Querying Transformer (Q-Former), which facilitates cross-modal interaction by projecting visual features into the language model's space. By providing visual cues to the language model, BLIP2 achieves a higher level of semantic alignment and finer granularity in multimodal representations, making it particularly effective for tasks requiring precise similarity search tasks.

These techniques highlight the ability of latent fusion approaches in capturing cross-modal relationships in MAA across online escort platforms. Vendors on these platforms often post text descriptions of escort profiles alongside images, creating an opportunity to align these modalities and harness complementary signals. By projecting the stylometric representations from writing and photometric patterns into a shared latent space, latent fusion techniques enable the effective linking of ads—even when writing styles or visual patterns differ.

## 4.3 DATASET

| Regions | Ads | Text | Images | % Faces | Vendors |
|---|---|---|---|---|---|
| South | 14088 | 13661 | 27423 | 0.4928 | 1450 |
| Midwest | 8564 | 8259 | 14883 | 0.5542 | 1008 |
| West | 3262 | 3153 | 5049 | 0.6052 | 507 |
| Northeast | 2599 | 2546 | 7760 | 0.6183 | 584 |
| All | 28513 | 27619 | 55115 | 0.5676 | 3549 |

Table 4.1: Number of advertisements, unique text descriptions, images, % of Faces in the image datasets, and vendors per region in the MATCHED dataset.

Lugo-Graulich and Meyer (2021) provide compelling evidence linking Backpage escort advertisements to human trafficking, motivating the focus of this study on Backpage ads. As illustrated in Table 4.1, a multimodal dataset of 28,513 ads is curated, comprising 27,619 unique text descriptions and 55,115 unique images associated with 3,549 vendors. Approximately 56% of the images feature an escort's face, while the remaining 44% display partial body images (without faces). The dataset spans seven major U.S. cities—Chicago, Houston, Detroit, Dallas, San Francisco, New York, and Atlanta—representing four geographic regions: South, Midwest, West, and Northeast. The South region dataset, containing the largest number of text and image ads, is the primary dataset for training and in-distribution evaluation. The Midwest, West, and Northeast datasets are utilized as OOD distribution datasets to evaluate model generalization.

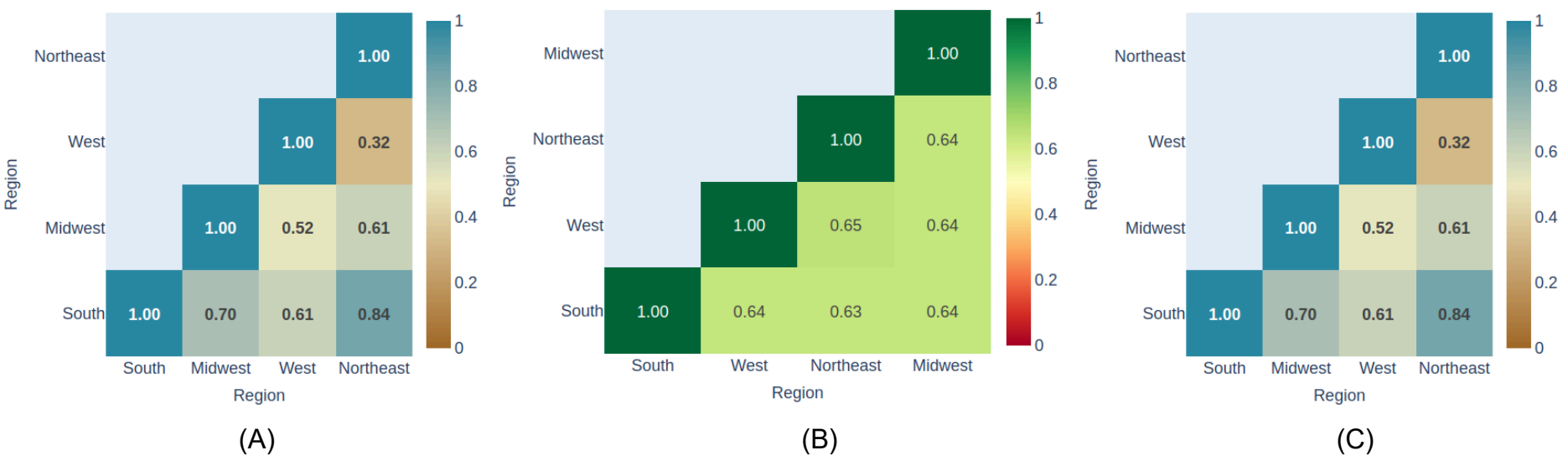

Figure 4.2: (A) % of vendors shared between different datasets, (B) Average text-to-text cosine similarity between datasets computed on the text representations from the pre-trained available checkpoints of DeCLUTR-small backbone, and (C) Average image-to-image cosine similarity between datasets computed on the image representations from the pre-trained available checkpoints of ViT-base-patch16 backbone.

Notably, many vendors appear across multiple regions, meaning the OOD distribution datasets include ads from vendors present in the South dataset as well as additional region-specific vendors. Figure 4.2(A) illustrates the percentage of shared vendors across different datasets. This cross-regional activity aligns with existing findings that the Backpage escort marketplace was frequently flagged for human trafficking activities, with vendors often advertising their services across multiple geographic regions (Lugo-Graulich and Meyer, 2021). However, this overlap also highlights a limitation in the OOD generalization experiments conducted in this study. These experiments evaluate model performance on data distributions that differ from the training data but may not fully replicate real-world conditions due to shared vendors across regions. To assess true OOD generalization, future work would require collecting ads from an entirely separate escort platform, enabling evaluation of model adaptability to a completely new distribution of ads—an approach that lies beyond the scope of this study. Figure 4.2(B) examines the average text-to-text similarity between ads from different datasets. Using a pre-trained DeCLUTR-small model (not fine-tuned on the MATCHED dataset), sentence embeddings are generated for each ad, and cosine similarity is calculated between pairs from different datasets. Given the high level of vendor overlap across regions, the text content exhibits considerable similarity, reflecting consistent writing styles. Similarly, Figure 4.2(C) illustrates the average image-to-image cosine similarity across ads from different datasets, computed using representations from a pre-trained ViT-base-patch16 model. In contrast to the relatively high text similarity, image similarity is significantly lower. This discrepancy suggests that while vendors often maintain consistent writing styles across regions, they frequently vary the images posted, potentially to depict different escorts or avoid detection.

Figure 4.3(A) and (B) illustrate the distributions of sentence and character lengths for text ads across the datasets. Sentence length is measured by counting the total number of tokens generated by the pre-trained DeCLUTR-small checkpoint after tokenization, while character length represents the total number of characters in each text ad. As shown, most text ads contain fewer than 512 tokens, prompting the truncation of all ads to this maximum length, which aligns with the sequence length limit of most transformer-based models.

Figure 4.3(C) depicts the frequency of text ads per vendor, revealing that most vendors post between 1 and 20 text ads. Unlike the AA approaches in

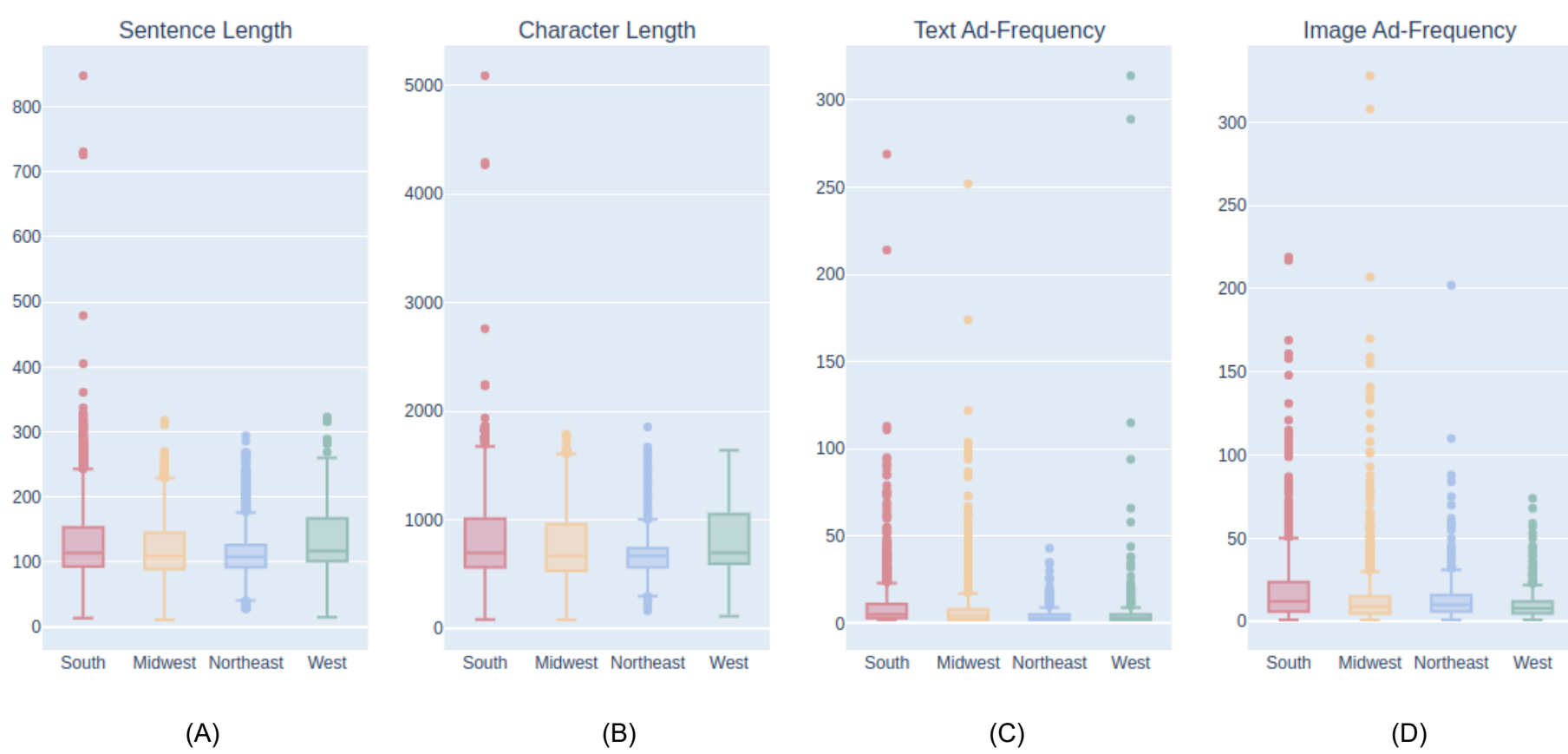

Figure 4.3: (A) Sentence length and (B) Character length distribution of the text ads, (C) Text-ad frequency and (D) Image-ad frequency, i.e. the number of text and image ads per vendor.

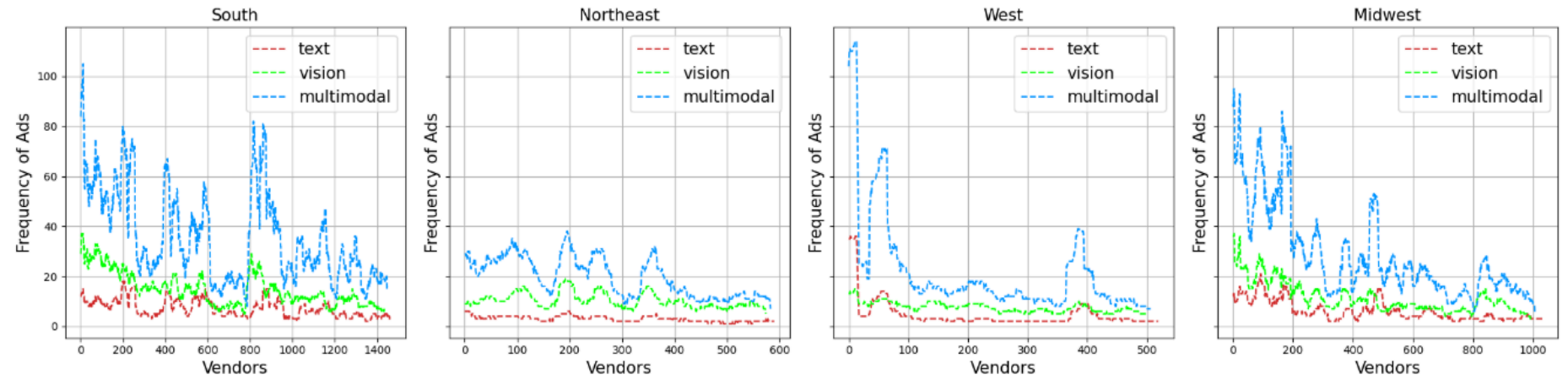

Figure 4.4: Frequency of text, image, and multimodal ads in South, Northeast, West, and Midwest-region datasets.

Chapter 2-3, which require a minimum of 5 or 20 ads for effective implementation, this study explores the applicability of AA techniques for vendors with as few as two ads. The distribution of ad frequency highlights a class imbalance in the dataset, necessitating the prioritization of Macro-F1 performance to ensure equal weighting across all classes in classification tasks. Similarly, Figure 4.3(D) illustrates the frequency of image ads per vendor, showing that most vendors post between 5 and 24 image ads. A detailed analysis of the frequency of text, image, and multimodal ads per vendor is provided in Figure 4.4. Note that these line plots are smoothed using a window size of 30 for improved readability. Finally, language analysis using the LangDetect model (Tamás et al., 2022) reveals that the vast majority of text ads are in English: 99.65% in the South dataset, 99.98% in the Midwest dataset, 99.88% in the West dataset, and 99.85% in the North-east dataset.

### 4.3.1 DATA PRE-PROCESSING

The dataset is curated by scraping titles, descriptions, and images for each advertisement. The text sequence for each entry is constructed by combining the title and description, separated by a "[SEP]" token. Since ads may contain multiple images, the text sequence is duplicated for each associated image to prepare the dataset for multimodal training.

To establish ground truth, similar tools from Chapter 3 (Hagberg et al., 2008; Nagpal et al., 2017; Chambers et al., 2019) are utilized to extract phone numbers and form vendor communities, which serve as vendor labels. Consistent with Chapter 3, most personal information—including phone numbers, escort ages, measurements, ad IDs, and posting dates—is masked. Despite these efforts, existing Named Entity Recognizers (NER) (Li et al., 2022b; Liu et al., 2023) struggle to extract escort names from the ads reliably. However, since escorts typically use pseudonyms in these ads (Carter et al., 2021; Lugo-Graulich, 2024), the risk of personal data misuse is inherently minimal.

4

For image anonymization, consideration is given to blurring faces to protect escort identities. However, manual inspection reveals that many images with blurred or cropped faces are already anonymously posted. Therefore, additional blurring is avoided to preserve these stylistic elements, as it could introduce visual biases. Similarly, other image augmentation techniques, such as flipping or rotating, are not applied, as these transformations could alter stylistic cues linked to specific vendors. Ads featuring naturally mirrored or rotated images are retained to prevent misattribution. To further analyze model behavior, the image dataset is categorized into "Face" and "No Face" subsets for each of the four regions—South, West, Midwest, and Northeast—using a pre-trained FaceNet model (Firmansyah et al., 2023). FaceNet detects and generates bounding boxes around faces in images, which are then assigned to the corresponding region's "Face" dataset.

## 4.4 EXPERIMENTAL SETUP

This study addresses two critical AA tasks for disrupting human trafficking networks: vendor identification (closed-set classification) and vendor verification (open-set metric learning). Vendor identification determines whether an advertisement originates from a known vendor in a predefined candidate set. In contrast, vendor verification assesses whether two ads belong to the same vendor, including vendors unseen during training. These tasks are evaluated using text-only, vision-only, and multimodal baselines on the South region dataset, with OOD generalization tested on the Midwest, West, and Northeast datasets.

### 4.4.1 Vendor Identification: A Classification Task

Vendor identification is framed as a multi-class classification task using pre-trained backbones with a classification head on the South region dataset. Models are optimized with cross-entropy (CE) loss (Juola and Baayen, 2005) and multitask joint objectives combining CE with supervised contrastive (SupCon) (Ye et al., 2023) and triplet losses (Hu et al., 2020). These joint objectives, referred to as CE+SupCon and CE+Triplet, enhance feature discrimination by aligning representations of ads from the same vendor while separating those from different vendors.

### 4.4.2 Vendor Verification: A Metric Learning Task

Vendor verification employs contrastive learning with triplet and SupCon losses (Kaya and Bilge, 2019; Wegmann et al., 2022) to learn discriminative ad embeddings. The metric learning task enables these embeddings to cluster ads from the same vendor while separating those from different vendors, allowing retrieval of all ads linked to a vendor—including those outside the training set—via FAISS-based similarity search (Douze et al., 2024).

### 4.4.3 Baselines

Following Chapter 3, text-only baselines utilize Style-Embedding (Wegmann et al., 2022) and DeCLUTR-small (Giorgi et al., 2021) backbones, while vision-only baselines employ VGG-16 (Simonyan and Zisserman, 2015), ResNet-50 (He et al., 2015), DenseNet-121 (Huang et al., 2018), InceptionNetV3 (Szegedy et al., 2015), EfficientNetV2 (Tan and Le, 2021), ConvNext-small (Woo et al., 2023), and ViT-base-patch16-244 (Dosovitskiy et al., 2021) backbones. Text-only and vision-only baselines are fine-tuned with CE, CE+Triplet, and CE+SupCon objectives for vendor identification and Triplet or SupCon objectives for vendor verification.

Multimodal baselines utilize VisualBERT (Li et al., 2019), ViLT (Kim et al., 2021), and a custom DeCLUTR-ViT backbone (combining DeCLUTR for text and ViT for images) with four latent-fusion strategies are explored: concatenation, mean pooling, self-attention, and adaptive auto-fusion via a neural network. Additionally, the DeCLUTR-ViT backbone is pre-trained on the combined dataset from all regions using three alignment strategies: Image-Text Contrastive (ITC, aka CLIP), ITC+ITM (Image-Text Contrastive and Image-Text Matching), and BLIP2. These alignment techniques ensure that text and images from the same ad are closely represented in the latent space, particularly when a single text ad is associated with multiple images. After being pre-trained on the three text-image alignment tasks, these architectures are fine-tuned with CE and CE+SupCon for vendor identification on the South region dataset.

### 4.4.4 EVALUATION

All baselines are evaluated for classification and retrieval tasks. Due to class imbalance, classifiers are assessed using the Macro-F1 metric. Retrieval tasks evaluate the model's ability to identify stylometric similarities between writing and photometric styles in escort ads. The dataset is split into training ("documents") and test ("queries") sets, with embeddings generated by trained models for cosine similarity computation via FAISS-based similarity search. Text-only and vision-only baselines extract embeddings directly from their respective encoders, while multimodal baselines combine text and vision embeddings from the DeCLUTR-ViT backbone using mean pooling. For ITC+ITM and BLIP2-based baselines, image embeddings are taken from the Q-Former encoder. Retrieval tasks are categorized as text-to-text, image-to-image, or multimodal based on query and document embeddings.

Similar to Chapter 3, retrieval performance is evaluated using R-Precision@X, which measures precision when the number of retrieved items equals the number of relevant ads per vendor, with higher scores reflecting more accurate representations of vendor activity. Mean Reciprocal Rank (MRR@10) evaluates the average ranking position of the first ten correctly retrieved ads for each query, with scores closer to 1 indicating higher relevance ranking and reduced manual search efforts (Striebel et al., 2024). Lastly, Macro-F1@X independently calculates and averages F1 scores for each vendor class, ensuring equal weight for all vendors regardless of sample size. In Macro-F1@X and R-Precision@X, X represents the cutoff, defined as the number of relevant items per vendor.

### 4.4.5 INFRASTRUCTURE & SCHEDULE

**Split Ratio:** The dataset is split into training, validation, and test sets using a standard ratio of 0.75:0.05:0.20 with a stratified sampling strategy. To ensure reproducibility, the seed parameter is set to 1111 during this process.

**Training:** Model training and evaluation are conducted on an NVIDIA H100 GPU with 80 GB of memory. The Adam optimizer is employed with $\beta 1$ and $\beta 2$ values of 0.9 and 0.999, respectively, and an L2 weight decay of 0.01. Learning rates of 0.01, 0.001, and 0.0001 were experimented with, with the best performance being observed at 0.001. A warm-up strategy for the first 100 steps is applied, followed by a linear decay schedule.

**Architectures and Hyperparameters:** Given computational constraints, text baselines are initialized using pre-trained checkpoints from DeCLUTR-small and Style-Embedding architectures. Vision baselines are initialized using pre-trained checkpoints from VGG-16, ResNet-50, DenseNet-121, InceptionNetV3, EfficientNetV2, ConvNext-small, and ViT-base-patch16-244 architectures. Face recognition models such as VGG-Face2 (Cao et al., 2018), ArcFace, FaceNet512 (Firmansyah et al., 2023), and GhostFaceNet (Alansari et al., 2023) from DeepFace (Serengil and Ozpinar, 2023) are also explored for vision baselines. However, these models struggle with vendor identification and verification tasks, likely due to their emphasis on facial features, making connecting multiple faces to a single vendor challenging. Training these models on the "Face" and "No Face" subsets of the dataset yields consistent results, confirming their unsuitability for these tasks. Multimodal baselines are initialized by combining the DeCLUTR-small and ViT-base-patch16-244 backbones to process text and vision modalities, with each model equipped with a sequence classification head for classification tasks. Due to resource limitations, all models are trained with a batch size of 32, the maximum feasible size, until convergence.

During training, five in-batch negatives are used for contrastive objectives such as Triplet, SupCon, CE+Triplet, and CE+SupCon. Increasing the number of in-batch negatives does not improve performance, likely due to the fixed batch size of 32 for the classification task. For the text-image alignment pre-training task, the Normalized Temperature-Scaled Cross-Entropy (NT-XENT) loss (Chen et al., 2020b) is employed for the Image-Text Contrastive (ITC) objective, with negatives sampled from regions outside the training dataset. In all multimodal experiments, negatives are strictly non-associated, ensuring text-image pairs are unrelated ads. Temperature coefficient values of 0.01, 0.1, and 0.3 are experimented with for the NT-XENT loss, with the best performance observed at 0.1.

All the experiments are conducted in Python 3.10 using frameworks such as scikit-learn (Pedregosa et al., 2011a), PyTorch (Paszke et al., 2019b), Hugging Face, timm, and Lightning 2.0 (Falcon and The PyTorch Lightning team, 2019). All the plots are developed using Matplotlib (Hunter, 2007) and Plotly (Inc., 2015).

## 4.5 RESULTS

This section evaluates text-only, vision-only, and multimodal baselines for vendor identification (classification) and verification (retrieval) tasks.

### 4.5.1 VENDOR IDENTIFICATION TASK

| Model | Param | Loss | Fusion | Epochs | Accuracy | Weighted F1 | Micro F1 | Macro F1 | Time (hrs.) |
|---|---|---|---|---|---|---|---|---|---|
| **Text-Baselines** | | | | | | | | | |
| **Style-Embedding** | 128M | CE | - | 28 | 0.6582 | 0.6883 | 0.6897 | 0.5210 | 01:07:12 |
| **DeCLUTR-small** | **86M** | CE | | 21 | 0.7647 | 0.7772 | 0.7777 | 0.6379 | 0:12:19 |
| | | CE+Triplet | | 10 | 0.6905 | 0.7068 | 0.7074 | 0.5503 | 0:07:32 |
| | | **CE+SupCon** | | **15** | **0.7786** | **0.7891** | **0.7898** | **0.6540** | **0:06:33** |
| **Vision-Baselines** | | | | | | | | | |
| **VGG-16** | 138M | CE | - | 9 | 0.6823 | 0.6873 | 0.6884 | 0.5262 | 0:15:33 |
| **ResNet-50** | 25M | | | 19 | 0.7741 | 0.7777 | 0.7789 | 0.6394 | 0:23:14 |
| **DenseNet-121** | 7M | | | 13 | 0.7624 | 0.7656 | 0.7673 | 0.6262 | 0:27:01 |
| **InceptionNetV3** | 23M | | | 12 | 0.7471 | 0.7510 | 0.7524 | 0.6047 | 0:20:26 |
| **EfficientNetV2** | 23M | | | 12 | 0.7652 | 0.7690 | 0.7703 | 0.6285 | 0:29:29 |
| **ConvNeXT-small** | 50M | | | 7 | 0.7593 | 0.7625 | 0.7646 | 0.6215 | 0:16:52 |
| **ViT-base-patch16** | **86M** | CE | | 8 | 0.7559 | 0.7593 | 0.7606 | 0.6142 | 0:13:16 |
| | | CE+Triplet | | 13 | 0.7729 | 0.7765 | 0.7771 | 0.6378 | 0:30:35 |
| | | **CE+SupCon** | | **13** | **0.7711** | **0.7709** | **0.7716** | **0.6294** | **0:31:41** |
| **Multimodal-Baselines** | | | | | | | | | |
| **ViLT** | 112M | CE | - | 12 | 0.8454 | 0.8327 | 0.8291 | 0.7369 | 01:18:00 |
| **VisualBERT** | 197M | | | 11 | 0.9652 | 0.9637 | 0.9641 | 0.9355 | 01:10:17 |
| **DeCLUTR-ViT** | 171M | CE | auto | 11 | 0.9344 | 0.9578 | 0.9565 | 0.9121 | 03:41:44 |
| | | | attention | 14 | 0.8774 | 0.9184 | 0.9217 | 0.8451 | 03:45:15 |
| | **169M** | | concat | 15 | 0.9422 | 0.9762 | 0.9781 | 0.9411 | 03:52:36 |
| | | | mean | 16 | 0.9713 | 0.9857 | 0.9861 | 0.9670 | 01:02:16 |
| | | **CE+SupCon** | **mean** | **17** | **0.9823** | **0.9911** | **0.9916** | **0.9802** | **01:15:56** |
| | | ITC+CE | | 18 | 0.9463 | 0.9744 | 0.9760 | 0.9466 | 01:17:20 |
| | 307M | ITC+ITM+CE | | 10 | 0.8456 | 0.9010 | 0.8995 | 0.8443 | 01:07:17 |
| | | BLIP2+CE | | 11 | 0.9101 | 0.9620 | 0.9644 | 0.9128 | 01:14:19 |
| | | BLIP2+CE+SupCon | | 13 | 0.9450 | 0.9702 | 0.9722 | 0.9420 | 01:30:57 |

Table 4.2: Performance metrics (Balanced Accuracy, Weighted-F1, Micro-F1, and Macro-F1) and computational details for text, vision, and multimodal classifier baselines trained on the South region dataset. Pre-training strategies—ITC, ITC+ITM, and BLIP2—are applied to DeCLUTR-small and ViT-base-patch16 backbones for the image-text alignment task. Fine-tuning is then conducted for the vendor identification task on the South region dataset, with classifiers optimized using CE, CE+Triplet, and CE+SupCon loss objectives.

As illustrated in Table 4.2 and consistent with prior findings in Chapter 3,

the DeCLUTR backbone achieves superior performance compared to Style-Embedding for text-based authorship attribution (AA). With cross-entropy (CE) loss, DeCLUTR attains a macro-F1 score of 0.6379, surpassing Style-Embedding (0.5210). The CE+SupCon joint objective further enhances DeCLUTR's performance, yielding the highest macro-F1 score of 0.6540 among text baselines.

For vision baselines, ResNet-50 with CE loss achieves the highest classification performance (macro-F1: 0.6394), followed by EfficientNetV2 (0.6285), DenseNet-121 (0.6262), ConvNext-small (0.6215), and ViT-base-patch16 (0.6141). Despite its slightly lower classification scores, ViT outperforms all other models in retrieval tasks for both in-sample and OOD distribution datasets (discussed in Section 4.5.2). This aligns with prior research (Gkelios et al., 2021; El-Nouby et al., 2021), which highlights ViT's ability to generate rich, contextualized embeddings that capture global stylistic patterns across diverse visual data (e.g., images with or without faces). The ViT backbone achieves its best classification performance (macro-F1: 0.6378) when trained with the CE+Triplet objective, with CE+SupCon close behind (0.6294).

Among multimodal baselines, the end-to-end DeCLUTR-ViT backbone with mean pooling fusion achieves the highest macro-F1 score (0.9670), surpassing VisualBERT (0.9355) and ViLT (0.7369). Pre-trained alignment strategies (CLIP, ITC+ITM, BLIP2) underperform compared to the end-to-end baseline, though BLIP2-aligned DeCLUTR-ViT approaches this performance (0.9420). When trained with the CE+SupCon objective, the DeCLUTR-ViT backbone demonstrates exceptional robustness (macro-F1: 0.9802), attributed to the dataset's structure—each text ad is paired with multiple images and vice versa, exposing the model to diverse combinations during training.

### Further Insights

| Seed | Acc. | Weighted-F1 | Micro-F1 | Macro-F1 |
|---|---|---|---|---|
| **100** | 0.9670 | 0.9862 | 0.9878 | 0.9630 |
| **500** | 0.9761 | 0.9914 | 0.9921 | 0.9755 |
| **1111** | 0.9823 | 0.9911 | 0.9916 | 0.9802 |
| **Mean** | 0.9751 | 0.9896 | 0.9905 | 0.9729 |
| **Std.** | 0.0077 | 0.0029 | 0.0024 | 0.0089 |

Table 4.3: Influence of random initialization on DeCLUTR-ViT classifier's performance

**Effects of Random Initializations:** Due to limited resources, we only examine the effects of different initializations on the model's performance for the established DeCLUTR-ViT benchmark with the CE+SupCon objective. Table 4.3 displays the mean and standard deviation in the model's performance against balanced accuracy, Micro-F1, Weighted-F1, and Macro-F1 scores. The results indicate minimal to no effects on these scores across different initializations.

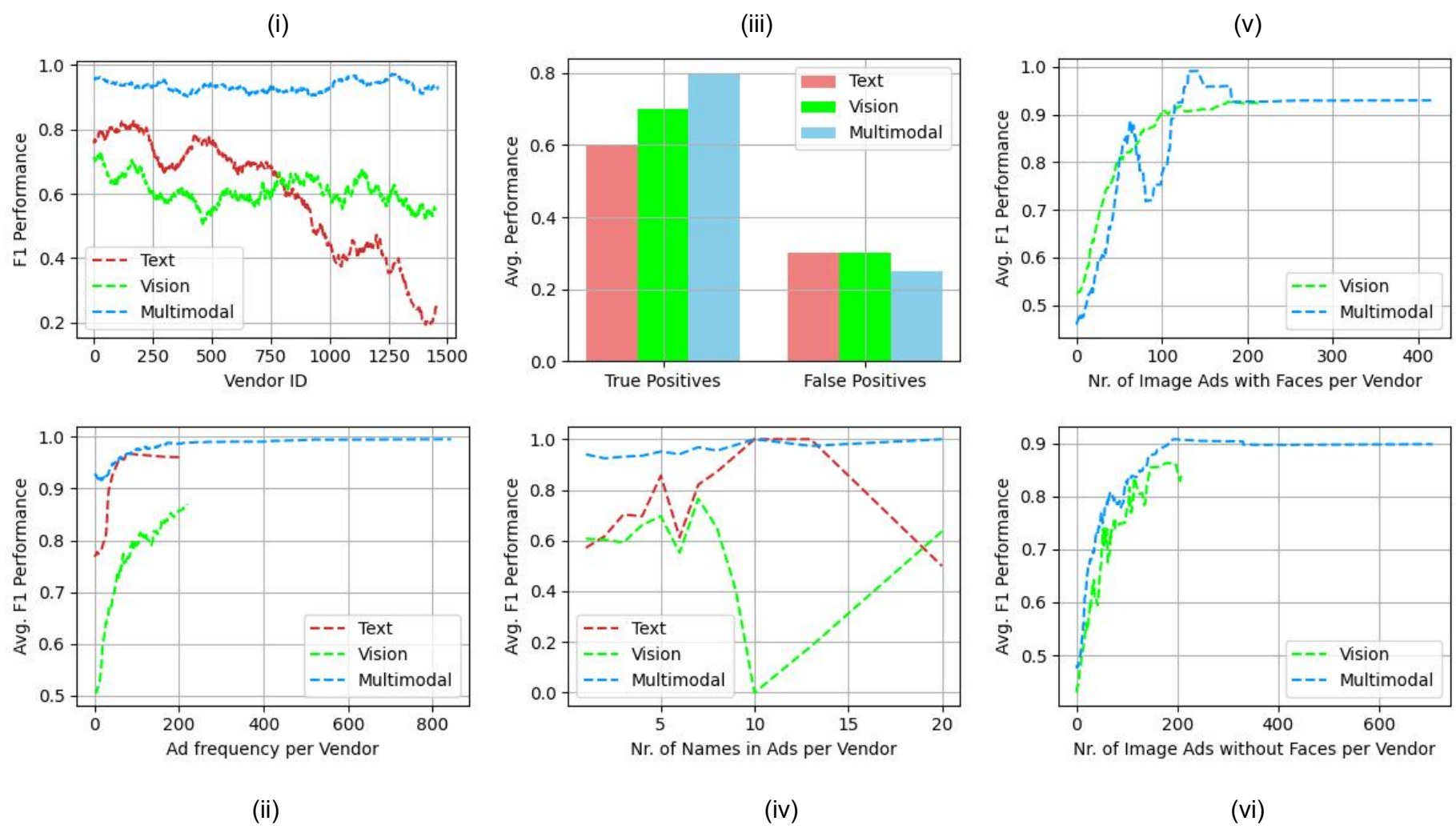

Figure 4.5: Comparison of model performance among text-only, vision-only, and multimodal classifiers trained on the South region test dataset: (i) F1 score across different vendor IDs, (ii) Average F1 score for vendors with varying ad frequencies, (iii) Analysis of true and false positives predictions, (iv) Average F1 score relative to the number of escort names (potentially representing different individuals) in vendor ads, and (v, vi) Average F1 score based on the number of vendor images with and without faces.

**Qualitative Analysis:** Figure 4.5(i) compares the average F1 performance of the DeCLUTR-small text-only, ViT-base-patch16-244 vision-only, and multimodal DeCLUTR-ViT classifiers for vendors in the South region dataset. The results demonstrate that the multimodal classifier consistently outperforms text- and vision-only baselines across all vendors. Further analysis, supported by the vendor frequency distribution in Figure 4.4 and Figure 4.5(ii), reveals that many vendors in the text-only and vision-only datasets have very few ads, likely contributing to lower model performance. In contrast, the multimodal classifier benefits from more training examples per vendor (at least five examples when combining text and vision data). This expanded training set enables the model to capture a broader range of stylistic and visual patterns, resulting in superior performance. These findings underscore the importance of multimodal integration in enhancing model effectiveness by capturing richer and more complementary stylometric cues, particularly for vendors with sparse data in individual modalities [1].

Figure 4.5(iii) compares the average number of true positives and false positives achieved by the text-only, vision-only, and multimodal DeCLUTR-ViT baselines across all vendors in the South region dataset. The results highlight a clear advantage for the multimodal baseline, which yields significantly more true positives while maintaining fewer false positives than the other baselines. This emphasizes the superiority of multimodal approaches in minimizing errors and improving prediction reliability.

Figure 4.5(iv) illustrates the average F1 performance of the text-only, vision-only, and multimodal baselines as a function of the number of escort names per vendor in the text ads. Since multiple escort names likely represent different individuals, this analysis evaluates the models' ability to link varying text descriptions and facial features to a single vendor. Escort names are extracted using (Li et al., 2022b), though manual inspection reveals occasional inaccuracies. However, the extracted entities remain consistent, serving as unique identifiers for escort names. The results indicate that the multimodal baseline consistently outperforms the text-only and vision-only baselines, demonstrating consistent performance even as the number of escort names per vendor increases. This suggests that the multimodal model is less affected by variations in physical descriptions of different escorts in the text ads.

Finally, Figure 4.5(v) and (vi) compare the average F1 performance of the vision-only and multimodal baselines as a function of the number of images with and without faces per vendor. In Figure 4.5(v), as the number of images with faces increases, the multimodal baseline underperforms the vision-only baseline up to approximately 120 images. Beyond this threshold, the multimodal baseline either outperforms or matches the vision-only baseline, indicating its adaptability in connecting images in larger datasets. In contrast, Figure 4.5(vi) shows that the multimodal base-line consistently outperforms the vision-only baseline for images without faces, demonstrating its superior ability to leverage text and image features even without facial features.

[1] All line plots are smoothed for clarity and readability using a window size 30.

### 4.5.2 VENDOR VERIFICATION TASK

The retrieval task evaluates the effectiveness of metric learning (Triplet and SupCon losses) and joint-objective classifiers in clustering ad representations by vendor-specific stylometric patterns. The Zero-Shot (ZS) average reflects retrieval performance across datasets without task-specific training, while the OOD average measures the generalization of South-trained models to unseen regions. The detailed results for the text-to-text, image-to-image, and multimodal retrieval performance of the individual baselines are included in the supplementary material, Section 4.10.

**Text-to-Text Retrieval:** Figure 4.6(A) compares the text-to-text retrieval performance of text-only pre-trained (●), fine-tuned, and multimodal baselines. Fine-tuning on the South region dataset significantly improves performance across all metrics. Among text-only baselines, the DeCLUTR backbone trained with the joint CE+SupCon objective (■) outperforms the CE-only baseline (■) and matches the SupCon-only baseline (■) on OOD average scores while surpassing it on the training dataset. Given its consistent performance in classification and retrieval tasks, the DeCLUTR backbone with CE+SupCon is established as the benchmark for the text-only modality. This benchmark is further compared against text representations from the multimodal DeCLUTR-ViT backbone trained end-to-end with CE+SupCon (✚) and the fine-tuned DeCLUTR-ViT backbone pre-trained for text-image alignment using the BLIP2 objective (✚). The end-to-end multimodal backbone with CE+SupCon consistently outperforms all baselines on both training and OOD datasets.

**Image-to-Image Retrieval:** As illustrated in Table 4.2, ResNet-50, DenseNet-121, and ConvNext-small architectures slightly outperform the ViT-base-patch16 baseline on the vision-only vendor identification task with CE-only objective. However, Table 4.4 demonstrates that the ViT-base-patch16 architecture consistently outperforms all other architectures in image-to-image retrieval tasks across both in-sample and OOD distribution datasets. Despite its underperformance in classification, ViT's superior retrieval performance aligns with its ability to capture global stylistic patterns and contextualized representations. Given the dual ob-jective of this study—helping law enforcement agencies connect existing vendors in their databases and link vendors from emerging markets—the ViT-base-patch16 backbone is established as the best-performing vision backbone on the MATCHED dataset. All the results are presented in x ± y format, where x represents the mean retrieval performance, and y represents the standard deviation between the performance of different vendor classes.

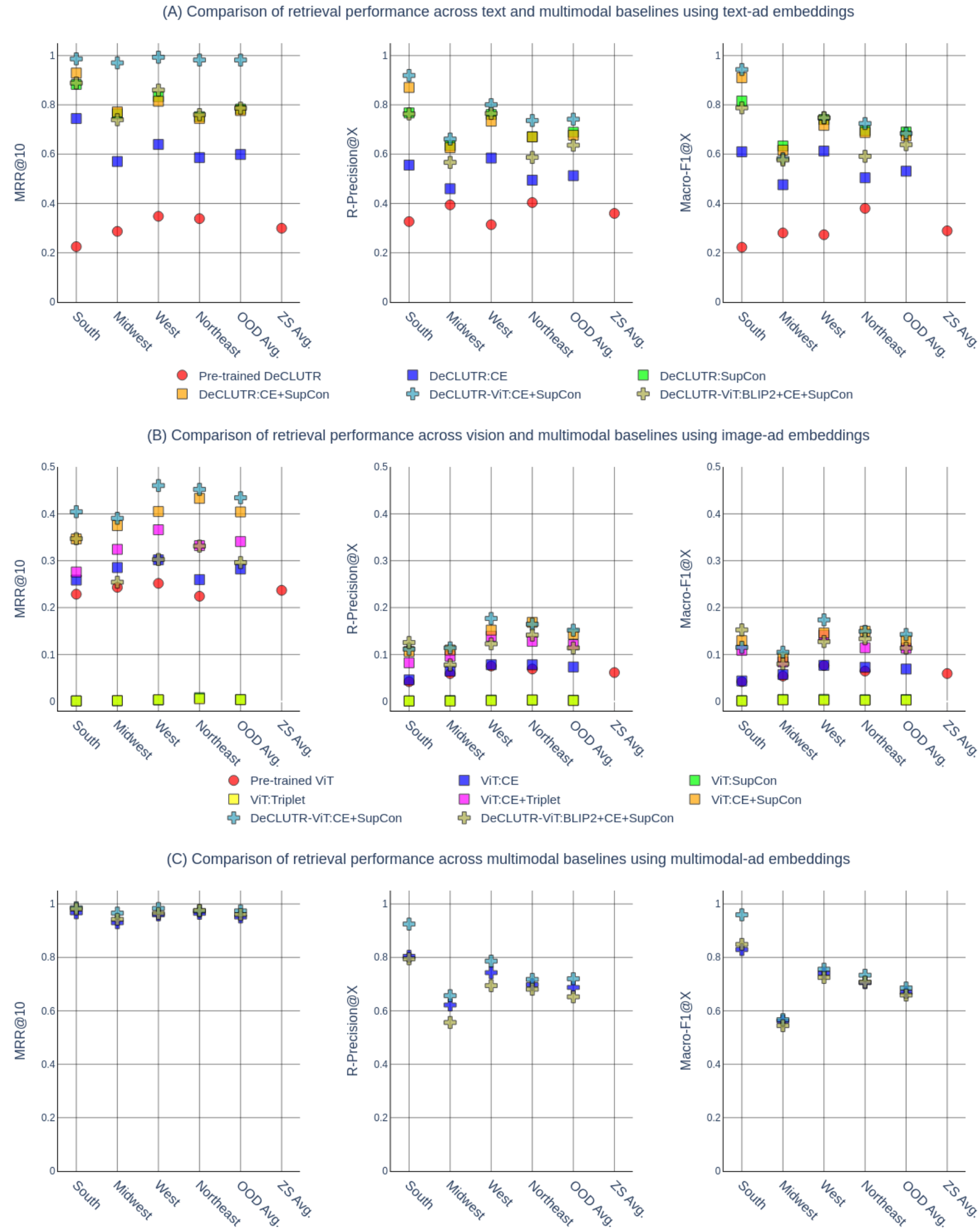

Figure 4.6: Comparison of ads retrieval performance across four regional datasets (South, Midwest, West, and Northeast) for three tasks: **(A) Text-to-Text Retrieval:** Baselines include the pre-trained DeCLUTR checkpoint (●), DeCLUTR classifiers fine-tuned with CE (■) and CE+SupCon losses (■), the DeCLUTR backbone trained with SupCon loss (■), an end-to-end DeCLUTR-ViT classifier (✚), and BLIP2-aligned DeCLUTR-ViT classifiers trained with CE+SupCon losses (✚), **(B) Image-to-Image Retrieval:** Baselines include the pre-trained ViT checkpoint (●), ViT classifiers fine-tuned with CE (■), CE+Triplet (■) and CE+SupCon losses (■), ViT backbones trained with SupCon (■) and Triplet (■) losses, an end-to-end DeCLUTR-ViT classifier (✚), and BLIP2-aligned DeCLUTR-ViT classifiers trained with CE and CE+SupCon losses (✚), and **(C) Multimodal Retrieval:** Baselines include end-to-end DeCLUTR-ViT classifiers trained with CE (✚), CE+SupCon (✚), and BLIP2-aligned CE+SupCon (✚) losses.

| Loss | South | Midwest | West | Northeast | OOD Avg. |
|---|---|---|---|---|---|
| **MRR@10** | | | | | |
| **VGG16** | 0.0069 ± 0.05 | 0.0098 ± 0.07 | 0.0491 ± 0.19 | 0.0172 ± 0.1 | 0.0254 ± 0.12 |
| **ResNet50** | 0.1026 ± 0.22 | 0.1569 ± 0.29 | 0.221 ± 0.35 | 0.125 ± 0.26 | 0.1676 ± 0.30 |
| **Densenet121** | 0.218 ± 0.32 | 0.2465 ± 0.35 | 0.2669 ± 0.37 | 0.1889 ± 0.32 | 0.2341 ± 0.35 |
| **InceptionNetV3** | 0.0477 ± 0.15 | 0.0583 ± 0.19 | 0.0684 ± 0.2 | 0.0625 ± 0.19 | 0.0631 ± 0.19 |
| **EfficientNetV2** | 0.2305 ± 0.32 | 0.2468 ± 0.35 | 0.2523 ± 0.36 | 0.2276 ± 0.35 | 0.2422 ± 0.35 |
| **ConvNext-small** | 0.0588 ± 0.17 | 0.0851 ± 0.22 | 0.0854 ± 0.23 | 0.0917 ± 0.24 | 0.0874 ± 0.23 |
| **ViT-base-patch16** | **0.2587 ± 0.33** | **0.2854 ± 0.37** | **0.3019 ± 0.39** | **0.2597 ± 0.36** | **0.2823 ± 0.37** |
| **R-Precision@X** | | | | | |
| **VGG16** | 0.0063 ± 0.03 | 0.0074 ± 0.03 | 0.0165 ± 0.05 | 0.0139 ± 0.06 | 0.0126 ± 0.05 |
| **ResNet50** | 0.0267 ± 0.05 | 0.0415 ± 0.09 | 0.0599 ± 0.1 | 0.0452 ± 0.09 | 0.0489 ± 0.09 |
| **Densenet121** | 0.0413 ± 0.08 | 0.0618 ± 0.11 | 0.0849 ± 0.11 | 0.0671 ± 0.11 | 0.0713 ± 0.11 |
| **InceptionNetV3** | 0.0084 ± 0.02 | 0.0176 ± 0.07 | 0.0224 ± 0.06 | 0.0143 ± 0.04 | 0.0181 ± 0.06 |
| **EfficientNetV2** | 0.0417 ± 0.07 | 0.0609 ± 0.1 | 0.0752 ± 0.11 | 0.0692 ± 0.11 | 0.0684 ± 0.11 |
| **ConvNext-small** | 0.0157 ± 0.04 | 0.026 ± 0.06 | 0.0299 ± 0.07 | 0.0291 ± 0.06 | 0.0283 ± 0.06 |
| **ViT-base-patch16** | **0.0459 ± 0.07** | **0.0645 ± 0.11** | **0.0781 ± 0.11** | **0.078 ± 0.13** | **0.0735 ± 0.12** |
| **Macro-F1@X** | | | | | |
| **VGG16** | 0.0091 ± 0.03 | 0.0151 ± 0.04 | 0.0171 ± 0.06 | 0.0158 ± 0.05 | 0.0160 ± 0.05 |
| **ResNet50** | 0.0276 ± 0.06 | 0.0407 ± 0.08 | 0.0565 ± 0.1 | 0.0468 ± 0.09 | 0.0479 ± 0.09 |
| **Densenet121** | 0.04 ± 0.07 | 0.0535 ± 0.09 | 0.0823 ± 0.12 | 0.0641 ± 0.11 | 0.0666 ± 0.11 |
| **InceptionNetV3** | 0.0083 ± 0.02 | 0.0154 ± 0.05 | 0.0215 ± 0.06 | 0.0147 ± 0.04 | 0.0172 ± 0.05 |
| **EfficientNetV2** | 0.042 ± 0.07 | 0.0546 ± 0.09 | 0.0764 ± 0.12 | 0.0648 ± 0.1 | 0.0653 ± 0.10 |
| **ConvNext-small** | 0.0159 ± 0.04 | 0.028 ± 0.06 | *0.0312 ± 0.07* | 0.0301 ± 0.06 | 0.0298 ± 0.06 |
| **ViT-base-patch16** | **0.0436 ± 0.07** | **0.0574 ± 0.09** | **0.077 ± 0.12** | **0.0727 ± 0.11** | **0.0690 ± 0.11** |

Table 4.4: Comparison of image-to-image retrieval performance for the vision-baselines trained on south region image ads with CE loss, evaluated on MRR@10, R-Precision@X, and Macro-F1@X metrics

Figure 4.6(B) highlights image-to-image retrieval performance, comparing vision-only pre-trained (●), fine-tuned, and multimodal baselines. Fine-tuning on image ads also improves retrieval performance. Among vision-only baselines, the ViT backbone trained with the CE+SupCon objective (■) achieves superior performance over other baselines on both training and OOD datasets, establishing itself as the benchmark for the vision-only modality. Despite its stronger classification performance, the ViT backbone with the CE+Triplet objective (■) underperforms in retrieval tasks. This vision benchmark is further compared against vision representations from the multimodal DeCLUTR-ViT backbone trained end-to-end

with CE+SupCon (✚) and the fine-tuned DeCLUTR-ViT backbone pre-trained for text-image alignment using the BLIP2 objective (✚). The end-to-end multimodal backbone with CE+SupCon consistently outperforms other baselines on OOD datasets but underperforms the fine-tuned BLIP2 baseline on R-Precision and Macro-F1 metrics for the training dataset.

**Multimodal Retrieval:** Figure 4.6(C) compares retrieval performance among multimodal baselines, evaluating the multimodal representation from the end-to-end multimodal DeCLUTR-ViT backbone trained with CE (✚) and CE+SupCon (✚) losses and the fine-tuned DeCLUTR-ViT backbone pre-trained for text-image alignment using the BLIP2 objective (✚). The end-to-end multimodal backbone with CE+SupCon consistently outperforms the other baseline across training and OOD datasets.

| Loss | South | Midwest | West | Northeast | Avg. |
|---|---|---|---|---|---|
| **Text-to-Image Alignment MRR@10** | | | | | |
| **ITC** | 0.0001 ± 0.01 | 0.0001 ± 0.01 | 0.0003 ± 0.02 | 0.0004 ± 0.02 | 0.0002 ± 0.01 |
| **ITC+ITM** | 0.0001 ± 0.01 | 0.0001 ± 0.01 | 0.0003 ± 0.02 | 0.0008 ± 0.03 | 0.0003 ± 0.02 |
| **BLIP2** | 0.001 ± 0.03 | 0.0027 ± 0.05 | 0.0063 ± 0.08 | 0.0098 ± 0.10 | 0.0050 ± 0.07 |
| **Text-to-Image Alignment R-Precision@X** | | | | | |
| **ITC** | 0.0001 ± 0.01 | 0.0002 ± 0.01 | 0.0013 ± 0.03 | 0.0005 ± 0.01 | 0.0005 ± 0.01 |
| **ITC+ITM** | 0.0002 ± 0.01 | 0.0002 ± 0.01 | 0.0006 ± 0.01 | 0.0007 ± 0.01 | 0.0004 ± 0.01 |
| **BLIP2** | 0.0017 ± 0.02 | 0.0049 ± 0.04 | 0.0103 ± 0.06 | 0.0104 ± 0.06 | 0.0068 ± 0.05 |
| **Text-to-Image Alignment Macro-F1@X** | | | | | |
| **ITC** | 0.0001 ± 0.01 | 0.0002 ± 0.01 | 0.0013 ± 0.03 | 0.0005 ± 0.01 | 0.0005 ± 0.02 |
| **ITC+ITM** | 0.0002 ± 0.01 | 0.0002 ± 0.01 | 0.0006 ± 0.01 | 0.0007 ± 0.01 | 0.0004 ± 0.01 |
| **BLIP2** | 0.0017 ± 0.02 | 0.0049 ± 0.04 | 0.0103 ± 0.06 | 0.0104 ± 0.06 | 0.0068 ± 0.05 |

Table 4.5: Text-to-Image retrieval results from the multimodal DeCLUTR-ViT backbone pre-trained on the text-image alignment task using CLIP (ITC), ITC+ITM (Image text matching loss), BLIP2 (ITC+ITM+Text generation loss).

**Text-to-Image Retrieval:** Table 4.5 highlights the text-to-image retrieval performance of pre-trained baselines, where the goal is to retrieve images associated with a given query text ad from the original advertisement. The underperformance of these models is attributed to the inherent lack of semantic alignment in escort ads. Unlike traditional multimodal tasks where text and images share a strong contextual relationship, escort ads often feature a disconnect between the two modalities. Text descriptions typically provide superficial details about the escorts' physiques—which may or may not be accurate—while the images primarily depict faces or partial body shots. This mismatch makes it challenging for models to establish meaningful connections between the textual and visual content. For instance, a text ad might describe an escort's height, hair color, or at-tire. However, the accompanying image may not reflect these attributes, or it might focus on facial features that are not explicitly mentioned in the text. This lack of direct correspondence between modalities undermines the effectiveness of text-image alignment strategies, leading to poor retrieval performance.

### FURTHER INSIGHTS

**Performance on Shared vs. Unique Vendors:** As indicated in Figure 4.2, despite the model not being exposed to OOD datasets during training, significant vendor overlap exists between the South region and OOD datasets. To confirm that retrieval performance generalizes beyond shared vendors, Table 4.6 presents a performance comparison of the end-to-end DeCLUTR-ViT baseline on text-to-text, image-to-image, and multimodal retrieval tasks. The evaluation focuses on two groups: shared vendors (present in both the South and OOD datasets) and unique vendors (exclusive to the OOD dataset). Representations for these tasks are derived from the end-to-end multimodal DeCLUTR-ViT classifier with mean pooling, trained on the South dataset. The results demonstrate consistent performance across both shared and unique vendors, highlighting the ability of the model to generalize to unseen scenarios effectively. This confirms the model's capability to link ads to emerging and previously unseen vendors, underscoring its practical utility in real-world human trafficking investigations, regardless of prior exposure to specific vendors or ads.

4

| Retrieval | Metric | Midwest | West | Northeast |
|---|---|---|---|---|
| **Text-to-Text** | **MRR@10** | Shared: 0.7164 ± 0.41<br>Unique: 0.7910 ± 0.37 | Shared: 0.8581 ± 0.33<br>Unique: 0.8498 ± 0.33 | Shared: 0.7859 ± 0.38<br>Unique: 0.7013 ± 0.42 |
| | **R-Precision@X** | Shared: 0.5027 ± 0.38<br>Unique: 0.6251 ± 0.37 | Shared: 0.7128 ± 0.36<br>Unique: 0.7234 ± 0.36 | Shared: 0.6553 ± 0.40<br>Unique: 0.5817 ± 0.44 |
| **Image-to-Image** | **MRR@10** | Shared: 0.3462 ± 0.36<br>Unique: 0.3583 ± 0.38 | Shared: 0.3506 ± 0.37<br>Unique: 0.3728 ± 0.37 | Shared: 0.3031 ± 0.38<br>Unique: 0.2432 ± 0.32 |
| | **R-Precision@X** | Shared: 0.0673 ± 0.09<br>Unique: 0.0914 ± 0.14 | Shared: 0.0896 ± 0.12<br>Unique: 0.1168 ± 0.16 | Shared: 0.0816 ± 0.13<br>Unique: 0.0807 ± 0.14 |
| **Multimodal** | **MRR@10** | Shared: 0.7862 ± 0.36<br>Unique: 0.8355 ± 0.31 | Shared: 0.8909 ± 0.28<br>Unique: 0.8693 ± 0.29 | Shared: 0.8138 ± 0.35<br>Unique: 0.7920 ± 0.29 |
| | **R-Precision@X** | Shared: 0.5026 ± 0.35<br>Unique: 0.6196 ± 0.34 | Shared: 0.7103 ± 0.33<br>Unique: 0.7266 ± 0.33 | Shared: 0.6436 ± 0.37<br>Unique: 0.5550 ± 0.41 |

Table 4.6: Text-to-Text, Image-to-Image, and multimodal retrieval performance for shared and unique vendors between South and Midwest, West, and Northeast region dataset. All the representations are extracted from the multimodal DeCLUTR-ViT backbone trained with CE+SupCon objective on the South region dataset.

**Qualitative Analysis:** This experiment analyzes retrieval performance by comparing the end-to-end multimodal DeCLUTR-ViT baseline against text-only (DeCLUTR-small) and vision-only (ViT-base-patch16-244) baselines, all trained with the CE+SupCon objective on the South (Figure 4.7), Midwest (Figure 4.8), West (Figure 4.9), and Northeast (Figure 4.10) region datasets. The comparison includes:

- **M-Text:** Text ad representations extracted from the DeCLUTR-ViT multimodal classifier.
- **M-Vision:** Image representations extracted from the DeCLUTR-ViT multimodal classifier.
- **Vision-Face:** Image representations from the ViT-base-patch16 classifier, evaluating performance on image datasets with faces (Face dataset) and without faces (No Face dataset).
- **Multimodal-Face:** Image representations from the DeCLUTR-ViT multimodal classifier, evaluating performance on image datasets with faces (Face dataset) and without faces (No Face dataset).
- **Multimodal:** Representations extracted from the DeCLUTR-ViT multimodal classifier by applying mean pooling of text and vision modalities.
- **Text:** Text representations extracted from the text-only DeCLUTR-small classifier.
- **Vision:** Image representations extracted from the vision-only ViT-base-patch16 classifier.

Consolidated insights across all regions are structured around key factors influencing performance: vendors, ad frequency, number of names, and the presence or absence of faces in images.

*(i) Performance per Vendor:* Across all regions, the multimodal baseline (Multimodal) consistently outperforms text-only (Text) and vision-only (Vision) baselines for both MRR@10 and R-Precision@X. This advantage underscores the power of integrating textual and visual cues, which capture complementary information. The M-Text and M-Vision representations, extracted from the multimodal model, also outperform the text-only and vision-only baselines. Notably, the text-only baseline performs better than the vision-only baseline, emphasizing the dominant role of text in vendor identification and retrieval tasks. However, the multimodal baseline exhibits lower performance variability than unimodal approaches, indicating its robustness across diverse vendors. This consistency is critical for real-world applications where vendor behaviors vary significantly.

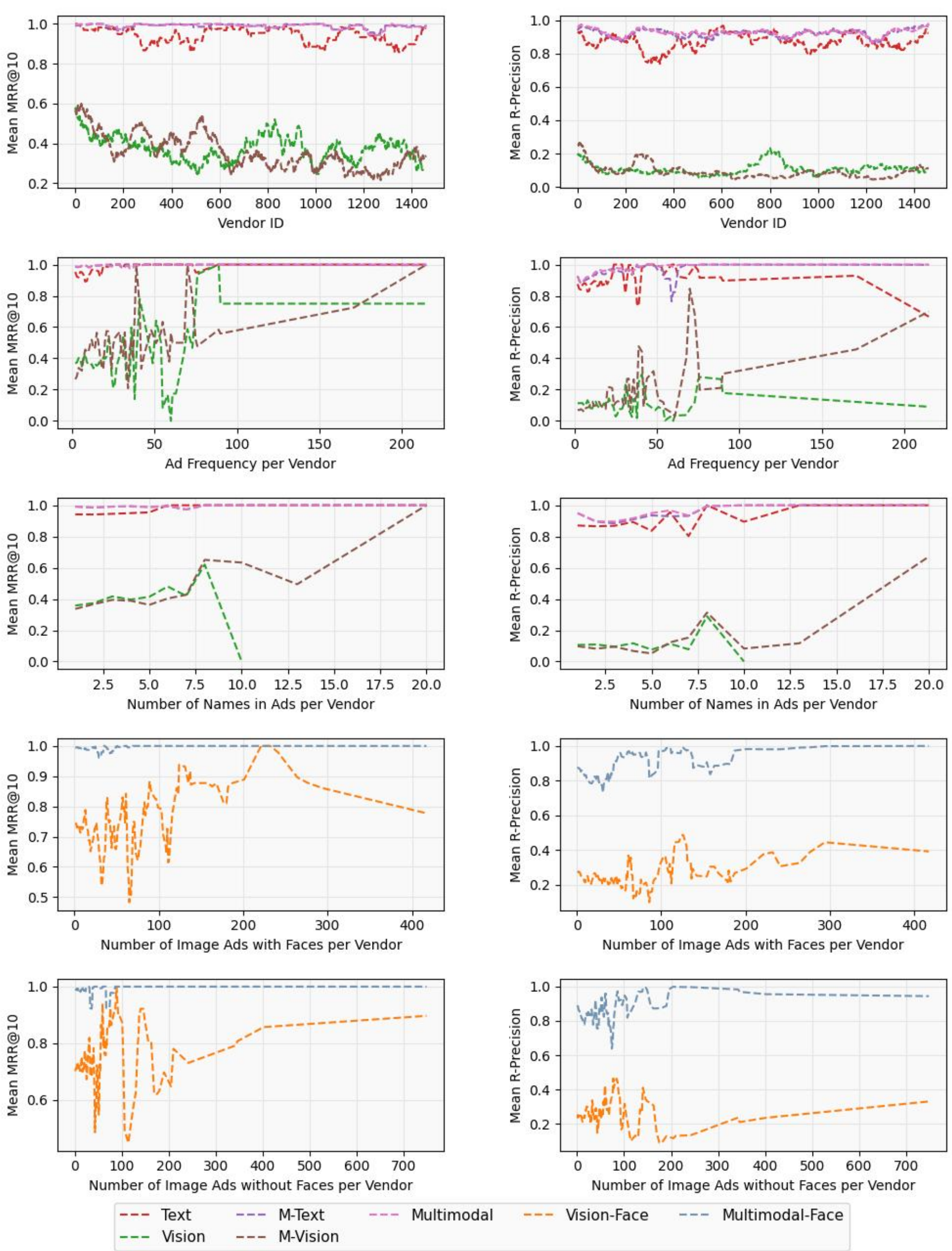

Figure 4.7: Comparison of retrieval performance on the South region test datasets. Text, vision, and multimodal baselines (DeCLUTR-small, ViT-base-patch16-224, and DeCLUTR-ViT, respectively) are trained end-to-end for vendor identification using the joint CE+SupCon objective on the South region dataset. M-Text and M-Vision represent text-only and image-only embeddings from the multimodal system. Vision-Face and Multimodal-Face denote evaluations of escort images with and without faces.

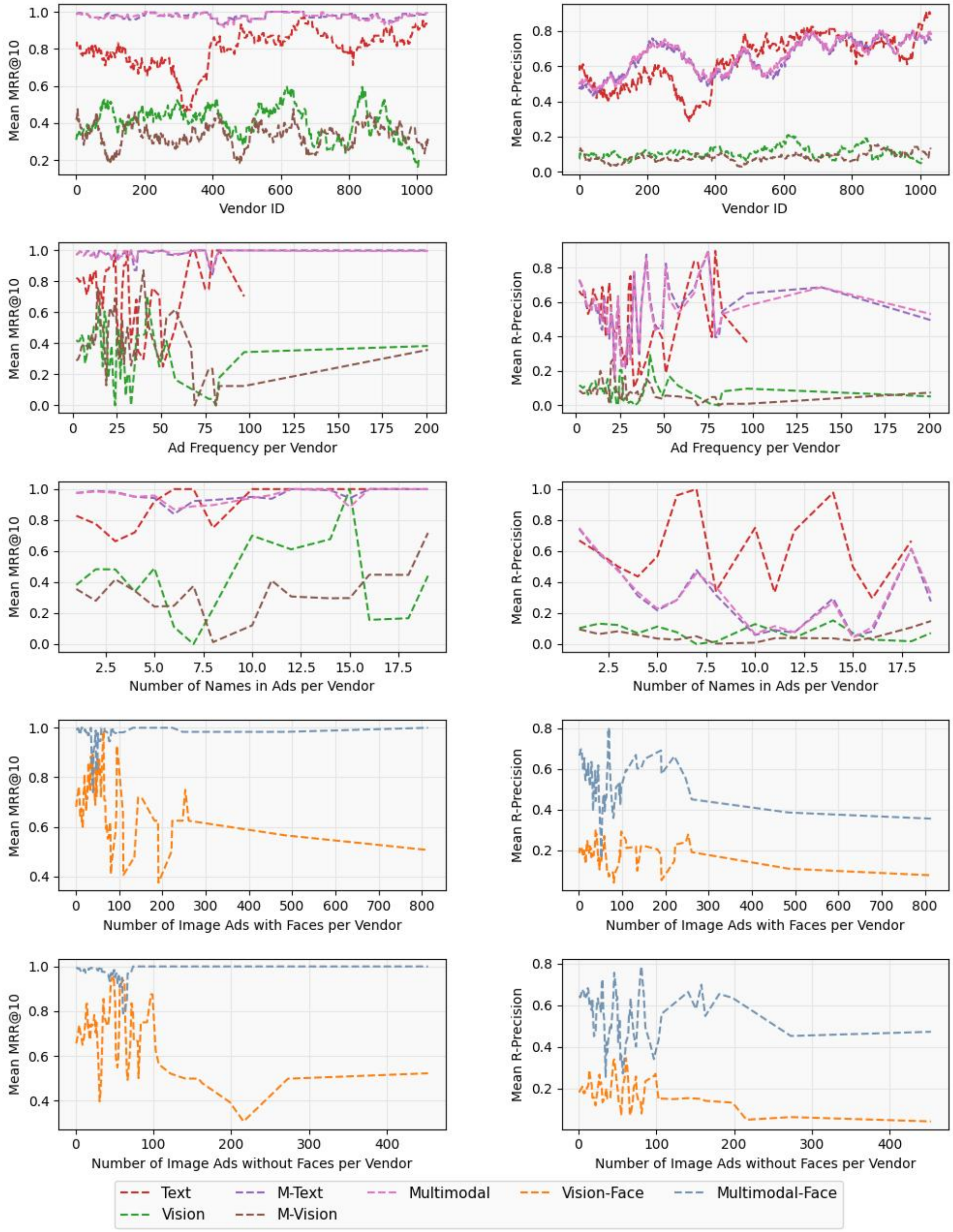

Figure 4.8: Comparison of retrieval performance on the Midwest region test datasets. Text, vision, and multimodal baselines (DeCLUTR-small, ViT-base-patch16-224, and DeCLUTR-ViT, respectively) are trained end-to-end for vendor identification using the joint CE+SupCon objective on the South region dataset. M-Text and M-Vision represent text-only and image-only embeddings from the multimodal system. Vision-Face and Multimodal-Face denote evaluations of escort images with and without faces.

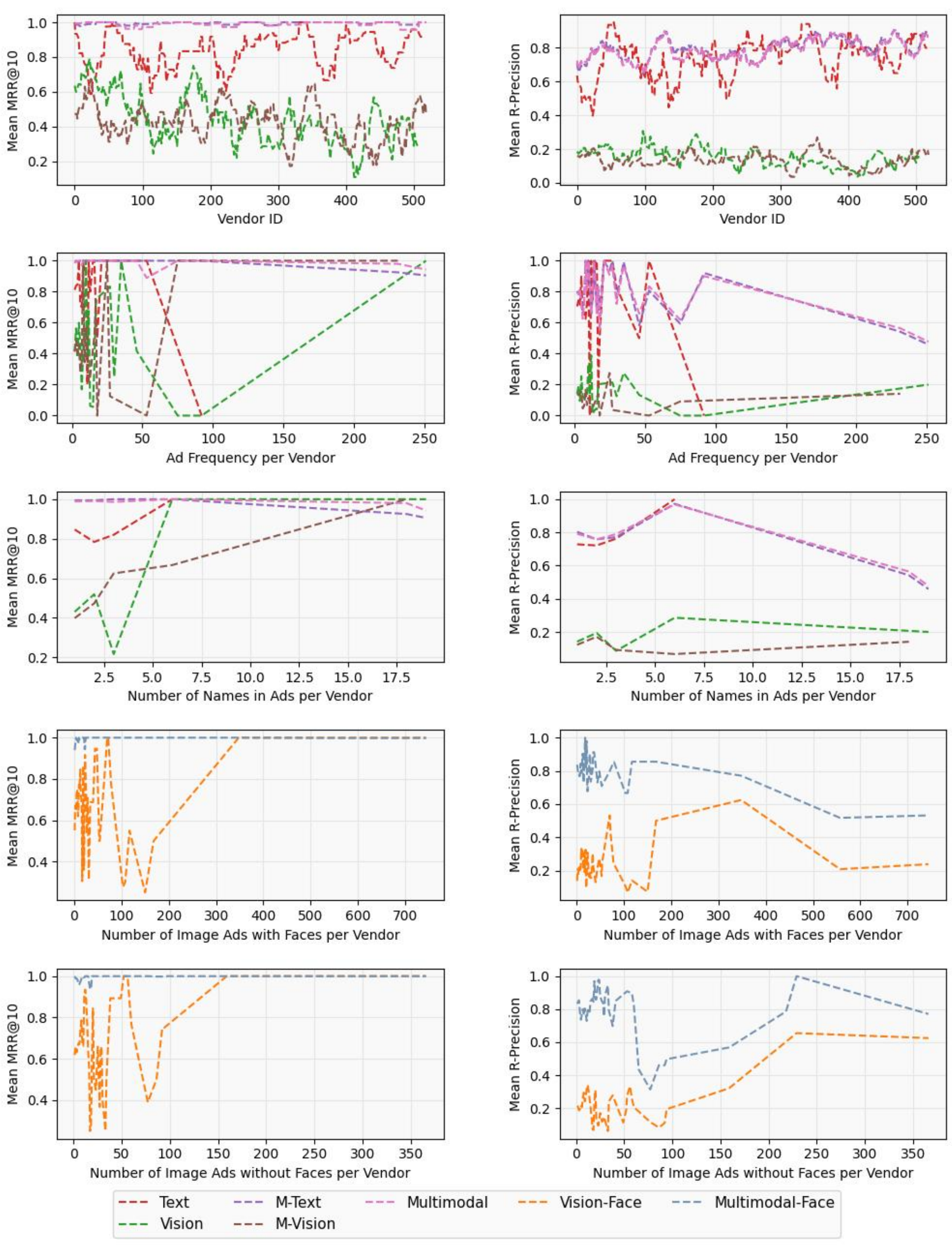

Figure 4.9: Comparison of retrieval performance on the West region test datasets. Text, vision, and multimodal baselines (DeCLUTR-small, ViT-base-patch16-224, and DeCLUTR-ViT, respectively) are trained end-to-end for vendor identification using the joint CE+SupCon objective on the South region dataset. M-Text and M-Vision represent text-only and image-only embeddings from the multimodal system. Vision-Face and Multimodal-Face denote evaluations of escort images with and without faces.

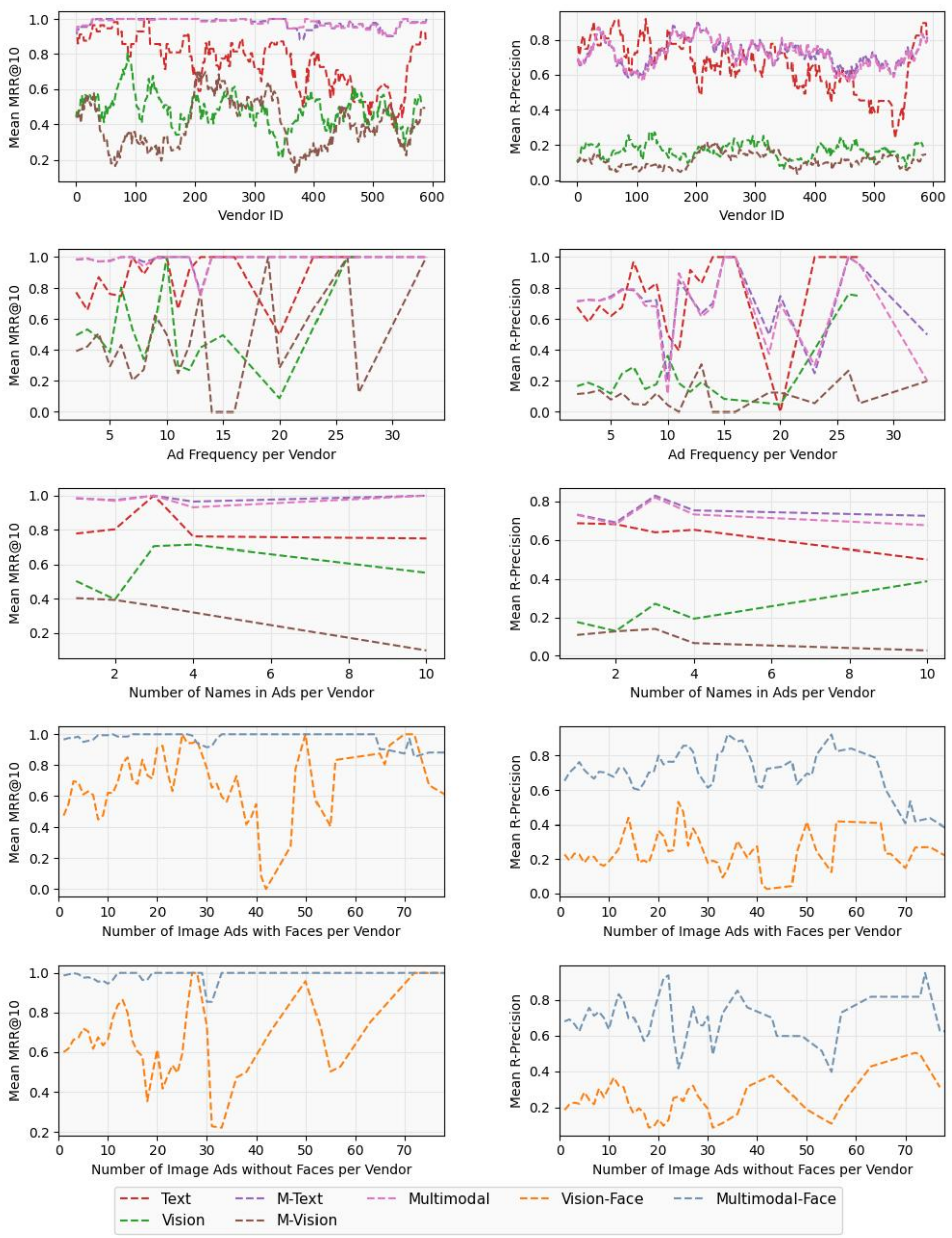

Figure 4.10: Comparison of retrieval performance on the Northeast region test datasets. Text, vision, and multimodal baselines (DeCLUTR-small, ViT-base-patch16-224, and DeCLUTR-ViT, respectively) are trained end-to-end for vendor identification using the joint CE+SupCon objective on the South region dataset. M-Text and M-Vision represent text-only and image-only embeddings from the multimodal system. Vision-Face and Multimodal-Face denote evaluations of escort images with and without faces.

*(ii) Performance by Ad Frequency:* The relationship between retrieval performance and ad frequency remains consistent across regions. The multimodal baseline achieves high performance across all ad frequencies, particularly excelling for vendors with fewer ads. This suggests that multimodal integration effectively compensates for data sparsity by leveraging both textual and visual features. The M-Text representation follows closely, showing a significant improvement over the text-only baseline, especially as ad frequency increases. While the vision-only baseline struggles with sparse data, the M-Vision representation extracted from the multimodal model provides a noticeable improvement, albeit still trailing behind M-Text. These results reinforce the strength of multimodal baselines in handling scenarios with limited vendor representation.

*(iii) Performance by Number of Names:* Analyzing retrieval performance by the number of names in the text ads associated with each vendor reveals the effectiveness of the multimodal baseline in linking ads with varied linguistic and visual patterns. These variations may arise from discrepancies in text descriptions or the presence of images with or without faces representing different individuals. Since many vendors post ads for multiple escorts, this experiment evaluates whether stylistic patterns can be consistently connected despite these discrepancies. Across all regions, the multimodal baseline maintains superior performance as the number of names increases, outperforming text-only and vision-only baselines. The M-Text representations also consistently surpass the text-only baseline, demonstrating that multimodal training enhances the robustness of textual representations. While the vision-only baseline experiences noticeable performance drops with increasing names, the M-Vision representation extracted from the multimodal model performs relatively better. These findings highlight the multimodal baseline's ability to capture stylistic and semantic variations better than unimodal baselines, which is crucial for identifying vendors with diverse aliases.

*(iv) Performance by Images with and without Faces:* The analysis of retrieval performance based on the presence or absence of faces in images provides critical insights into the multimodal baseline's ability to leverage facial and other existing features. Across all regions, the Multimodal-Face baseline consistently outperforms the Vision-Face baseline for both MRR@10 and R-Precision@X, demonstrating the advantages of combining facial and textual cues. For images with faces, the multimodal baseline initially struggles when connecting fewer images with faces but eventually outperforms the vision-only baseline as the number of faces increases. This trend reflects the model's ability to adapt and utilize visual information effectively when sufficient samples are present. For images without faces, the Multimodal-Face baseline consistently surpasses the Vision-Face baseline, leveraging non-facial visual patterns and textual information to improve retrieval performance.

## 4.6 PRACTICAL UTILITY: GENERATING KNOWLEDGE GRAPHS

To demonstrate the practical utility of this research, the multimodal DeCLUTR-ViT model, trained with the CE+SupCon objective on the South region dataset, is employed to create knowledge graphs using retrieval-based methods. The choice of representations for constructing these graphs is informed by the retrieval performance of text, vision, and multimodal embeddings on R-Precision and MRR@10 metrics. Since text-only representations from the multimodal baseline exhibit superior retrieval performance across both metrics for the dataset, they are utilized for retrieval analysis.

Figure 4.11a and Figure 4.11b illustrate knowledge graphs generated for vendor-IDs 784 and 1101 from the South region dataset, respectively. To construct these graphs, a query advertisement (highlighted in red) is used to retrieve all relevant ads from the training dataset based on R-Precision performance. Each advertisement is represented as a node in the graph and labeled with its unique advertisement ID, which serves as an anonymous identifier since all personally identifiable information has been removed using comprehensive masking techniques. Edges in the graph encode the similarity scores between connected nodes and the query advertisement, providing a quantifiable measure of relatedness. The graphs on the left depict all retrieved ads for a given query, visualizing the comprehensive network of connected advertisements for a specific vendor.

To provide flexibility for practitioners, an alternative approach using MRR@K is proposed. This all ows stakeholders to retrieve the top-K most relevant ads based on similarity, enabling focused analysis depending on similarity scores or a chosen investigative threshold. For demonstration purposes, the threshold is set to 0.50, meaning all retrieved ads from the top-K (where K = 10 in Figure 4.11a and K = 5 in Figure 4.11b) predictions must have a similarity score of at least 0.5. Illustrated on the right side of the figures, the resulting knowledge graphs present a filtered view, facilitating efficient examination of high-confidence matches. By leveraging these knowledge graphs, stakeholders can visualize vendor activity across ads, identify patterns, and establish connections, using them to initiate investigations into identifying human trafficking indicators.

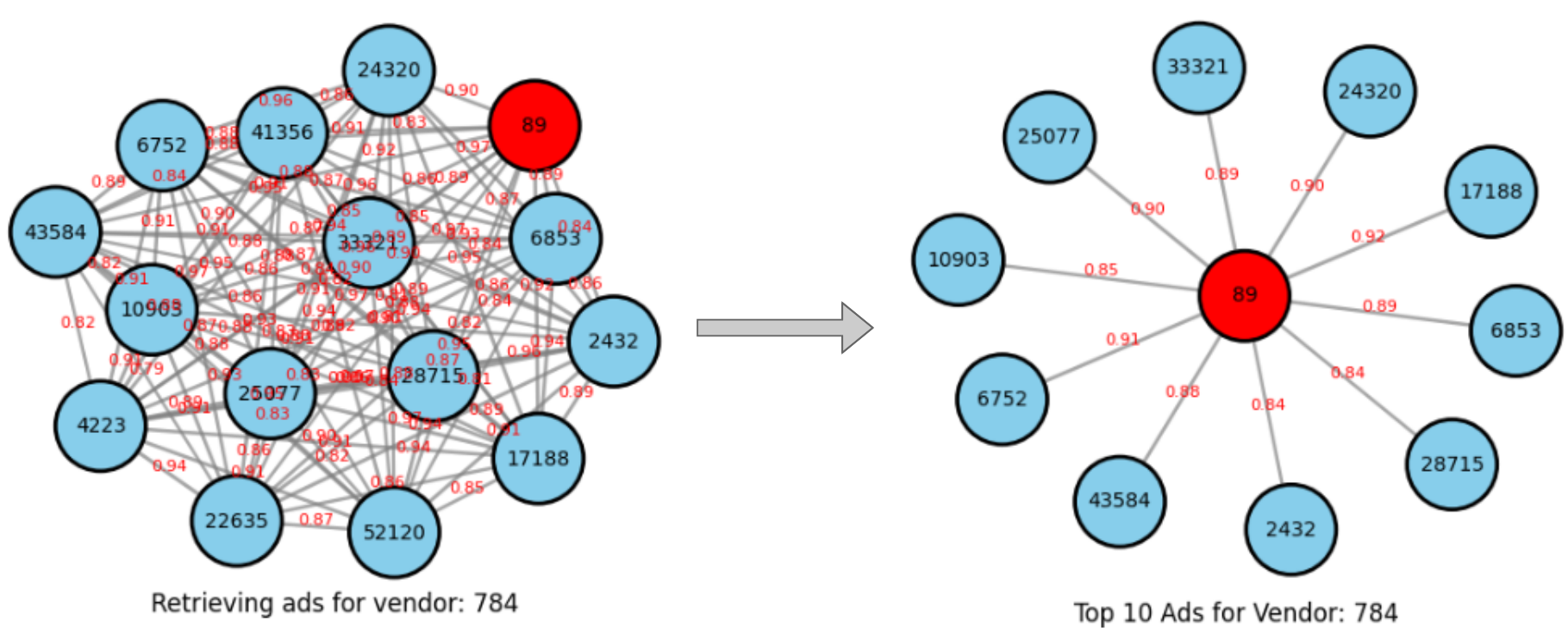

(a) Vendor 784

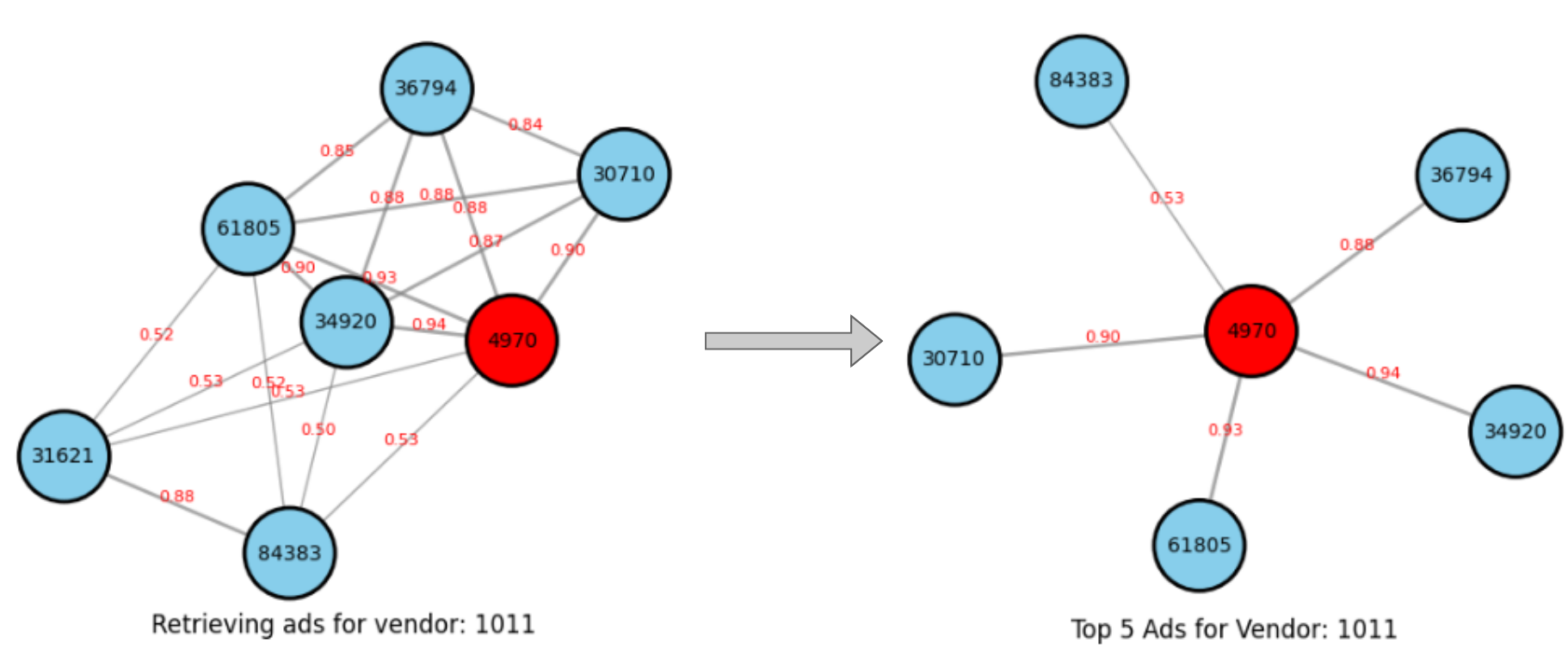

(b) Vendor 1101

Figure 4.11: Knowledge graph representation generated using AA retrieval for Vendor labels 784 and 1101 from the South region dataset. The left graph utilizes R-Precision metrics to link all relevant ads for a query ad (highlighted in red), while the right graph applies (a) MRR@10 and (b) MRR@5 to identify the top-10 and top-5 most likely relevant ads. Nodes represent advertisement IDs, and edges denote the similarity between ads, both in relation to each other and the query ad, showcasing the effectiveness of AA retrieval in constructing relational insights.

## 4.7 KEY TAKEAWAYS

**1. Dataset Value and Fine-Tuning:** Fine-tuning on the MATCHED dataset significantly enhances retrieval performance, underscoring its value and exposing the limitations of existing pre-trained checkpoints in adapting to the unique linguistic and stylistic patterns of escort ads.

**2. Benefits of Dual Objective training:** Given the dual objective of achieving in-sample and OOD distribution performance, the CE+SupCon joint objective consistently outperforms or matches other training objectives, demonstrating effectiveness and generalization. This dual focus enables models to effectively address closed-set vendor identification (linking ads to known vendors in LEA databases) and open-set vendor verification (connecting ads from emerging vendors on new platforms). While some baselines excel at one task, the benchmarks are established based on their ability to perform well across both objectives, ensuring practical utility for researchers, LEAs, and practitioners in tracking known and emerging human trafficking networks.

**3. Benefits from Multimodal Integration:** Multimodal integration significantly enhances AA performance by leveraging complementary textual and visual features to capture richer authorship patterns. Beyond quantitative improvements (Table 4.2, Table 4.3 and Figure 4.6), qualitative analysis in Figure 4.5 reveals that multimodal training improves classification performance across all vendors, including those with lower class frequency. It also better connects images without faces, performs more effectively for vendors advertising multiple escorts, and increases true positive rates while reducing false positives. These patterns are corroborated in retrieval tasks (Figure 4.7-4.10).

**4. Modality-Specific Performance:** While integrating text and vision features enhances vision retrieval performance compared to vision-only baselines, vision remains less reliable (Figure 4.6(B) and Table 4.6). Conversely, integrating vision features into text representations significantly boosts text retrieval performance, with text consistently outperforming vision and multimodal representations ( Figure 4.6(A) and (C)). This highlights the superiority of the text representations from the DeCLUTR-ViT backbone, making it the most effective option for retrieval tasks on the MATCHED dataset.

**5. OOD Generalization Performance:** While the multimodal DeCLUTR-ViT classifier achieves a strong macro-F1 score (0.9802), this performance reflects its ability to learn discriminative in-sample distribution patterns from paired text-image samples during training. However, retrieval results—particularly on OOD distribution—reveal inherent generalization challenges on the same scale. As shown in Figure 4.6 and Appendix Section 4.10, the model achieves average R-Precision scores of 0.7418 (text-to-text), 0.1518 (image-to-image), and 0.7202 (multimodal) for OOD retrieval, highlighting a notable performance gap. This discrepancy stems from the challenge of linking novel text-image combinations unseen during training. The model is trained by associating individual text descriptions with multiple images, learning stylistic and visual patterns across modalities. In OOD scenarios, the model encounters entirely new pairs, requiring it to infer authorship from subtle cross-modal cues rather than relying on memorized associations. For instance, a vendor might reuse a new image with text that shares stylistic similarities to prior ads. The model's retrieval performance under such conditions demonstrates its ability to leverage these complementary signals. This distinction between classification (closed-set identification) and retrieval (open-set verification) is critical for real-world applications. In practice, LEAs may frequently encounter OOD cases where vendors alter content across platforms or regions to evade detection. The model's design—emphasizing OOD generalization and cross-modal linking—addresses a crucial gap in AA and HT investigations, where robustness to evolving evasion tactics is essential.

**6. Vendor Overlap and Generalization:** Figure 4.2(A) highlights significant vendor overlap across the four geographic regions, raising concerns about model generalization on OOD distribution. However, similarity analysis of the datasets (Figure 4.2(a)(B)-(C)) and retrieval performance on shared versus unique vendors (Table 4.6) demonstrate that the end-to-end multimodal DeCLUTR-ViT backbone performs equally well on both shared and unique vendors. This indicates strong generalization capabilities in scenarios with overlapping or region-specific vendor activity.

**Text-Image Alignment Challenges:** To improve alignment between text descriptions and escort images, three text-alignment strategies—CLIP, CLIP with Image-Text Matching (ITM), and BLIP2—are experimented. While these models show improved retrieval performance over pre-trained checkpoints (Appendix Table 4.7–4.9), they consistently underperform compared to the end-to-end DeCLUTR-ViT baseline, even after fine-tuning for the vendor identification task. This underperformance is attributed to the low semantic overlap between noisy text (vague descriptions) and images (e.g., partial or absent faces) in escort ads (Table 4.5). Given these findings, using state-of-the-art (SoTA) multimodal models like LLaVA-OV 7B (Li et al., 2024a), Gemini Flash 8B (Team, 2024), and Pixtral 12B (Agrawal et al., 2024) presents discouragements. These models have significantly larger parameter sizes, making them impractical within the computational constraints and unfair compared to the 169M-parameter DeCLUTR-ViT backbone. Additionally, they are optimized for unrelated tasks like knowledge reasoning and Q&A, which do not align with our AA objectives. Lastly, our BLIP2 results show that projecting visual features into language space, as also used by models like LLaVA, does not resolve the alignment challenges caused by low semantic overlap. Therefore, pursuing these larger general-purpose multimodal models for our AA tasks is deemed unsuitable.

## 4.8 LIMITATIONS

**Assumption:** Similar to existing research, this study assumes that each class label corresponds to a distinct vendor during the classification task, enabling the model to leverage domain knowledge effectively. However, the manual analysis identifies cases where the trained multimodal classifier misclassifies ads, likely due to similarities in writing style and content, suggesting the possibility that multiple vendors might belong to the same entity. While definitive ground truth is unavailable to confirm this hypothesis, it represents a notable challenge in ensuring label accuracy. Improving the quality of vendor labels would likely enhance benchmark performance and enable more robust model evaluations.

**Dataset Limitations and Generalization Challenges:** This research uti-

lizes escort ads collected from the Backpage platform between December 2015 and April 2016, spanning seven U.S. cities in four regions. While the dataset provides valuable insights into MAA approaches for sex trafficking investigations, it also presents several limitations. Notably, significant vendor overlap across regions (Figure 4.2) and near-duplicate ads—challenging to identify and remove due to noise and variability—complicate the evaluation of the model's generalization capabilities. Although this study evaluates OOD generalization on shared and unique vendors, more comprehensive assessments would benefit from data collected from multiple escort platforms and diverse geographical regions to better simulate cross-platform generalization.

While the Global Organized Crime Index highlights regions worldwide for human trafficking activities, human trafficking manifests in various forms, such as labor, organ, and sex trafficking, as well as forced servitude. This research focuses specifically on addressing sex trafficking within escort advertisements. Expanding data collection beyond U.S.-based ads to encompass a wider range of geographical regions and demographics is crucial, but identifying escort platforms directly linked to human trafficking operations remains a significant challenge, as such connections often require verification through law enforcement investigations or court rulings. To date, beyond Backpage and Craigslist, only a few escort platforms have been explicitly linked to human trafficking activities.

Data collection for this study was conducted under strict ethical oversight, with approval from the ethics committee largely due to the relatively dated nature of the dataset, which reduces privacy risks. It is suspected that many victims and perpetrators have since moved from these platforms or changed their personal information to avoid identification. Furthermore, this research is not an active investigation but rather an effort to develop tools that may assist researchers, LEAs, and practitioners in identifying and linking escort ads to disrupt trafficking networks. Future work aims to explore methods for ethically collecting data from additional escort platforms—particularly those with verifiable connections to human trafficking operations—to enhance generalizability across diverse demographics and regions. This expansion will be crucial for developing more robust, globally representative AA models for human trafficking investigations. That said, the current dataset remains a valuable benchmark for future research, offering critical insights into how online vendors facilitated human trafficking on Backpage escort platforms during 2015–2016.

It will serve as a reference point for analyzing how criminal behavior and evasion tactics have evolved over time and across platforms, helping researchers, LEAs, and practitioners track shifts in trafficking strategies and adapt investigative approaches accordingly.

**Selective Feature Extraction and Fine-Tuning:** This work extracts text and vision representations exclusively from the final layers of the models, which may not fully capture nuanced features learned at earlier layers. Representations extracted from intermediate layers could yield different or potentially better outcomes. Additionally, while fine-tuning pre-trained text-image alignment models, all layers are fine-tuned uniformly, which may not be optimal. Techniques like Centered Kernel Alignment (CKA) (Kornblith et al., 2019) can provide insights into which layers learn the most relevant features, enabling more informed decisions about representation extraction and selective layer freezing during fine-tuning. Addressing these concerns is beyond the scope of this research but will be explored in future work.

**Computational Constraints:** While this research employs relatively large model architectures and advanced training strategies, it is limited by the computing resources available. Larger model architectures could potentially enhance performance across classification and retrieval tasks. However, when applied to text-image alignment tasks, the computational demands of scaling these models exceeded resource capacity. As a result, smaller, more efficient architectures were chosen to fit within computational constraints, ensuring a fair and balanced comparison across baselines. Similarly, this research relies heavily on contrastive learning objectives, and prior studies (Gao et al., 2021; Vaessen and van Leeuwen, 2024) highlight the benefits of larger batch sizes for such tasks. However, to maintain consistency and fairness among baselines, the batch size was limited to 32, as larger sizes led to memory errors, particularly with text-image alignment models. This computational limitation also influenced the decision to forego fine-tuning pre-trained CLIP and BLIP2 checkpoints, as the memory requirements for fine-tuning the BLIP2 architecture caused GPU crashes. These decisions reflect deliberate trade-offs made to ensure the reproducibility and fairness of experimental comparisons while working within resource limitations.

**Model Explainability:** While this study does not explicitly address model explainability, it remains crucial in fostering trust among researchers, practitioners, and law enforcement agencies. Prior chapters (Chapter 2 and Chapter 3) have explored interpretability using local feature attribution techniques applied to text ads. In these unimodal contexts, frameworks such as LIME (Ribeiro et al., 2016) and SHAP (Lundberg and Lee, 2017) have been widely used to explain model predictions. However, extending these methods to a multimodal setting presents additional challenges, as existing techniques struggle to capture complex interactions between textual and visual features. Furthermore, research highlights limitations in current explainability approaches, including susceptibility to adversarial attacks, network sparsity, and inconsistencies in attribution results (Das and Rad, 2020; Krishna et al., 2022). To enhance transparency and reliability in multimodal authorship attribution, future work plans to develop a robust explainability framework capable of quantifying the contributions of both modalities in decision-making processes. Such advancements are crucial in ensuring the interpretability of multimodal models in real-world investigative applications.

## 4.9 SUMMARY

This study addresses RQ2 by demonstrating the potential of multimodal authorship attribution approaches in addressing the complexities of vendor identification and verification within online escort markets. Using the novel MATCHED dataset, text-only, vision-only, and multimodal approaches are extensively benchmarked, showcasing the advantages of joint CE+SupCon multitask training objectives. Analysis reveals that this dual-objective consistently outperforms single-task approaches across in-sample and OOD distribution datasets, enabling researchers, LEAs, and practitioners to identify known vendors while linking emerging ones in new markets. Additionally, multimodal integration significantly enhances model performance by capturing complementary patterns across text and images. While text remains the dominant modality, integrating image data alongside text descriptions adds stylistic cues that enrich the model's capabilities. Among text, vision, and multimodal representations, text representations from the DeCLUTR-ViT backbone emerge as the most effective for retrieval tasks, achieving the best results across all modalities. Furthermore, pre-trained text-image alignment strategies like CLIP and BLIP2 fail to establish meaningful cross-modal connections due to low semantic overlap and ineffective use of stylistic features, highlighting the limitations of general-purpose alignment techniques in this domain. In contrast, end-to-end multitask training proves to be a more effective approach for leveraging multimodal data in AA tasks. Finally, the performance gap between pre-trained checkpoints and fine-tuned baselines underscores the importance of domain-specific adaptations and task-specific training, providing a strong foundation for future research. By addressing real-world challenges and emphasizing scalability, this study aims to equip researchers, LEAs, and practitioners with actionable tools to uncover and disrupt trafficking networks effectively.

# 4.10 APPENDIX: SUPPLEMENTARY MATERIAL

| Backbone | Loss | South | Midwest | West | Northeast | OOD Avg. | ZS Avg. |
|---|---|---|---|---|---|---|---|
| **Text MRR@10** | | | | | | | |
| **DeCLUTR** | **CE+SupCon** | 0.9290 ± 0.23 | 0.7716 ± 0.38 | 0.8145 ± 0.36 | 0.7449 ± 042 | 0.7770 ± 0.39 | |
| **End2End** | **CE** | 0.9850 ± 0.10 | 0.9693 ± 0.14 | 0.9900 ± 0.07 | 0.9778 ± 0.12 | 0.9790 ± 0.11 | - |
| **DeCLUTR-ViT** | **CE+SupCon** | **0.9866 ± 0.09** | **0.9704 ± 0.14** | **0.9932 ± 0.07** | **0.9821 ± 0.12** | **0.9819 ± 0.11** | - |
| **Aligned DeCLUTR-ViT** | **ITC** | 0.4097 ± 0.43 | 0.4289 ± 0.45 | 0.5404 ± 0.47 | 0.5034 ± 0.47 | - | 0.4909 ± 0.46 |
| | **ITC+ITM** | 0.8192 ± 0.37 | 0.7990 ± 0.39 | 0.8600 ± 0.35 | 0.5914 ± 0.48 | - | 0.7674 ± 0.40 |
| | **BLIP2** | 0.7551 ± 0.41 | 0.7226 ± 0.44 | 0.8400 ± 0.37 | 0.5376 ± 0.49 | - | 0.7140 ± 0.43 |
| | **BLIP2-Cond** | 0.7672 ± 0.41 | 0.7203 ± 0.44 | 0.8400 ± 0.37 | 0.4946 ± 0.49 | - | 0.7055 ± 0.43 |
| **Fine-tuned DeCLUTR-ViT** | **ITC+CE** | 0.8613 ± 0.34 | 0.6623 ± 0.46 | 0.8600 ± 0.35 | 0.6263 ± 0.48 | 0.7162 ± 0.43 | - |
| | **ITC+ITM+CE** | 0.4239 ± 0.39 | 0.2851 ± 0.37 | 0.3417 ± 0.42 | 0.3600 ± 0.41 | 0.3289 ± 0.40 | - |
| | **BLIP2+CE** | 0.8866 ± 0.30 | 0.7226 ± 0.44 | 0.8400 ± 0.37 | 0.7292 ± 0.44 | 0.7639 ± 0.42 | - |
| | **BLIP2+CE+SupCon** | 0.8886 ± 0.31 | 0.7397 ± 0.43 | 0.8600 ± 0.35 | 0.7604 ± 0.42 | 0.7867 ± 0.40 | - |
| **Text R-Precision@X** | | | | | | | |
| **DeCLUTR** | **CE+SupCon** | 0.8706 ± 0.24 | 0.6264 ± 0.38 | 0.7339 ± 0.37 | 0.6699 ± 0.41 | 0.6767 ± 0.39 | |
| **End2End** | **CE** | 0.8687 ± 0.19 | 0.6500 ± 0.30 | 0.7934 ± 0.24 | 0.7300 ± 0.28 | 0.7245 ± 0.27 | - |
| **DeCLUTR-ViT** | **CE+SupCon** | **0.9193 ± 0.16** | **0.6612 ± 0.31** | **0.8008 ± 0.25** | **0.7365 ± 0.28** | **0.7418 ± 0.28** | - |
| **Aligned DeCLUTR-ViT** | **ITC** | 0.2337 ± 0.28 | 0.2936 ± 0.34 | 0.4035 ± 0.37 | 0.3779 ± 0.38 | - | 0.3583 ± 0.36 |
| | **ITC+ITM** | 0.4964 ± 0.34 | 0.5679 ± 0.38 | 0.7093 ± 0.33 | 0.4818 ± 0.45 | - | 0.5639 ± 0.38 |
| | **BLIP2** | 0.4230 ± 0.34 | 0.5094 ± 0.39 | 0.6354 ± 0.37 | 0.3913 ± 0.41 | - | 0.4898 ± 0.38 |
| | **BLIP2-Cond** | 0.4341 ± 0.35 | 0.5142 ± 0.39 | 0.6644 ± 0.36 | 0.3729 ± 0.42 | - | 0.4964 ± 0.38 |
| **Fine-tuned DeCLUTR-ViT** | **ITC+CE** | 0.6378 ± 0.33 | 0.4885 ± 0.37 | 0.6825 ± 0.35 | 0.3770 ± 0.35 | 0.5160 ± 0.36 | - |
| | **ITC+ITM+CE** | 0.1462 ± 0.19 | 0.0818 ± 0.14 | 0.1292 ± 0.18 | 0.1359 ± 0.19 | 0.1156 ± 0.17 | - |
| | **BLIP2+CE** | 0.7131 ± 0.32 | 0.5569 ± 0.39 | 0.7280 ± 0.36 | 0.5627 ± 0.41 | 0.6159 ± 0.39 | - |
| | **BLIP2+CE+SupCon** | 0.7632 ± 0.32 | 0.5666 ± 0.40 | 0.7652 ± 0.31 | 0.5869 ± 0.40 | 0.6362 ± 0.37 | - |
| **Text Macro-F1@X** | | | | | | | |
| **DeCLUTR** | **CE+SupCon** | 0.9102 ± 0.21 | 0.6162 ± 0.37 | 0.7169 ± 0.33 | 0.6879 ± 0.40 | 0.6737 ± 0.37 | |
| **End2End** | **CE** | 0.8726 ± 0.20 | 0.5653 ± 0.33 | 0.7374 ± 0.26 | 0.7261 ± 0.31 | 0.6763 ± 0.30 | - |
| **DeCLUTR-ViT** | **CE+SupCon** | **0.9433 ± 0.16** | **0.5819 ± 0.34** | **0.7466 ± 0.26** | **0.7242 ± 0.31** | **0.6841 ± 0.30** | - |
| **Aligned DeCLUTR-ViT** | **ITC** | 0.3039 ± 0.31 | 0.3756 ± 0.35 | 0.4887 ± 0.33 | 0.4173 ± 0.39 | - | 0.4272 ± 0.36 |
| | **ITC+ITM** | 0.5079 ± 0.32 | 0.5659 ± 0.36 | 0.7281 ± 0.29 | 0.5136 ± 0.44 | - | 0.5946 ± 0.35 |
| | **BLIP2** | 0.4283 ± 0.33 | 0.5279 ± 0.38 | *0.6552 ± 0.34* | 0.4216 ± 0.39 | - | 0.5605 ± 0.38 |
| | **BLIP2-Cond** | 0.4356 ± 0.34 | 0.5251 ± 0.38 | 0.6720 ± 0.34 | 0.4249 ± 0.42 | - | 0.5125 ± 0.38 |
| **Fine-tuned DeCLUTR-ViT** | **ITC+CE** | 0.6805 ± 0.32 | 0.5054 ± 0.37 | 0.6877 ± 0.32 | 0.3790 ± 0.34 | 0.5240 ± 0.34 | - |
| | **ITC+ITM+CE** | 0.1438 ± 0.20 | 0.0748 ± 0.12 | 0.1218 ± 0.18 | 0.1214 ± 0.17 | 0.1060 ± 0.16 | - |
| | **BLIP2+CE** | 0.7215 ± 0.31 | 0.5774 ± 0.38 | 0.7391 ± 0.32 | 0.5499 ± 0.39 | 0.6221 ± 0.36 | - |
| | **BLIP2+CE+SupCon** | 0.7879 ± 0.29 | 0.5762 ± 0.39 | 0.7482 ± 0.29 | 0.5912 ± 0.38 | 0.6385 ± 0.35 | - |

Table 4.7: Comparison of text-to-text retrieval performance for the multimodal, DeCLUTR-ViT backbone, evaluated on the text-only modality using MRR@10, R-Precision@X, and Macro-F1@X metrics. The DeCLUTR-small model serves as the text-only baseline. End2End baselines denote DeCLUTR-ViT models trained directly for vendor identification tasks, while Aligned baselines represent DeCLUTR-ViT backbone pre-trained for text-image alignment tasks using ITC, ITC+ITM, and BLIP2 objectives. Fine-tuned baselines build upon pre-trained aligned models by fine-tuning them for vendor identification tasks on the South region ads.

| Backbone | Loss | South | Midwest | West | Northeast | OOD Avg. | ZS Avg. |
|---|---|---|---|---|---|---|---|
| **Vision MRR@10** | | | | | | | |
| **ViT** | **CE+SupCon** | 0.3464 ± 0.37 | 0.3749 ± 0.40 | 0.4049 ± 0.42 | 0.4330 ± 0.42 | 0.4041 ± 0.41 | - |
| **End2End DeCLUTR-ViT** | **CE** | 0.2257 ± 0.33 | 0.1716 ± 0.32 | 0.2142 ± 0.35 | 0.1866 ± 0.32 | 0.2575 ± 0.33 | - |
| | **CE+SupCon** | **0.4045 ± 0.38** | **0.3905 ± 0.40** | **0.4603 ± 0.45** | **0.4521 ± 0.42** | **0.4343 ± 0.42** | - |
| **Aligned DeCLUTR-ViT** | **ITC** | 0.2329 ± 0.30 | 0.2336 ± 0.33 | 0.2984 ± 0.39 | 0.2964 ± 0.37 | - | 0.2761 ± 0.36 |
| | **ITC+ITM** | 0.3281 ± 0.37 | 0.3434 ± 0.39 | 0.3683 ± 0.43 | 0.3442 ± 0.40 | - | 0.3324 ± 0.38 |
| | **BLIP2** | 0.2119 ± 0.32 | 0.2055 ± 0.33 | 0.2674 ± 0.40 | 0.2858 ± 0.39 | - | 0.2425 ± 0.36 |
| | **BLIP2-Cond** | 0.2049 ± 0.32 | 0.1855 ± 0.31 | 0.2488 ± 0.39 | 0.2450 ± 0.36 | - | 0.2211 ± 0.35 |
| **Fine-tuned DeCLUTR-ViT** | **ITC** | 0.4157 ± 0.38 | 0.3512 ± 0.39 | 0.3818 ± 0.43 | 0.3792 ± 0.41 | 0.3707 ± 0.41 | - |
| | **ITC+ITM** | 0.4239 ± 0.39 | 0.2851 ± 0.37 | 0.3417 ± 0.42 | 0.3600 ± 0.41 | 0.3289 ± 0.40 | - |
| | **BLIP2** | 0.3677 ± 0.38 | 0.2629 ± 0.36 | 0.3229 ± 0.41 | 0.3128 ± 0.39 | 0.2995 ± 0.39 | - |
| | **BLIP2-CE+SupCon** | 0.3470 ± 0.38 | 0.2542 ± 0.35 | 0.3026 ± 0.41 | 0.3312 ± 0.39 | 0.2960 ± 0.39 | - |
| **Vision R-Precision@X** | | | | | | | |
| **ViT** | **CE+SupCon** | 0.1064 ± 0.16 | 0.1095 ± 0.16 | 0.1519 ± 0.20 | 0.1685 ± 0.21 | 0.1433 ± 0.19 | - |
| **End2End DeCLUTR-ViT** | **CE** | 0.0862 ± 0.16 | 0.0567 ± 0.12 | 0.0915 ± 0.14 | 0.0676 ± 0.11 | 0.0719 ± 0.12 | - |
| | **CE+SupCon** | **0.1115 ± 0.15** | **0.1141 ± 0.16** | **0.1768 ± 0.21** | **0.1646 ± 0.19** | **0.1518 ± 0.19** | - |
| **Aligned DeCLUTR-ViT** | **ITC** | 0.0537 ± 0.09 | 0.0752 ± 0.13 | 0.1275 ± 0.17 | 0.1143 ± 0.16 | - | 0.1057 ± 0.16 |
| | **ITC+ITM** | 0.0650 ± 0.10 | 0.0826 ± 0.14 | 0.1218 ± 0.17 | 0.1003 ± 0.14 | - | 0.0924 ± 0.14 |
| | **BLIP2** | 0.0645 ± 0.15 | 0.0641 ± 0.13 | 0.1197 ± 0.20 | 0.1492 ± 0.24 | - | 0.0994 ± 0.18 |
| | **BLIP2-Cond** | 0.0563 ± 0.13 | 0.0569 ± 0.13 | 0.1001 ± 0.18 | 0.1115 ± 0.20 | - | 0.0812 ± 0.16 |
| **Fine-tuned DeCLUTR-ViT** | **ITC** | 0.1247 ± 0.17 | 0.0957 ± 0.15 | 0.1461 ± 0.18 | 0.1383 ± 0.17 | 0.1267 ± 0.17 | - |
| | **ITC+ITM** | 0.1462 ± 0.19 | 0.0818 ± 0.14 | 0.1292 ± 0.18 | 0.1359 ± 0.19 | 0.1156 ± 0.17 | - |
| | **BLIP2** | 0.1370 ± 0.19 | 0.0775 ± 0.14 | 0.1217 ± 0.18 | 0.1393 ± 0.21 | 0.1128 ± 0.18 | - |
| | **BLIP2-CE+SupCon** | 0.1256 ± 0.19 | 0.0777 ± 0.14 | 0.1228 ± 0.17 | 0.1414 ± 0.20 | 0.1140 ± 0.17 | - |
| **Vision Macro-F1@X** | | | | | | | |
| **ViT** | **CE+SupCon** | 0.1296 ± 0.21 | 0.0948 ± 0.14 | 0.1460 ± 0.20 | 0.1497 ± 0.20 | 0.1302 ± 0.18 | - |
| **End2End DeCLUTR-ViT** | **CE** | 0.1028 ± 0.21 | 0.0600 ± 0.11 | 0.0960 ± 0.15 | 0.0657 ± 0.11 | 0.0859 ± 0.14 | - |
| | **CE+SupCon** | **0.1152 ± 0.17** | **0.1049 ± 0.14** | **0.1739 ± 0.21** | **0.1493 ± 0.18** | **0.1427 ± 0.19** | - |
| **Aligned DeCLUTR-ViT** | **ITC** | 0.0689 ± 0.11 | 0.0892 ± 0.14 | 0.1415 ± 0.19 | 0.1072 ± 0.15 | - | 0.1118 ± 0.18 |
| | **ITC+ITM** | 0.0614 ± 0.10 | 0.0675 ± 0.11 | 0.1070 ± 0.15 | 0.0933 ± 0.13 | - | 0.0837 ± 0.13 |
| | **BLIP2** | 0.0938 ± 0.20 | 0.0908 ± 0.17 | 0.1281 ± 0.22 | 0.1458 ± 0.24 | - | 0.1146 ± 0.21 |
| | **BLIP2-Cond** | 0.0805 ± 0.18 | 0.0776 ± 0.16 | 0.1074 ± 0.20 | 0.1088 ± 0.19 | - | 0.0936 ± 0.18 |
| **Fine-tuned DeCLUTR-ViT** | **ITC** | 0.1319 ± 0.18 | 0.0914 ± 0.14 | 0.1485 ± 0.20 | 0.1333 ± 0.17 | 0.1244 ± 0.17 | - |
| | **ITC+ITM** | 0.1438 ± 0.20 | 0.0748 ± 0.12 | 0.1218 ± 0.18 | 0.1214 ± 0.17 | 0.1060 ± 0.16 | - |
| | **BLIP2** | 0.1517 ± 0.23 | 0.0837 ± 0.14 | 0.1277 ± 0.19 | 0.1367 ± 0.20 | 0.1160 ± 0.18 | - |
| | **BLIP2-CE+SupCon** | 0.1526 ± 0.24 | 0.0799 ± 0.14 | 0.1276 ± 0.19 | 0.1335 ± 0.20 | 0.1137 ± 0.18 | - |

Table 4.8: Comparison of image-to-image retrieval performance for the multimodal, DeCLUTR-ViT backbone, evaluated on the vision-only modality using MRR@10, R-Precision@X, and Macro-F1@X metrics. The ViT-base-patch16-244 model serves as the vision-only baseline. End2End baselines denote DeCLUTR-ViT models trained directly for vendor identification tasks, while Aligned baselines represent DeCLUTR-ViT backbone pre-trained for text-image alignment tasks using ITC, ITC+ITM, and BLIP2 objectives. Fine-tuned baselines build upon pre-trained aligned models by fine-tuning them for vendor identification tasks on the South region ads.

| Backbone | Loss | South | Midwest | West | Northeast | OOD Avg. | ZS Avg. |
|---|---|---|---|---|---|---|---|
| **Multimodal MRR@10** | | | | | | | |
| **End2End** | **CE** | 0.9669 ± 0.13 | 0.9297 ± 0.20 | 0.9592 ± 0.17 | 0.9650 ± 0.14 | 0.9513 ± 0.17 | - |
| **DeCLUTR-ViT** | **CE+SupCon** | **0.9859 ± 0.10** | **0.9658 ± 0.15** | **0.9834 ± 0.11** | **0.9735 ± 0.13** | **0.9742 ± 0.13** | - |
| **Aligned DeCLUTR-ViT** | **ITC** | 0.6574 ± 0.35 | 0.6822 ± 0.36 | 0.7396 ± 0.36 | 0.6750 ± 0.38 | - | 0.6886 ± 0.36 |
| | **ITC+ITM** | 0.9375 ± 0.18 | 0.9389 ± 0.19 | 0.9601 ± 0.16 | 0.9715 ± 0.14 | - | 0.9520 ± 0.17 |
| | **BLIP2** | 0.6142 ± 0.36 | 0.6136 ± 0.39 | 0.6108 ± 0.41 | 0.5921 ± 0.42 | - | 0.6077 ± 0.40 |
| | **BLIP2-Cond** | 0.6052 ± 0.36 | 0.6006 ± 0.39 | 0.5975 ± 0.41 | 0.5657 ± 0.42 | - | 0.5923 ± 0.40 |
| **Fine-tuned DeCLUTR-ViT** | **ITC** | 0.9650 ± 0.13 | 0.8331 ± 0.29 | 0.7313 ± 0.36 | 0.7641 ± 0.34 | 0.7762 ± 0.33 | - |
| | **ITC+ITM** | 0.9739 ± 0.12 | 0.9285 ± 0.20 | 0.9498 ± 0.19 | 0.9655 ± 0.15 | 0.9480 ± 0.23 | - |
| | **BLIP2** | 0.9774 ± 0.11 | 0.9378 ± 0.20 | 0.9559 ± 0.18 | 0.9690 ± 0.14 | 0.9542 ± 0.17 | - |
| | **BLIP2-CE+SupCon** | 0.9814 ± 0.10 | 0.9426 ± 0.19 | 0.9648 ± 0.15 | 0.9759 ± 0.12 | 0.9602 ± 0.19 | |
| **Multimodal R-Precision@X** | | | | | | | |
| **End2End** | **CE** | 0.8040 ± 0.20 | 0.6217 ± 0.26 | 0.7429 ± 0.24 | 0.6980 ± 0.27 | 0.6875 ± 0.26 | - |
| **DeCLUTR-ViT** | **CE+SupCon** | **0.9248 ± 0.14** | **0.6567 ± 0.30** | **0.7861 ± 0.25** | **0.7178 ± 0.30** | **0.7202 ± 0.28** | - |
| **Aligned DeCLUTR-ViT** | **ITC** | 0.1797 ± 0.16 | 0.2373 ± 0.20 | 0.3330 ± 0.23 | 0.3076 ± 0.24 | - | 0.2644 ± 0.21 |
| | **ITC+ITM** | 0.4939 ± 0.24 | 0.5705 ± 0.26 | 0.7046 ± 0.23 | 0.6747 ± 0.26 | - | 0.6109 ± 0.25 |
| | **BLIP2** | 0.1708 ± 0.22 | 0.1847 ± 0.22 | 0.2182 ± 0.24 | 0.2841 ± 0.32 | - | 0.2145 ± 0.25 |
| | **BLIP2-Cond** | 0.1455 ± 0.20 | 0.1602 ± 0.20 | 0.1830 ± 0.21 | 0.2324 ± 0.29 | - | 0.1803 ± 0.23 |
| **Fine-tuned DeCLUTR-ViT** | **ITC** | 0.7377 ± 0.21 | 0.3716 ± 0.22 | 0.2844 ± 0.22 | 0.3700 ± 0.26 | 0.3420 ± 0.23 | - |
| | **ITC+ITM** | 0.7282 ± 0.22 | 0.4968 ± 0.23 | 0.6109 ± 0.23 | 0.6419 ± 0.27 | 0.5832 ± 0.24 | - |
| | **BLIP2** | 0.7723 ± 0.2 | 0.5524 ± 0.25 | 0.6759 ± 0.23 | 0.6691 ± 0.27 | 0.6325 ± 0.25 | - |
| | **BLIP2-CE+SupCon** | 0.7950 ± 0.19 | 0.5564 ± 0.25 | 0.6943 ± 0.23 | 0.6809 ± 0.26 | 0.6524 ± 0.25 | |
| **Multimodal Macro-F1@X** | | | | | | | |
| **End2End** | **CE** | 0.8294 ± 0.21 | 0.5618 ± 0.29 | 0.7408 ± 0.24 | 0.7053 ± 0.29 | 0.6693 ± 0.27 | - |
| **DeCLUTR-ViT** | **CE+SupCon** | **0.9595 ± 0.12** | **0.5671 ± 0.33** | **0.7560 ± 0.26** | **0.7333 ± 0.30** | **0.6855 ± 0.29** | - |
| **Aligned DeCLUTR-ViT** | **ITC** | 0.2519 ± 0.23 | 0.3254 ± 0.26 | 0.4687 ± 0.27 | 0.3493 ± 0.26 | - | 0.3488 ± 0.26 |
| | **ITC+ITM** | 0.4809 ± 0.27 | 0.5239 ± 0.28 | 0.7023 ± 0.23 | 0.6934 ± 0.27 | - | 0.6001 ± 0.26 |
| | **BLIP2** | 0.3263 ± 0.35 | 0.3408 ± 0.35 | 0.4612 ± 0.37 | 0.4190 ± 0.38 | - | 0.3868 ± 0.37 |
| | **BLIP2-Cond** | 0.2724 ± 0.32 | 0.2850 ± 0.32 | 0.3649 ± 0.33 | 0.3353 ± 0.35 | - | 0.3144 ± 0.33 |
| **Fine-tuned DeCLUTR-ViT** | **ITC** | 0.7698 ± 0.23 | 0.4008 ± 0.25 | 0.4003 ± 0.27 | 0.3881 ± 0.28 | 0.3964 ± 0.27 | - |
| | **ITC+ITM** | 0.7313 ± 0.25 | 0.4538 ± 0.26 | 0.6275 ± 0.24 | 0.6591 ± 0.28 | 0.5801 ± 0.27 | - |
| | **BLIP2** | 0.7973 ± 0.22 | 0.5325 ± 0.28 | 0.7050 ± 0.24 | 0.6944 ± 0.29 | 0.6440 ± 0.27 | - |
| | **BLIP2-CE+SupCon** | 0.8487 ± 0.20 | 0.5446 ± 0.29 | 0.7250 ± 0.24 | 0.7077 ± 0.29 | 0.6591 ± 0.27 | |

Table 4.9: Comparison of multimodal retrieval performance for the DeCLUTR-ViT backbone evaluated on the multimodal (text and image) ads using MRR@10, R-Precision@X, and Macro-F1@X metrics. The End2End baselines represent the DeCLUTR-ViT backbone trained directly on the vendor identification task, while the Pre-trained baselines involve an image-text alignment task aligning text and images from the same advertisements. The Fine-tuned baselines build upon the Pre-trained models by performing vendor identification on the South region multimodal ads.

| Loss | South | Midwest | West | Northeast | OOD Avg. | ZS Avg. |
|---|---|---|---|---|---|---|
| **MRR@10** | | | | | | |
| **Pre-trained** | 0.2248 ± 0.30 | 0.2866 ± 0.36 | 0.3479 ± 0.41 | 0.3385 ± 0.38 | - | 0.2995 ± 0.36 |
| **CE** | 0.7445 ± 0.39 | 0.5703 ± 0.46 | 0.6394 ± 0.45 | 0.5862 ± 0.48 | 0.5986 ± 0.46 | - |
| **Triplet** | 0.4282 ± 0.45 | 0.3200 ± 0.43 | 0.4074 ± 0.46 | 0.3503 ± 0.45 | 0.3592 ± 0.45 | - |
| **SupCon** | 0.8829 ± 0.29 | 0.7636 ± 0.39 | 0.8331 ± 0.35 | 0.7520 ± 0.42 | 0.7829 ± 0.39 | - |
| **CE+Triplet** | 0.8891 ± 0.28 | 0.6410 ± 0.45 | 0.6969 ± 0.43 | 0.6561 ± 0.45 | 0.6647 ± 0.44 | - |
| **CE+SupCon** | **0.9290 ± 0.23** | **0.7716 ± 0.38** | **0.8145 ± 0.36** | **0.7449 ± 0.42** | **0.7770 ± 0.39** | - |
| **R-Precision@X** | | | | | | |
| **Pre-trained** | 0.3265 ± 0.47 | 0.3943 ± 0.49 | 0.3139 ± 0.46 | 0.4037 ± 0.49 | - | 0.3596 ± 0.48 |
| **CE** | 0.5557 ± 0.36 | 0.4596 ± 0.40 | 0.5842 ± 0.41 | 0.4944 ± 0.43 | 0.5127 ± 0.41 | - |
| **Triplet** | 0.3200 ± 0.34 | 0.2443 ± 0.33 | 0.3365 ± 0.38 | 0.3032 ± 0.38 | 0.2947 ± 0.36 | - |
| **SupCon** | 0.7673 ± 0.29 | 0.6346 ± 0.37 | 0.7612 ± 0.35 | 0.6707 ± 0.41 | 0.6888 ± 0.38 | - |
| **CE+Triplet** | 0.8055 ± 0.30 | 0.5000 ± 0.40 | 0.5890 ± 0.4 | 0.5410 ± 0.42 | 0.5433 ± 0.41 | - |
| **CE+SupCon** | **0.8706 ± 0.24** | **0.6264 ± 0.38** | **0.7339 ± 0.37** | **0.6699 ± 0.41** | **0.6767 ± 0.39** | - |
| **Macro-F1@X** | | | | | | |
| **Pre-trained** | 0.2224 ± 0.30 | 0.2804 ± 0.36 | 0.2731 ± 0.36 | 0.3801 ± 0.39 | - | 0.2890 ± 0.37 |
| **CE** | 0.6098 ± 0.35 | 0.4760 ± 0.38 | 0.6123 ± 0.35 | 0.5042 ± 0.42 | 0.5308 ± 0.38 | - |
| **Triplet** | 0.4135 ± 0.37 | 0.2892 ± 0.35 | 0.4337 ± 0.35 | 0.3121 ± 0.39 | 0.3450 ± 0.36 | - |
| **SupCon** | 0.8157 ± 0.27 | 0.6333 ± 0.36 | 0.7408 ± 0.31 | 0.6950 ± 0.39 | 0.6897 ± 0.35 | - |
| **CE+Triplet** | 0.8680 ± 0.26 | 0.5198 ± 0.39 | 0.5789 ± 0.35 | 0.5612 ± 0.41 | 0.5533 ± 0.38 | - |
| **CE+SupCon** | **0.9102 ± 0.21** | **0.6162 ± 0.37** | **0.7169 ± 0.33** | **0.6879 ± 0.40** | **0.6737 ± 0.37** | - |

Table 4.10: Comparison of text-to-text retrieval performance for the text-only benchmark, DeCLUTR-small backbone, with different objectives (losses), evaluated across MRR@10, R-Precision@X, and Macro-F1@X metrics.

# 5

# Responsible Guidelines for Authorship Attribution Tasks in Sensitive Domains

**This chapter is based on the following research:**

- **Vageesh Saxena**, Aurelia Tamò-Larrieux, Gijs van Dijck, and Gerasimos Spanakis. 2025a. Responsible Guidelines for Authorship Attribution Tasks in NLP. Ethics and Information Technology 27, 16 (2025), Springer Nature.

While Chapter 2–4 present technical advancements in AA methods for high-risk and sensitive domains—such as darknet forums and online escort marketplaces—they do not address the ethical foundations necessary to guide the responsible development and deployment of such systems. Although numerous ethical frameworks exist within ML and NLP, few directly confront the unique challenges and trade-offs inherent to AA applications, particularly in domains where privacy, consent, and potential misuse are critical concerns. This chapter aims to bridge that gap by outlining a set of responsible guidelines for creating and using NLP-based AA datasets and models. Special emphasis is placed on aligning AA research with ethical principles related to privacy, fairness, transparency, accountability, and harm prevention. While the scope of this chapter is centered on NLP-based AA applications, the proposed guidelines are intended to serve as a foundation for future extensions to vision-based and multimodal AA systems.

## 5.1 Introduction

AA approaches in NLP involve analyzing stylometric and linguistic features from textual segments of multiple authors, focusing on word choice, syntactic patterns, and writing styles. Recent advancements in NLP have enabled the development of ML approaches for authorship verification (Koppel and Schler, 2004; Bevendorff et al., 2022; Lei et al., 2022b) and identification (Mohsen et al., 2016; Kale and Prasad, 2017; Yülüce and Dalkılıç, 2022). These techniques are applied across various domains, including forensic linguistics (Iqbal et al., 2008; Yang and Chow, 2014a; Fobbe, 2020), cybersecurity (Nirkhi and Dr.R.V.Dharaskar, 2013), cybercrime detection (Zhang et al., 2019; Zheng et al., 2003; Kumar et al., 2020; Manolache et al., 2022), and anomaly detection (Neme et al., 2011b; Boukhaled and Ganascia, 2014), among others.

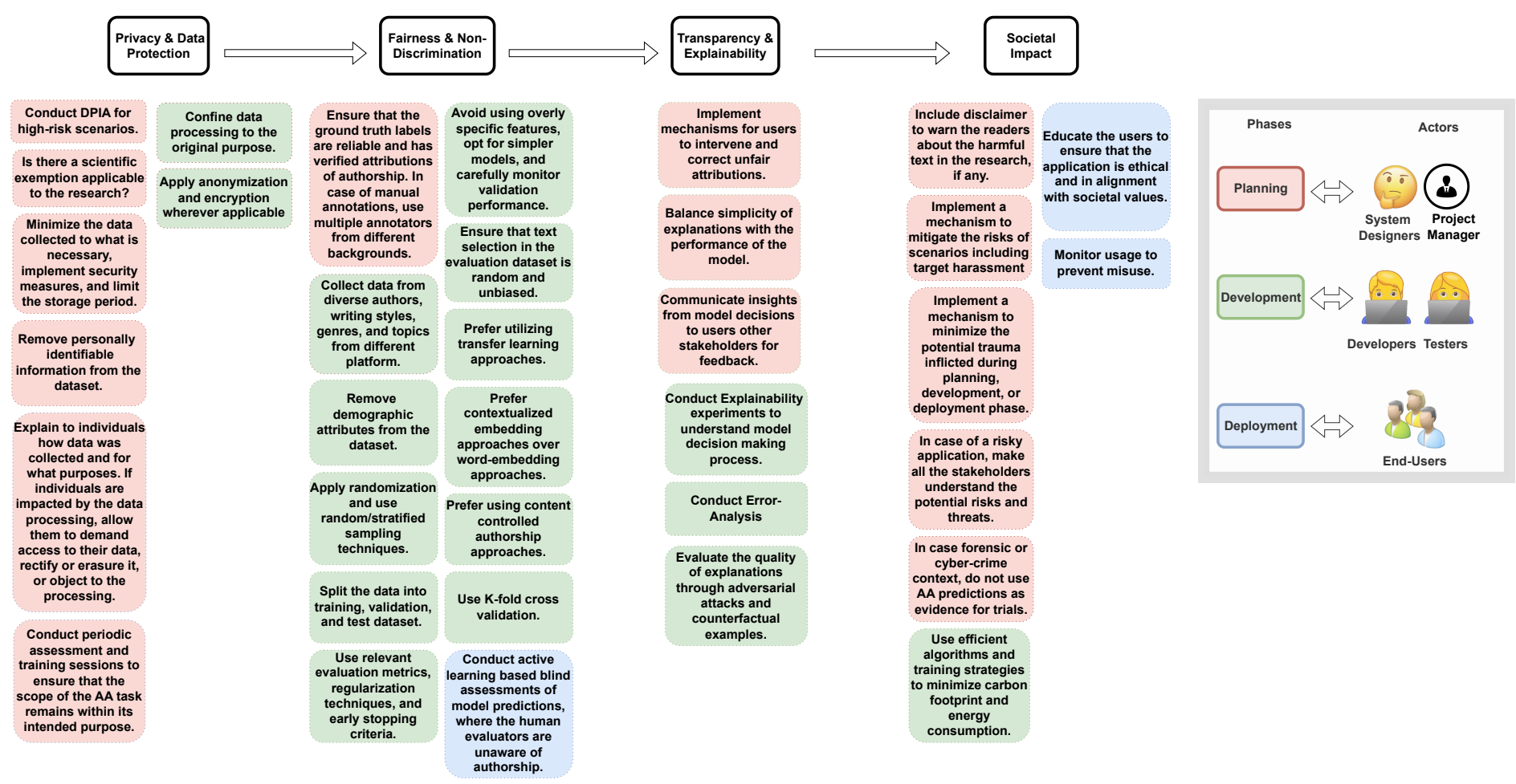

Figure 5.1: The framework of responsible guidelines for the Authorship Attribution (AA) approaches in Natural Language Processing (NLP), encompassing aspects like Privacy & Data Protection, Fairness & Non-discrimination, Transparency & Explainability, and Social Impact. The guidelines are established throughout the Design and Planning, Development and Validation, and Deployment and Feedback phases of the Software Development Life Cycle (SDLC), involving System Designers, Project Managers, Developers, QA Testers, and End-Users as key stakeholders.

AA methods have evolved from traditional stylometric methods to sophisticated ML and DL models, leveraging data to identify unique linguistic fingerprints. Stylometric approaches rely on lexical, syntactic, and character-level features such as word frequency, grammar patterns, and n-grams, offering effective solutions for small datasets where interpretability is crucial (Prasad et al., 2015). As textual data complexity increases, ML approaches like Support Vector Machines, Naive Bayes, and Random Forests automate feature learning, enabling scalability for larger datasets (Diederich et al., 2003; Altheneyan and Menai, 2014; Khonji et al., 2015). These methods combine linguistic markers for optimal performance, but their success depends on high-quality training data and careful feature engineering (Stamatatos, 2009b; Sapkota et al., 2015b). On the other hand, DL models such as RNNs, LSTMs (Jafariakinabad et al., 2019), CNNs (Zhang et al., 2015), and transformers like BERT (Devlin et al., 2019a) and GPT (Yenduri et al., 2024) excel at capturing hierarchical and nuanced text patterns, making them highly effective for large-scale tasks while reducing the need for manual feature extraction. However, these DL models often lack transparency, posing challenges in domains like forensic linguistics, where interpretability is critical (Rudin, 2019). In contrast, traditional and ML methods remain relevant in scenarios requiring lower computational resources or high explainability (Strubell et al., 2020).

While AA methods advance applications in politics, law enforcement, and cybersecurity, they raise ethical and societal concerns. Risks include unauthorized surveillance, identity exposure, and bias against demographic groups, which may lead to systemic discrimination, misuse, or reputational harm (Juola, 2020; Lund et al., 2023). Transparency and accountability are particularly challenging in high-stakes scenarios, such as legal or academic investigations (Boenninghoff et al., 2019b), and AA systems can be weaponized to suppress dissent or manipulate public opinion (Shamsi et al., 2016). Given these risks, the development and deployment of AA technologies must be grounded in strong ethical foundations, rigorous oversight, and a commitment to minimizing harm. Research by Jobin et al. (2019); Benzie and Montasari (2023) highlights key ethical principles, including Transparency, Justice and Fairness, Non-Maleficence, Responsibility, and Privacy. Adapting these principles, this study establishes guidelines tailored for stakeholders involved in the Software Development Life Cycle (SDLC) (Ruparelia, 2010) of AA tasks. These stakeholders include system designers, project managers, developers, testers, and end-users, encompassing the three phases: Design and Planning, Development and Validation, and Deployment. Based on existing state-of-the-art practices, the proposed framework focuses on four key areas: privacy and data protection, fairness and non-discrimination, transparency and explainability, and consideration of broader societal impact. By addressing these as-pects, the framework consolidates responsible AA practices in NLP, guiding stakeholders in understanding and managing the inherent trade-offs in AA tasks. Figure 5.1 illustrates the framework overview as a roadmap for implementing responsible AA practices. Finally, to demonstrate the practical application of the framework, the guidelines are applied to the published IDTraffickers research (Saxena et al., 2023a), which aims to identify potential human-trafficking vendors in online escort markets. [1]

## 5.2 Literature Research

### 5.2.1 Foundation of the Framework

Ethical considerations in AI development have been rigorously debated, with scholars identifying critical challenges and foundational principles. A foundational study by Jobin et al. (2019) outlines core ethical priorities for AI, including Transparency, Justice and Fairness, Non-Maleficence, Responsibility, and Privacy. In their research, Jobin et al. (2019); Benzie and Montasari (2023) examines the translation of these principles into AI systems, emphasizing their role in fostering responsible innovation. Transparency, for instance, requires that AA processes, algorithms, and data usage be clear to users and stakeholders. The decision-making process must be explainable, especially when determining the author of a sensitive artifact, with insights into how conclusions were drawn, which features were most effective, and what data was utilized. Justice and fairness necessitate mitigating biases, particularly when AA models trained on skewed datasets risk misrepresenting underrepresented groups. Nonmaleficence prioritizes harm prevention, such as avoiding erroneous authorship claims in legal contexts that could lead to wrongful accusations. Responsibility mandates accountability among stakeholders, whereas privacy safeguards are vital to protect individuals from unintended exposure through AA techniques. While existing ethical AI frameworks (Fjeld et al., 2020; Loi and Spielkamp, 2021; Loi et al., 2021; John Albert and Müller, 2022; Mollen, 2023; Charles Radclyffe, 2023) provide broad guidance, their relevance to AA-specific challenges remains underdeveloped, underscoring the necessity for a tailored framework.

Following Jobin et al. (2019), this study adopts the ethical principles for cyber-enabled AA approaches by emphasizing four key responsible pillars: Privacy and Data Protection, Fairness and Non-Discrimination, Transparency and Explainability, and Societal Impact. The principle of Privacy and Data Protection aligns with the General Data Protection Regulation (GDPR) (Voigt and Bussche, 2017), addressing privacy concerns in AA applications through data minimization, limited processing, purpose limitation, and responsible handling of personal information (Klymenko et al., 2022; Habernal et al., 2023; Sousa and Kern, 2023). Fairness and Non-Discrimination align with the principles of Justice and Fairness, ensuring that AA models avoid perpetuating biases or discriminating against specific groups (Blodgett et al., 2020; Czarnowska et al., 2021; Halvani and Graner, 2021; Murauer and Specht, 2021a). This commitment to fairness intersects with Transparency and Explainability, a cornerstone of the framework, ensuring AA processes and decisions are clear, understandable, and accessible to stakeholders, fostering trust and accountability (Escart'in et al., 2021). The principle of Societal Impact encompasses the ethical considerations of Non-Maleficence, focusing on mitigating risks, harm, and malicious use while addressing the environmental impact of the deployment of AA systems. It encourages stakeholders to evaluate the broader implications of AA technologies to achieve positive social outcomes (Hovy and Spruit, 2016; Bender et al., 2021). Finally, Responsibility is addressed by providing clear guidelines to all stakeholders involved in the AA development cycle, ensuring accountability and ethical practices. By laying a structured questionnaire, this framework guides stakeholders in promoting ethical practices that safeguard individual rights, enhance fairness and transparency, and contribute to the responsible and equitable use of AA models.

[1] It should be noted that responsible data-sharing practices are beyond the scope of this study. However, readers are strongly encouraged to follow the extensively detailed practices outlined by Gebru et al. (2021).

### 5.2.2 BACKGROUND ON THE ETHICAL PRINCIPLES

The rapid advancement of ML applications has sparked significant debates about its benefits and risks, prompting the development of frameworks for responsible AI (Floridi and Cowls, 2019; Fjeld et al., 2020; Floridi et al., 2021). These frameworks advocate for fairness, accountability, and transparency (FAccT AI) (Simbeck, 2022; Laufer et al., 2022; Young et al., 2022), with a strong emphasis on socio-technical considerations (Dignum, 2019). However, translating these principles into actionable guidelines for AA applications remains a critical gap. This study builds on existing ethical AI frameworks (Mittelstadt, 2019; Felzmann et al., 2020; Prem, 2023) to address specific Ethical, Legal, and Social Issues (ELSI/ELSA) in AA applications, offering a questionnaire designed to guide stakeholders in navigating these challenges. This questionnaire serves as an awareness tool rather than a checklist, facilitating discussions on ELSI/ELSA issues and providing strategies to address them.

**Privacy & Data Protection:** AA tasks in NLP often involve handling large volumes of textual data and processing personal information, even when pseudonymized (Sjöberg, 2021). For example, identifying the authorship of emails or social media posts may inadvertently expose sensitive personal details through writing styles or contextual clues (Sennewald et al., 2020). This highlights the necessity of strict compliance with data protection and privacy regulations, including lawfulness, purpose limitation, data minimization, and data security (De Terwangne, 2020; Gellman, 2023). Privacy-by-design approaches must be integrated early in AA development to balance utility with protecting individual rights (Tamò-Larrieux et al., 2018). Therefore, this study proposes guidelines to ensure AA applications adhere to privacy and data protection norms while effectively fulfilling their intended purposes.

**Fairness & Non-discrimination:** The presence of biases in NLP, including AA models, poses a significant challenge, often leading to unfair or discriminatory outcomes (Shah et al., 2019; Chang et al., 2019; Blodgett et al., 2020; Hovy and Prabhumoye, 2021; Lalor et al., 2022). For instance, if an AA model is trained primarily on texts from a specific demographic, it may disproportionately misattribute authorship when analyzing texts from underrepresented groups. This perpetuates societal inequalities, reinforces stereotypes, and undermines the fairness of the attribution process (Caliskan et al., 2017; Dev et al., 2022). Biases can infiltrate the AA pipeline at multiple stages, including data collection (Dixon et al., 2018), feature selection (Bolukbasi et al., 2016; Ai et al., 2022), model training (Zafar et al., 2017), and evaluation (Delobelle et al., 2022). While existing research has examined biases in AA tasks (Bevendorff et al., 2019; Bischoff et al., 2020; Murauer and Specht, 2021a; Brad et al., 2021; Ai et al., 2022), this study leverages these insights to propose strategies for mitigating biases across the various phases of AA application development.

**Transparency and Explainability:** Ensuring transparency in AA systems involves clarifying decision-making processes, providing information on the features utilized, and detailing how data is processed. This is critical for building trust and enabling users to understand the rationale behind authorship decisions, particularly in law enforcement applications. Transparency ensures AA systems can be audited and justified, which is essential for maintaining integrity and fairness in algorithmic decision-making. The importance of transparency and explainability has been widely discussed across various fields, including NLP (Chiticariu et al., 2015; Kim et al., 2020; Ma et al., 2020; Chen et al., 2020a; Saxon et al., 2021; Ethayarajh and Jurafsky, 2021; Balkir et al., 2022b; Bhatt et al., 2022) and law and ethics (John-Mathews et al., 2022; Weinberg, 2022; Zhang et al., 2023). In the context of AA, transparency addresses two primary concerns: prospective transparency, which involves disclosing how data will be used and processed, and retrospective transparency, which focuses on explaining the outcomes of algorithmic processes (Felzmann et al., 2019). Retrospective transparency is particularly relevant to research on the explainability of algorithmic decision systems, ensuring that the steps leading to a decision can be understood and scrutinized (Ding et al., 2015; Boenninghoff et al., 2019a; Bogdanova and Romanov, 2021; Theophilo et al., 2022; Kondyurin, 2022). This aspect is closely tied to accountability, as it verifies the system's compliance with ethical standards and responsiveness to user concerns (Procter et al., 2020; Qian et al., 2021; Rawal et al., 2021; Angelov et al., 2021; Balkir et al., 2022c). While extensive research on transparency exists in the broader field of NLP (Shook et al., 2017; Tubella et al., 2019; Bogina et al., 2021; Angerschmid et al., 2022; Hacker et al., 2022), specific, actionable guidelines for implementing transparency in AA tasks remain limited. This study addresses this gap by providing comprehensive guidelines for transparent and fair data processing and decision-making, fostering accountability and trustworthiness in AA systems.

**Societal Impact:** Despite their potential, the perceived accuracy and reliability of AA models make them susceptible to misuse and misrepresentation (Potthast et al., 2016; Suresh and Guttag, 2021; Zhai et al., 2022; Uchendu et al., 2023). For instance, AA models may falsely attribute texts to incorrect authors, leading to significant ethical and legal consequences. While AA approaches hold promise for social welfare in fields such as forensic linguistics, cybersecurity, and cybercrime detection, they also carry risks and potential for misuse similar to other AI technologies (Juola, 2020). To mitigate these risks, it is essential to understand the broader societal implications of AA, particularly the potential for harm and ethical breaches. The broader context of risks and harms associated with NLP technologies underscores the need for careful consideration in the design, implementation, and deployment of AA models (Banko et al., 2020; Weidinger et al., 2021; Suresh and Guttag, 2021; Shmueli et al., 2021; Hovy and Spruit, 2016; Kirk et al., 2022; Haduong et al., 2023). Additionally, the environmental impact of AA models is a critical societal consideration. The efficiency and robustness of trained AA models directly influence their carbon footprint and resource consumption. As the demand for larger and more complex models grows, so does their environmental impact, raising concerns about sustainability and the responsible use of resources (Van Wynsberghe, 2021; Wu et al., 2022). Addressing these environmental considerations ensures that AA technologies are ethically sound and environmentally responsible. Building on these discussions, this study aims to provide comprehensive guidelines addressing AA applications' specific threats, environmental impacts, and broader societal implications. By integrating these considerations into each phase of the AA application's life cycle, this study promotes the responsible and ethical use of AA methods, ensuring they contribute positively to society while minimizing risks and harms (Martin et al., 2020a; Madiega, 2021; Gerards et al., 2022; Tabassi, 2023).

### 5.2.3 Stakeholders of the Framework

To address the responsibility concerns outlined by Jobin et al. (2019), this study adopts the standard SDLC waterfall model (Ruparelia, 2010), which allows for designing a structured and sequential framework enabling the clear identification of stakeholders and the systematic assignment of responsibilities throughout each phase of AA system development. Key stakeholders include system designers, developers, project managers, quality assurance testers, and end-users, each playing a critical role in ensuring the ethical and responsible deployment of AA technologies. The framework categorizes these stakeholders into three distinct SDLC phases:

**Design and Planning Phase:** In this phase, the emphasis is on data collection and processing, ensuring that system designers and project managers integrate privacy considerations by embedding ethical guidelines into the architecture. This integration establishes a robust foundation for subsequent phases.

- **System Designers:** Engaged in the early stages, system designers are responsible for incorporating ethical principles into the AA cycle's architecture. Their focus on privacy, fairness, and transparency establishes a solid ethical basis guiding the development process.
- **Project Managers:** Collaborating with system designers, project managers oversee the entire development lifecycle. They ensure eth-

ical standards are maintained from requirements gathering through deployment, coordinate between teams, manage timelines, and hold stakeholders accountable for ethical practices.

**Development and Validation Phase:** During this phase, developers and testers convert ethical designs into functional models by addressing potential biases and ensuring compliance with privacy regulations. This phase encompasses training, testing, evaluation, and continuous monitoring, demonstrating a commitment to ethical development.

- **Developers:** Developers implement the algorithms defined by system designers, ensuring that the resulting systems adhere to ethical guidelines, mitigate potential biases, and protect user privacy.
- **Quality Assurance Testers:** Testers evaluate the AA systems to verify that they function as intended and comply with ethical guidelines. Their role involves identifying and rectifying issues before deployment, serving as the final checkpoint to confirm that the system is ethically sound and ready for use.

**Deployment and Feedback Phase:** In the final phase, AA applications are deployed, and end-users provide critical feedback essential for refining the system and ensuring ongoing adherence to ethical guidelines.

- **End-Users:** End-users—including forensic analysts, researchers, and media professionals—interact directly with the deployed AA systems. Their feedback proves invaluable in assessing system performance and ethical impact, identifying areas for improvement, and ensuring that the system remains aligned with ethical standards and its intended purpose.

By organizing stakeholders according to their roles within the SDLC and their respective phases, this study ensures a clear flow of responsibilities and accountability throughout the AA lifecycle. This structured approach integrates ethical principles at every stage, maintaining the integrity, trustworthiness, and societal value of AA technologies.

## 5.3 FRAMEWORK

This study emphasizes integrating ethical principles into a practical framework, addressing the critical need for responsible AA system development. By employing the SDLC waterfall model, the study systematically incorporates four key ethical principles—privacy and data protection, fairness and non-discrimination, transparency and explainability, and societal impact—throughout each phase of AA system development. This structured approach defines the roles and responsibilities of various stakeholders, including designers, project managers, developers, testers, and end-users, across the design and planning, development and validation, and deployment phases. To operationalize these principles, role-based guidelines and an accompanying color-coded guiding questionnaire are proposed to guide various SDLC stakeholders through the project phases: design and planning, development and validation, and deployment and feedback. The guidelines associated with Privacy and Data Protection are abbreviated as PD, Fairness and Non-Discrimination as FB (for Fairness and Biases), Transparency and Explainability as TE, and Societal Impact as SI in the questionnaire. Practical examples are illustrated through the case study performed on the published IDTraffickers research (Saxena et al., 2023a). The case study highlights how these guidelines address privacy risks, biases, transparency gaps, and societal harms, offering actionable recommendations through detailed explanations and suggestions. The proposed questionnaire serves several purposes:

- **Operationalizing Ethical Principles:** The questionnaire translates abstract ethical principles into concrete, actionable steps by linking specific questions to the ethical guidelines outlined in each section.

- **Guiding Stakeholders:** Each question is color-coded and aligned with the roles and responsibilities of relevant stakeholders (e.g., system designers, project managers, developers, testers, and end-users), ensuring that ethical considerations are systematically addressed throughout the AA development and deployment process. This role-specific structured approach offers novel insights to stakeholders to evaluate their decisions, document their adherence to ethical standards, and enhance transparency and accountability. While existing literature does not explicitly define these role-based guidelines, the assignment is carefully structured based on each stakeholder's established duties and responsibilities in the AA system lifecycle.

- **Ensuring Practical Utility:** The explanations and recommendations in the questionnaires clarify specific aspects of the framework, emphasizing responsible and ethical practices. These details enable readers and practitioners to quickly identify actionable insights and areas requiring further attention.

### 5.3.1 PRIVACY & DATA PROTECTION

#### SYSTEM DESIGNER AND PROJECT MANAGERS

Privacy constitutes a fundamental theme in ethical AI guidelines (Fjeld et al., 2020) and remains a central concern due to its inherently elusive nature (Gasser, 2016; Nissim and Wood, 2018). In the evolving landscape of digital technologies and a data-driven society, the focus on privacy research has shifted toward data privacy and protection (Lynskey, 2017). Regulations such as the GDPR in the European Union (Regulation, 2018) are pivotal in addressing privacy challenges related to AA techniques.

To ensure privacy-based compliance standards, system designers conduct Data Protection Impact Assessments (DPIAs) to evaluate potential risks associated with data processing (Demetzou, 2019b; Martin et al., 2020b). Mandated by the GDPR, DPIAs identify and mitigate privacy risks by examining how data is collected, stored, and used while incorporating safeguards such as encryption and anonymization to protect against unauthorized access or breaches. These assessments may also serve as exemptions for scientific research under GDPR Article 89 (Staunton et al., 2019). Designers can employ a range of templates and tools to conduct these DPIAs effectively (De Hert, 2012; Bieker et al., 2016; Mantelero, 2018). DPIAs further support compliance with emerging AI regulations such as the EU AI Act (Edwards, 2021), which categorizes AA systems based on risk levels (De Cooman, 2022; Hupont et al., 2023; Neuwirth, 2023). For example, systems that predict behavior solely through profiling are prohibited (Article 5), whereas those supporting human assessments may qualify for exemptions (Art. 5(d)). High-risk applications, including those employed in law enforcement, require compliance documentation and registration with the appropriate authorities (Arts. 6(2-4), 49 AI Act).

In addition to conducting DPIAs, system designers are responsible for implementing secure and compliant data storage practices within AA systems. Encryption and pseudonymization are employed to safeguard data at rest and in transit, thereby reducing the risk of unauthorized access or breaches (Voigt and Von dem Bussche, 2017). To adhere to data minimization principles, designers ensure that data is retained only for the duration necessary to fulfill its intended purpose (Mondschein and Monda, 2019). Furthermore, robust authentication protocols and access controls are integrated into the system architecture to restrict data access exclusively to authorized personnel (Kennedy and Millard, 2016). By incorporating these measures into the system design, adherence to privacy principles can foster trust and reduce risks throughout the AA system lifecycle.

Project managers are crucial in ensuring that AA systems align with privacy and data protection principles. Their responsibilities encompass facilitating stakeholder communication to promote awareness and understanding of relevant privacy laws affiliated with GDPR and the EU AI Act. Furthermore, they coordinate with system designers and compliance officers to ensure that data handling practices remain secure and compliant, protecting sensitive information from unauthorized access or breaches. Project managers also document the processes of data collection, processing, and storage to ensure that all practices remain transparent and adhere to established privacy principles (Demetzou, 2019a). Additionally, they uphold user rights—including access, rectification, objection, and erasure (Ausloos et al., 2019b)—while recognizing that the GDPR permits automated decisions under certain conditions, such as explicit consent or legal requirements, provided appropriate safeguards are in place (Tamo-Larrieux, 2021). Through vigilant oversight, project managers ensure AA systems respect privacy, maintain high data protection standards, and foster trust, promoting responsible AI development.

**PD1. Does the AA research/application under study involve a high level of risk, necessitating a Data Protection Impact Assessment (DPIA)? High-risk scenarios may include biometric identification, law enforcement, or justice system usage, etc.**
– The research does not explicitly state whether a DPIA was conducted.
**Suggestion:** Given the application's high-risk nature, conducting a DPIA is suggested to identify and mitigate risks associated with data processing proactively. This would align with GDPR requirements and provide a structured approach to handling data responsibly. DPIAs are essential for evaluating the potential impact of data processing activities, particularly when dealing with high-risk scenarios like profiling individuals based on writing styles or patterns. Implementing a DPIA can help anticipate privacy issues, establish controls to mitigate risks, and ensure ongoing compliance with regulatory standards.

**PD2. Is there a scientific purpose or objective justifying exemptions from GDPR provisions?**
– The research aims to combat human trafficking, a significant societal issue that could be considered a legitimate scientific purpose. The research provides these details under Section 8, Privacy Considerations and Potential Risk paragraph.

**PD3. What measures are in place to comply with privacy principles regarding data collection, processing, and storage in the AA system?**
- While Appendix sections A.2.3 and A.2.4 provide detailed information about data collection and processing, there is a lack of information regarding data storage practices.
**Suggestion:** To fully comply with privacy principles, it is important to include comprehensive details on data storage practices in addition to data collection and processing. The documentation should also address how data is stored, including security measures and retention policies.

**PD4. Are there specific protocols to protect sensitive/personal information from unauthorized access or breaches throughout the AA system?**
– The personal data within the dataset, including phone numbers and email addresses, is masked, minimizing the risk of individual identification. The research provides these details under Section 8, Privacy Considerations and Potential Risk paragraph. However, while personal identifiers are masked, the inherent nature of the AA task in NLP involves linking writing styles (as a unique signature in the raw data) to individuals. The research does not explicitly address how to mitigate this potential privacy concern.
**Suggestion:** To minimize the risks and misuse of the AA predictions, regular audits and training sessions must be established. Furthermore, incorporating consent mechanisms and allowing individuals to opt out of data use would further enhance privacy protection.

**PD5. Is the information provided to individuals about data processing clear, complete, and correct?**
– The research doesn't mention whether or not individuals are informed about data processing. For GDPR compliance and ethical considerations, ensuring transparency about data use is critical, allowing individuals to understand and, where applicable, contest the use of their data.

**Suggestion:** To comply with GDPR and uphold ethical standards, providing individuals with clear, complete, and accurate information about how their data is being processed is essential. This includes detailing the purpose of data collection, the specific data being used, how long it will be retained, and individuals' rights over their data. Implementing transparent communication practices, such as privacy notices or consent forms, ensures that individuals are informed and can contest or opt out of data processing if necessary. Regularly updating these communications to reflect any changes in data processing practices is also recommended.

**PD6. Is there periodic assessment and review to ensure the ongoing relevance of data usage and AA application?**
- No such details are provided in the research.
**Suggestion:** The project manager must conduct regular assessments and training sessions to ensure data usage remains relevant and compliant throughout the AA application cycle. These periodic reviews will help maintain the integrity of the data, adapt to any changes in the context or requirements, and ensure ongoing alignment with privacy and data protection standards.

#### Developers and QA Testers

Developers and testers play a crucial role in ensuring that AA systems adhere to GDPR data privacy principles, including purpose limitation, data minimization, storage limitation, and data integrity and confidentiality (De Terwangne, 2020; Bincoletto, 2020). They must design and evaluate AA systems with clear data processing policies that help users understand what data is being collected and how it will be used. They must also create user interfaces and consent mechanisms that provide straightforward controls over data usage.

To align with purpose limitation and data minimization requirements, developers must ensure that data is only processed for predefined objectives, using anonymized or aggregated data where possible. Storage limitation practices should be enforced by implementing automatic data deletion protocols and robust security measures, such as encryption, secure communication channels, stringent access controls, and real-time threat detection mechanisms. By integrating these privacy principles into the system's design, developers can safeguard user privacy, foster trust, and ensure compliance with GDPR standards.

Throughout the AA development cycle, testers verify compliance with GDPR data privacy principles (De Terwangne, 2020; Bincoletto, 2020). Their role involves assessing whether the system adheres to fairness, transparency, purpose limitation, data minimization, storage limitation, and data integrity and confidentiality. This includes ensuring that data is processed solely for the defined purposes (purpose limitation), only relevant data is collected (data minimization), and storage limitation practices are effectively implemented by verifying automatic data removal protocols to ensure personal data is deleted when no longer necessary. Furthermore, testers must evaluate the effectiveness of security mechanisms, including encryption and access control protocols, to maintain data integrity and confidentiality (Sweeney et al., 2015; Sion et al., 2021). Through rigorous testing, they help ensure that AA systems comply with GDPR requirements, thereby upholding individual privacy rights throughout the system's lifecycle.

**PD7. Is data processing confined to the original purpose for which it was collected? Is there periodic assessment and review to ensure ongoing relevance?**
**-** Yes, the dataset is specifically compiled for AA to identify human trafficking operations, aligning with the original data collection purpose. However, periodic assessments and reviews are not explicitly mentioned yet could be essential for maintaining data relevance and compliance.
**Suggestion:** The project manager must conduct periodic assessments and training to maintain data relevance and compliance throughout the AA application cycle.

**PD8. Have adequate safeguards like anonymization, encryption, data minimization, and security procedures been implemented to minimize risks and protect individual rights? Are these measures in line with guidelines from research and academic organizations, with ethical oversight?**
**-** The research outlines anonymization and data minimization measures, such as masking personal details. The researchers control access to the data through the DataverseNL portal, where only the meta-data is publicly available. However, the detailed security procedures are not specified. Implementing comprehensive security procedures and seeking ethical oversight would further enhance data protection. The research provides these details under Section 8, Privacy Considerations and Potential Risk paragraph.
**Suggestion:** To enhance data protection, comprehensive security procedures must be implemented alongside the existing anonymization and data minimization practices. Additionally, obtaining ethical oversight and regular audits is crucial. These audits play a significant role in ensuring that all privacy considerations are adequately addressed, instilling a sense of security and confidence in the data protection measures.

### 5.3.2 Fairness & Non-discrimination

Like other NLP applications, AA approaches are vulnerable to unintended biases that can influence accuracy and fairness (Halvani and Graner, 2021; Murauer and Specht, 2021b). Various forms of biases can affect these systems, including label bias, selection bias, demographic and population bias, sampling bias, domain and genre bias, representation bias, model overfitting, evaluation bias, and user-interaction bias. Identifying and mitigating these biases is essential to ensuring the reliability and ethical deployment of AA technologies.

#### System Designers

**Label Bias** in NLP-based AA tasks emerges when training data contains biased or incorrect authorship labels due to systematic errors or human annotation inaccuracies (Shah et al., 2020). Therefore, it is essential to ensure that training data is reliable, with authorship attributions rigorously validated and cross-checked (Shah et al., 2020). Involving multiple domain experts to assign authorship labels to documents can help mitigate individual biases or errors (Søgaard et al., 2014; Shah et al., 2020). Consensus-based annotations further reduce the risk of inaccuracies (Søgaard et al., 2014). Additionally, employing active learning strategies allows for periodic review and updates to the training data, incorporating corrections or new insights to maintain its quality and reliability (Zhang et al., 2022).

**FB1. Is there a specific label or target for each data instance, and how were these labels obtained? For manually annotated data, please provide details about the number of annotators, their backgrounds, and any measures taken to mitigate label bias.**
– Vendor labels are generated based on phone number connections among ads, not manual annotation. This approach aims to reduce bias associated with manual labeling.

### DEVELOPERS

**Selection Bias** occurs when the training dataset systematically over-represents or under-represents certain groups or characteristics, leading to a dataset that does not accurately reflect the target distribution (Cawley and Talbot, 2010). This bias stems from systematic factors influencing data selection, such as collecting data from a specific platform or genre, which may fail to capture the complete landscape of authorship in the broader population. While semi-supervised approaches can help mitigate selection bias, they may prove ineffective if the gap between the source and target domains is too wide, potentially exacerbating the issue (Plank et al., 2014; Søgaard et al., 2014). To counteract this, developers must ensure that datasets encompass diverse authors, writing styles, genres, and topics, reflecting the real-world distribution of texts (Shah et al., 2020). Combining data from multiple platforms reduces the risk of over-reliance on a single source.

**Sampling Bias** refers to any bias resulting from a training sample that inadequately represents the entire population of interest (Hacker, 2018; Prabhu et al., 2019; Mehrabi et al., 2021). This can stem from systematic selection issues or random anomalies in the sampling process. Unlike selection bias, a specific form of sampling bias, sampling bias broadly encompasses any imbalance in the data distribution, hindering the model's ability to generalize to new authors or writing styles. To address this, developers must employ random or stratified sampling methods to ensure the training data reflects the population (Prabhu et al., 2019). While generalizing to every possible author is unfeasible, these sampling methods promote more balanced and representative model learning (Prabhu et al., 2019; Mehrabi et al., 2021).

**Demographic and Population Bias** arise from systematic patterns in the dataset that correlate with authors' demographic attributes, such as age, gender, ethnicity, or geographical location (Mehrabi et al., 2021). Variations in writing styles, vocabulary, or linguistic patterns associated with different demographic groups can sway AA models, leading to unfair or inaccurate predictions based on demographics rather than writing style (Ai et al., 2022). Although removing personally identifiable information and demographic attributes can help (Dev et al., 2022; Bender and Friedman, 2018), it does not fully address the underlying biases influencing model predictions. To mitigate these biases effectively, developers must enrich datasets with diverse writing samples across demographic groups or implement algorithmic adjustments to compensate for potential biases.

**Domain and Genre Bias** in AA relates to dataset patterns associated with specific text domains or genres rather than writing style (Julian et al., 2017). These biases result from language use and conventions specific to different domains or genres, leading to incorrect authorship attributions. To address this, the developer must ensure a balanced representation of authors from diverse domains and genres in the training dataset. Transfer learning, where models are pre-trained on broad, diverse datasets and fine-tuned for specific tasks, can help reduce domain and genre bias (Barlas and Stamatatos, 2021b). Content-controlled authorship approaches, which normalize thematic content during training, further assist in focusing on writing style rather than domain-specific cues (Wegmann et al., 2022).

**Overfitting** occurs when an AA model performs exceptionally well on training data but struggles to generalize to unseen texts in test or OOD datasets (Schaffer, 1993; Lawrence and Giles, 2000). This high-variance problem arises when the model becomes overly tailored to the specific patterns of the training data, rendering its predictions unreliable for unseen data. To mitigate overfitting, developers must split datasets into training, validation, and test sets, using k-fold cross-validation for evaluation (Jabbar and Khan, 2015). Avoiding overly specific features, opting for simpler models when appropriate (Cawley and Talbot, 2010), employing regularization and early stopping techniques (Cawley and Talbot, 2010), and monitoring validation performance are additional strategies to detect and address overfitting.

**Underfitting** occurs when a model is too simplistic to capture the complex patterns in authorship styles, leading to high bias in its predictions (Jabbar and Khan, 2015). In this case, the model fails to model the un-derlying relationships in the data, resulting in poor performance on both the training and test data. To avoid underfitting, developers should avoid overly simplistic models (Wolfe and Caliskan, 2021), create ensemble models to leverage diverse perspectives (Jabbar and Khan, 2015), employ sophisticated feature representation techniques (Ai et al., 2022; Wegmann et al., 2022), fine-tune hyperparameters for optimal performance (Locatelli et al., 2022), apply regularization methods (Wolfe and Caliskan, 2021; Locatelli et al., 2022), and use cross-validation for robust generalization assessments (Jabbar and Khan, 2015).

It is important to note that in AA tasks, "bias" can refer to two distinct concepts: algorithmic bias, which pertains to the model's performance, and ethical bias, which relates to fairness and potential discrimination. Underfitting and overfitting are examples of algorithmic bias linked to the bias-variance trade-off (Belkin et al., 2019). This trade-off explains how AA models balance their ability to fit training data and generalize to new, unseen data. Underfitting leads to high bias, where a model is too simplistic, failing to capture patterns in the data. Overfitting results in high variance, where a model becomes overly complex and fits the training data too closely, reducing its ability to generalize. Both types of bias affect the model's accuracy but are distinct from ethical concerns about fairness or discrimination.

**FB2. Does the training dataset sufficiently represent the entire authorship landscape, and what steps were taken to mitigate selection bias?**
– The dataset comprises a significant number of ads from the Backpage escort market from 14 states and 41 cities of the U.S., aiming to capture a broad spectrum of authorship styles linked to potential human trafficking. The research provides these details under Section 3, Dataset and Appendix A.4, Datasheet.
**Suggestion:** Although the dataset includes ads from a broad geographical region in the U.S., all the text ads used in this study were sourced from a single online escort platform. Expanding the data collection to include ads from multiple platforms could enhance model generalization and help mitigate potential biases.

**FB3. Are there correlations between authors in the dataset and**

**specific demographic attributes or population characteristics?**
– The research does not analyze the authors' demographic attributes or population characteristics, potentially missing insights into bias related to these factors.
**Suggestion:** To gain a deeper understanding, it would be beneficial to investigate the correlations between the authors and specific demographic attributes or population characteristics. By identifying these correlations, the research can assess whether the model performs equitably across different demographic groups, thereby ensuring fairness and reducing the risk of bias. This analysis can also provide valuable insights into how representative the dataset is of the broader population and inform adjustments to improve the model's overall effectiveness.

**FB4. Does the dataset cover multiple text genres or domains, and if not, what actions were taken to prevent biases related to the domain and genre?**
**-** The dataset focuses on text escort ads from a single source (Backpage), indicating a specific domain focus. The document does not describe actions taken to address potential biases arising from this focus, such as incorporating or analyzing ads from varied platforms or text genres to enhance generalizability.
**Suggestion:** To address potential domain and genre biases, it would be useful to expand the dataset to include text from multiple sources and platforms, covering various genres beyond the Backpage escort ads.

**FB5. Is there a class imbalance in the dataset, and what measures were implemented to avoid over-representing certain authors? Describe any sampling strategies used to address potential sampling bias?**
– The research mentions the distribution of ads per vendor but does not detail specific measures to address potential class imbalance or sampling bias. To consider class imbalance, the authors emphasize using Macro-F1 and R-Precision metrics for evaluation.
**Suggestion:** Strategies like stratified sampling or synthetic data generation could help mitigate these issues, ensuring a balanced representation of different authors. The research provides these details under Section 3, Dataset.

**FB6. What feature extraction techniques were employed during training, and was fine-tuning performed on the target data?**

– The research employs various contextualized (transformers-based) models, focusing on style representations extracted from classified ads. Fine-tuning on the target dataset (IDTraffickers) is conducted to adapt the model to the specific domain, indicating an effort to capture domain-specific authorship styles effectively. The research provides these details under Section 4, Experimental Setup, and Section 5, Results.

**FB7. What precautions were taken to prevent overfitting and underfitting during model training?**
- The research outlines the data-split of 0.75:0.05:0.20 for training, validation, and test datasets. Measures like cross-validation, regularization, and early stopping address potential overfitting and underfitting issues, ensuring model robustness. The research provides these details under Appendix A.4, Datasheet.

### QA TESTERS

**Evaluation Bias** in NLP leads to incorrect conclusions about a model's capabilities (Suresh and Guttag, 2021). It can manifest when datasets lack diverse authorship, favoring frequent authors and affecting the performance of minority ones, which can compromise model reliability and performance. Therefore, the testers must ensure that the evaluation datasets mirror real-world scenarios with random and unbiased text selection (Cawley and Talbot, 2010). Choosing appropriate evaluation metrics aligned with the task's objectives is crucial for accurate assessments and informed deployment decisions (Powers, 2008; Hämäläinen and Alnajjar, 2021; Mehrabi et al., 2021).

**FB8. Do the chosen evaluation metrics align with the primary task objectives? What insights can be provided about model generalization and robustness?**
– The research highlights macro-F1 and mean r-precision scores for evaluating model performance, focusing on identifying authorship patterns across different vendors. The research provides these details under Section 4, Experimental Setup, and Section 5, Results. While the chosen metrics are relevant to the task objectives, no insights into model generalization are presented in the research.
**Suggestion:** Additional analysis of model performance on the OOD

dataset can provide deeper insights into model generalization and Robustness.

#### End Users

**User-Interaction Bias** in AA stems from human involvement during model design, development, or evaluation, which can introduce unintentional biases into training data and evaluation setups (Mehrabi et al., 2021). User interactions, like annotators' expectations or feedback (Hämäläinen and Alnajjar, 2021), can influence the model's training and evaluation, skewing predictions. Therefore, deployers must conduct blind assessments, involve diverse users for broader perspectives, and utilize explainability tools for unbiased analysis (Srinivasan and Chander, 2021; Mehrabi et al., 2021; Hämäläinen and Alnajjar, 2021). These steps help ensure that user interactions do not unduly influence the model's outcomes.

**FB9. Were any independent blind assessments conducted by external evaluators, and was the information about their backgrounds and diversity provided?**
- Details about any independent blind assessments by external evaluators are not mentioned in the research.
**Suggestions:** These blind assessments can enhance the credibility of the research findings. Providing information about the backgrounds and diversity of these evaluators would further strengthen the research, ensuring that a wide range of perspectives is considered. This approach can help identify potential biases in the model's evaluation and promote a more thorough and unbiased assessment of its performance.

### 5.3.3 Transparency and Explainability

Transparency is a critical component in AA tasks, particularly in ensuring compliance with regulations such as the GDPR (Ausloos et al., 2019a; Felzmann et al., 2019; Hacker and Passoth, 2020; Grünewald and Pallas, 2021) and the AI Act (Hacker et al., 2022; Madiega, 2021). Clear and comprehensive disclosures aligned with regulatory requirements must employ multi-channel communication strategies to meet data subject expectations and uphold fairness. This intersection of transparency and fairness is essential to prevent unjust or discriminatory practices (Bincoletto, 2020).

Explainable AI (XAI) techniques play a pivotal role in identifying and mitigating unfairness in AA models (Theophilo et al., 2022; Ai et al., 2022), particularly in sensitive applications such as cybercrime and forensic text analysis (Solanke, 2022). Interactive XAI methods enable researchers to detect and rectify unfair attributions while providing textual explanations for model decisions (Stevens et al., 2020; Alikhademi et al., 2021; Theophilo et al., 2022; Balkir et al., 2022c; Ai et al., 2022). These approaches empower practitioners to promote transparency by intervening and correcting unfair attributions on a case-by-case basis (Shrestha et al., 2017; Manolache et al., 2021; Huertas-Tato et al., 2022). By generating explanations for predictions, XAI techniques can highlight instances where problematic correlations may influence outcomes incorrectly. Additionally, AA tools must include user instructions detailing the provider's identity, contact information, and the system's characteristics, capabilities, and limitations.

Effective implementation of transparency requires mechanisms that allow users to intervene and correct unfair attributions generated by the AA system. Active learning strategies (Abbas et al., 2023) involve engaging users or experts to provide feedback when incorrect attributions are detected. This iterative process enhances the model's accuracy and fairness over time. Another critical aspect of transparency is the communication of AA decisions to users and stakeholders, which ensures that all parties understand the decision-making processes and their implications.

However, achieving explainability in AA systems is not without challenges (Balkir et al., 2022a). One key challenge is balancing simplicity and accuracy in explanations (Barredo Arrieta et al., 2019; Agarwal, 2020; Crook et al., 2023; Saeed and Omlin, 2023). Overly simplistic explanations may fail to capture the complexity of model logic, potentially leading to misunderstandings or the introduction of new biases. Additionally, evaluating the quality of explanations and their impact on the fairness of authorship analysis remains an ongoing challenge (Ai et al., 2022). Continuous research and development are necessary to ensure that XAI techniques effectively support the goal of fair and transparent AA tasks.

#### SYSTEM DESIGNERS AND PROJECT MANAGERS

**TE1. What steps are taken to balance simplicity and accuracy in the explanations provided by the AA system?**
– The research utilizes local and global feature attribution techniques for qualitative analysis. These techniques allow the AA system to highlight the most relevant features contributing to a decision, simplifying complex model logic without compromising accuracy.

**TE2. How are the AA decisions communicated to users and stakeholders?**
– The chosen case study provides no specific details on communication methods for AA decisions. This is expected in a research environment where all participants often share stakeholder responsibilities.
**Suggestion:** In a typical SDLC, it is crucial to establish clear roles and responsibilities for communication. This ensures a secure and organized communication process and helps avoid any potential misunderstandings. To address this, project managers can implement a structured communication plan that defines how AA decisions are conveyed to different stakeholders at each phase of the SDLC. For example, system designers and developers can provide technical documentation and visualizations to explain decision-making processes. At the same time, end-user training sessions can help non-technical stakeholders understand the implications of these decisions. Regular updates and feedback loops can ensure that stakeholders remain informed and engaged throughout the development and deployment of the AA system.

#### DEVELOPER AND QA TESTERS

**TE3. Are any explainability/interpretability experiments conducted to understand model decision-making processes and predictions? –** The research utilizes local and global feature attribution techniques for qualitative analysis, indicating an effort toward explainability. The research provides these details under Section 5.3, Qualitative Analysis.

**TE4. Are any error-analysis experiments conducted to understand false-positive and true-positive predictions?**
– The research provides a systematic error analysis of true and false-positive predictions to understand the underlying causes of these outcomes. The research provides these details under Section 5.3, Qualitative Analysis.

**TE5. Are there mechanisms for users to intervene and correct unfair attributions made by the AA system?**
– No such details are provided in the chosen case study.
**Suggestion:** To address this gap, the AA system could use active learning strategies (Abbas et al., 2023) to engage users or experts when incorrect attributions arise, allowing user input to improve accuracy. An interface for users to flag or correct unfair attributions can help ensure ongoing fairness and reliability, creating a feedback loop that enhances the system over time.

**TE6. Is there a process for evaluating the quality of explanations provided by the AA system?**
No such details are provided in the chosen case study.
**Suggestion:** To assess the quality of explanations in the AA system, robustness checks can be conducted using techniques such as adversarial attacks (Zhang et al., 2020b) or counterfactual examples (Stepin et al., 2021). These methods evaluate how well the explanations withstand intentional disturbances or hypothetical changes, ensuring the system's explanations are dependable and resistant to manipulation. Implementing these checks helps maintain the integrity and trustworthiness of the AA system's outputs.

### 5.3.4 SOCIETAL IMPACT

#### SYSTEM DESIGNERS

AA approaches offer significant value across diverse domains. However, they also introduce notable risks and harms that necessitate careful considerations (Hovy and Spruit, 2016; Juola, 2020). According to Kirk et al. (2022), harm in the context of AA encompasses both content-related issues and risks that may negatively impact the emotional, psychological, and physical well-being of individuals, groups, or society. A primary concern involves privacy and confidentiality breaches, where AA models may inadvertently disclose sensitive information about authors, leading to identity exposure, reputational harm, or legal consequences (Chaski, 2005b; Kirk et al., 2022; Banko et al., 2020). Additionally, the misuse of AA algorithms poses significant risks, enabling malicious activities such as targeted harassment, social engineering, or the creation of deceptive content falsely attributed to others (Banko et al., 2020). In forensic and cybercrime applications, the use of AA techniques raises concerns about the potential trauma experienced by individuals involved in the design, development, and deployment phases, as they may encounter harmful or criminal content (Pyevich et al., 2003; Dubberley et al., 2015; Banko et al., 2020; Duran and Woodhams, 2022; Birze et al., 2023).

To reduce the risk of exposing individuals to harmful content, it is essential for all stakeholders to clearly understand research objectives and potential threats (Renda et al., 2021; Kirk et al., 2022). AA researchers must prioritize the protection of subjects when publishing their work, highlighting potential risks associated with sensitive data and distancing their research from harmful viewpoints Kirk et al. (2022). Collaborative teamwork can help mitigate individual exposure to harmful content by fostering clear communication and providing necessary support. Offering mental health and psychological resources to team members dealing with harmful text is crucial to help them manage potential challenges (Kirk et al., 2022).

To minimize potential misuse, it is critical to ensure that the model's scope aligns with its intended purpose (Koops, 2021; Moraes et al., 2021). Incorporating human oversight and intervention mechanisms is vital, allowing for thorough review and rejection of content with ethical concerns (Kirk et al., 2022). This human-in-the-loop approach helps prevent the inclusion of harmful or ethically problematic content in the training data. However, human oversight is imperfect, as individuals may overlook issues, introduce bias, or lack the expertise to detect problematic content. Therefore, routine audits and updates are essential to address potential ethical challenges proactively. Regularly reviewing and refining the AA model ensures alignment with evolving ethical standards and societal expectations, reflecting a commitment to responsible AI development (Tabassi, 2023).

Finally, the use of AA techniques as legal evidence in forensic and criminal cases faces challenges in meeting the admissibility standards established by the 1993 Daubert ruling (Gold et al., 1993). These standards require scientific testimony to include attributes such as calculated er-ror rates, empirical testing, standardized procedures, peer review, and acceptance within the scientific community. Traditional AA techniques often fall short of these criteria, limiting their use to investigations and preventing their admission as legal evidence (Chaski, 1997; Yang and Chow, 2014b). To improve AA's acceptance in courtrooms, Chaski (2005a); Frye and Wilson (2018) propose approaches that enhance accuracy, establish error rates, and meet the stringent Daubert criteria (Gold et al., 1993). Additionally, all documentary evidence submitted to the court must undergo manual review to comply with authentication and expert testimony standards (Howard, 2008).

**SI1. Is there a disclaimer to alert readers to potentially harmful content in the research?**
– The research does not mention a specific disclaimer regarding potentially harmful content.
**Suggestion:** Given the sensitive nature of AA research, including a clear disclaimer alerting readers to potentially harmful content is advised. This disclaimer would help prepare readers and mitigate any distress caused by exposure to sensitive information or topics discussed in the research.

**SI2. What measures are in place to minimize the potential trauma experienced by individuals during the design, development, and deployment stages? Are there regular check-ins amongst team members to ensure clear communication and support maintaining a healthy and safe working environment? Is mental health and psychological support offered to team members dealing with harmful text?**
– No details are provided in the research regarding this matter.
**Suggestion:** To address the potential trauma associated with handling sensitive content, it is crucial to implement measures such as providing access to mental health and psychological support services for all team members. Regular check-ins and open communication channels should be established to monitor individuals' well-being and promptly address any concerns. Additionally, incorporating training on resilience and coping strategies, as well as rotating tasks to limit prolonged exposure to harmful content, can further support maintaining a healthy and safe working environment.

**SI3. Does the scope of the AA model align with its intended purpose to minimize potential misuse?**
– The model's development and application specifically aim to identify potential human trafficking operations, suggesting alignment with the intended purpose. However, outlining explicit use cases and restrictions could further minimize potential misuse.
**Suggestion:** To minimize potential misuse of the AA model, we recommend clearly defining its scope and intended purpose by specifying explicit use cases, limitations, and restrictions. This can help ensure the model is used ethically and aligns with its intended goals, reducing the risk of misuse.

**SI4. Does the AA processing encompass systematic and extensive automated processing that leads to decisions with legal or significant effects on individuals? Are measures in place to prevent identity disclosure, reputational damage, or legal ramifications?**
– While the research involves automated processing to identify potential human trafficking operations, specific measures to prevent identity disclosure and reputational damage are mentioned, such as data anonymization and personal information masking. Furthermore, the authors emphasize that law enforcement agencies and researchers should not solely depend on our analysis as evidence for criminal prosecution. However, the research does not detail safeguards against potential legal consequences for individuals incorrectly identified. The research provides these details under Section 8, Privacy Considerations and Potential Risk, and Legal Impact paragraphs.
**Suggestion:** To enhance safeguards, we recommend establishing protocols to verify AA results before any legal action is taken and issuing disclaimers on their use as evidence. If AA results are to be used for prosecution, the methods and findings must be scientifically validated, empirically tested, and presented by qualified experts who can clearly explain them to jurors and withstand cross-examination (Howard, 2008).

**SI5. Is there a mechanism to mitigate the risk of potential misuse and abuse, including scenarios involving targeted harassment, social engineering, or the creation of deceptive content falsely attributed to others?**

– While the research outlines privacy-preserving measures under Section 8, Privacy Considerations and Potential Risk paragraph, it does not specifically address mechanisms to mitigate misuse and abuse, such as targeted harassment. Implementing comprehensive ethical guidelines and employing restrictions is essential for minimizing such risks.
**Suggestion:** Implementing comprehensive restrictions such as user access controls, monitoring systems, and reporting protocols can help prevent misuse and abuse, including targeted harassment, social engineering, and the creation of deceptive content. Incorporating a robust consent framework, where stakeholders are informed of potential risks and agree to ethical usage, can also be valuable. Regular training sessions for users on ethical considerations and potential risks, combined with periodic audits and updates to the system, would ensure ongoing alignment with ethical standards and reduce the likelihood of misuse. Finally, establishing a clear process for detecting and responding to misuse can provide a practical response to ethical breaches.

#### PROJECT MANAGER

Human oversight and intervention mechanisms are critical in sensitive AA applications to mitigate ethical issues and potential harm. AA systems may inadvertently disclose sensitive information, misattribute content, or be exploited for harmful purposes, such as targeted harassment or social engineering. Without proper oversight, these risks can result in significant ethical violations, privacy breaches, and data misuse, potentially causing harm to individuals or groups, damaging reputations, or leading to legal consequences. Oversight mechanisms ensure that the model's outputs align with ethical standards and societal norms, reducing the likelihood of unintended negative outcomes.

The project manager is responsible for establishing and integrating these oversight processes into the project. This includes coordinating ethical review boards, scheduling regular audits, and creating feedback channels. Their primary role involves fostering a culture of ethical awareness and accountability, ensuring that all team members understand and adhere to established ethical guidelines. By embedding these practices into the project framework, project managers help safeguard against ethical risks and support achieving both technical and societal objectives.

**SI6. Are there mechanisms for human oversight and intervention to review and reject content with ethical concerns?**
– The research does not detail mechanisms for human oversight and intervention.
**Suggestion:** To ensure ethical integrity, it is recommended to implement mechanisms for human oversight and intervention. This could include establishing an ethical review board, regular audits, and a pre-release content review process, allowing for identifying and rejecting content with ethical concerns. These measures will help prevent misuse and uphold ethical standards.

#### Developer

Minimizing the carbon footprint and energy consumption during the training of AA models is essential for aligning AI development with sustainability goals (Strubell et al., 2019; Bannour et al., 2021). Developers can adopt proactive measures by prioritizing efficient algorithms and training strategies that reduce computational demands, optimizing model architectures through techniques like knowledge distillation (Beyer et al., 2022; Panov et al., 2022), and leveraging data-efficient methods such as transfer learning (Gupta et al., 2020; Barlas and Stamatatos, 2021b; Hessenthaler et al., 2022; Silva et al., 2023). Additionally, monitoring and quantifying carbon emissions during model training using carbon tracking tools (Lacoste et al., 2019; Anthony et al., 2020) can offer valuable insights for implementing optimization and offsetting strategies.

**SI7. Are efficient algorithms and training strategies given priority to minimize the carbon footprint and energy consumption? Is carbon tracking employed to monitor and quantify carbon emissions during Authorship Attribution model training, aiding in optimization and offsetting strategies?**
- The authors perform fine-tuning on the pre-trained checkpoints to perform the AA task. The environmental impact of the research, including carbon footprint and energy consumption, is described in Section 8.

#### End-User

End-users are critical in ensuring AA tools are employed ethically and aligned with societal values. They must adhere to strict ethical guidelines and limit their use to legitimate purposes such as research, security, or intellectual property protection. Misuse of AA applications—such as for targeted harassment, unauthorized surveillance, spreading misinformation, or falsely attributing content to harm others—poses serious ethical risks and can result in legal consequences. Vigilance is essential to prevent abuse, promote transparency, and safeguard privacy and individual rights. This responsibility includes remaining alert to potential misuse, taking steps to prevent unethical behavior, and ensuring that all usage respects the privacy and rights of individuals. By adhering to these ethical standards, end-users help ensure AA applications are deployed responsibly, contributing positively to society while upholding trust and integrity.

**SI8. How are end-users educated and held accountable to ensure that AA applications are used ethically and aligned with societal values? Are there mechanisms to monitor usage and prevent misuse, such as targeted harassment, unauthorized surveillance, or spreading misinformation?**
- No such details are provided in the research.
**Suggestion:** To ensure the ethical use of AA applications, end-users should receive mandatory training on ethical guidelines and societal values. Additionally, implementing monitoring systems to track usage and clear policies and consequences for misuse can help prevent unethical behavior such as targeted harassment, unauthorized surveillance, or spreading misinformation. Regular audits and user compliance checks can further enhance accountability.

This framework demonstrates how the proposed questionnaire is a practical tool for translating ethical guidelines into actionable steps throughout the AA lifecycle. The case study underscores the complexities of applying these principles by highlighting the necessity of flexibility in addressing domain-specific challenges, such as balancing data privacy with system utility and ensuring fairness despite resource constraints. Importantly, the structured questionnaire approach fosters stakeholder reflection, accountability, and continuous improvement, reinforcing the alignment of technical decisions with ethical priorities. This exercise emphasizes the importance of integrating ethical considerations early and iteratively, recognizing that responsible AA development is an evolving process shaped by context-specific trade-offs and emerging challenges.

## 5.4 Broader Discussion & Challenges

While the proposed responsible AA guidelines are comprehensive, they are not without limitations. Every AA task involves a unique combination of factors—such as text data type, the specific domain, and the intended application—that can complicate their implementation. Although the framework offers a structured approach, it may not fully capture every nuance or address all considerations in every scenario. To truly assess its effectiveness in promoting responsible AA, rigorous testing is needed to evaluate how well it can be operationalized. Real-world applications will be critical for validating and refining the framework's practices. Additionally, the diversity of AA applications means that certain aspects of responsibility may need to be prioritized over others, and practical constraints—like limited data availability, resource limitations, and contextual factors—may require trade-offs. Researchers must carefully evaluate these trade-offs against the potential benefits of their models and methodologies.

For instance, the IDTraffickers case study, which used AA to identify potential human trafficking vendors, revealed key insights and significant challenges. Implementing privacy measures, such as data masking and anonymization, was relatively straightforward and helped reduce identification risks, aligning with GDPR standards. However, ensuring comprehensive data protection across the entire data lifecycle—including secure storage and handling—proved more complicated due to the variability of the data. Addressing fairness and non-discrimination also posed challenges. While it was possible to mitigate algorithmic biases, creating a diverse dataset by collecting data from multiple platforms and demographics was resource-intensive and, in some cases, impractical. Balancing transparency through explainability was another hurdle, as the complexity of neural networks made it difficult to generate reliable explanations or debug certain outcomes. Additionally, providing mental health support to team members working with sensitive data underscored the need for structured support systems and ethical oversight. While crucial, implementing such support was challenging in a university-based research environment with a small team and limited resources.

These examples illustrate that the trade-offs in implementing responsible AA practices can vary widely depending on the application. Given the diversity of AA use cases and their unique challenges, expecting a one-size-fits-all set of recommendations for every stakeholder and scenario is unrealistic. Practitioners must recognize that while the guidelines provide a solid foundation, context-specific adjustments, and trade-offs are often necessary to align with the ethical considerations and practical needs of each application. This emphasis on flexibility allows practitioners to adapt the framework to their specific circumstances while promoting responsible AA practices. Ultimately, ensuring responsible AA is not a one-size-fits-all endeavor. While the framework offers valuable guidance, practitioners must remain adaptable to the unique characteristics and requirements of their AA tasks. Continuous evaluation and refinement of the framework, informed by real-world experiences and evolving ethical standards, will be essential for addressing AA's dynamic challenges and responsibilities.

## 5.5 SUMMARY

This study responds to RQ3 by introducing a framework of a comprehensive set of responsible guidelines to address ELSI/ELSA considerations in AA within NLP. It proposes a framework of structured questionnaires to aid in navigating the complexities and trade-offs inherent in AA applications by focusing on privacy and data protection principles, fairness and non-discrimination, transparency and explainability, and greater societal good. This framework aims to guide researchers and stakeholders at all stages of an AA application's lifecycle, from initial design to deployment, helping identify and address potential ethical issues. Furthermore, this study demonstrates the application of these guidelines through a case study on sensitive AA research, providing a practical example of how they can inform the ethical assessment of AA projects. It is important to note that while the framework aims for comprehensiveness, it may not cover every unique AA application, necessitating rigorous real-world testing to ensure its effectiveness. Different scenarios and AA applications may require varying priorities and trade-offs, demanding flexibility and adaptation to specific AA task characteristics. Therefore, continuous evaluation and refinement of the framework, informed by practical experiences and evolving ethical standards, are essential.

# 6

# DISCUSSION

This chapter discusses a datasheet for the IDTraffickers and MATCHED datasets, outlining their motivation, structure, collection process, and intended use. Furthermore, the limitations of this work are examined on a broader scale, assessing the challenges of AA approaches in online criminal platforms, computational constraints, explainability issues, and the lack of responsible guidelines for multimodal systems. The chapter suggests future research directions to tackle these gaps. Finally, ethical considerations are addressed, emphasizing data protocols, the steps to ensure privacy and minimize harm, legal and environmental impact, and compliance with institutional review board guidelines.

## 6.1 DATASHEET

Following the framework proposed by Gebru et al. (2021), this section presents a detailed datasheet for the IDTraffickers and MATCHED datasets.

### 6.1.1 MOTIVATION

**For what purpose was the dataset created? Was there a specific task in mind? Was there a specific gap that needed to be filled? Please provide a description.**

- LEAs and practitioners rely on sex trafficking indicators to distinguish between online escort advertisements and human trafficking cases. These investigations involve connecting ads to specific individuals or trafficking networks, often by examining phone numbers or email addresses. However, existing methods for linking these escort ads have limitations. Therefore, this research presents IDTraffickers and MATCHED, two authorship attribution datasets, to identify and link potential human trafficking vendors by analyzing and linking distinctive stylistic cues in online escort ads.

**Who created the dataset (e.g., which team, research group) and on behalf of which entity (e.g., company, institution, organization)?**
- Both the datasets are created by Law & Tech Lab at Maastricht University.

### 6.1.2 COMPOSITION

**What do the instances that comprise the dataset represent (e.g., documents, photos, people, countries)? Are there multiple types of instances (e.g., movies, users, and ratings; people and interactions between them; nodes and edges)? Please provide a description.**
- The IDTraffickers dataset comprises instances of raw text sequences generated by merging the title and description of escort ads using the [SEP] token. Similarly, the MATCHED dataset includes two components: (1) a raw text sequence formed by combining the title and description of the escort ad, separated by a [SEP] token, and (2) one or more images associated with the ad, typically depicting the escort being advertised. Each ad instance in both datasets is linked to a vendor ID (e.g., "Vendor 10"), a unique identifier representing the individual or organization responsible for posting the ad.

**How many instances are there in total (of each type, if appropriate)? What data does each instance consist of? “Raw” data (e.g., unprocessed text or images)or features? Is there a label or target associated with each instance?**
- The IDTraffickers dataset consists of 87,595 unique text ads linked to 5,244 vendors. The dataset is structured as a Pandas DataFrame and stored in a .csv file featuring two primary columns: "TEXT" and "VENDOR." The MATCHED dataset comprises 28,513 ad instances, including 27,619 unique text descriptions and 55,115 escort images associated with 3,549 unique vendors. Each instance in the dataset contains raw, unprocessed text and image data. The dataset is provided as a pandas DataFrame in a .csv format, with three main columns: "TEXT," "IMAGES," and "VENDOR." The "TEXT" column contains input text sequences in string format, created by merging the title and description of the ad. The "IMAGES" column stores the local file paths for images associated with the ad. The "VENDOR" column includes class labels as integer IDs corresponding to specific vendors.

**Does the dataset contain all possible instances or is it a sample (not necessarily random) of instances from a larger set? If the dataset is a sample, then what is the larger set? Is the sample representative of the larger set (e.g., geographic coverage)? If so, please describe how this representativeness was validated/verified. If it is not representative of the larger set, please describe why not (e.g., to cover a more diverse range of instances because instances were withheld or unavailable).**
- The IDTraffickers dataset consists of advertisements from the Backpage escort market, covering locations across 14 states and 41 cities in the United States. Figure 6.1 illustrates the density of unique advertisements collected across these states. Initially, a total of 513,705 ads were collected. However, ads lacking phone numbers are filtered out, as these phone numbers serve as the ground truth for establishing vendor identities. This filtering process resulted in a refined dataset of 202,439 ads. The MATCHED dataset represents a subset of the IDTraffickers dataset, with ads collected from seven cities (Chicago, Atlanta, Houston, Dallas, Detroit, New York, and San Francisco) across five U.S. states. To ensure a reliable ground truth for AA tasks, the dataset is further filtered to include only ads containing phone numbers (used to establish vendor labels) and at least one associated image. This filtering process yielded a final set of 28,513 ads. While the dataset does not encompass the entirety of the Backpage escort market, it focuses on instances where both text and image modalities are available, which is critical for exploring MAA approaches.

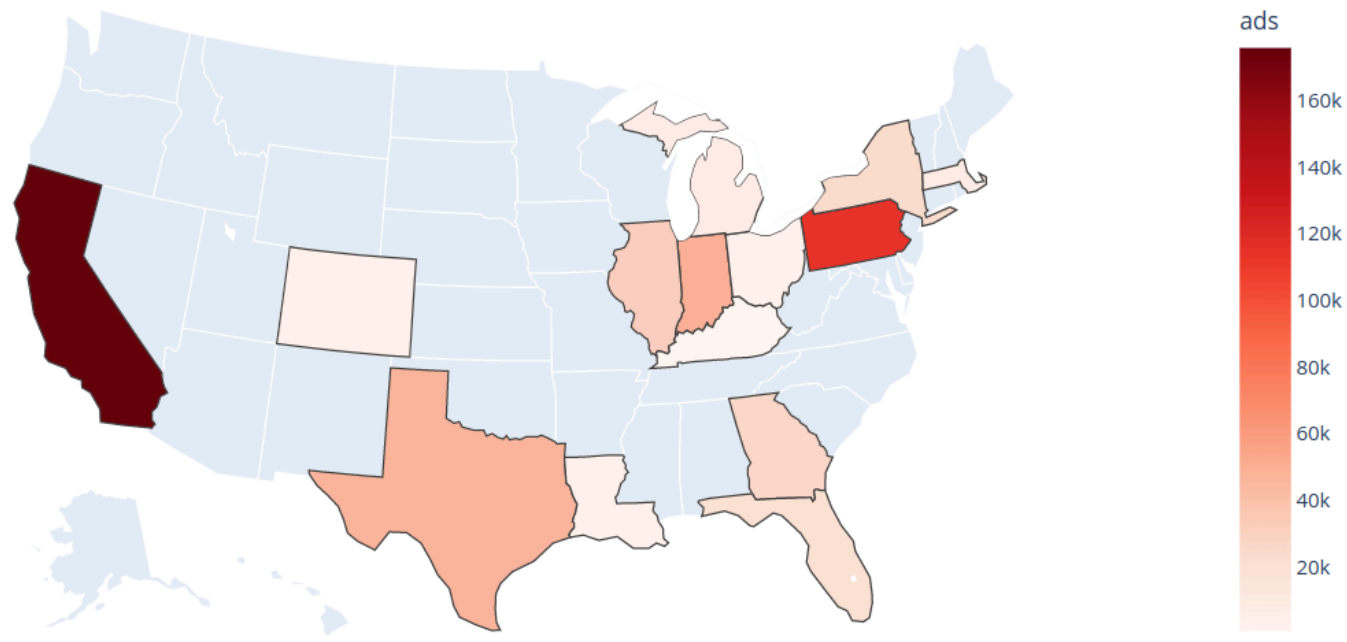

Figure 6.1: Density of all advertisements collected from Backpage Escort Market across American states

**Is any information missing from individual instances? If so, please provide a description explaining why this information is missing (e.g.,**

**because it was unavailable). This does not include intentionally removed information but might include, e.g., redacted text.**
- No.

**Are relationships between individual instances made explicit (e.g., users' movie ratings, social network links)? If so, please describe how these relationships are made explicit.**
- Ground truth labels for the authorship task in both the IDTraffickers and MATCHED datasets are generated by extracting phone numbers from the ads using the TJBatchExtractor (Nagpal et al., 2017) and CNN-LSTM-CRF classifier (Chambers et al., 2019). These phone numbers serve as vendor identifiers, enabling vendor communities to be created using NetworkX (Hagberg et al., 2008). Each community is assigned a unique label ID, which forms the basis for vendor labels.

**Are there recommended data splits (e.g., training, development/validation, testing)? If so, please provide a description of these splits, explaining the rationale behind them.**
- The dataset is divided into training, validation, and test sets using a 0.75:0.05:0.20 split ratio. This allocation ensures a substantial training set (75%) for effective model learning, a validation set (5%) for hyperparameter tuning and preventing overfitting, and a test set (20%) to evaluate model generalization and in-sample distribution performance.

**Are there any errors, sources of noise, or redundancies in the dataset? If so, please provide a description.**
- As illustrated in Figure 3.4, a significant amount of noise is present in the Backpage escort ads. Text descriptions in these ads often contain extra punctuation, emojis, irregular white spaces, and random characters, likely used by vendors to evade automated detection systems. These irregularities complicate text processing and increase the complexity of data-cleaning efforts. Manual inspection of the image data also reveals visual noise, such as intentionally blurred areas and white noise, which adds further challenges to the analysis. Quantifying the extent of this noise in images remains difficult. Despite these issues, the noise and irregularities reflect the original conditions under which the ads were posted, providing a realistic foundation for developing robust AA models capable of handling similar real-world scenarios. Additionally, analysis in Figure 3.10–3.11 suggests the possibility of two or more vendor labels being associated with the same entity. However, without absolute ground truth, definitive conclusions cannot be drawn.

**Is the dataset self-contained, or does it link to or otherwise rely on external resources (e.g., websites, tweets, other datasets)? If it links to or relies on external resources, a) are there guarantees that they will exist, and remain constant, over time; b) are there official archival versions of the complete dataset (i.e., including the external resources as they existed at the time the dataset was created); c) are there any restrictions (e.g., licenses, fees) associated with any of the external resources that might apply to a dataset consumer? Please provide descriptions of all external resources and any restrictions associated with them, as well as links or other access points, as appropriate.**
**-** No. Both datasets are self-contained.

**Does the dataset contain data that might be considered confidential (e.g., data that is protected by legal privilege or by doctor-patient confidentiality, data that includes the content of individuals' nonpublic communications)? If so, please provide a description.**
**-** Measures have been implemented to mitigate the potential privacy risks associated with escort ads across the IDTraffickers and MATCHED datasets. To protect privacy, sensitive details such as phone numbers, email addresses, age information, post IDs, dates, and links are entirely masked. Phone numbers are anonymized by replacing digits (the numbers) with the letter "N," while email addresses are replaced with $<EMAIL_ID>$, post IDs with $POST_ID : NNNNN$, dates with $<DATES>$, and links with $<LINK>$. Furthermore, entity recognition techniques (Li et al., 2022b; Liu et al., 2023) are employed in an attempt to mask escort names and posted locations, though noise in the data results in some false positive predictions. Since the existing research indicates that escorts often use pseudonyms in advertisements (Carter et al., 2021; Lugo-Graulich, 2024), and no public records of these ads exist after the seizure of Backpage Escort Markets in 2016, reducing the likelihood of personal data being exploited to harm individuals, escort names and locations of the posted ads are left unmasked at the end.

Privacy risks associated with images in the MATCHED dataset are more complicated to address. Initially, face blurring enhances anonymity, but manual inspection reveals that many images already have faces blurred or cropped by the posters. Additional blurring is avoided to preserve natural stylistic cues critical for AA tasks, as it could introduce unintended biases.

A sanity check using the FairFace (Karkkainen and Joo, 2021) and Deep-Face (Serengil and Ozpinar, 2023) models confirms that these tools cannot extract ethnicity or age-related information from the noisy image dataset. These measures collectively ensure that privacy risks are minimized while maintaining the integrity of the datasets for research purposes.

**Does the dataset contain data that, if viewed directly, might be offensive, insulting, threatening, or might otherwise cause anxiety? If so, please describe why.**
- Yes, the datasets comprise text and (semi-nude) images from escort ads that contain sexual descriptions.

**Does the dataset identify any subpopulations (e.g., by age, gender)? If so, please describe how these subpopulations are identified and provide a description of their respective distributions within the dataset.**
- The datasets do not explicitly identify subpopulations by age, as all age information has been masked in the text ads. However, some ads include descriptions of the escorts' ethnicities, which remain unmasked to preserve the original stylometric features for AA tasks. Additionally, most ads in the datasets correspond to women-based escort services. It is important to note that while we have not provided age or ethnicity labels, malicious users could potentially infer such details by applying automated systems to the images. This potential for inference underscores the importance of responsible dataset usage and adherence to ethical guidelines to prevent misuse.

**Is it possible to identify individuals (i.e., one or more natural persons), either directly or indirectly (i.e., in combination with other data) from the dataset? If so, please describe how.**
- While the possibility of identifying individuals through the dataset cannot be entirely ruled out, extensive privacy measures are implemented to minimize this risk. In text ads, private identifiers such as phone numbers, email addresses, and other personal information are masked to protect individual identities. The dataset consists of ads from the Backpage escort market collected between December 2015 and April 2016, a period for which no public records exist following the website's seizure. However, a residual risk remains associated with the images in the dataset, as they may still allow for indirect identification of individuals.

To address this risk, access to the IDTraffickers and MATCHED datasets is restricted to approved researchers or agencies with legitimate research objectives, particularly those focused on combating human trafficking or conducting academic (non-commercial) AA research. Access is granted through a Dataverse data portal, subject to approval from an ethics review board, ensuring that the dataset is used solely for its intended purposes. Unauthorized use of the dataset, especially for purposes beyond AA or human trafficking research, is strictly prohibited under ethical guidelines and carries legal consequences.

**Does the dataset contain data that might be considered sensitive in any way (e.g., data that reveals race or ethnic origins, sexual orientations, religious beliefs, political opinions or union memberships, or locations; financial or health data; biometric or genetic data; forms of government identification, such as social security numbers; criminal history)? If so, please provide a description.**
**-** Despite extensive masking efforts, both datasets retain some sensitive information. While private identifiers such as phone numbers and email addresses are successfully masked, challenges persist in anonymizing other potentially sensitive details, including escort names, ad locations, ethnicities, and sexual orientations. These details are embedded in the text descriptions and could be extracted from the images using automated systems. The inherent noise in the data further complicates the accurate masking of these elements. Consequently, despite significant precautions, there remains a possibility that sensitive information could still be inferred from the dataset.

6

### 6.1.3 COLLECTION PROCESS

**How was the data associated with each instance acquired? Was the data directly observable (e.g., raw text, movie ratings), reported by subjects (e.g., survey responses), or indirectly inferred/derived from other data (e.g., part-of-speech tags, model-based guesses for age or language)? If the data was reported by subjects or indirectly inferred/derived from other data, was the data validated/verified? If so, please describe how.**
**-** The text descriptions are Images are directly scraped from the raw ads of Backpage Escort markets. Vendor labels in both the IDTraffickers and MATCHED datasets are generated by extracting phone numbers from the ads using the TJBatchExtractor (Nagpal et al., 2017) and CNN-LSTM-CRF classifier (Chambers et al., 2019). These phone numbers serve as identifiers to group ads into vendor communities, which are constructed using NetworkX (Hagberg et al., 2008). Each community is assigned a unique label ID, forming the basis for vendor labels. Given the large datasets and the prevalence of phone numbers in the advertisements, manual validation of each extracted phone number is infeasible. To address this, the CNN-LSTM-CRF classifier is trained using five random initializations to assess the robustness of the predictions. Comparisons across these initializations demonstrate high performance, with 99.86% Levenshtein accuracy, 99.82% perfect accuracy, and 99.50% digit accuracy, indicating strong robustness in the model's predictions. This approach ensures reliable grouping of ads by vendors, supporting the AA task.

**What mechanisms or procedures were used to collect the data (e.g., hardware apparatuses or sensors, manual human curation, software programs, software APIs)? How were these mechanisms or procedures validated?**
- The raw data is provided by Bashpole Software, Inc.

**Did you collect the data from the individuals in question directly, or obtain it via third parties or other sources (e.g., websites)? Over what timeframe was the data collected? Does this timeframe match the creation timeframe of the data associated with the instances (e.g., recent crawl of old news articles)? If not, please describe the timeframe in which the data associated with the instances was created.**
- The IDTraffickers dataset comprises ads from 41 cities across 14 U.S. states, collected from online postings on the Backpage Escort Markets between December 2015 and April 2016. The MATCHED dataset, a subset of the IDTraffickers dataset, includes ads from seven U.S. cities, also scraped from online postings on the Backpage Escort Markets during the same time period.

**Were the individuals in question notified about the data collection? If so, please describe (or show with screenshots or other information) how notice was provided, and provide a link or other access point to, or otherwise reproduce, the exact language of the notification itself.**
- The individuals featured in the ads were not notified about the data collection. Given that the ads were posted on Backpage between December 2015 and April 2016, obtaining consent from these individuals is infeasible. Since the Backpage escort market was seized and shut down, reconnecting with these individuals—many of whom used pseudonyms and transient contact information such as phone numbers or email addresses—is impractical after such a long period. Additionally, as Backpage

Escort Marketplaces no longer exists as a platform, contacting the original posters would be challenging and unlikely to yield responses.

**Did the individuals in question consent to the collection and use of their data? If so, please describe (or show with screenshots or other information) how consent was requested and provided, and provide a link or other access point to, or otherwise reproduce, the exact language to which the individuals consented.**
- No.

**If consent was obtained, were the consenting individuals provided with a mechanism to revoke their consent in the future or for certain uses? If so, please provide a description, as well as a link or other access point to the mechanism (if appropriate).**
- Not Applicable.

### 6.1.4 PREPROCESSING, DATA CLEANING, AND LABELING

**Was any preprocessing/cleaning/labeling of the data done (e.g., discretization or bucketing, tokenization, part-of-speech tagging, SIFT feature extraction, removal of instances, processing of missing values)? If so, please provide a description. If not, you may skip the remaining questions in this section.**
- Extensive measures are implemented to protect sensitive information in both the IDTraffickers and MATCHED datasets to prioritize privacy and minimize the potential for misuse. In the IDTraffickers dataset, identifiable elements such as phone numbers, email addresses, age details, post IDs, dates, and links are masked within the ad descriptions. Additionally, all instances lacking phone numbers are removed. Similarly, the same masking procedures are applied to text descriptions in the MATCHED dataset to anonymize sensitive details. However, images are retained in their original form to preserve stylometric cues. Finally, instances without phone numbers or images are excluded to align with the research objectives. These safeguards maintain individuals' confidentiality while enabling valuable research and analysis.

**Was the "raw" data saved in addition to the preprocessed/cleaned/labeled data (e.g., to support unanticipated future uses)? If so, please provide a link or other access point to the "raw" data.**
- No.

6

**Is the software that was used to preprocess/clean/label the data available? If so, please provide a link or other access point.**
- The code to pre-process, clean, and label is available at IDTraffickers and IDTraffickers repositories.

### 6.1.5 USES

**Has the dataset been used for any tasks already? If so, please provide a description.**
- This research introduces IDTraffickers and MATCHED, two novel datasets comprising text descriptions and images from Backpage escort markets developed explicitly for AA tasks. While these datasets have not been used in prior studies, several works have reportedly employed text descriptions or images from Backpage escort marketplaces for similar analyses (Alvari et al., 2016; Portnoff et al., 2017; Alvari et al., 2017b), among others. However, due to these datasets' unavailability, verifying whether any ads overlap with those included in the IDTraffickers and MATCHED datasets is challenging.

**What (other) tasks could the dataset be used for?**
- Both datasets are specifically designed to enable AA methods, and it is crucial that they are used strictly within their defined scope and purpose. Any use outside of AA research or non-commercial applications is prohibited to ensure ethical and responsible utilization of the data.

**Is there anything about the composition of the dataset or the way it was collected and preprocessed/cleaned/labeled that might impact future uses? For example, is there anything that a dataset consumer might need to know to avoid uses that could result in unfair treatment of individuals or groups (e.g., stereotyping, quality of service issues) or other risks or harms (e.g., legal risks, financial harms)? If so, please provide a description. Is there anything a dataset consumer could do to mitigate these risks or harms?**
- Both datasets are scraped from raw ads posted on Backpage Escort Markets within the U.S. region between 2015 and 2016. They do not represent the complete scope of human trafficking activities and are specific to U.S.-based activities during that period. As such, the datasets are not generalizable to regions outside the U.S. or to the present day. Furthermore, extensive precautions mask sensitive information; however, the datasets still include potentially sensitive details such as escort pseudonyms, posted locations, ethnicity, and sexual preferences. While these details are unlikely to harm individuals directly, unethical applications—such as those aimed at re-identifying individuals, profiling, targeting, or stereotyping—are strongly discouraged. To mitigate risks, dataset users are advised to adhere strictly to ethical guidelines, focusing solely on the dataset's intended purpose of combating human trafficking through academic research. Further anonymization techniques, particularly for images, are recommended to prevent unintentional exposure or misrepresentation of individuals or groups. Users are urged to avoid practices that could compromise privacy or perpetuate harm.

### 6.1.6 DISTRIBUTION

**Will the dataset be distributed to third parties outside of the entity (e.g., company, institution, organization) on behalf of which the dataset was created? If so, please provide a description.**
- Yes, the datasets accessible to third parties via the DataverseNL data repository. To mitigate risks of illegal or unethical use, access will be granted under specific conditions, including mandatory signing of a non-disclosure agreement (NDA) and data protection agreements. Each application for access will be evaluated by our ethics committee to ensure alignment with the dataset's intended purpose. These agreements will prohibit data redistribution and restrict its use exclusively to ethical, non-commercial research, especially in contexts that support combating human trafficking.

**How will the dataset will be distributed (e.g., tarball on website, API, GitHub)? Does the dataset have a digital object identifier (DOI)?**
- Access to the datasets is restricted and will only be granted to individuals or organizations conducting academic research on AA-based non-commercial applications. Requests for access can be made through the dataset DOIs: IDTraffickers and MATCHED.

**Have any third parties imposed IP-based or other restrictions on the data associated with the instances? If so, please describe these restrictions, and provide a link or other access point to, or otherwise reproduce, any relevant licensing terms, as well as any fees associated with these restrictions.**
- No.

6

### 6.1.7 MAINTAINENCE

**Will the dataset be updated (e.g., to correct labeling errors, add new instances, delete instances)? If so, please describe how often, by whom, and how updates will be communicated to dataset consumers (e.g., mailing list, GitHub)?**
- As the research in this field progresses, efforts will be focused on exploring advanced NLP-based entity extraction techniques to enhance the privacy of individuals in the dataset. These methods can mask escort pseudonyms, posted locations, and ethnicities to ensure anonymity and confidentiality. Additionally, plans are underway to expand the dataset by incorporating ads from multiple escort platforms, enabling the evaluation of the model's generalization capabilities on real-to-close-world OOD datasets. These updates aim to improve the privacy measures and the dataset's utility for robust, cross-platform AA tasks. Progress and findings will be shared with the research community through publications, and updates will be reflected in the dataset's description on the respective DataverseNL portals.

6

**If others want to extend/augment/build on/contribute to the dataset, is there a mechanism for them to do so? If so, please provide a description. Will these contributions be validated/verified? If so, please describe how. If not, why not? Is there a process for communicating/distributing these contributions to dataset consumers? If so, please provide a description.**
- Researchers are encouraged to collaborate on extending and improving the dataset through extensions, augmentations, or related enhancements. To safeguard the privacy and well-being of individuals represented in the dataset, sharing rights are restricted, meaning contributors cannot freely distribute the dataset. However, direct collaboration is welcomed, and validated contributions will be reviewed and integrated responsibly to enhance the dataset's utility. All validated contributions and updates will be acknowledged and communicated to the research community, ensuring transparency and recognition for collaborative efforts.

## 6.2 LIMITATIONS

This section examines the wider limitations and challenges of applying AA approaches across the entire research, going beyond the specific constraints outlined in each individual chapter.

**Authorship Attribution – A Supervised Task:** A current limitation of the proposed AA methods is their dependence on labeled data, requiring ground truth labels for model training and evaluation. Without these labels, extracting these stylistic patterns or cues that differentiate one author from another becomes challenging. This issue becomes crucial in cases involving pseudonymous or anonymous figures—such as online escort vendors—where labeled data is either extremely limited or sometimes nonexistent. As a result, unsupervised or semi-supervised AA approaches might struggle to capture these stylistic features essential for accurate attribution, leading to unreliable performance.

As discussed earlier, another major limitation of AA methods, when applied to cyber-enabled criminal markets, is the lack of verified ground truth to determine whether the same entity controls two vendors or represents distinct individuals. This creates uncertainty in the classification task, as models are trained on the assumption that all the vendors in the dataset represent individual entities. For instance, two vendors might exhibit similar writing styles due to intentional mimicry (or copycats), shared templates, or coincidental resemblances, which can result in false positives. Alternatively, a single vendor using multiple aliases can show varied stylistic patterns, leading to false negatives. Without definitive ground truths, evaluating these models becomes speculative, thereby casting doubt on the reliability and applicability of this research's findings in real-world investigative scenarios.

**Scalability & Computational Constraints:** While larger transformer-based backbones offer the potential for improved performance and robustness, their implementation is often constrained by limited computational resources during this research. Similarly, techniques such as larger batch sizes and mining hard negatives in contrastive learning training are well-documented for enhancing performance. However, these approaches demand substantial computational power, often exceeding the capabilities of the available DSRI infrastructure. To ensure efficiency and reproducibility, this research makes trade-offs by employing various optimization strategies, including DeepSpeed optimizers and mixed precision quantization strategies (Rasley et al., 2020), prioritizing smaller distilled models and smaller batch sizes to mitigate memory limitations. These constraints also reflect the practical realities LEAs face, which may lack access to state-of-the-art GPUs for model training. Frequent model updates are essential, given the dynamic nature of criminal activities on on-line platforms. Therefore, this research focuses on developing lightweight models that maintain competitive performance under minimal computational requirements. Future work will explore strategies such as knowledge distillation (Li et al., 2021a; Huang et al., 2023), transfer learning (Barlas and Stamatatos, 2021b; Fedotova et al., 2024), zero and few-shot learning (Chinea-Rios et al., 2022; Hung et al., 2023), and class incremental learning (Rahgouy et al., 2024) approaches to further optimize these models for deployment on resource-constrained hardware, including LEA-issued local systems.

**Scope and Age of the Datasets:** The scope and age of the datasets impose significant limitations on the generalizability and applicability of the findings to current and future scenarios. AA approaches are demonstrated using Darknet marketplace ads (2011–2018) and online escort marketplace ads (2015–2016); however, criminal entities frequently modify their behaviors to avoid detection, and vendors may increasingly employ LLMs to obscure writing and photometric styles. Although mere speculation with no definitive proof, these developments challenge the effectiveness of AA methods that rely on stylistic cues for linking vendors on illicit platforms. While Darknet market ads capture international trades, Backpage escort ads are predominantly localized to the U.S. and confined to a single platform. To assess the generalizability of AA approaches, it is crucial to analyze the ability of these AA approaches to link ads across markets spanning multiple platforms and countries, which requires data from diverse sources reflecting the global nature of online criminal activities. Moreover, the temporal scope of the datasets restricts the exploration of methodologies for dynamically updating models to accommodate emerging market trends, as vendors frequently alter their writing styles and behaviors. Future work will investigate strategies such as transfer learning and few-shot learning to determine how models can be updated with minimal data to capture new writing styles, with plans to recollect new data and conduct experiments addressing these limitations.

**Model Explainability:** This research employs feature attribution techniques in VendorLink and IDTraffickers studies to extract local features from the text ads—highlighting the influence of individual tokens—and global features that capture aggregated textual patterns across vendor ads, contributing to model predictions. Although these methods yield valuable insights into model behavior, they fail to capture complex feature interactions comprehensively. Furthermore, as demonstrated in Chapter 2 and Chapter 3, the resulting explanations frequently exhibit inconsistencies across different explainability frameworks and within samples from the same vendor class, likely due to network sparsity and the absence of sufficient granularity and contextual information (Das and Rad, 2020; Krishna et al., 2022). In multimodal AA approaches such as MATCHED, explainability techniques encounter additional challenges; current methods struggle to integrate textual and photometric modalities into unified, coherent explanations, thereby complicating the interpretation of interdependent signals across data types. Future work plans to extend research by developing advanced explainability and interpretability approaches that link patterns in writing and photometric styles to establish greater trust among LEAs.

**Responsible Guidelines for Multimodal Systems:** While Chapter 5 establishes responsible guidelines for authorship attribution (AA) in NLP, their extension to multimodal systems remains unexplored. This presents significant challenges, particularly within vision modalities, where images may contain private identifiers such as human faces, raising concerns about the misuse and potential harm to individuals in escort markets. As discussed in Chapter 4, anonymization techniques have been considered, such as adding Gaussian noise to blur images. However, such modifications introduce biases within AA tasks. Some images inherently feature blurred faces, reflecting a vendor-specific photometric style and behavior. Introducing artificial noise may alter these characteristics, leading to misattribution and impacting model performance. Therefore, while the proposed responsible AA framework provides a foundation for textual data in NLP-driven AA tasks, its adaptation to multimodal domain or, more specifically, vision-based modalities remains a critical avenue for future research.

## 6.3 ETHICAL CONSIDERATIONS

This section discusses mandatory data collection protocols, ethical considerations, potential risks, and legal, societal, and environmental impacts.

### 6.3.1 DATA PROTOCOLS

**VendorLink:** Ethical concerns associated with web scraping do not apply to this research, as the online darknet data in this study is obtained through a signed Memorandum of Agreement (MoA) with the IMPACT Cy-ber Trust portal (ICC). This data is freely available, legally collected, and distributed for large-scale cybersecurity analytics, enabling researchers to advance the state-of-the-art in cyber-risk R&D and decision support.

**IDTraffickers and MATCHED:** The datasets in these studies is collected from the Backpage Escort Markets, with ads posted between December 2015 and April 2016 across various locations in the United States. In their work, Krotov et al. (2020) characterize the ethical implications of web scraping using seven guidelines. Adhering to these guidelines, no explicit prohibitions regarding the use policy against data scraping on the Backpage website are stated. Furthermore, the meta-data to the dataset is available through the DataverseNL data repository to mitigate any potentially illegal or fraudulent use. Access to the data is subject to specific conditions, including the requirement for researchers to sign a non-disclosure agreement (NDA) and data protection agreements. These agreements prohibit sharing the data and its use for unethical commercial purposes. Considering that the data was collected before the seizure of the Backpage escort markets, it does not likely pose any substantial harm to the website or its web server.

### 6.3.2 Privacy Considerations and Potential Risks

**VendorLink:** Using vendor names in the study raises potential privacy concerns. However, a Data Privacy Impact Assessment (DPIA) conducted at Maastricht University concludes that the vendor names used are pseudonyms and do not reflect any individual's identity. Research indicates that the lifespan of Darknet vendors and marketplaces typically ranges from a few months to a couple of years (Booij et al., 2021; Broadhurst et al., 2021b; UNDOC, 2020). Given that the market ads in the dataset span from 2011 to 2018, the likelihood of any vendor existing with the same username is very low. Under Article 6 of the GDPR, the processing of personal data is considered lawful when carried out in the public interest. This study aims to benefit LEAs and potentially save lives by addressing illegal activities on the Darknet markets. While using vendor names promotes transparency and reproducibility, vendors are encouraged to reach out with concerns, and immediate action will be taken to remove their information if necessary.

While the study acknowledges the potential for misuse, particularly in matching user activity across public platforms, the extreme difference in language between Darknet and surface web websites makes such exploita-tion unlikely (Choshen et al., 2019). Despite this, the possibility of privacy infringement outside criminal markets exists, particularly in matching user activity across public social media platforms. Ill-intentioned third parties or organizations could exploit the study to circumvent an individual's identity. Therefore, readers are urged to consider the ethical implications when using the research for authorship technologies in cybersecurity and beyond.

**IDTraffickers:** The study acknowledges potential privacy concerns associated with the information within escort advertisements. Extensive measures are taken to mitigate these risks, including masking personal information such as phone numbers, email addresses, age details, post IDs, dates, and links. This masking process eliminates personal information to prevent reverse engineering. For example, an email address like xzy@gmail.com is transformed into <EMAILID-23>, a phone number such as +00 00 0000 000 becomes +NN NN NNNN NNN, and a link like www.google.com is rendered as <LINK> within the dataset. While entity recognition techniques are experimented with to mask escort names and posted locations mentioned in the advertisements, the noise in the data presents challenges in accurately masking these segments, resulting in false positive entity predictions.

In their research, Carter et al. (2021); Lugo-Graulich (2024) indicates that escorts often use pseudonyms in their ads. Additionally, no public records of these ads exist after the seizure of Backpage Escort Markets in 2016. This reduces the likelihood of personal data being exploited to harm individuals. The masking methodology aligns with the spirit of GDPR Article 6, which acknowledges the importance of balancing privacy concerns with societal benefits. The study contributes to addressing critical issues such as human trafficking and enhancing law enforcement efforts to reduce harm and save lives.

To prevent public misuse, access to the IDTraffickers dataset is granted via the DataverseNL data portal, subject to stringent restrictions. Researchers seeking access must undergo an application process requiring approvals and assurances. These assurances include non-disclosure and data protection agreements, ensuring responsible and secure use and data release.

**MATCHED:** Like IDTraffickers, extensive measures are taken to mask personal identifiers within the dataset. Phone numbers, email addresses, post IDs, dates, and links in text data are transformed into generalized formats such as "<EMAILID-23>" or "<LINK>" to minimize the risk of reverse engineering and personal identification. SoTA entity recognition tools are explored to mask names (Li et al., 2022b; Liu et al., 2023) and locations, though inherent noise in the data leads to inaccuracies and false positives.

Privacy risks are more challenging to mitigate for image data, as stylistic cues must be preserved for analysis. While initial consideration is given to blurring faces, this approach is ultimately rejected to avoid introducing biases that could compromise the authenticity of stylometric patterns. Many ads already feature images with blurred or cropped faces, suggesting an attempt by individuals to maintain anonymity. Similarly, other image augmentations, such as flipping or rotating, are avoided to preserve stylistic features tied to individual vendors.

Finally, like IDTraffickers, access to the MATCHED dataset is restricted and provided to researchers and organizations with legitimate research goals, particularly those focused on anti-trafficking and public welfare. Each access request undergoes a thorough review by an ethics review board, assessing the legitimacy of the research goals and the adequacy of the applicant's security measures. Applicants must sign non-disclosure and data protection agreements, legally binding them to ethical and secure usage standards. Only metadata on the DataverseNL platform provides a high-level overview without compromising sensitive information.

### 6.3.3 LEGAL IMPACT

This research aims to bring structure and meaning to the vast online data available on Darknet and escort markets to enhance the knowledge and tooling regarding investigative processes. While the specific impact of the research on law enforcement efforts cannot be predicted, the intent is to identify potential connections between vendors of illegal goods or services and provide LEAs with a broader information base for their internal processes. It is important to note that the research does not claim to provide the evidence necessary for prosecuting any criminal activity. Instead, the findings are intended to serve as tools to aid investigations, offering insights into vendor connections within online markets. LEAs, practitioners, and researchers are strongly advised not to rely solely on this analysis as direct evidence for criminal prosecution, but to use it as a supplementary resource to guide and enhance their investigative efforts.

### 6.3.4 ENVIRONMENTAL IMPACT

All the estimations below are conducted using and Machine Learning Impact calculator (Lacoste et al., 2019) and Equation (6.1):

$$\text{Carbon Emission (kg } CO_2) = \text{Power (kW)} \times \text{Training Time (hours)} \times \text{Carbon Intensity (kg } CO_2/\text{kWh)} \tag{6.1}$$

**VendorLink:** This study acknowledges that not all LEAs have the resources to train computationally expensive architectures. To address this, the study investigates using knowledge transfer to train low-compute-resource models. The transfer-BiGRU classifier demonstrates a carbon efficiency of 0.07 $kgCO_2eq/kWh$ and 2.25 $kgCO_2eq/kWh$ on the Vallhalla-Berlusconi and Traderoute-Agora datasets. In contrast, the BERT-cased classifier has a carbon efficiency of 0.12 $kgCO_2eq/kWh$ and 4.21 $kgCO_2eq/kWh$ on the same datasets. In total, approximately 377:52 hours of training are conducted for the statistical Bag-of-Words (BoW) models on a 20-core CPU with a Thermal Design Power (TDP) of 300W and a carbon intensity of 0.432 $kgCO_2/kWh$ for Europe, resulting in carbon emissions equivalent to 48.93 $kgCO_2/kWh$. For all neural network models, GPU training is employed on a Tesla V100-SXM2-32GB GPU for 513:33 hours with the same TDP and carbon intensity, resulting in carbon emissions equivalent to 55.44 $kgCO_2/kWh$.

**IDTraffickers:** All experiments in this study are conducted on a private infrastructure equipped with a Tesla V100-SXM2-32GB GPU (TDP of 300W), which has a carbon efficiency of 0.432 $kgCO_2eq/kWh$. The cumulative training time for establishing all baselines in the study amounts to 191 hours. The total estimated emissions for these experiments are calculated to be 24.62 $kgCO_2eq$.

**MATCHED:** The experiments for this research are conducted on a private infrastructure featuring an NVIDIA H100 80GB GPU (TDP of 350W) with a carbon efficiency of 0.475 $kgCO_2eq/kWh$. Establishing all baselines required a cumulative training time of 45:79 hours. The total estimated emissions for these experiments are approximately 16.625 $kgCO_2eq$.

## 6.4 DECLARATION ON THE USE OF LLMS

This research employs LLMs such as ChatGPT and Deepseek to rephrase and paraphrase sentences within the dissertation. They are also used to

6

translate the English summary of the research to its Dutch equivalent. However, these models are not used to generate original content, develop experimental results, or contribute to the conceptual framework of the research. Additionally, these LLMs are occasionally used to debug code, but they are never relied upon to create complete code implementations. All experimental designs, analyses, and findings are developed independently to ensure the integrity and originality of the research.

## 6.5 ETHICAL APPROVAL

This research has undergone rigorous ethical scrutiny by the Ethical Review Committee Inner City faculties (ERCIC) at Maastricht University and has received formal approval (`ERCIC_407_5_1_2023_van_Dijck`) to proceed with the project. The committee raised no ethical objections, confirming that the study adheres to established ethical standards and guidelines.

# 7
# Conclusions

This research introduces responsible AA approaches to support practitioners (researchers and LEAs) in analyzing, linking, and investigating cyber-enabled trafficking operations on online platforms. It presents two novel datasets for developing and evaluating AA solutions that leverage writing and photometric styles on ads from online criminal marketplaces. By utilizing these datasets—alongside existing ones—the study benchmarks the performance of SoTA AA models on vendor identification and verification tasks tailored explicitly to Darknet and Online Escort marketplaces, a common avenue for weapons, drugs, and human (sex) trafficking operations. It also explores advanced techniques, including transfer learning, multitask learning, and multimodal fusion of text and images, demonstrating significant improvements over unimodal baselines. Finally, this research establishes an ethical framework for the responsible AA application, emphasizing key principles such as privacy and data protection, fairness and non-discrimination, transparency and explainability, and the broader societal impact. This chapter synthesizes the key findings and presents the results for the three research questions outlined in Section 1.6.

***RQ1: How can authorship attribution methodologies assist in analyzing writing styles to identify and connect online vendors involved in cyber-enabled trafficking operations?***

This research demonstrates how foundational AA methods can extract distinctive linguistic signatures from text ads, enabling the systematic analysis and linking of vendor behaviors in the context of cyber-enabled traf-ficking operations. Together, Chapter 2 and Chapter 3 provide practitioners with computationally efficient AA tools to trace, connect, and study online criminal networks by profiling vendors based on their writing patterns. These chapters demonstrate how AA methods can be effectively applied to darknet and online escort marketplaces, enabling the identification and verification of vendors operating on multiple markets and under multiple aliases, enhancing the ability to disrupt trafficking operations across diverse digital platforms.

***- RQ1(a): How can authorship identification and verification approaches analyze and compare writing styles to connect existing (known) and emerging (upcoming) across darknet marketplaces?***

Chapter 2 introduces *VendorLink*, an NLP-based AA approach that demonstrates the effectiveness of fine-tuned transformer models—particularly cased contextualized transformers, which differentiate between upper and lower case text. These models significantly outperform traditional stylometric, statistical, and conventional neural network-based methods in capturing and linking writing styles from textual advertisements across seven darknet markets, including Alphabay, Dream, Silk Road 1, Valhalla, Berlusconi, Traderoute, and Agora. By identifying stylistic markers such as vocabulary usage, grammar, syntax, symbol patterns, and phrasing in the text ads, these models learn to extract distinctive linguistic fingerprints and attribute them to individual vendors on darknet markets. The study employs a BERT-cased architecture to perform vendor identification, a closed-set classification task where the model learns to attribute a text advertisement to a predefined set of vendors. Beyond classification, the learned representations are used to perform vendor verification, an open-set text similarity task aimed at linking aliases that vendors adopt to evade law enforcement detection across and within markets. Unlike identification, this verification task relies on cosine similarity to measure the similarity of writing styles between potential aliases and a parent vendor profile. Finally, the study demonstrates how these models can adapt and learn from emerging darknet markets using low-resource and compute-efficient transfer learning strategies, thereby balancing scalability with environmental sustainability as new platforms and writing styles emerge.

***- RQ1(b): How can authorship attribution approaches be applied in the absence of ground truth on online escort market-***

***places? How can the authorship verification approach be systematically designed and evaluated to ensure computational efficiency?***

Building on the findings of Chapter 2, Chapter 3 introduces a novel dataset, *IDTraffickers*, consisting of 87,595 text advertisements collected from the Backpage Escort Market—a platform widely recognized for facilitating human trafficking activities—between April 2015 and December 2016. These ads span 14 states and 41 cities in the United States. Unlike the Darknet markets, these escort ads lack vendor handles (vendor names) that can serve as ground truth for AA tasks. Therefore, the study extracts phone numbers from the ad descriptions and employs network analysis to construct vendor communities by linking these phone numbers to unique vendor IDs. Extensive benchmarking is performed using SoTA transformer baselines, with DeCLUTR-small emerging as the best-performing model for vendor identification (a closed-set classification task) during model training and vendor verification (an open-set retrieval task) during inference. Unlike the pairwise text similarity approach employed in Chapter 2, this study reframes vendor verification as a retrieval problem, significantly improving computational and memory efficiency—key for scaling to emerging markets. Instead of comparing one advertisement to another iteratively, the retrieval-based method uses FAISS-based similarity search to match a query ad against a pool of vendor embeddings, effectively forming clusters in the representation space. This shift enhances scalability and allows for robust performance evaluation using standard retrieval metrics such as Precision@K, Recall@K, Mean Average Precision (MAP@K), and R-Precision.

**Future Research:** Although the studies above examined feature-attribution-based methods for explaining model decisions, the results indicate that these explanations can be inconsistent across different frameworks and even for similar samples from the same vendor, likely due to network sparsity and a lack of granularity and contextual detail. For law enforcement to trust and transparently use these models, it is essential to develop explainability techniques that consistently and reliably highlight the stylometric signatures influencing the model decisions. Consequently, future work will focus on creating explainability and interpretability methods that provide clear, consistent attributions across vendor ads. Additionally, while transfer learning has effectively adapted to the language and trends of emerging markets and performs well in low-resource scenar-ios, its scalability varies across markets. For example, the BiGRU-transfer model does not generalize as effectively on the Traderoute-Agora market (high-resource dataset). Therefore, future research will explore improved training strategies, including few-shot learning, to better update models and enhance their adaptability to emerging markets.

> ***RQ2: How does integrating text with images and advanced training objectives affect authorship attribution generalization and effectiveness within online escort advertisements?***

Most existing AA research, including Chapter 3, relies primarily on NLP-based methods that focus exclusively on textual data, overlooking the inherently multimodal nature of online advertisements—especially within escort platforms where images accompany text. To bridge this gap, Chapter 4 introduces *MATCHED*, a multimodal dataset built upon IDTraffickers, comprising 27,619 unique text ads and 55,115 corresponding images sourced from Backpage across seven U.S. cities, organized into four geographic regions (South, West, Midwest, and Northeast). This study systematically benchmarks text-only, image-only, and multimodal AA models to assess their effectiveness in vendor identification (closed-set classification) and vendor verification (open-set metric learning) tasks. By incorporating multimodal fusion, contrastive learning, and multitask training, the study demonstrates that models integrating text and images significantly outperform their unimodal counterparts in classification and retrieval settings. Specifically, the most effective model fuses DeCLUTR-small (text encoder) and ViT-base-patch16-224 (image encoder) using mean pooling (latent-fusion technique) and is trained using a dual-objective loss: Cross-Entropy (CE) for discriminative learning in vendor identification, and Supervised Contrastive (SupCon) loss for aligning same-author representations in the latent space for vendor verification. This joint optimization enhances performance on in-sample data (South region) and improves generalization across out-of-distribution (West, Midwest, and Northeast) datasets—demonstrating robustness in real-world deployment scenarios. The results highlight that while textual features remain primary signals for authorship attribution, visual cues—such as photometric styles and background aesthetics—offer complementary information that strengthens overall model performance. Multimodal integration enables the models to capture richer stylistic patterns, enhancing their ability to detect known vendors and uncover emerging ones in unfamiliar markets. Beyond these technical contributions, the study also explores practical implications by supporting the construction of vendor-level knowledge graphs and offering practitioners scalable tools for mapping and investigating trafficking networks. By addressing key limitations of prior AA studies and focusing on generalization and real-world applicability, this study lays the groundwork for future research on multimodal AA in the context of cyber-enabled human trafficking.

**Future Research:** Despite the promising results, several limitations highlight directions for future research. First, although the datasets were segregated into training and OOD groups, a substantial overlap of shared vendors persists. While experiments indicate that model performance is consistent for both shared and unique vendors in OOD datasets, a rigorous test of generalization is needed—evaluating models on escort ads from platforms other than Backpage and extending the analysis beyond U.S.-specific data to capture the global nature of human trafficking. Second, the current work employs AA models on online ads collected from 2011 to 2018. However, the rise of LLMs suggests that vendors online may now use these automated systems to evade stylometric detection. Future studies should, therefore, evaluate model performance on recent data to accurately assess their applicability in today's landscape. Lastly, although local and global feature attribution techniques have been used to explain decisions in text-based AA, applying explainability methods to multimodal systems presents further challenges. Developing techniques that can effectively correlate textual features with corresponding visual cues—such as stylistic symbols, grammatical structures, image poses, or quality—remains an essential area for future exploration.

> ***RQ3: What ethical considerations must guide the development and deployment of authorship attribution systems applied to sensitive (high-risk) domains?***

Chapter 5 outlines a comprehensive responsible framework for developing and deploying NLP-based AA systems in sensitive domains, anchored in four core pillars. First, the framework emphasizes privacy and data protection by mandating rigorous safeguards such as data minimization, anonymization, and secure storage, alongside proactive measures like DPIAs in high-risk scenarios to ensure compliance with regulations such as GDPR and secure handling of sensitive authorship data. Second, it addresses fairness and non-discrimination by ensuring training datasets represent diverse authorship landscapes, mitigating selection bias through stratified sampling, expanding datasets to cover multiple text genres (e.g., escort ads from varied platforms), and analyzing demographic correlations to prevent inequitable outcomes. Third, the framework insists on transparency and explainability via local and global feature attribution methods, systematic error analysis of predictions, and documentation of model limitations to enable stakeholder understanding and intervention. Fourth, the framework prioritizes positive societal impact by mandating continuous evaluation of potential risks—including misuse, exposure of anonymized individuals, reputational harm, and environmental costs—and by advocating for stakeholder's education, monitoring, and providing mental health services for individuals handling sensitive content. Finally, accountability is established by assigning clear roles across the SDLC, including oversight mechanisms, regular audits, and protocols to manage liability (e.g., verifying results before taking action). To demonstrate the practical utility of these guidelines, the framework is evaluated through a case study on the published IDTraffickers research (Saxena et al., 2023a). A questionnaire is designed to apply the proposed guidelines, illustrating how ethical considerations can inform real-world decision-making. This includes managing priorities and trade-offs by privacy and utility, ensuring fairness despite limited demographic data, and mitigating risks of misuse while providing actionable recommendations for responsible development and deployment. Together, these pillars collectively ensure that AA systems are efficient, ethically sound, and aligned with broader societal and human rights values.

**Future Research:** As outlined in Chapter 5, while the proposed framework offers a structured foundation for ethical AA development, it has limitations. The variability of AA tasks—spanning context and type of text, domains, and applications—poses challenges in creating universally applicable guidelines. For instance, systems analyzing political documents versus social media content may face distinct demands and challenges. Practical constraints, such as limited access to representative datasets, resource disparities among developers, and different priorities (e.g., privacy versus utility), further restrict the framework's adaptability. These scenarios necessitate context-specific adjustments, where stakeholders must weigh ethical imperatives against feasibility. Future efforts aim to refine the framework through iterative real-world testing, particularly in scenarios with scarce resources or conflicting objectives. Only through ongoing evaluation and continuous improvement can the framework remain robust to ensure AA systems uphold ethical standards across diverse and dynamic contexts. Furthermore, the designed guidelines are only appli-cable to AA applications within the context of NLP. Future research would require to recalibrate these guidelines to accomodate multimodal AA, especially for vision driven applications.

# 8

# Academic and Real-World Relevance

Beyond addressing the abovementioned research questions, this research attempts to contribute to the academic literature and real-world practice in the fight against cyber-enabled trafficking. The work advances academic debates on responsible ML-based multimodal AA methods, while simultaneously offering practical tools, datasets, and frameworks that can support practitioners—including LEAs, intelligence analysts, and industry stakeholders—in analyzing, linking, and disrupting illicit online activities. This section, therefore, reflects on how the findings of this research can strengthen theoretical understanding, address gaps in existing research, and provide actionable value to those working on cybercrime investigation.

## 8.1 Advancing the Authorship Attribution Research

This work aims to broaden the scope of AA research by demonstrating how these methods can operate effectively in short, noisy, and deliberately obfuscated artifacts (text and images), where other traditional stylometric approaches perform poorly. By emphasizing the use of SoTA transformer models and their contextualized learning capabilities on large and small-scale datasets of real-world illicit ads, this research provides empirical evidence about the potential and limits and potential of modern AA methods beyond their traditional use cases.

Furthermore, this work advances AA research by initiating an academic discussion on how current AA methods can be effectively transformed post-hoc to perform an open-set authorship verification task, an area where ground truth is frequently unavailable or incomplete, especially relevant to applications designed for illicit online marketplaces where many vendors and markets emerge every day. The transformation of verification from computationally intensive pairwise-similarity analysis to a scalable retrieval problem offers a new methodological pathway for AA researchers seeking computationally efficient and practically deployable solutions. Therefore, this contributes not only a new empirical strategy but also a conceptual reframing of how authorship methods can be practically deployed and inferred in close-to-real-world open-ended scenarios.

## 8.2 ADVANCING THE STUDY AND ANALYSIS OF TRAFFICKING ACTIVITIES ON ONLINE ILLICIT MARKETS

This work deepens understanding of cyber-enabled trafficking by introducing two large-scale datasets—IDTraffickers and MATCHED—designed explicitly for AA research on illicit online markets. Whereas prior studies have often relied on smaller, mostly single-platform datasets, this research shows how computationally efficient AA methods can uncover behavioral consistencies within and across marketplaces, revealing hidden vendor structures that qualitative analyses alone are unlikely to detect.

These contributions advance the broader academic discussion on the use of machine learning to study criminal ecosystems. Moreover, the findings illustrate how computational profiling can complement criminological theories of offender behaviour, particularly in contexts where identities are anonymized or intentionally deceptive. Although the applied models are evaluated on Darknet and online escort markets, the underlying methodological principles are transferable to other similarly structured online platforms, with potential relevance for a wide range of criminal investigations. In doing so, this work helps bridge disciplinary divides between NLP, criminology, and cybercrime research.

## 8.3 ADVANCING THE MULTIMODAL MACHINE LEARNING RESEARCH

This research also contributes to multimodal ML research by demonstrating the utility of fusing textual writing styles with visual photometric cues for AA tasks. Existing AA literature is overwhelmingly text-centric, and multimodal approaches remain largely unexplored. Through this research, we empirically confirm that visual features, including image composition, background artifacts, and stylistic signatures, can carry meaningful authorship information that complement linguistic cues.

By benchmarking numerous established architectures, fusion strategies, and training objectives, this research attempts to provide empirical insights into how multimodal AA methods can enhance model performance and generalization, particularly in obfuscated real-world scenarios. This adds to the academic discussion on the effects of multimodal ML methods on robustness, domain generalization, and better model representations, showing how these techniques can be repurposed for authorship tasks rather than the usual vision-language applications such as classification or retrieval.

## 8.4 ADVANCING THE ETHICAL DISCOURSE IN RESPONSIBLE AI FOR HIGH-RISK APPLICATIONS

Finally, through this research, we contribute to the ethical and societal dimensions of AA research. As the use of AI applications rapidly evolves, discussions around the responsible and ethical deployment of these applications remain fragmented and underdeveloped, particularly in the field of AA tasks. This research proposes one of the first structured ethical frameworks specifically tailored for AA systems, grounded in privacy, fairness, explainability, societal impact, and accountability.

By operationalizing these principles through role-based and stakeholder-specific guidelines, detailed SDLC mappings, and an applied case study, this research attempts to move beyond abstract generalized ethical recommendations toward actionable procedures. This addresses a significant gap in the literature, where high-level principles often fail to translate into practice. In doing so, the research showcases how AI systems should be governed when deployed in domains involving vulnerable populations, contested identities, and high-stakes decision-making.

## 8.5 THE EMERGING INTEREST FROM THIS RESEARCH

While scientific contributions are meaningful outcomes in themselves, we are happy to report that the datasets, methodologies, and frameworks developed through this research are attracting attention from researchers and practitioners seeking ethical access to the resulting resources. Multiple academic groups have requested permission to use the IDTraffickers and MATCHED datasets for further studies in authorship attribution, cybercrime analytics, online harms, and responsible AI, underscoring the datasets' value as foundational resources for the field.

The reach of this research also extends into operational domains. Inspired by the potential demonstrated through our research, we are now collaborating with **WEB-IQ**, a company specializing in structured, interconnected, and actionable OSINT (Open Source Intelligence) applications for more than 100 law enforcement agencies worldwide. This collaboration aims to convert our research's findings into deployable tools that support real-world investigations of cyber-enabled trafficking and related illicit online activities. We are glad that our research has started moving beyond the academic environment toward practical impact, informing the design of scalable systems that can assist law enforcement in identifying vendor networks, understanding behavioral patterns, and allocating investigative resources more effectively.

Together, these developments highlight that the contributions of this research may have extended beyond theoretical advancement.

# Impact Paragraph

The doctoral regulations at Maastricht University require the inclusion of an impact paragraph in the thesis, which reflects on the scientific and, where relevant, societal impact of the research.[1] Scientific impact refers to the contribution of the research findings to advancing knowledge, methods, theories, and applications within and across disciplines, both in the short and long term. Societal impact pertains to the influence of the research on addressing societal challenges or driving changes and developments in various societal sectors, also considering short- and long-term effects. This addendum addresses the four questions outlined in the regulations to structure the reflection on these impacts.

### *1.Research: What is the main purpose of the research described in the thesis and what are the main results and conclusions?*

Darknet marketplaces and online escort platforms are hubs for illegal activities, including drug trafficking, weapon sales, and human trafficking. To evade detection, criminals often create multiple accounts—either as migrants (across different platforms) or as aliases (using different handles on the same platform). This fragmentation makes it challenging for LEAs to assess the full scope of their operations, enabling these activities to persist undetected. Traditional methods for linking criminal accounts that rely on private identifiers, near-identical ad content, or specific keyword phrases are increasingly ineffective, as vendors frequently modify these features to avoid detection. Therefore, this research proposes leveraging AA approaches to extract and analyze unique linguistic and photometric signatures from online advertisements. On Darknet marketplaces, AA can identify stylistic similarities to connect migrants and aliases, offering a robust and scalable solution for mapping criminal networks. Similarly, in online escort marketplaces—where human traffickers exploit victims through force, deception, or coercion—AA can help link ads even when traditional identifiers like phone numbers or email addresses are absent. By connecting these ads, AA can enable law enforcement to better study human trafficking indicators, distinguishing trafficking instances from legitimate escort services and enhancing efforts to disrupt these criminal networks.

[1] Doctoral Regulations by Maastricht University, retrieved on December, 2025.

The findings of this research highlight the potential of responsible AA applications in combating cyber-enabled trafficking operations on Darknet and online escort marketplaces. By structuring AA tasks into vendor identification and verification tasks, this research provides researchers, LEAs, and practitioners with a dual approach—connecting vendors to existing databases while enabling the discovery of their presence in emerging markets. The results demonstrate that writing style analysis is an effective tool for linking criminals across illegal online platforms. Moreover, integrating multimodal features from text and images, leveraging contrastive learning strategies, and optimizing joint multitask objectives significantly enhance the performance of AA models, improving their ability to connect and retrieve ads even in unseen datasets. Finally, by illustrating how AA models can be used to construct knowledge graphs, the research provides researchers, LEAs, and practitioners with a strategic tool to better allocate resources, analyze patterns, and systematically disrupt trafficking networks.

### *2.Relevance: What is the (potential) contribution of the results of this research to science, and if applicable to societal sectors and societal challenges?*

This research makes contributions to both scientific advancements and societal impact. From a scientific perspective, it advances the field of forensic stylistics by enabling AA approaches on online criminal marketplaces and introducing two novel datasets — IDTraffickers and MATCHED — that facilitate the analysis of AA in high-stakes human trafficking investigations. The research systematically benchmarks traditional and transformer-based AA models, evaluates multimodal integration of text and images, and explores advanced training strategies such as contrastive learning and multitask optimization, providing a foundation for future research on generalization, robustness, and scalability of AA approaches in closed-set and open-set environments. Although not a contribution, the research also emphasizes the limitations of existing AA approaches in evolving digital spaces, stressing the need for continuous adaptation in reaction to emerging challenges, such as using AI-generated content by criminals to evade detection.

Beyond academia, this research focuses on societal impact by empowering law enforcement efforts to combat cyber-enabled trafficking. The development of scalable and generalizable AA models can help equip researchers and LEAs with advanced tools to monitor, identify, link, and study illegal vendors operating across darknet and escort platforms. By replacing reliance on easily manipulated identifiers (e.g., phone numbers, keyword phrases, and special symbols) with stylistic and photometric patterns, these models can enable researchers, LEAs, and practitioners to uncover hidden connections between fragmented accounts, revealing the true scale of criminal networks and allowing law enforcement to prioritize high-impact cases through research and evidentiary linkages efficiently. Furthermore, the proposed retrieval-based vendor verification approach and its role in constructing knowledge graphs enable faster, more scalable, and more efficient investigations. Finally, by proposing an ethical framework for responsible AA deployment, the research ensures that these technologies are used to uphold ethical, legal, and societal standards.

Overall, this research advances AA methods with practical tools to combat cyber-enabled trafficking and cybercrime. It enables law enforcement to link criminal accounts through writing, photometric styles, and multimodal patterns, strengthening investigative precision and efficiency. The ethical framework ensures these tools align with societal values, balancing innovation with accountability. Ultimately, it highlights the critical role of collaboration between AI experts, policymakers, and practitioners in turning technical breakthroughs into real-world impact.

### *3.Target Group: To whom are the research findings interesting and/or relevant? And why?*

This research addresses three primary groups: researchers, practitioners, and LEAs, with implications for advancing theoretical and practical efforts to combat cyber-enabled trafficking operations.

Researchers will benefit from introducing two novel datasets, IDTraffickers and MATCHED, designed to enable AA tasks in sensitive, real-world contexts. By benchmarking performance across diverse ML architectures—from statistical models and neural networks to transformer-based architectures and multimodal systems—this research establishes critical baselines for closed-set authorship identification and open-set authorship verification tasks. Furthermore, applying advanced training strategies, such as contrastive learning and joint multitask optimization, offers actionable insights into improving model generalization and scalability of these AA approaches beyond the training dataset. These contributions fill gaps in existing literature, providing a foundation for future studies on stylistic analysis beyond the text-based AA norms in adversarial environments.

Practitioners, including cybercrime experts and forensic analysts, gain practical tools for linking criminal accounts across platforms like Darknet and online escort marketplaces. Traditional linking methods rely on keywords, identifiers (e.g., phone numbers), or near-identical ad content, which are easily circumvented by online criminals altering these features. This research addresses these limitations by proposing AA techniques that analyze writing styles and photometric patterns to connect accounts even when explicit identifiers are absent. The joint multitask framework further enables the simultaneous identification of known vendors (in existing databases) and the detection of emerging ones in new markets, enhancing practical utility.

For LEAs, the findings translate to actionable tools for disrupting trafficking networks. By linking ads through stylistic (writing and photometric styles) and multimodal patterns, agencies can construct knowledge graphs to map online criminal operations, prioritize resource allocation, and identify high-value targets. For instance, connecting aliases or migrants through stylistic patterns can help uncover the true scale of trafficking activities masked by fragmented accounts. Furthermore, creating a knowledge graph enables LEAs to study trafficking indicators on a broader scale, providing them with a better understanding of online criminal behavior.

Finally, the responsible framework embedded in this research—addressing privacy, fairness, transparency, and societal impact—ensures these tools align with legal and human rights standards, enabling accountability and mitigating risks. Note that while the research is designed for scalability across online platforms, further testing is needed to ensure the adaption across diverse contexts (e.g., social media or emerging marketplaces). Nevertheless, this research bridges critical gaps between technical innovation and real-world application, offering stakeholders a cohesive strategy to study, disrupt, and combat cyber-enabled trafficking.

***4.Activity: In what way can these target groups be involved and informed about the research findings so that the knowledge gained can be used in the future?***

Most of the research presented in this thesis has been documented in several scientific publications, including three peer-reviewed papers published in conference proceedings and journals, with one additional paper currently under review. These works have been presented at their respective conferences through oral presentations or poster sessions. To promote transparency and reproducibility, the source code for all contributions is publicly available on GitHub, and the datasets introduced in this thesis are accessible via the DataverseNL portal. Furthermore, the findings of this research were also shared with a broader audience through a poster presented at CLIN32 2022 and a tutorial presented at the AMLD EPFL 2024 workshops.

# About the Author

Vageesh Saxena was born in Lucknow, India, on December 25, 1992. Between 2004 and 2010, he completed his secondary education at Kendriya Vidyalaya, Shahibaug, Ahmedabad, and Rani Laxmi Bai School, Lucknow. In 2015, he earned his Bachelor of Technology in Electronics and Communication Engineering from Dr. A.P.J. Abdul Kalam University, India, graduating with first-class honors.

After 2+ years of industrial experience in various research-based roles in India, Vageesh continued his academic journey by pursuing a Master of Science in Cognitive Systems: Language, Learning, and Reasoning at the University of Potsdam, Germany, and graduated in 2020 with honors. His Master's thesis (Saxena et al., 2024) focused on extending a self-developed interpretability framework, TX-Ray (Rethmeier et al., 2020), to visualize, quantify, and analyze knowledge transfer in Multilingual Machine Translation within a continual learning setup.

In November 2020, Vageesh began his Ph.D. at the Law & Tech Lab, Maastricht University, under the supervision of Prof. Gerasimos Spanakis and Prof. Gijs van Dijck. His doctoral research was partially supported by the Sector Plan Digital Legal Studies, funded by the Dutch Ministry of Education, Culture, and Science. His Ph.D. work centered on developing responsible authorship attribution techniques and applying them to identify and connect vendors in trafficking networks operating on illegal online marketplaces. His findings have been published in leading peer-reviewed conference proceedings, including ACL, EMNLP, and COLING, as well as in journals like Ethics and Information Technology, Springer.

Beyond his research, Vageesh contributed as a teaching assistant for Exploratory Data Analysis, Legal Analytics, and Advanced Legal Analytics courses. He also supervised Bachelor's and Master's theses at the Faculty of Law and the Department of Advanced Computing Sciences, Maastricht University. Furthermore, he actively engaged with the academic community as a regular reviewer for esteemed conferences (EMNLP 2023, JURIX 2023, CLSR 2023, IJAIT 2023, EWAF 2023) and workshops (NLLP@EMNLP

2023-24).